\documentclass[SYS,biber,plain]{nowfnt} %
\usepackage[utf8]{inputenc}

\usepackage{amsmath}
\usepackage{amssymb}
\usepackage{mathtools}
\usepackage{ marvosym }
\usepackage{booktabs}
\usepackage{paralist}

\usepackage{hyperref}
\usepackage{todonotes}
\usepackage{standalone}
\usepackage{xcolor}

\usepackage{acronym}

\usepackage{mathtools}
\usepackage{amsfonts}
\usepackage{fixltx2e}
\usepackage{tikz}
\usepackage{scalerel}
\usepackage{pict2e}
\usepackage{tkz-euclide}
\usetikzlibrary{patterns,arrows,arrows.meta,calc,shapes,shadows,decorations.pathmorphing,decorations.pathreplacing,automata,shapes.multipart,positioning,shapes.geometric,fit,circuits,trees,shapes.gates.logic.US,fit}
\usetikzlibrary{backgrounds,scopes}
\usetikzlibrary{shadows}
\usetikzlibrary{external}
\usetikzlibrary{shapes}
\usetikzlibrary{positioning, arrows.meta, bending, decorations.text}
\usepackage{pgfplots}
\pgfplotsset{compat=newest}
\usepgfplotslibrary{statistics}
\usepgfplotslibrary{fillbetween}
\usepackage{mdframed}
\usepackage{parskip}
\usepackage[acronym]{glossaries}

\usepackage{pgfplotstable}
\usepackage{colortbl}
\usepackage{arydshln}
\usepgfplotslibrary{groupplots}

\usepackage{nicefrac}

\usepackage{xspace}
\usepackage{microtype}
\usepackage{colonequals}
\usepackage{diagbox} 

\usepackage{eucal} 	 	%
\usepackage{verbatim}      	%
\usepackage{makeidx}       	%

\usepackage{epsfig}         	%
\usepackage{multirow}
\usepackage{multicol}

\usepackage{paralist}
\usepackage{subcaption}
\usepackage{algpseudocode}
\usepackage{url}
\usepackage{graphicx}

\usepackage{algorithm}
\usepackage{listings}

\makeatletter
\def\blfootnote{\gdef\@thefnmark{}\@footnotetext}
\makeatother

\usepackage{amsmath}
\usepackage{amssymb}
\usepackage{wasysym}
\DeclareMathOperator{\scope}{\mathbin{.}}

\newcommand{\satisfies}{\models}
\newcommand{\always}{\square}
\newcommand{\eventually}{\Diamond}
\newcommand{\nextop}{\ocircle}
\newcommand{\until}{\mathcal{U}}

\newcommand{\true}{\relax\ifmmode \mathit{True} \else \em True \/\fi}
\newcommand{\false}{\relax\ifmmode \mathit{False} \else \em False \/\fi}
\newcommand{\aand}{\wedge}
\newcommand{\oor}{\vee}
\newcommand*\cox{\ensuremath{
    \tikz[baseline] {
      \node at (0,0.12) [circle,scale=0.8,draw] {};
      \node at (0,0.12) [diamond, scale=0.5,draw] {};
    }
  }
  \hspace{0.5mm}
}

\newcommand{\probpmdp}{\mathcal{P}}

\newcommand{\control}{u}

\newcommand{\siFLTL}{\textrm{si-FLTL}_{\mathsf{G}}}
\newcommand{\siFLTLX}{\textrm{si-FLTL}_{\mathsf{G_X}}}
\newcommand{\vanish}{\mathsf{vanish}}
\newcommand{\priority}{\rho}
\newcommand{\pspec}{\mathcal{P}}

\newcommand{\denext}{\mathsf{denext}}

\newcommand{\PX}{P_{\mathsf{X}}}

\newcommand{\automaton}{\mathcal{A}}

\newcommand{\game}{{\mathcal{G}}}
\newcommand{\dom}{\mathit{dom}}

\newcommand{\num}[1]{\relax\ifmmode \mathbb #1\else $\mathbb #1$\fi}
\newcommand{\naturals}{{\num N}}
\newcommand{\reals}{{\num R}}
\newcommand{\naturalsle}[1]{\naturals_{\leq #1}}

\newcommand{\union}{\hspace{1mm}\cup\hspace{1mm}}
\newcommand{\intersect}{\hspace{1mm}\cap\hspace{1mm}}

\newtheorem{assumption}{Assumption}[chapter]

\usepackage{ltl}
\usepackage{xspace}
\usepackage{paralist}
\usepackage{wrapfig}
\usepackage{subcaption}

\newcommand{\trueval}{\mathsf{true}}

\newcommand{\nats}{\mathbb{N}}
\newcommand{\natsbot}{\nats \cup \{\bot\}}

\newcommand{\defeq}{\ensuremath{\mathrel{\raisebox{-.3ex}{$\stackrel{\text{\tiny def}}=$}}}\xspace}

\newcommand{\subf}[1]{\mathsf{subf}(#1)}

\newcommand{\trans}{\mathcal{T}}

\newcommand{\val}[2]{\mathit{val}(#1,#2)}

\newcommand{\autA}{\mathcal{A}}
\newcommand{\autB}{\mathcal{B}}

\newcommand{\lang}[1]{\mathcal{L}(#1)}
\newcommand{\relaxfg}{\mathit{Relax}_{\tiny\fg}}

\newcommand{\rej}{\mathsf{rej}}

\newcommand{\props}{\mathcal{AP}}
\newcommand{\alphabet}{\Sigma}
\newcommand{\ialphabet}{{2^\mathcal{I}}}
\newcommand{\oalphabet}{{2^\mathcal{O}}}
\newcommand{\inpv}{\mathcal{I}}
\newcommand{\outv}{\mathcal{O}}
\newcommand{\inpval}{{\sigma_{I}}}
\newcommand{\outval}{{\sigma_{O}}}

\newcommand{\spec}{{\varphi}}
\newcommand{\softSpec}{{\varphi}}
\newcommand{\fg}{\LTLfinally\LTLglobally}
\newcommand{\gf}{\LTLglobally\LTLfinally}
\newcommand{\gphi}{{\LTLglobally\varphi}}

\newcommand{\gphij}{{\LTLglobally\varphi_j}}
\newcommand{\fgphi}{{\LTLfinally\LTLglobally\varphi}}

\newcommand{\gfphi}{{\LTLglobally\LTLfinally\varphi}}

\newcommand{\anot}{\lambda}

\newcommand{\office}{\mathit{office}}
\newcommand{\occupied}{\mathit{occupied}}
\newcommand{\passage}{\mathit{passage}}

\newcommand{\library}{\mathit{library}}
\newcommand{\entr}{\mathit{entrance}}
\newcommand{\corr}{\mathit{corridor}}
\newcommand{\exh}{\mathit{exhibition}}

\def\mdp{\mathcal{M}}
\def\st{S}
\def\sinit{s_{init}}
\def\act{\Act}
\def\tr{\mathcal{P}}
\def\rew{r}

\def\env{\mathcal{E}}
\def\AP{{\mathcal{AP}}} %
\def\Lb{\bar{\mathcal{L}}}
\def\L{\mathcal{L}} %
\def\Lh{\hat{\mathcal{L}}}
\def\obs{\mathcal{O}}
\def\z{\mathcal{Z}}
\def\dfa{\mathcal{D}}
\def\Q{\mathcal{Q}}
\def\qinit{q_{init}}
\def\Facc{\mathcal{F}} %
\def\del{\delta}
\def\delp{\bar{\delta}}
\def\risk{\mathcal{R}}
\def\Dist{{\mathit{Dist}}}
\def\P{\mathrm{Pr}}
\def\E{\mathbb{E}}
\def\dkl{D_{\mathcal{KL}}}
\def\jsd{D_{\mathcal{JSD}}}
\newcommand{\abs}[1]{\lvert #1 \rvert}

\def\Eventually{{\Diamond}}
\def\Always{{\Square}}
\def\Until{{\mathcal{U}}}
\def\Next{\kern0.25ex\vcenter{\hbox{$\scriptstyle\bigcirc$}}\kern0.25ex}

\newcommand{\tool}[1]{\textrm{#1}\xspace}

\newcommand{\dtmc}{\mathcal{D}}

\newcommand{\p}{\ensuremath{\mathbb{P}}}

\newcommand{\pr}{\ensuremath{\mathrm{Pr}}}

\newcommand{\reachProp}[2]{\ensuremath{\p_{\leq #1}(\finally #2)}}
\newcommand{\reachProplT}{\ensuremath{\reachProp{\lambda}{T}}}

\newcommand{\ltlformula}{\psi}
\newcommand{\reachPropSymbol}{\varphi}
\newcommand{\ereachPropSymbol}{\psi}

\newcommand{\er}{\ensuremath{\mathrm{ER}}}

\newcommand{\expRewProp}[2]{\ensuremath{\er_{\leq #1}(\finally #2)}}

\newcommand{\expRewPropkT}{\ensuremath{\expRewProp{\kappa}{T}}}
\newcommand{\rewFunction}{\ensuremath{{c}}}

\newcommand{\finally}{\lozenge}

\newcommand{\sj}[1]{}

\newcommand{\R}{\mathbb{R}}

\newcommand{\Reals}{\ensuremath{\mathbb{R}}\xspace}    %

\newcommand{\Ireal}{[0,\, 1]\subseteq\mathbb{R}}  %
\newcommand{\Irat}{[0,\, 1]\subseteq\mathbb{Q}}  %
\newcommand{\Ex}{\ensuremath{\mathbb{E}}\xspace}        %

\newcommand{\Distr}{\mathit{Distr}}

\newcommand{\distDom}{X}

\newcommand{\distFunc}{\mu}
\newcommand{\distDomElem}{x}
\newcommand{\supp}{\mathit{supp}}

\newcommand{\Finally}{\lozenge \, }
\newcommand{\Ever}{\Diamond \, }

\newcommand{\Paramvar}{\ensuremath{{V}}\xspace}        %

\newcommand{\parammdp}{\mdp_{\Paramvar}}
\newcommand{\MdpInit}[1][]{\ensuremath{\mdp{#1}=(S{#1},\sinit{#1},\Act,\probmdp{#1})}}
\newcommand{\pMdpInit}[1][]{\ensuremath{\parammdp{#1}=(S{#1},\,\sinit{#1},\Act,\Paramvar,\probmdp{#1})}}
\newcommand{\probmdp}{\mathcal{P}}

\newcommand{\epsgraph}{\varepsilon_\text{graph}}

\renewcommand{\Pr}{\ensuremath{\textnormal{Pr}}}

\newcommand{\sched}{\ensuremath{\sigma}}
\newcommand{\Sched}{\ensuremath{\mathit{Pol}}}

\newcommand{\Act}{\ensuremath{\mathit{Act}}}
\newcommand{\ActS}{\ensuremath{\mathit{Act(s)}}}

\newcommand{\pmdp}{\ensuremath{\mathcal{P}}}

\newcommand{\pathset}{\mathsf{Paths}}

\newcommand{\pathsfin}{\pathset_{\mathit{fin}}}

\renewcommand{\path}{\pi}

\DeclareMathAlphabet{\mathpzc}{OT1}{pzc}{m}{it}
\def\presuper#1#2%
  {\mathop{}%
   \mathopen{\vphantom{#2}}^{#1}%
   \kern-\scriptspace%
   #2}

\newcommand{\paramspace}[1][]{\ensuremath{{V}_{#1}}}

\newcommand{\reachPropg}[2]{\ensuremath{\p_{\leq #1}(\finally #2)}}

\newcommand{\satprob}{F(\vmdp,\varphi)}
\newcommand{\satreachprob}{F(\vmdp,\varphi_r)}
\newcommand{\satreachprobmc}{F(\vmc,\varphi_r)}

\newcommand{\falseprob}{F(\vmdp,\neg\varphi)}

\newcommand{\reachPropgT}{\ensuremath{\reachPropg{\lambda}{T}}}
\newcommand{\vmdp}{\mdp_{\probdist}}

\newcommand{\probdist}{\ensuremath{\mathbb{P}}}
\newcommand{\PP}{\mathbb{P}}

\newcommand{\minimize}{\textnormal{minimize}}
\newcommand{\subjectto}{\textnormal{subject to}}
\newcommand{\vmc}{\dtmc_{\probdist}}

\newcommand{\instantiation}{\mathbf{v}}

\newcommand{\probumdp}{\mathcal{P}}
\newcommand{\intervals}{\mathbb{I}}
\newcommand{\osched}{\ensuremath{\mathit{\sigma}}}
\newcommand{\oSched}{\ensuremath{\Sigma}}
\newcommand{\states}{\ensuremath{S}}
\newcommand{\ObsSym}{{Z}}
\newcommand{\ObsFun}{{O}}
\newcommand{\PomdpInit}[1][]{\nompomdp{#1}=(\mdp{#1},\ObsSym{#1},\ObsFun{#1})}
\newcommand{\nompomdp}{\mathcal{M}_{\ObsSym}}
\newcommand{\pomdp}{\mathcal{M}_{\ObsSym}}
\newcommand{\upomdp}{\mathcal{M}_{\ObsSym,\probmdp}}

\newcommand{\fsc}{\ensuremath{\mathcal{A}}}

\DeclareMathOperator*{\argmax}{\text{arg}\,\text{max}}

\newcommand{\policy}{\osched}
\newcommand{\Policy}{\oSched}

\usepackage{aligned-overset}

\def\addlegendimage{\csname pgfplots@addlegendimage\endcsname}

\newcommand{\obsSeq}{\mathsf{ObsSeq}_{\mathit{fin}}}
\newcommand{\obsSeqFin}{\obsSeq}
\newcommand{\obspath}{\ensuremath{\path_o}}
\newcommand{\gridsizeparam}{\ensuremath{c}}
\newcommand{\gridScale}{1}

\newcommand{\fillGridAt}[3]{
	\node [xshift=.5*\gridScale cm,yshift=.5*\gridScale cm] at (#1,#2){#3};	
}

\newcommand{\RNNfun}{\ensuremath{\hat{\osched}}}

\tikzstyle{state}=[
	circle,
	minimum size=20pt,
	draw=black,
	thick,
	fill=blue!25,
	text=black,
	inner sep=0pt
	]

\newcommand{\ie}{\emph{i.e.}}
\newcommand{\eg}{\emph{e.g.}}
\newcommand{\etc}{\emph{etc}}

\newtheorem{thm}{Theorem}

\newtheorem{problem}{Problem}
\numberwithin{problem}{chapter}

\title{Formal Methods for~Autonomous~Systems}

\maintitleauthorlist{
  Tichakorn Wongpiromsarn \\
  Iowa State University \\
  nok@iastate.edu
\and
Mahsa Ghasemi\\
Purdue University\\
mahsa@purdue.edu\\
\and
Murat Cubuktepe\\
Georgios Bakirtzis\\
Steven Carr\\
Mustafa O. Karabag\\
Cyrus Neary\\
Parham Gohari\\
Ufuk Topcu \\
  The University of Texas at Austin \\
  utopcu@utexas.edu
}

\date{Working paper}

\newacronym{ltl}{LTL}{linear temporal logic}
\newacronym{bdd}{BDD}{binary decision diagram}
\newacronym{tla}{TLA}{temporal logic of actions}
\newacronym{promela}{PROMELA}{process meta-language}
\newacronym{wts}{WTS}{weighted transition system}
\newacronym{fltl}{FLTL}{finite linear temporal logic}
\newacronym{gr[1]}{GR[1]}{generalized reactivity[1]}
\newacronym{maxsat}{MaxSAT}{maximum satisfiability}
\newacronym{cnf}{CNF}{conjunctive normal form}
\newacronym{lp}{LP}{linear programming}
\newacronym{scp}{SCP}{sequential convex programming}
\newacronym{mc}{MC}{Markov chain}
\newacronym[plural=MDPs, firstplural=Markov decision processes (MDPs)]{mdp}{MDP}{Markov decision process}
\newacronym{pmdp}{pMDP}{parametric Markov decision process}
\newacronym{umdp}{uMDP}{uncertain Markov decision process}
\newacronym[plural=POMDPs, firstplural=partially observable Markov decision processes (POMDPs)]{pomdp}{POMDP}{partially observable Markov decision process}
\newacronym{upomdp}{uPOMDP}{uncertain partially observable Markov decision process}
\newacronym{fsc}{FSC}{finite state controller}
\newacronym{nmt}{NMT}{natural motion trajectory}
\newacronym{roz}{ROZ}{restricted operating zone}
\newacronym{scltl}{scLTL}{syntactically co-safe linear temporal logic}
\newacronym{slam}{SLAM}{simultaneous localization and mapping}
\newacronym[plural=RNNs, firstplural=recurrent neural networks (RNNs)]{rnn}{RNN}{recurrent neural network}
\newacronym{lstm}{LSTM}{long short-term memory}
\newacronym{milp}{MILP}{mixed-integer linear programming}
\newacronym{mpc}{MPC}{model predictive control}
\newacronym{rrt}{RRT}{rapidly-exploring random trees}
\newacronym{ros}{ROS}{robot operating system}
\newacronym{ompl}{OMPL}{open motion planning library}
\newacronym{darpa}{DARPA}{defense advanced research projects agency}
\newacronym{rddf}{RDDF}{route data definition file}
\newacronym{dfa}{DFA}{deterministic finite automaton}
\newacronym{dtmc}{DTMC}{discrete-time Markov chain}

\newacronym{rl}{RL}{reinforcement learning}

\usepackage{mwe}

\author[1]{Wongpiromsarn, Tichakorn}
\author[2]{Mahsa Ghasemi}
\author[3]{Murat Cubuktepe}
\author[3]{Georgios Bakirtzis}
\author[3]{Steven Carr}
\author[3]{Mustafa O. Karabag}
\author[3]{Cyrus Neary}
\author[3]{Parham Gohari}
\author[3]{Topcu, Ufuk}

\affil[1]{Iowa State University; nok@iastate.edu}
\affil[2]{Purdue University; mahsa@purdue.edu}
\affil[3]{The University of Texas at Austin; utopcu@utexas.edu}

\articledatabox{\nowfntstandardcitation}

\makeglossaries

\begin{document}

\makeabstracttitle

\begin{abstract}
  Formal methods refer to rigorous, mathematical approaches to system development and have played a key role in establishing the correctness of safety-critical systems.
  The main building blocks of formal methods are models and specifications, which
  are analogous to behaviors and requirements in system design and give us the means to verify and synthesize system behaviors with \emph{formal guarantees}.
  
  This monograph provides a survey of the current state of the art on applications of formal methods in the autonomous systems domain. We consider correct-by-construction synthesis under various formulations, including closed systems, reactive, and probabilistic settings.
  Beyond synthesizing systems in known environments, we address the concept of \emph{uncertainty} and bound the behavior of systems that employ learning using formal methods. Further, we examine the synthesis of systems with \emph{monitoring}, a mitigation technique for ensuring that once a system deviates from expected behavior, it knows a way of returning to normalcy. We also show how to overcome some limitations of formal methods themselves with learning. We conclude with future directions for formal methods in reinforcement learning, uncertainty, privacy, explainability of formal methods, and regulation and certification.
\end{abstract}

\chapter{Introduction}
\glsresetall
\label{chap:intro} %

This monograph is about a class of formal methods that verify \emph{systems} properties we care about, such as ``bad things never happen'' (\emph{safety}) and ``good things will eventually happen'' (\emph{liveness}), for autonomous system analysis and synthesis. Unlike traditionally engineered systems, autonomous systems need to readily react to changing environments and operational situations, often by coordinating and adapting. This ability of autonomous systems makes them both valuable and challenging to certify simultaneously. For example, compared to traditional industrial control systems where the operational contexts are static, autonomous systems must behave acceptably in various environments, often partially unknown~\citep{bakirtzis:2022}.

Finding design bugs that occur at the interaction of subsystems is particularly challenging~\citep{bakirtzis:2022a}. In traditionally engineered systems testing and simulation are often forms of sufficient certification. However, these techniques are inadequate in ensuring the absence of design bugs in autonomous systems. Formal methods applied to autonomous systems bound the uncertainty arising from the unknown physical environments they deploy in and, therefore, assure that they will not misbehave.

Standard systems engineering goes through multiple steps of iteration to produce a system. Late in the lifecycle, system designs go through verification \& validation. It is at verification \& validation where formal methods provide proof of correct behavior for autonomous systems. The most significant barrier to certifying systems is ensuring that the assumptions made at the previous stages of systems engineering agree with the formal model. In subsequent sections, we introduce some of these models in increasing levels of expressivity. From simple to complex, system designers use these different models based on the system's expected behavior and deployment context. Therefore, applying formal methods to autonomous system design is not merely picking the most expressive model but, rather, the most appropriate one for a given application.

The impact of formal methods can be twofold. First, they provide provable guarantees that the system satisfies the desired properties \citep{BK08,holzmannSPIN,Holzmann94Theory}. This is the usual way we use formal methods, and it requires that both the system and the properties we are trying to prove are well-defined within a formalism. In addition, provable guarantees relate to the model and not the actual system. In the past, the congruence between the actual system and its model was done manually in an ad-hoc manner. Today, synthesis methods exist that automatically output the assured behavior of the system under examination from a formal model with provable guarantees. Second, we use formal methods to interrogate our assumptions for the system design~\citep{lamport:2002,newcombe:2015}. Often, it is more useful to think about a design problem formally and use formal models as decision guides for the rest of the development of the system. This way of using formal methods is done from the early lifecycle all the way up to deployment. The expectation is not that the system contains provable guarantees via formal methods but that the system developers are informed about potential issues they need to address at the design phase, e.g., clashing requirements and beyond, modeling the open world environments we deploy autonomous systems.

\section{Autonomous Systems}

While the public\footnote{This section partially follows~\citet{Wongpiromsarn:2020:Journey}.}  is most familiar with misbehavior in the context of autonomous vehicles, similar inconsistencies arise in a range of applications of autonomous systems.

\subsubsection{Historical roadmap}
\begin{figure}[!t]
  \centering
  \begin{tikzpicture}[
    font=\small,
    inout/.style={text width=3.0cm,text centered, font=\footnotesize},
    block/.style={minimum width=3.0cm,draw,text width=2.2cm,text centered},
    forked/.style={to path={-| (\tikztotarget) \tikztonodes}},
    node distance=2cm
    ]

    \node [block] (world)   at (-4.0,  0.0) {World};
    \node [block] (sense)   at (0.0, 0.0)   {Sensors};
    \node [block] (locperc) at ( 4.0,  0.0) {Localization \& Perception};
    \node [block] (planctr) at ( 4.0,  -2.0) {Planning \& Control};
    \node [block] (physical) at ( -4.0, -2.0) {Physical Component};

    \draw[-stealth,semithick]
    (world) edge (sense)
    (sense) edge (locperc)
    (locperc) edge (planctr)
    (physical) edge (world)
    (planctr) edge (physical);
  \end{tikzpicture}
  \caption{A typical architecture of autonomous vehicles.}
  \label{fig:architecture}
\end{figure}
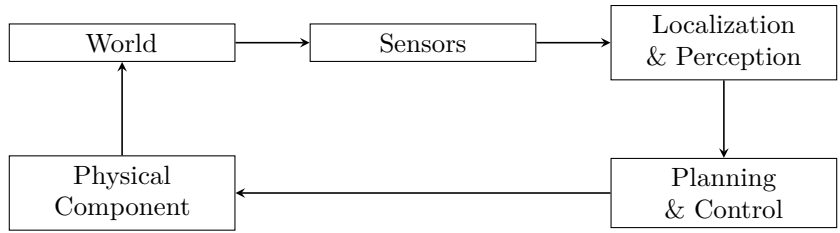

The modern development roadmap of autonomous systems can be illustrated by that of modern autonomous vehicles revealing that there are several design challenges to overcome. In particular, building and deploying reliable autonomy requires designers and engineers to consider the interactive nature between subsystems. While individually subsystems may be well understood, their composition can give rise to emergent behavior \citep{Campbell:2010:Autonomous}. Briefly, most autonomous vehicles can be modeled via two main subsystems: perception and localization, and planning and control (Figure~\ref{fig:architecture}). The essential sensors can include GPS, IMU, odometry, lidar, radar, and cameras. The perception and localization subsystem use the measurements from these sensors to localize the vehicle within the map, detect and track objects and relevant features (e.g., road, lane marking, stop lines, etc.), and create a parsable representation of the world around the vehicle. The planning component uses this information to compute a vehicle's trajectory. Finally, the control component sends actuation commands, including brake, throttle, steering, gear shifting, to keep the vehicle within this calculated trajectory.

One of the main key technological developments to steer the improvement of autonomous system was the introduction Velodyne’s HDL-64E sensor. This spinning lidar and its successors, including the more affordable models with 16 and
32 beams and the high-end models with 128 beams, are still a key component of many autonomous vehicles today.

The planning component is typically decomposed into three levels---the mission, the behavioral, and the trajectory planners---although naming and detail of responsibilities and algorithms
varies between implementations. Roughly, the mission planner computes a high-level route for the vehicle
to complete its mission. The behavioral planner is responsible for making local decisions (e.g.,
whether to stay in lane, proceed through an intersection, etc.) and typically is implemented as a
finite-state machine. The trajectory planner then translates the decision into a trajectory for the
vehicle to follow, using variations of optimization-based (e.g., model predictive control (MPC)) and graph-based (e.g., rapidly-exploring
random trees (RRT) and
probabilistic roadmap (PRM)) approaches. The early controllers of this era were typically based on pure pursuit and proportional-integral-derivative (PID) control. More details about planning and control algorithms can be found in \citet{Paden:2016:Survey}.

Similarly, to the development of modern autonomous vehicles, another example is that of unmanned aerial vehicles, which can be perceived as having the same components, except that the calculated dynamics concern with flight rather than driving. Unmanned aerial vehicle software design has improved from manual definition of state machines and ad hoc tuning of control parameters in a fly-fix-fly fashion to automated synthesis from verified and validated formal models, such as state machine representations. The hardware side has also seen both an increase in capabilities and a reduction in size, giving rise to, for example, networked clusters of very small unmanned aerial vehicles cooperating for common goals. Larger unmanned aerial vehicles have become more precise, less perturbed by uncertainties in the environmental parameters, and with faster and smaller motors.

One useful delineation for understanding modern autonomous systems is, therefore, the following: (1) there is increasing development in high-precision sensors and controllers with reduced cost and (2) there are new software synthesis techniques that can make those controllers highly reliable and modular. In this work we will consider the second part of the modern development of autonomous systems as it related from the translation of formal specification to the synthesis of behavior on fabric, abstracting away from particular implementation details in the systems and microcontrollers.

\subsubsection{Major Technological Challenges}
The challenges associated with individual components (e.g., developing scalable algorithms for perception, planning, control, and contingency management) and their integration into a holistic system are further intensified by the safety-critical nature of autonomous systems. In particular, subtle design bugs may arise from the unforeseen interactions among different components and manifest as undesirable behavior only under a specific set of conditions, making them very hard to catch using simulation and testing. For example, consider an implementation of an autonomous vehicle that composes of the following components.
\begin{itemize}
  \setlength{\parsep}{0pt}
  \setlength{\itemsep}{0pt}
\item The trajectory planner, which generated a path for the vehicle to follow.
\item The safety system, which rapidly decelerated the vehicle when it deviated too much from the
  planned path and got too close to an obstacle.
\item The low-level steering controller, which limited the steering rate at low speeds to protect the vehicle steering system.
\end{itemize}

Each of these functionalities is straightforward to implement in isolation. However, when combined, they can leed to unsafe behavior under specific circumstances and contexts, e.g., when the vehicle has to make a tight turn while merging into traffic facing a concrete barrier next to the major road. In this case, the vehicles planned path contains a sharp turn. Accelerating from a low speed, the controller cannot execute the turn closely due to the limited steering rate. As a result, the vehicle may deviate from the path and head instead towards the concrete barrier. 

Further safety systems are necessary to avoid hazardous situations like the above, but even then the safety system may be a cause of concern as well. By activating it may slow the vehicle down as it is taking the sharp turn, leading to an even stricter limit on the steering rate. This cycle can cause the vehicle to be stuck at the corner of a sharp turn, dangerously stuttering in the middle of an intersection.
The analysis presented in \citet{WMML:HSCC09} reveals that the software design was not inherently flawed.
The undesirable behavior was caused by an unfortunate choice of certain parameters relating to the
geometric properties of the planner-generated paths, compounding the challenge of safety when deploying autonomy.

In short, the key technical challenges in autonomous systems evolved around the following factors: uncertainties, complex tasks, and interconnection of computing, communication, and physical components. The uncertain and unstructured nature of environments lead to an unreasonably large number of test scenarios and give rise to the question of how to address edge cases. The complex interaction of different components can cause any change in one component to engender interaction faults once integrated with others. Finally, complex tasks are primarily handled by handcrafted implementations, e.g., via increasingly complicated finite state machines, which ends up hosting several hacks to handle corner cases encountered during testing. In particular, a naive behavioral planner architecture can be implemented as a finite state machine with less than five states. Still, to address corner cases such state machines at a minimum might need three interacting finite state machines, each containing more than 20 states, making it almost impossible to analyze or debug.

\section{Overview of Verification and Synthesis}

This article aims to explain frameworks for formal verification and synthesis of autonomous systems to provide a formal, mathematical guarantee of the correctness of such a system with respect to its desired properties.
These systems typically consist of both low-level (continuous) dynamics associated with the physical hardware and the high-level (discrete) logics that govern the overall behavior of the systems.
Synthesis and verification of these systems thus require integration of reasoning about discrete and continuous behaviors within a single framework.
A common approach to enable such integration is to construct a finite state model that serves as an abstract model of the physical system (which typically has infinitely many states) and apply formal verification and synthesis to the resulting finite state model.
Several abstraction methods have been proposed based on a fixed abstraction \citep{kress-gazit07,conner07rsj,Kloetzer08,wongpiromsarn12tac,Tabuada04lineartime,Girard09Hierarchical} and sampling \citep{karaman09cdc,Castro:2013:CDC,Wongpiromsarn:2021:Minimum}.
The focus of this article, however, is on the high-level logics and we assume that the system can be
abstracted using a finite state model.

The research presented here consists of three key components: specification, synthesis, and
verification. Specification refers to a precise description of the system and its desired
properties. However, this precise description of the system does not need to capture all the details
of the actual implementation itself. To simplify the analysis of the system, one may want to capture
only the essential aspects and abstract the actual implementation in this description.
In this article, we use the term ``desired properties'', ``desired behaviors'', and ``requirements''
interchangably to refer to what we want the system to do.

Verification is the process of checking the correctness of the system. Here, correctness is only
defined relative to the desired properties.
Specifications of both the system and its desired properties are essential in this process.
As previously discussed, verifying the correctness of complex systems such as autonomous vehicles can
be very difficult due to the interleaving between their continuous and their discrete
components. Although much work exists in this domain, verifying such systems remains time-consuming
and requires some level of expertise.

There has been a growing interest in the automatic synthesis of autonomous systems that provide formal system correctness guarantees to complement system verification. This type of automated synthesis can potentially reduce the time and cost of the system development cycle. It will ultimately help reduce the number of iterations between redesigning the system and verifying the new design.

\section{What Makes It Hard?}

Autonomous systems typically feature a tight interaction between the computing and the physical components. A combination of model-based and data-driven approaches may be employed to design these components. Furthermore, autonomous systems are composed of multiple heterogeneous components whose complex interactions may cause any change in one component to affect others in unexpected ways. 

\begin{figure}[!t]
  \centering
  \includegraphics[width=0.7\textwidth,trim={5cm 6.5cm 4.5cm 8cm},clip]{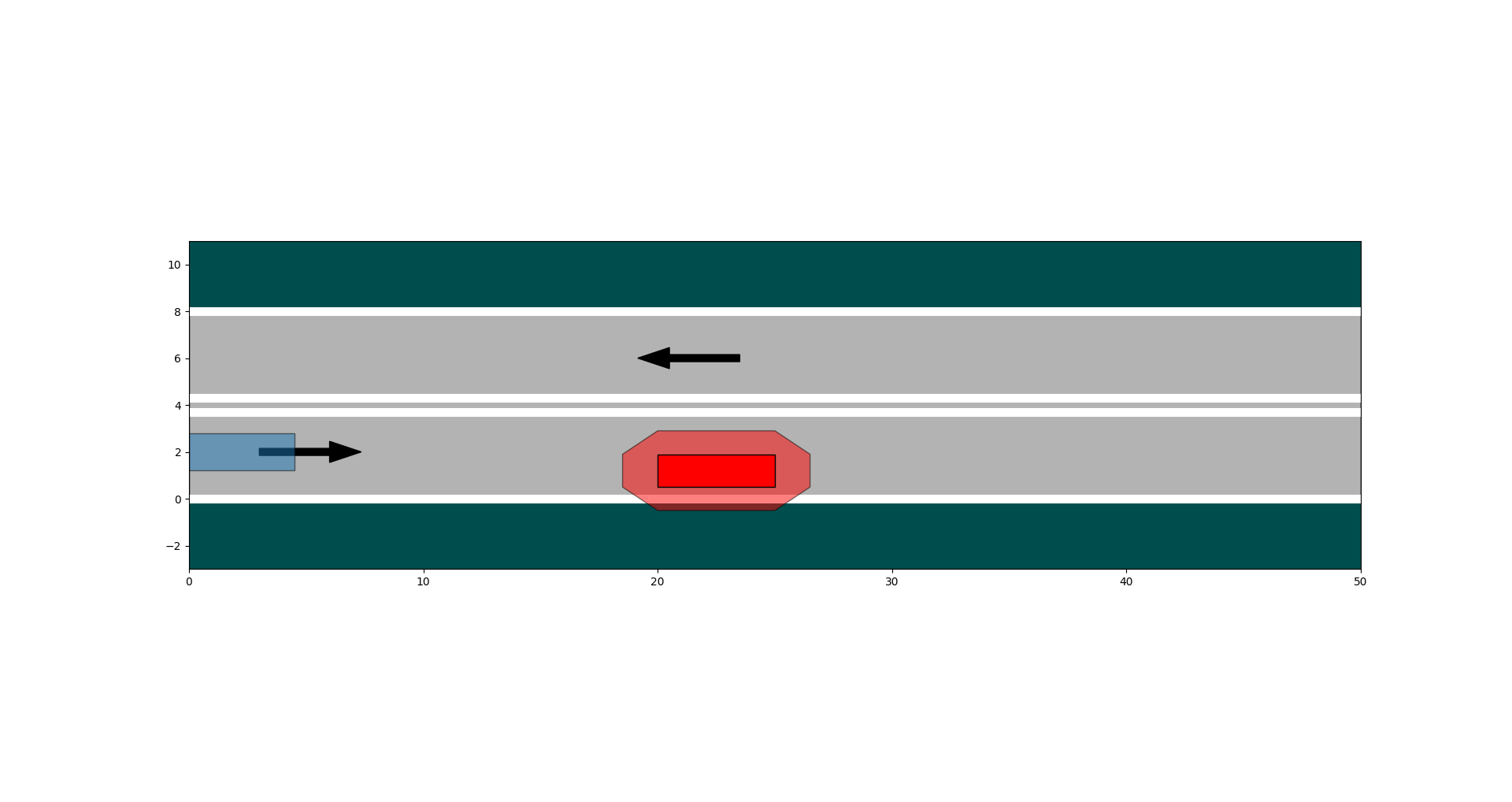}
  \caption{
    The autonomous vehicle (blue rectangle) encounters a stationary vehicle (red rectangle)
    on a two-lane road with a double white lane divider.
    The red octagon represents the clearance zone around the stationary vehicle.
  }
  \label{fig:setup-overtaking}
\end{figure}

Many autonomous systems need to perform complex tasks and are safety-critical. As a result, they are subject to strict regulations. A software bug in these systems can lead to a violation of law and morality. The complex tasks autonomous systems have to perform and regulatory requirements form a set of complex rules that the system needs to satisfy. Under certain situations, these rules may be conflicting, i.e., they cannot be simultaneously satisfied. That means that there can be requirement misalignment with our values that then clash in the eventual implementation of the system. For example, item 221 of Singapore's final theory of driving~\citep{SG-FTD} suggests keeping a safe gap of one meter when passing by a parked vehicle. In contrast, item 52 of Singapore's basic theory of driving~\citep{SG-BTD} prohibits crossing a solid double white lane divider. As a result, when encountering a vehicle that is improperly parked in a lane with a solid double white lane divider (Figure~\ref{fig:setup-overtaking}), an autonomous vehicle may need to violate either of these rules unless the lane is wide enough to laterally accommodate two cars with a buffer of one meter.

The uncertain and unstructured nature of environments in which the systems operate further amplifies the complexity of the verification and synthesis of these systems. In particular, the sociotechnical environments may change abruptly, drastically, and unexpectedly and may include adversaries. Such an open-world challenge leads to an unreasonably large number of test scenarios. It drives the question of edge cases from the safety and verification and the certification and regulation aspects.

\section{Organizational Outline}

This monograph is segmented to incrementally introduce different types of formalisms for the analysis and synthesis of autonomous systems. In particular, we first introduce basic concepts of models---what they are, why the exist, and how they are used---and specifications---formal semantics for modeling autonomous system behavior (Section 2). Then, we show how those formal specifications can verify certain system requirements that are capturable within a logic (Section 3). We continue by showing how those specifications can synthesize behavior equipped with formal guarantees in varying mathematical settings that capture different systems scenarios (Sections 4, 5 and 6). Using these problem formulation models we show how to deal with partiality in the information the system can perceive (Section 7). We extend the synthesis problem to monitoring for runtime assurance of correct behavior (Section 8). We address the addition of learning in the verification and synthesis problem (Section 9). We conclude with the open problems in the intersection of formal methods and autonomy (Section 10).

\section{A Note on the Coverage of the Article}

In the rest of this article, we follow the exposition of several earlier publications by the authors. We list such publications at the beginning of the associated sections. Additionally, while the upcoming sections strive to give an objective coverage of the existing work, they do not provide a complete literature survey and, at times, are biased toward the references that had influenced the authors' own work.

\chapter{Models and Specifications}
\glsresetall
\label{chap:models_and_specifications}

We present\blfootnote{At times, the exposition in this section follows~\citet{BK08}.} a series of formal verification and synthesis problems and their applications in the context of autonomous systems. This section provides a high-level view on formal verification and synthesis and their main building blocks: {\it models} and {\it specifications}. 

Models are representations of our knowledge---and the limitations in it---on, e.g., the capabilities of the system of interest, the environment in which it is to operate, the uncertainties to which it is subject to, and the resources it has access. Specifications are representations of the requirements that we impose on the system's operation and the assumptions we place on the aspects that influence the system's behavior but cannot be controlled by the system.

\begin{figure}
  \centering
  \input{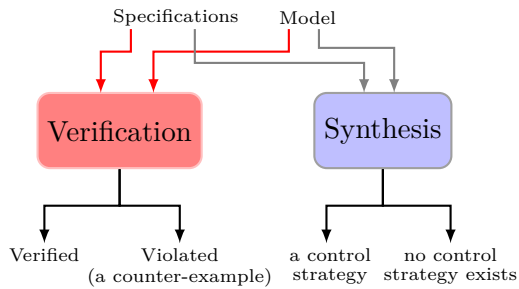}
  \caption{Inputs and potential outcomes of formal verification and synthesis.}
  \label{fig:ver-syn-chart}
\end{figure}

Models and specifications are used to formally verify systems (\figurename~\ref{fig:ver-syn-chart}). Roughly speaking, verification is concerned with whether all (or most) executions of the system generated by implementing a given control strategy will respect a specification. The answer to a verification question is either affirmative, i.e., all possible executions of the system respect the specification, or negative, i.e., counter-example traces that don't respect the specification are output. Synthesis is concerned with the possibly more ambitious question of whether and how a system, e.g., by choice of a control strategy, can ensure the satisfaction of the specifications in all system executions. The outcome of synthesis is a strategy that, when implemented, ensures that the system satisfies the specifications or evidence that no such strategy exists.

The choice of mathematical representations and specification languages for verification and synthesis depends on several factors. Factors include the objective of verification and synthesis, the type of knowledge about the underlying system, and the uncertainties in this knowledge. A hierarchy of mathematical models and specification languages exists to address these factors. For example, one sequence of models that we will encounter in the upcoming sections is from finite transition systems to Markov decision processes to partially observable Markov decision processes to uncertain partially observable Markov decision processes. This evolution will help account for the introduction of stochastic uncertainties, limitations in the availability of run-time information, and limitations in the knowledge about the stochastic uncertainties in the way the system behaves or perceives its environment.

\begin{figure}
  \centering
  \input{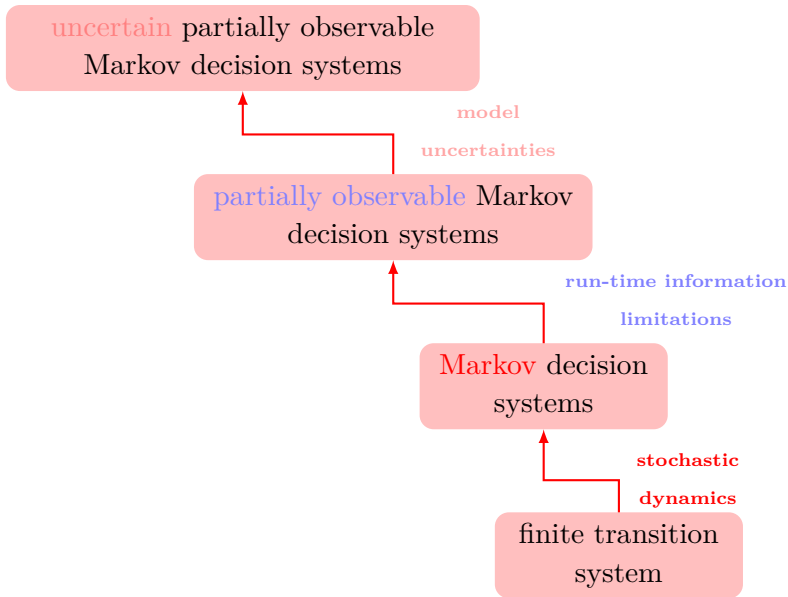}
  \caption{A sequence of models with increasing expressivity.}
  \label{fig:model-hier}
\end{figure}

We introduce the first type of model in the sequence, (finite) {\it transition systems} as the basis for modeling as well as {\it linear temporal logic} as an example of a formal specification language. As they become relevant, we will introduce extensions of such models and variants of temporal logic.

It is impossible to present all models and specification languages exhaustively. For example, one of the aspects of formally verifying autonomous systems that we do not consider is verification and synthesis using models with continuous (or hybrid) state and action spaces, which is illustrated elsewhere \citep{kress-gazit07,conner07rsj,Kloetzer08,wongpiromsarn12tac,Tabuada04lineartime,Girard09Hierarchical,karaman09cdc,Castro:2013:CDC,Wongpiromsarn:2021:Minimum,DBLP:conf/cdc/SrinivasanCE18}.

\section{Transition Systems}

We first introduce some notation. Given a set $X$, let $X^*$, $X^\omega$ and $X^+$ denote the set of
finite, infinite and nonempty finite strings, respectively, of $X$ and let
$|X|$ denote the cardinality of $X$.
For sequences $\pi$, $\pi_1$ and $\pi_2$, let
$\pi_1 \pi_2$ denote a sequence obtained by concatenating $\pi_1$ and $\pi_2$
and let $\pi^\omega$ denote an infinite sequence obtained by
concatenating $\pi$ infinitely many times.

A transition system is a mathematical description of the behavior of systems with discrete inputs, outputs, internal states, and transitions between states. \emph{Atomic propositions} that express essential characteristics of individual states of the system formalize the behavior of transition systems. Roughly, a \emph{proposition} is a statement that can be either true or false, but not both. An \emph {atomic proposition} is a proposition whose truth or falsity does not depend on the truth or falsity of any other proposition. For example, a statement ``traffic light is green'' is an atomic proposition, whereas a statement ``traffic light is either green or red'' is not an atomic proposition.

\begin{definition}
  \label{def:transition_system}
  A (labeled) \emph{transition system} $TS$ is a tuple
  $$TS = (S, Act, \to, I, AP, L)$$ is composed of the following data.
  \begin{itemize}
  \item A set of states, $S$.
  \item A  set of actions, $Act$.
  \item A transition relation, $\to \subseteq S \times Act \times S$.
  \item A set of initial states, $I \subseteq S$.
  \item A set of atomic propositions, $AP$.
  \item A labeling function $L\colon S \to 2^{AP}$.
  \end{itemize}
  We use the relation notation, $s \stackrel{\alpha}{\to} s'$, to denote $(s, \alpha, s') \in \to$.
  The transition system $TS$ is called \emph{finite} if $S$, $Act$ and $AP$ are finite. Note that a transition system $TS$ may consist of a subset of these elements
\end{definition}

The following example is borrowed from \citep{BK08}.

\begin{example}
  \label{ex:light_model}
  Consider a traffic light that can be either red or green.
  Let $g$ denote an atomic proposition stating that the light is green.
  This traffic light can be modeled by a transition system
  $$T = (S, Act, \to, I, AP, L),$$
  where
  $S = \{s_{1}, s_{2}\}$,
  $Act = \{\alpha\}$,
  $\to = \{(s_{1}, \alpha, s_{2}), (s_{2}, \alpha, s_{1})\}$,
  $I = \{s_{1}\}$,
  $AP = \{g\}$ and
  $L : S \to 2^{AP}$ is defined by $L(s_{1}) = \emptyset$ and $L(s_{2}) = \{g\}$.
\end{example}

Given a transition system $TS = (S, Act, \to, I, AP, L)$, $s \in S$ and $\alpha \in Act$, we let
$$Act(s) = \{\alpha \in Act : \exists s' \in S \hbox{ such that } s \stackrel{\alpha}{\to} s'\}$$
denote the set of enabled actions in $s$,
\[Post(s, \alpha) = \{s' \in S : s \stackrel{\alpha}{\to} s'\} \text{ and }
  Post(s) = \bigcup_{\alpha \in Act} Post(s, \alpha)\]
denote the set of direct successors of $s$.
We say that $TS$ is \emph{action-deterministic} if and only if
$|I| \leq 1$ and $|Post(s, \alpha)| \leq 1$ for all $s \in S$ and $\alpha \in Act$.
A sequence of states, either finite $\pi = s_0s_1 \cdots s_n$, or infinite $\pi = s_0s_1 \cdots $,
is a \emph{path fragment} if $s_{i+1} \in Post(s_i)$ for all $i \geq 0$.
A \emph{path} is a path fragment such that $s_0 \in I$ and
it is either a finite path fragment that ends in a state $s$ with $Post(s) = \emptyset$ %
or an infinite path fragment.
We denote the set of paths in $TS$ by $Path(TS)$.
The \emph{trace} of an infinite path fragment $\pi = s_0s_1 \ldots $ is defined by
$trace(\pi) = L(s_0)L(s_1)\cdots$.
The set of traces of $TS$ is defined by
$$Trace(TS) = \{trace(\pi) : \pi \in Path(TS)\}.$$

\begin{remark}
  Transition systems that are not action-deterministic are those in which
  some action, when applied in some state, leads to several possible next states.
  Hence, they can be used to capture uncertainties in the system, especially
  those that arise from difference choices of valid environment behaviors over which the system does
  not have control.
\end{remark}

\begin{example}
  \label{ex:light_path}
  Consider a traffic light $T$ (Example~\ref{ex:light_model}).
  An infinite sequence $\pi = (s_{1}s_{2})^{\omega}$ is a path of $T$
  with the corresponding trace $trace(\pi) = (\emptyset \{g\})^{\omega}$.
  The set of paths and the set of traces of $T$ are given by
  $Paths(T) = \{\pi\}$ and $Traces(T) = \{trace(\pi)\}$, respectively.
\end{example}

Complex systems are typically composed of
multiple components that can be executed at the same time.
Suppose a transition system can model each component of a system. Composing the complete system amounts to composing the finite transition systems representing individual components. There are several composition techniques, depending on how the components interact, e.g., no communication to synchronous and asynchronous message transfer \citep{BK08}. For example, hand-shaking is a mode of communication between the components that leads to synchrony. The composition of finite transition systems by hand-shaking is defined as follows.

\begin{definition}
  Let two transition systems
  \[TS_1 = (S_1, Act_1, \to_1, I_1, AP_1, L_1) \text{ and } TS_2 = (S_2, Act_2, \to_2, I_2, AP_2, L_2)\]
  be given.
  Their composition (by hand-shaking), denoted by $TS_1 || TS_2$, is the transition system defined by
  \begin{equation*}
    TS_1 || TS_2 = (S_1 \times S_2, Act_1 \cup Act_2, \to, I_1 \times I_2, AP_1 \cup AP_2, L),
  \end{equation*}
  where $L(\langle s_1, s_2 \rangle ) = L_1(s_1) \cup L_2(s_2)$ and $\to$ is defined by the following rules:
  \begin{itemize}
  \item If $\alpha \in Act_1 \cap Act_2$, $s_1 \stackrel{\alpha}{\to} s_1'$ and $s_2 \stackrel{\alpha}{\to} s_2'$, then
    $\langle s_1, s_2 \rangle \stackrel{\alpha}{\to} \langle s_1', s_2' \rangle$.
  \item If $\alpha \in Act_1 \setminus Act_2$ and $s_1 \stackrel{\alpha}{\to} s_1'$, then
    $\langle s_1, s_2 \rangle \stackrel{\alpha}{\to} \langle s_1', s_2 \rangle$.
  \item If $\alpha \in Act_2 \setminus Act_1$ and $s_2 \stackrel{\alpha}{\to} s_2'$, then
    $\langle s_1, s_2 \rangle \stackrel{\alpha}{\to} \langle s_1, s_2' \rangle$.
  \end{itemize}
\end{definition}

\begin{figure}[!t]
  \centering
  \includegraphics[width=0.27\textwidth]{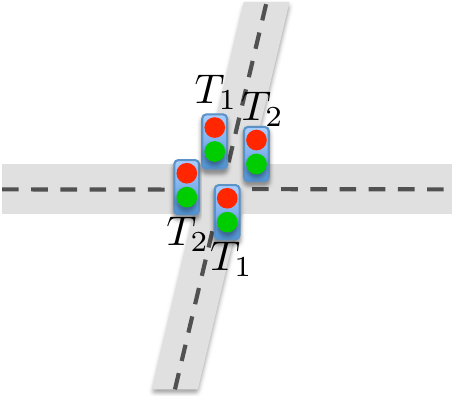}
  \caption{A system of traffic lights considered in Example \ref{ex:lights_model}.}
  \label{fig:lights}
\end{figure}

\begin{figure}[!t]
  \centering
  \input{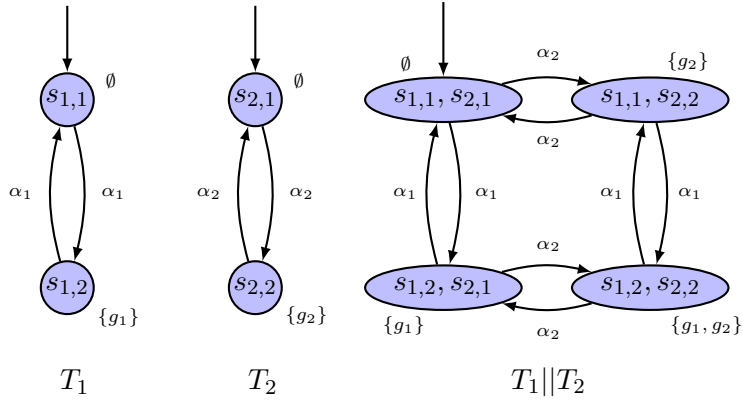}
  \caption{The transition systems representing the models of traffic lights in Example \ref{ex:lights_model}.}
  \label{fig:lights_model}
\end{figure}

\begin{example}
  \label{ex:lights_model}
  Consider the system composed of 2 sets of traffic lights (\figurename~\ref{fig:lights}).
  For $i \in \{1, 2\}$, we let $g_i$ denote an atomic proposition stating that light $T_i$ is green.
  Then, $$T_1 = (S_1, Act_1, \to_1, I_1, AP_1, L_1)$$ is a finite transition system where
  $S_1 = \{s_{1,1}, s_{1,2}\}$,
  $Act_1 = \{\alpha_1\}$,
  $\to_1 = \{(s_{1,1}, \alpha_1, s_{1,2}), (s_{1,2}, \alpha_1, s_{1,1})\}$,
  $I_1 = \{s_{1,1}\}$,
  $AP_1 = \{g_1\}$ and
  $L : S_1 \to 2^{AP_1}$ is defined by $L(s_{1,1}) = \emptyset$ and $L(s_{1,2}) = \{g_1\}$.
  Also, $$T_2 = (S_2, Act_2, \to_2, I_2, AP_2, L_2)$$ is a finite transition system where
  $S_2 = \{s_{2,1}, s_{2,2}\}$,
  $Act_2 = \{\alpha_2\}$,
  $\to_2 = \{(s_{2,1}, \alpha_2, s_{2,2}), (s_{2,2}, \alpha_2, s_{2,1})\}$,
  $I_2 = \{s_{2,1}\}$,
  $AP_2 = \{g_2\}$ and
  $L : S_2 \to 2^{AP_2}$ is defined by $L(s_{2,1}) = \emptyset$ and $L(s_{1,2}) = \{g_2\}$.
  \figurename~\ref{fig:lights_model} shows the graphical representation of $T_1$ and $T_2$ and their composition
  $T_1 || T_2$.
  Note that $T_1 || T_2$, is action-deterministic because at every state, the actions uniquely determines the next state.
\end{example}

In settings where two players, one representing the controllable system and one representing the non-controllable environment, determine the properties characterized by the atomic propositions independently, one may resort to an alternative definition of transition system.
An \emph{input-output transition system}, different from the transition system in Definition \ref{def:transition_system}, treats the \emph{input} atomic propositions controlled by the environment and the \emph{output} atomic propositions controlled by the system separately. The overall set of atomic propositions is $\props = \inpv \cup \outv$, where $\inpv$ and $\outv$ are disjoint sets, denoting input and output atomic propositions, respectively.

\begin{definition}
  \label{def:io_transition_system}
  An \emph{input-output transition system} over a set of input propositions $\inpv$ and a set of output propositions $\outv$ is a tuple $\trans = (S,s_0,\tau)$, where $S$ is a set of states, $s_0$ is the initial state, and the transition function $\tau : S \times \ialphabet \to S \times \oalphabet$ maps a state $s$ and a valuation $\inpval \in \ialphabet$ of the input propositions to a successor state $s'$ and a valuation $\outval \in \oalphabet$ of the output propositions. For any letter $\sigma$, the projection to input propositions is $\inpval \defeq \sigma \cap \inpv$ and to output propositions is $\outval \defeq \sigma \cap \outv$.
\end{definition}

An \emph{execution} of $\trans$ is an infinite sequence $s_0, (\inpval_0 \cup \outval_0), s_1, (\inpval_1 \cup \outval_1), s_2\ldots$ such that $s_0$ is the initial state, and $(s_{i+1},\outval_i) = \tau(s_i,\inpval_i)$ for every $i \geq 0$. The corresponding sequence $(\inpval_0 \cup \outval_0),(\inpval_1 \cup \outval_1),\ldots \in \alphabet^\omega$ is considered a trace. 

The input-output transition system is applicable in modeling of reactive systems, as discussed in Section \ref{sec:max}.

\section{Formal Specification Languages}
\label{sec:spec}

A first step toward providing guarantees on the behavior of the design artifacts is formally expressing what operational requirements and safety constraints the system ought to satisfy. These specifications shall be in a language that is (i) mathematically based to allow automated tool development, (ii) rich enough to capture diverse set of properties, and (iii) relatively natural to allow the designers to express their intent at a high level, potentially with domain-specific terms. One such family of possible specification languages is temporal logic which is used for specifying properties of infinite sequences of states. For example, for an autonomous system, snapshots of the relevant variables needed to model the behavior of the system including its own position, availability of its actuation and motion capabilities, quality of its sensors, position and status of the other systems can all be modeled with temporal logic.

\subsubsection{Linear Temporal Logic}

The broad family of temporal-logic-based specifications provides a basis for expressing specifications for autonomous systems in a formal language. Temporal logics is a branch of logic that implicitly incorporates temporal aspects and
can be used to reason about a time line~\citep{BK08,allen90,huth04,zohar92}.
Its use as a specification language was introduced by \citet{Pnueli77}.
Since then, temporal logic has been demonstrated to be an appropriate
specification formalism for reasoning about various kinds of systems,
especially those of concurrent programs.
It has been used to formally specify and verify behavioral properties
in various applications, including concurrent systems, reactive systems, discrete event systems,
robotics and aerospace~\citep{Clarke86AutomaticVerification,Pnueli86Applications,Galton87TemporalLogics,Lin93,Holzmann94Theory,Bouma94,Seow94,Gabbay95Handbook,Jagadeesan95Formal,Cerrito98,Jiang01failurediagnosis,Schneider98Validating,Havelund01Formal,Holzmann14Mars}.

We consider a version of temporal logic, namely \gls{ltl}.
An \gls{ltl} formula is built up from a set of atomic propositions
and two kinds of operators: logical connectives and temporal modal operators.
The logic connectives are those used in propositional logic:
{\em negation\/} ($\neg$), {\em disjunction\/} ($\oor$),
{\em conjunction\/} ($\aand$) and
{\em material implication\/} ($\implies$).
The temporal modal operators include {\em next\/} ($\nextop$), {\em always\/} ($\always$),
{\em eventually\/} ($\eventually$) and {\em until\/} ($\until$).
Specifically, an \gls{ltl} formula over a set $AP$ of atomic propositions is inductively defined as follows.
\begin{enumerate}[(1)]
\item $\true$ is an \gls{ltl} formula,
\item any atomic proposition $p \in AP$ is an \gls{ltl} formula and
\item given \gls{ltl} formulas $\varphi$, $\varphi_1$ and $\varphi_2$,
  $\neg \varphi$, $\varphi_1 \oor \varphi_2$, $\nextop \varphi$ and
  $\varphi_1 \until \varphi_2$ are also \gls{ltl} formulas.
\end{enumerate}
Additional operators can be derived from the logical connectives $\oor$ and $\neg$ and
the temporal modal operator $\until$.
For example,
$\varphi_1 \aand \varphi_2 = \neg(\neg\varphi_1 \oor \neg \varphi_2)$,
$\varphi_1 \implies \varphi_2 = \neg \varphi_1 \oor \varphi_2$,
$\eventually\varphi = \true \ \until \ \varphi$ and
$\always\varphi = \neg\eventually\neg\varphi$.

\gls{ltl} formulas are interpreted on infinite strings $\sigma = \sigma_0 \sigma_1 \sigma_2 \cdots$ where $\sigma_i \in 2^{AP}$ for all $i \geq 0$.
Such infinite strings are referred to as {\em words\/}.
The satisfaction relation is denoted by $\models$, i.e.,
for a word $\sigma$ and an \gls{ltl} formula $\varphi$, we write $\sigma \models \varphi$ if and only if $\sigma$ satisfies $\varphi$.
The satisfaction relation is defined inductively as follows.
\begin{itemize}
\item $\sigma \models \true$,
\item for an atomic proposition $p \in AP$, $\sigma \models p$ if and only if $p \in \sigma_0$,
\item $\sigma \models \neg \varphi$ if and only if $\sigma \not\models \varphi$,
\item $\sigma \models \varphi_1 \aand \varphi_2$ if and only if $\sigma \models \varphi_1$ and $\sigma \models \varphi_2$,
\item $\sigma \models \nextop\varphi$ if and only if $\sigma_1 \sigma_2\cdots \models \varphi$ and
\item $\sigma \models \varphi_1 \until \varphi_2$ if and only if there exists $j \geq 0$ such that
  $\sigma_j \sigma_{j+1} \cdots \models \varphi_2$ and for all $i$ such all $0 \leq i < j$, $\sigma_i \sigma_{i+1}\cdots \models \varphi_1$.
\end{itemize}

Let $\varphi$ be an \gls{ltl} formula over $AP$. The linear-time property induced by $\varphi$ is defined as
$Words(\varphi) = \{\sigma \in (2^{AP})^\omega \colon \sigma \models \varphi\}$.
Given a transition system $TS$, its infinite path fragment $\pi$ and an \gls{ltl} formula $\varphi$ over $AP$,
we say that $\pi$ satisfies $\varphi$, denoted $\pi \models \varphi$, if $trace(\pi) \models \varphi$.
Finally,
we say that $TS$ satisfies $\varphi$, denoted $TS \models \varphi$, if $Trace(TS) \subseteq Words(\varphi)$.

The decision problem of determining whether there exists a transition system that satisfies an \gls{ltl} formula is called the \emph{realizability problem for \gls{ltl}}. If an \gls{ltl} formula $\varphi$ is realizable, the goal of \emph{\gls{ltl} synthesis problem} is to construct a transition system $TS$ such that $TS \models \varphi$.

\subsubsection{Examples of \gls{ltl} formulas}

Given propositional formulas $p$ and $q$, %
important and widely used properties
can be defined in terms of their corresponding \gls{ltl} formulas as follows.

  \begin{description}
  \item [Safety (invariance)] A safety formula is of the form $\always p$,
    which asserts that the property $p$ remains
    invariantly true throughout an execution.
    Typically, a safety property ensures that nothing bad happens.
    A classic example of safety property frequently used in the robot
    motion planning domain is obstacle avoidance.

  \item [Guarantee (reachability)] A guarantee formula is of the form
    $\eventually p$,
    which guarantees that the property $p$ becomes
    true at least once in an execution.
    Reaching a goal state is an example of a guarantee property.

  \item [Obligation] An obligation formula is a disjunction of safety and
    guarantee formulas, $\always p \oor \eventually q$.
    It can be shown that any safety and progress property can be
    expressed using an obligation formula.
    (By letting $q \equiv \false$, we obtain a safety formula and by
    letting $p \equiv \false$, we obtain a guarantee formula.)

  \item [Progress (recurrence)] A progress formula is of the form
    $\always \eventually p$,
    which essentially states that the property $p$
    holds infinitely often in an execution.
    As the name suggests, a progress property typically ensures that the
    system makes progress throughout an execution.

  \item [Response] A response formula is of the form
    $\always (p \implies \eventually q)$,
    which states that following any point
    in an execution where the property $p$ is true, there exists a point
    where the property $q$ is true.
    A response property can be used, for example, to describe how the system should
    react to changes in the operating conditions.

  \item [Stability (persistence)] A stability formula is of the form
    $\eventually \always p$,
    which asserts that there is a point in an
    execution where the property $p$ becomes invariantly true for the
    remainder of the execution.
    This definition corresponds to the definition of stability in the
    controls domain since it ensures that eventually, the
    system converges to a desired operating point and remains there for
    the remainder of the execution.
\end{description}

\begin{example}
  \label{ex:ltl:lights}
  Consider the system of traffic lights (Example~\ref{ex:lights_model}).
  Its desired properties might include the following.
  \begin{itemize}
  \item ``At least one of the lights is always on'' is a safety property and can be expressed in \gls{ltl} as
    $\always (g_1 \oor g_2)$.
  \item ``Two lights are never green at the same time'' is also a safety property and can be expressed in \gls{ltl} as
    $\always (\neg g_1 \oor \neg g_2)$.
  \item ``$T_1$ will turn green infinitely often'' is known as a ``progress'' property and can be expressed in \gls{ltl} as
    $\always \eventually g_1$.
  \end{itemize}
\end{example}

\begin{example}
  \label{chap:introduction:ex:ltl}
  Consider a robot motion planning problem where the robot is moving in
  an environment that is partitioned into six regions (\figurename~\ref{chap:introduction:F:LTLexample}).

  \begin{figure}
    \centering
    \input{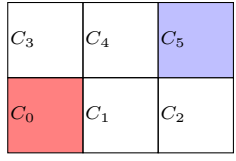}
    \caption{The robot environment of Example~\ref{chap:introduction:ex:ltl}.}
    \label{chap:introduction:F:LTLexample}
  \end{figure}

  Let $s$ represent the position of the robot and $C_0, \cdots, C_5$
  represent the polygonal regions in the robot environment.
  Suppose the robot receives an externally triggered PARK signal.
  Consider the following desired behaviors.
  \begin{enumerate}[(a)]
  \item Visit region $C_5$ infinitely often.
  \item Eventually go to region $C_0$ when a PARK signal is received.
  \end{enumerate}
  Assuming that infinitely often, a PARK signal is not received,
  the desired properties of the system can be expressed in \gls{ltl} as
  \begin{equation}
    \always\eventually(\neg \mathit{park}) \implies
    \big(\always\eventually(s \in C_5) \aand
    \always(\mathit{park} \implies \eventually(s \in C_0)) \big).
    \label{eq:ex:chap:introduction:ex:ltl}
  \end{equation}
  Here, $\mathit{park}$ is a Boolean variable that indicates whether a PARK signal is received.

  Let
  $S_{1} = \{s_{0}, s_{1}, \cdots, s_{5}\}$ be a finite set of the (discretized) positions of the
  robot such that $s_{i} \in C_{i}$ for all $i$,
  Let $S_{2} = \{p, p'\}$, where $p$ and $p'$ indicates that the PARK signal is received and is not
  received, respectively,
  $Act = \{l, r, u, d\}$, where $l, r, u, d$ represent the left, right, up, and down movement of
  the robot, respectively,
  $I = \{s_{0}\}$, and
  $AP = \{C_{1}, C_{2}, \cdots, C_{5}, park\}$.
  We can model the complete system as a finite transition system $TS = (S, Act, \to, I, AP, L)$
  where $S = S_{1} \times S_{2}$,
  $((s_{i}, \tilde{p}), \alpha, (s_{j}, \hat{p})) \in \to$ if and only if
  $C_{i}$ and $C_{j}$ are adjacent, $\alpha$ corresponds to the location of $C_{j}$ relative to
  $C_{i}$, and $\tilde{p}, \hat{p} \in \{p, p'\}$.
  The labeling function $L \colon S \to 2^{AP}$ is defined by
  $L(s_{i}, p) = \{C_{i}, park\}$ and $L(s_{i}, p') = \{C_{i}\}$ for all $i$.

  A path
  $\pi = \big((s_{0}, p) , (s_{1}, p), (s_{2}, p'), (s_{5}, p), (s_{4}, p), (s_{3},
  p')\big)^{\omega}$
  of $TS$ satisfies the \gls{ltl} formula \ref{eq:ex:chap:introduction:ex:ltl} while
  $\pi' = \big((s_{0}, p) , (s_{1}, p')\big)^{\omega}$
  does not satisfy this formula.
\end{example}

\begin{figure}
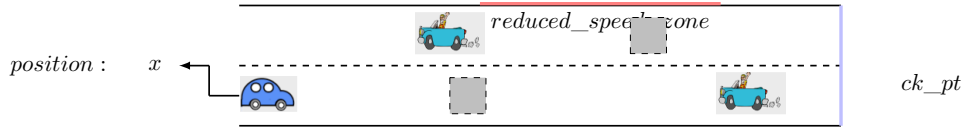

  \centering

  \includestandalone[scale=.8]{figures/figure2_6}
  \caption{A simplified autonomous driving problem considered in Example \ref{ex:autonomous}.}
  \label{fig:autonomous}
\end{figure}

\begin{example}
  \label{ex:autonomous}
  An autonomous vehicle competing in the DARPA Urban Challenge is required to follow traffic rules
  as well as completing a task specified by a sequence of checkpoints that the vehicle has to cross
  (\figurename~\ref{fig:autonomous}).
  We define the state of the autonomous vehicle as $(x, \theta, v)$ where $x \in \reals^{2}$
  represents the center of its front bumper, $\theta \in \reals$ represents its heading, and $v \in
  \reals$ represents its speed.
  Additionally, let $\mathsf{Obs} \subset \reals^{2}$ represent the union of the footprints of all the
  vehicles and obstacles in the environment.
  The state of the complete system (autonomous vehicle and the environment) is then given by $(x,
  \theta, v, \mathsf{Obs})$.
  Examples of some desired properties and \gls{ltl} formulas expressing those properties are given
  below.

  \begin{itemize}
  \item [\textbf{Traffic rule 1.}] No collision:
    $$\always\big(\mathsf{FP}(x, \theta) \intersect \mathsf{Obs} = \emptyset\big),$$
    where for any $x, \theta \in \reals$, $\mathsf{FP}(x,\theta) \subset \reals^{2}$ is the
    footprint of the autonomous vehicle when the center of its front bumper is at $x$ and its
    heading is $\theta$.
  \item [\textbf{Traffic rule 2.}] Obey speed limits: $$\always\big( (x \in \hbox{Reduced\_Speed\_Zone}) \implies (v \leq v_{reduced}) \big),$$
    where $v_{reduced}$ is a pre-specified parameter for the maximum speed in Reduced\_Speed\_Zone.
  \item [\textbf{Goal.}]Eventually visit the checkpoint: $\eventually(x = \hbox{ck\_pt})$
    where $\hbox{ck\_pt}$ denote the position of the checkpoint.
  \end{itemize}

  The following are considered to be atomic propositions in this example as
  they are evaluated to be either true or false, given the state $(x, \theta, v, \mathsf{Obs})$ of
  the system:
  \begin{align*}
    & \mathsf{FP}(x, \theta) \intersect \mathsf{Obs} = \emptyset,\\
    &x \in \hbox{Reduced\_Speed\_Zone},\\
    &v \leq v_{reduced}, \text{ and} \\
    &x = \hbox{ck\_pt}.
  \end{align*}
\end{example}

\begin{remark}
 \gls{ltl} offers an extension to the properties, e.g., safety (in the form of constraints on the system state) and reachability (in the form of convergence to a desired state) that have typically been used in the controls literature.

\end{remark}

\subsubsection{Automata Representation of \gls{ltl} Formulas}

There are several ways for checking whether the behaviors (i.e., traces) that can be generated by a transition system satisfy a given temporal logic specification. One notion that will appear frequently in subsequent sections is that of an {\it automaton that witnesses the satisfaction of a temporal logic formula}.

The language accepted by an \gls{ltl} formula can equivalently be represented by a nondeterministic (or universal) B\"uchi (or co-B\"uchi) automaton \citep{buchi1990decision}.
A \emph{B\"uchi automaton} over a finite alphabet $\alphabet$ is a tuple $\autA = (Q,Q_0,\delta,F)$, where $Q$ is a finite set of states, $Q_0$ is the set of initial states, $\delta \subseteq Q \times \alphabet \times Q$ is the transition relation, and $F \subseteq Q$ is a subset of states. A run of $\autA$ on an infinite word $w=\sigma_0\sigma_1\cdots \in \alphabet^\omega$ is an infinite sequence $q_0,q_1,\ldots$ of states, where $q_0 \in Q_0$ is an initial state and for every $i \geq 0$ it holds that $(q_i,\sigma_i,q_{i+1}) \in \delta$.

A run of a B\"uchi automaton is accepting if it contains infinitely many occurrences of states in $F$. A \emph{co-B\"uchi automaton} $\autA = (Q,Q_0,\delta,F)$ differs from a B\"uchi automaton in the accepting condition: a run of a co-B\"uchi automaton is accepting if it contains only \emph{finitely many} occurrences of states in $F$. For a B\"uchi automaton the states in $F$ are called \emph{accepting states}, while for a co-B\"uchi automaton they are called \emph{rejecting states}.
A \emph{nondeterministic} automaton $\autA$ accepts a word $w \in \alphabet^\omega$ if \emph{some} run of $\autA$ on $w$ is accepting.
A \emph{universal} automaton $\autA$ accepts a word $w \in \alphabet^\omega$ if \emph{every} run of $\autA$ on $w$ is accepting.

Figure~\ref{fig:aut-ex1} and Figure~\ref{fig:aut-ex2} represent examples of B\"uchi automata for the \gls{ltl} formulas in Example \ref{chap:introduction:F:LTLexample} and Example \ref{ex:autonomous}, respectively.

\begin{figure}[!t]
  \centering
  \input{./figures/figure2_7.tikz}
  \caption{B\"uchi automaton for the \gls{ltl} formula $\always\eventually(\neg \mathit{park}) \implies \big(\always\eventually(s \in C_5) \aand \always(\mathit{park} \implies \eventually(s \in C_0)) \big)$ in Example \ref{chap:introduction:F:LTLexample}.}
  \label{fig:aut-ex1}
\end{figure}

\begin{figure}[!t]
  \centering
  \input{./figures/figure2_8.tikz}
  \caption{B\"uchi automaton for the \gls{ltl} formula $\always\big( \mathsf{FP}(x, \theta)
    \intersect \mathsf{Obs} = \emptyset \big) \aand \always\big( (x \in \hbox{Reduced\_Speed\_Zone})
    \implies (v \leq v_{reduced}) \big) \aand \eventually(x = \hbox{ck\_pt})$ in Example
    \ref{ex:autonomous}. Here, $ck\_pt$, $collision$, $speed\_zone$, and $v_{reduced}$ represent
    $x = \hbox{ck\_pt}$, $\mathsf{FP}(x, \theta) \intersect \mathsf{Obs} \not= \emptyset$,
    $x \in \hbox{Reduced\_Speed\_Zone}$, and $v \leq v_{reduced}$, respectively.
  }
  \label{fig:aut-ex2}
\end{figure}

While the class of nondeterministic B\"uchi automata are sufficient to represent any \gls{ltl} formula, there are other classes of automata that are widely used. The major variations in the definition of these different automata relate to their input, states, transition function, and acceptance condition. Other automata types include deterministic and nondeterministic finite-state automata \citep{mcculloch1943logical}, deterministic B\"uchi automata \citep{buchi1990decision}, Rabin automata \citep{rabin1959finite}, and parity automata \citep{gradel2002automata}. For instance, deterministic and nondeterministic finite-state automata are expressive enough to represent regular languages. Therefore, if an application requires only finite-horizon specifications, its \gls{ltl} formulas can be represented by finite-state automata that are simpler than B\"uchi automata.

\chapter{Verification and Model Checking}
\glsresetall

This section reviews existing approaches to system verification.
These methods perform automated analysis of the correctness of the system's abstract mathematical model relative to the system's requirements.
As a result, these approaches provide a formal guarantee that the desired system properties hold over all of its possible executions, provided that the actual execution of the system respects its model.

\section{Model Checking}
\label{sec:model-checking}

Model checking is a well-established technique for system verification based on exhaustive
exploration of the state space.
The key requirement of this technique is that the description of the system and its requirements be
formulated in some precise mathematical language.
From the description of the system, all of its possible behaviors can be derived.
In addition, all the valid and invalid behaviors can be obtained from the system requirements.
A model checker then checks whether an intersection of all the possible behaviors of the system
and all the invalid behaviors is empty.
It terminates with a yes/no answer and provides an error trace in case of a negative result.
This technique is very attractive because it is automatic, fast and requires no human interaction.
However, as it is based on exhaustive exploration of the state space,
model checking is limited to finite state systems.
It also faces a combinatorial blow up of the state space, commonly known as the state explosion
problem \citep{holzmannSPIN,BK08}.

\begin{figure}
  \centering
  \includegraphics[width=0.5\textwidth]{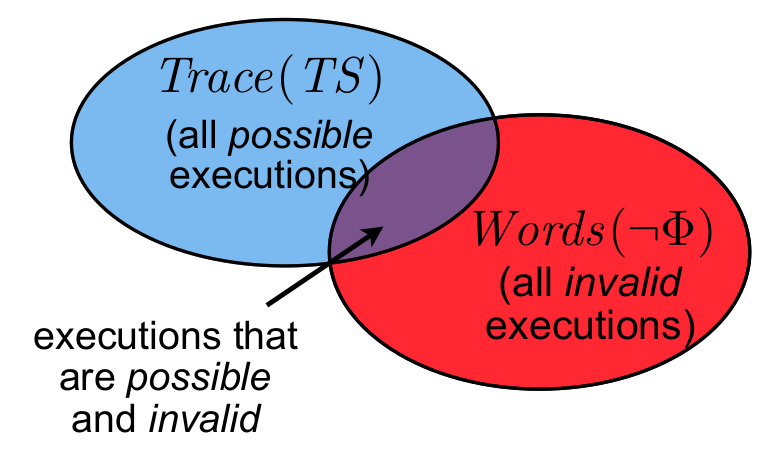}
  \caption{The basic idea behind model checking.}
  \label{fig:model_checking}
\end{figure}

A model checking problem is to find a path in the finite transition system $TS$ that violates the
specification $\varphi$.
All the possible behaviors of $TS$ can be captured by its trace, $Trace(TS)$ whereas all the invalid behaviors of the system can be captured by the linear-time property $Words(\neg \varphi)$ (Chapter~\ref{chap:models_and_specifications}).
The satisfiability problem -- determining whether $TS$ satisfies $\varphi$ -- can then be solved by claiming that $Trace(TS) \cap Words(\neg \varphi) =
\emptyset$ as shown in Figure \ref{fig:model_checking}.
In case of negative result, a word in $Trace(TS) \cap Words(\neg \varphi)$ is a counterexample.
A positive result means $Trace (TS) \cap Words(\neg \varphi) = \emptyset$, i.e., a path $\pi$ of
$TS$ that violates $\varphi$ does not exist; hence, we can conclude that $\varphi$ is satisfied.

For simplicity of the presentation, we assume that $TS$ only has one valid initial state.
For $TS$ with multiple valid initial states, the procedure described below can be applied to each initial state separately.
To check whether $Trace(TS) \cap Words(\neg \varphi) = \emptyset$, we first compute a
non-deterministic B\"{u}chi automaton $\mathcal{A} = (Q, \delta, Q_0, F)$ over $2^{AP}$ that accepts
all and only words over $AP$ that satisfy $\neg \varphi$.
The product $TS_p = TS \otimes \mathcal{A}$ can then be constructed based on the following definition.

\begin{definition}
  \label{def:product}
  Let $TS = (S, Act, \to, I, AP, L)$ be a transition system
  and $\mathcal{A} = (Q, \delta, Q_0, F)$ be a non-deterministic B\"{u}chi automaton over $2^{AP}$.
  The product of $TS$ and $\mathcal{A} $ is the transition system
  $TS_p = TS \otimes \mathcal{A}$ defined by $TS_p = (S \times Q, Act, \to_p, I_p, Q, L_p)$ where
  \begin{enumerate}[(i)]
  \item for any $s, t \in S$, $\alpha \in Act$ and $p, q \in Q$,
    $\langle s, p \rangle \stackrel{\alpha}{\to_p} \langle t, q \rangle$ if and only if
    $s \stackrel{\alpha}{\to} t$ and $p \stackrel{L(t)}{\to} q$,
  \item $I_p = \{ \langle s_0, q_0 \rangle : s_0 \in I \hbox{ and } \exists q \in Q_0 \hbox{ such that } q \stackrel{L(s_0)}{\to} q_0\}$ and
  \item $L_p : S \times Q \to 2^Q$ is given by $L_p(\langle s, q \rangle ) = \{q\}$.
  \end{enumerate}
\end{definition}

\begin{figure}
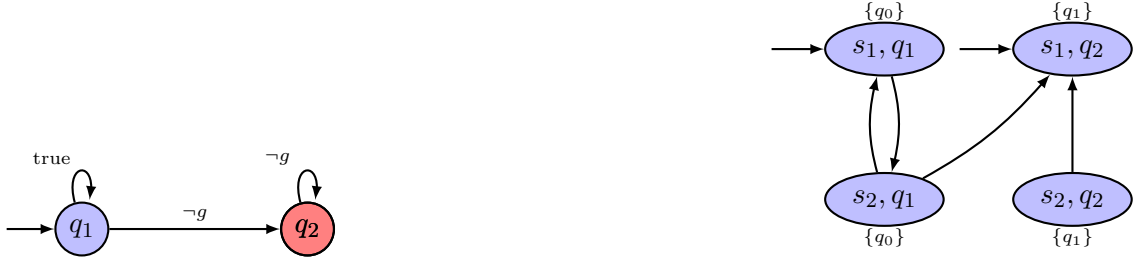

  \centering
    \input{./figures/03-verify/nba_example.tikz}
  \hfill
 \input{./figures/03-verify/product_automaton.tikz}
  \caption{(Left) A non-deterministic B\"{u}chi automaton $\mathcal{A}$ that accepts all and only words
    over $AP = \{g\}$ that satisfy $\neg \always\eventually g$.
    (Right) The product $T \otimes \mathcal{A}$ for Example~\ref{ex:light_product}}.
  \label{fig:light_model_checking}
\end{figure}

\begin{example}
  \label{ex:light_product}
  Consider the traffic light $T$ in Example \ref{ex:light_model}.
  Suppose we want to ensure that the light is green infinitely often.
  This requirement can be expressed in \acrshort{ltl} as $\varphi = \always \eventually g$.
  A non-deterministic B\"{u}chi automaton $\mathcal{A}$ that accepts all and only words over
  $AP = \{g\}$ that satisfy $\neg\varphi$ as well as the product $T \otimes \mathcal{A}$
  are shown in Figure \ref{fig:light_model_checking}.
\end{example}

Consider a path $\pi_p = \langle s_0, q_0 \rangle \langle s_1, q_1 \rangle \ldots$ on $TS_p$.
We say that $\pi_p$ is \emph{accepting} if and only if
there exist infinitely many $j \geq 0$ such that $q_j \in F$.
Stepping through Definition \ref{def:product} shows that given a path $\pi_p$ on $TS_p$,
the corresponding path $\pi = s_0 s_1 \ldots$ on $TS$ generates a word $L(s_0) L(s_1) \ldots$
that satisfies $\neg \varphi$ if and only if $\pi_p$ is accepting.
Hence, an accepting path of $TS_p$ uniquely corresponds to a path of $TS$ that violates $\varphi$.
As a result, model checking can be reduced to a graph search problem to find a state
$\langle s, q \rangle$ in $TS_p$ satisfying the following conditions:
\begin{enumerate}
\item[(MC1)] $L_p(\langle s, q \rangle ) \in F$.
\item[(MC2)] $\langle s, q \rangle$ is reachable, i.e.,
  there exists a finite path fragment $\pi_p^p$ from some $\langle s_0, q_0 \rangle \in I_p$ to $\langle s, q \rangle$ in $TS_p$.
\item[(MC3)] $\langle s, q \rangle$ is on a direct cycle, i.e.,
  there exists a finite path fragment $\pi_p^c$ from some $\langle s', q' \rangle \in Post(\langle s, q \rangle)$
  to $\langle s, q \rangle$ in $TS_p$.
\end{enumerate}
If such $\langle s, q \rangle$ does not exists, we can conclude that $Trace(TS) \cap Words(\neg
\varphi) = \emptyset$ and therefore $TS \models \varphi$.
Otherwise, an accepting path $\pi_p$ on $TS_p$ can be simply defined by
$\pi_p = \pi_p^p (\pi_p^c)^\omega$.
A path $\pi = s_0 s_1 \ldots$ on $TS$ corresponding to $\pi_p$ is a counterexample of a run on $TS$ that
violates $\varphi$.

\begin{example}
  \label{ex:light_model_checking}
  Let us revisit Example \ref{ex:light_product}.
  Recall that $F = q_{2}$ and the set of states of $TS_{p}$ is given by $\{s_{1}, s_{2}\} \times \{q_{1},
  q_{2}\}$ as shown in Figure \ref{fig:light_model_checking}.
  First, consider the states $(s_{1}, q_{1})$ and $(s_{2}, q_{1})$.
  Here, $L((s_{1}, q_{1})) = L((s_{2}, q_{1})) = \{q_{1}\} \not\subset F$; thus, condition (MC1) fails.
  Next, the state $(s_{1}, q_{2})$ is not on a direct cycle; thus, condition (MC3) fails.
  Finally, the state $(s_{2}, q_{2})$ is not reachable; thus, condition (MC2) fails.
  We thus conclude that there is no accepting path $\pi_{p}$ on $TS_{p}$ and so the transition system $TS$ modeling the traffic light $T$ satisfies the specification $\varphi = \always \eventually g$.
\end{example}

\section{Computational Complexity}
The number of states of $\mathcal{A}$ is exponential in the length of $\varphi$, i.e., the number of
operators in $\varphi$.
Hence, the number of states of the product transition system $TS_p$ is $O(|S|) 2^{O(|\varphi|)}$
where $|S|$ is the number of states in $TS$ and $|\varphi|$ is the length of $\varphi$.
A nested depth-first search algorithm \citep{BK08} can be used to detect accepting cycles efficiently,
with the worst-case time complexity that is linear in the number of states and transitions of $TS_p$.

Several reduction techniques have been proposed to allow model checkers to handle large state spaces.
One example is state compression, including lossy compression (e.g., hash-compact and bitstate hashing),
lossless compression, and alternate state representation methods, help reduce memory requirements by
reducing the amount of memory required to store each state.
In contrast, partial order reduction strategies avoid computing equivalent paths, which
helps reduce the number of states that needs to be explored.
Symbolic model checkers such as SMV and NuSMV use compressed representation of the state space known
as \gls{bdd}.
On the other hand, Spin avoids constructing the complete state space by employing on-the-fly
construction of the finite transition system, the non-deterministic B\"{u}chi automaton, and the product
automaton.
We refer the reader to \citep{holzmannSPIN} for more details on these reduction techniques.

\section{Tools for Model Checking}
There are various model checkers for different specification languages.
TLC \citep{yu99model} is  a model checker for specifications written in
\acrshort{tla}+, which is a specification language based on
\acrfull{tla} \citep{abadi94oldfashioned,lamport83,lamport94temporal}.
\acrshort{tla} introduces new kinds of temporal assertions \citep{lamport94temporal}
to traditional \gls{ltl} to make it practical to describe a system by a single formula
and to make the specifications simpler and easier to understand.

The Spin model checker deals with specifications written in \acrfull{promela}~\citep{holzmannSPIN}.
This language was influenced by Dijkstra, Hoare's CSP language and C.
It emphasizes the modeling of process synchronization and coordination,
not computation and is not meant to be analyzed by a human.
Spin can be run in two modes---simulation and verification.
The simulation mode performs random or iterative simulations of the
modeled system's execution while the verification mode generates a C
program that performs a fast exhaustive verification of the system
state space.
Spin is mainly used for checking for deadlocks, livelocks,
unspecified receptions, unexecutable code, correctness of system
invariants and non-progress execution cycles. It also supports the
verification of linear time temporal constraints.
Spin has been used in many applications, especially in proving correctness
of safety-critical software~\citep{Havelund01Formal,Gluck02UsingSpin}.
Other popular model checkers include Symbolic Model Verifier (SMV)~\citep{McMillanSMV} and
its successor NuSMV~\citep{NuSMV2}.

\section{Model Checking for Autonomous Vehicles} \label{sec:model_checking_for_Alice}
\begin{figure}
  \centering

  \input{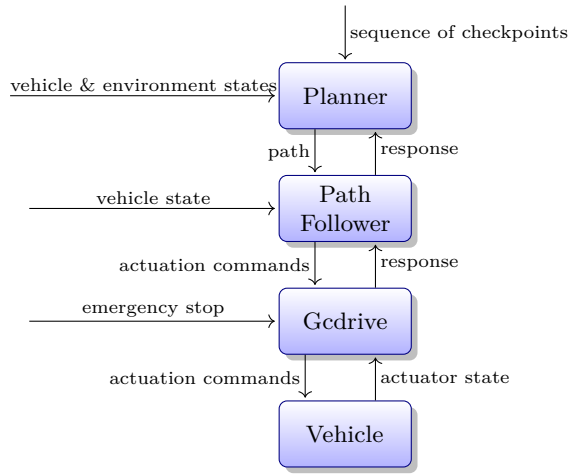}

  \caption{The embedded control component of Alice.}
  \label{chap:introduction:F:Alice-planner-controller}
\end{figure}

This section illustrates the applications of model checking (Section~\ref{sec:model-checking})
to the embedded control component of Alice,
an autonomous vehicle built at the California Institute of Technology for the DARPA Urban Challenge.
Alice was equipped with 25 CPUs and utilized a networked control system architecture
to provide high performance and modular design.
The embedded control component of Alice is shown in
\figurename~\ref{chap:introduction:F:Alice-planner-controller}.
We refer the reader to \citep{Burdict07Sensing,duToit08Situational,Wongpiromsarn08Distributed} for
more details on this hierarchical control architecture.

The case studies presented in this section focus on the low-level
module, namely Gcdrive, which is the overall driving software for Alice.
It takes independent commands from Path Follower
and DARPA and sends appropriate commands to the actuators.
Commands from Path Follower include control signals to
throttle, brake and transmission.
Commands from DARPA include estop pause, estop run and estop disable.
An estop pause command should cause the vehicle to be brought quickly and safely to a rolling stop
and reject commands to any actuator.
An estop run command resumes the operation of the vehicle.
An estop disable command is used to stop the vehicle and put it in the disable mode.
A vehicle that is in the disable mode may not restart in response to an estop run
command.

\begin{figure}
   \centering
   \includegraphics[width=0.85\textwidth]{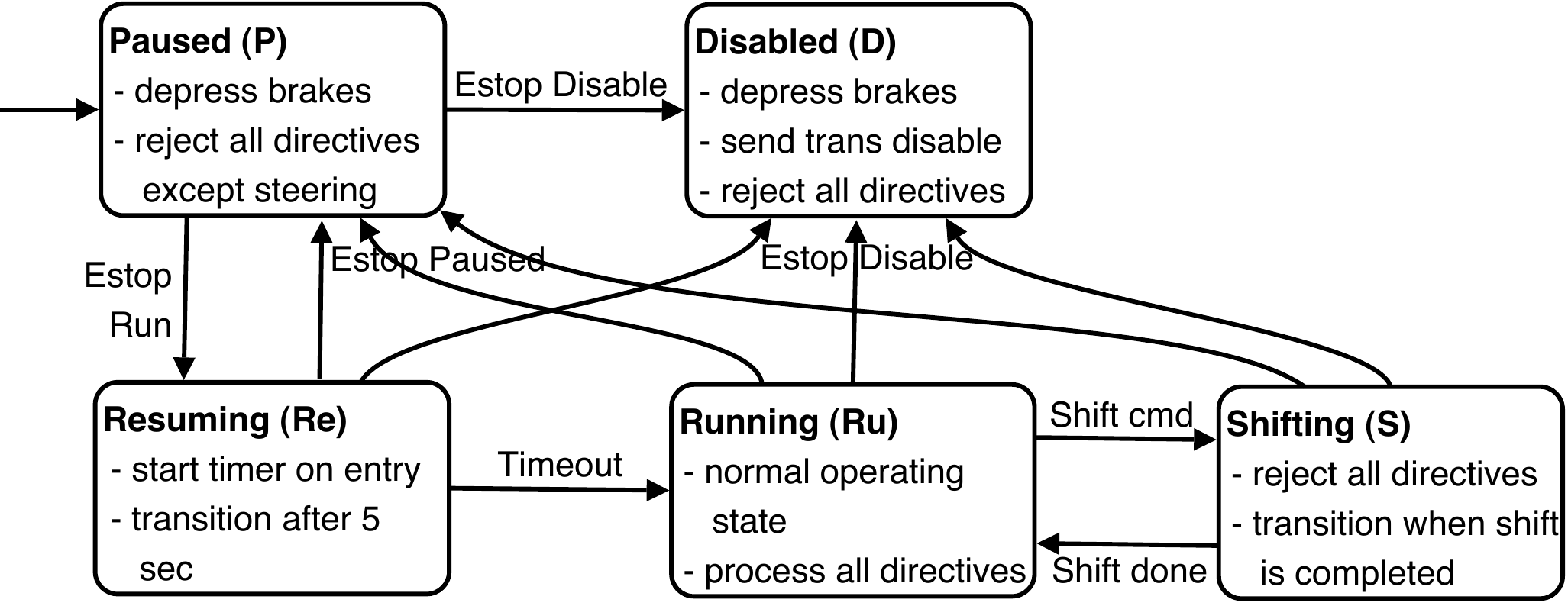}
   \caption{Finite state machine implemented in Actuation Interface.}
   \label{chap:alice:F:GcdriveFSM}
\end{figure}

\begin{figure}
  \centering
  \includegraphics[width=0.75\textwidth]{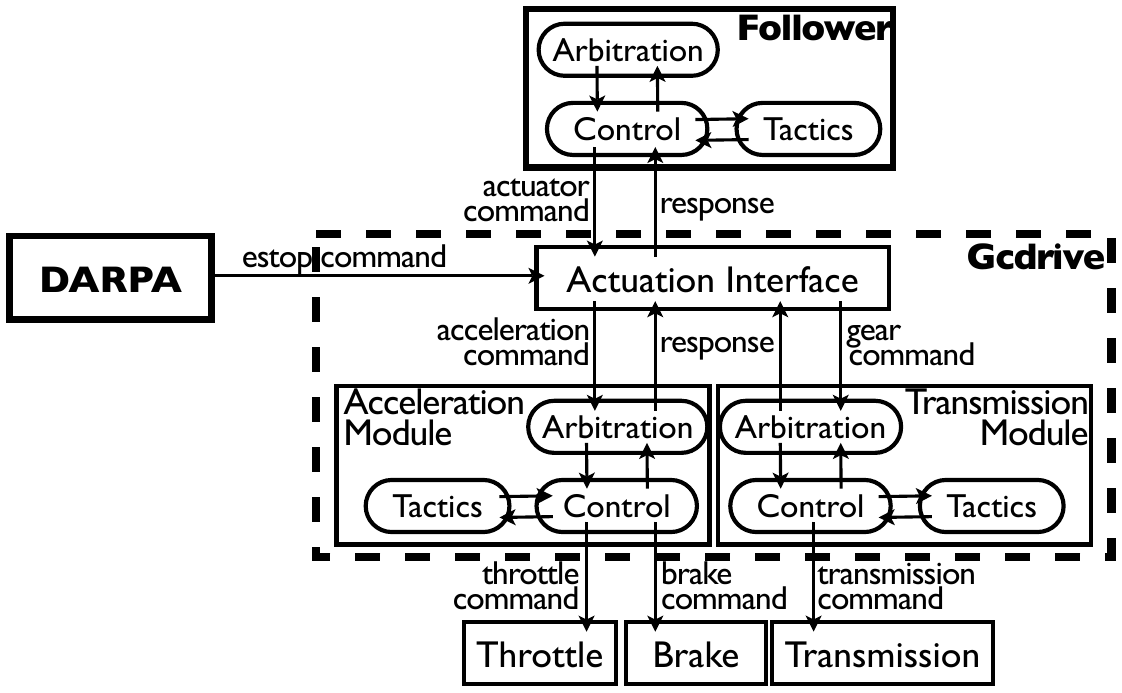}
  \caption{The components involved in the Gcdrive FSM example.}
  \label{chap:alice:F:followerGcdriveDarpa}
\end{figure}

The logic in Gcdrive to handle these concurrent commands can be described by a finite state machine
shown in \figurename~\ref{chap:alice:F:GcdriveFSM}, which is implemented in the Actuation Interface
component of Gcdrive (See \figurename~\ref{chap:alice:F:followerGcdriveDarpa}).
This example illustrates the use of model checking
in proving the correctness of the implementation of this finite state machine.
We model Follower, Gcdrive, and DARPA (see Figure~\ref{chap:alice:F:followerGcdriveDarpa})
in the Spin model checker with the following global variables.

\begin{itemize}
\item $state \in \{$DISABLED (D), PAUSED (P), RUNNING (Ru), RESUMING (Re), SHIFTING (S)$\}$
  is the state of the finite state machine as described in Figure~\ref{chap:alice:F:GcdriveFSM}.
\item $estop \in \{$DISABLE (0), PAUSE (1), RUN (2)$\}$ is the emergency stop command sent by DARPA.
\item $acc \in [-1, 1]$ and $acc\_cmd \in [-1, 1]$ are the acceleration commands sent from
  Actuation Interface to Acceleration Module and from Follower to Actuation Interface,
  respectively.
\item $gear \in \{-1, 0, 1\}$ and $gear\_cmd \in \{-1, 0, 1\}$ are the gear commands
  sent from Actuation Interface to Transmission Module and
  from Follower to Actuation Interface, respectively.
\item $timer \in \{0, 1, 2, 3, 4, 5\}$ keeps track of the time after which the latest estop run command
  is received.
\end{itemize}

The desired properties can be expressed in \acrshort{ltl} as follows.
\begin{enumerate}[(1)]
\item If DARPA sends an estop disable command, Gcdrive state will
  eventually stay at DISABLED and Acceleration Module will
  eventually command full brake forever.
  \begin{equation}
    \label{chap:alice:Eq:gcdriveExDisabled}
    \always\big((estop = 0) \implies \eventually\always(state =
    \mbox{D} \aand acc = -1)\big)
  \end{equation}
\item If DARPA sends an estop pause command while the vehicle is not disabled, eventually Gcdrive
  state will be PAUSED.
  \begin{equation}
    \label{chap:alice:Eq:gcdriveExPaused}
    \always\big((estop = 1 \aand state \not= \mbox{D}) \implies
    \eventually(state = \mbox{P})\big)
  \end{equation}
\item If DARPA sends an estop run command while the vehicle is not disabled, eventually Gcdrive state
  will be RUNNING or RESUMING or DARPA will send an estop disable or estop pause command.
  \begin{equation}
    \label{chap:alice:Eq:gcdriveExRun}
    \begin{array}{c}
      \always\big((estop = 2 \aand state \not= \mbox{D}) \implies
      \eventually(state \in \{\mbox{Ru}, \mbox{Re}\} \oor
      estop \not= 2)\big)
    \end{array}
  \end{equation}
\item If the current state is RESUMING, eventually the state will be
  RUNNING or DARPA will send an estop disable or pause command.
  \begin{equation}
    \label{chap:alice:Eq:gcdriveExResuming}
    \begin{array}{c}
      \always\big((state = \mbox{Re}) \implies
      \eventually(state = \mbox{Ru}
      \oor estop \in \{0, 1\})\big)
    \end{array}
  \end{equation}
\item The vehicle is disabled only after it receives an estop disable command.
  \begin{equation}
    \label{chap:alice:Eq:gcdriveExDisabled2}
    \big((state \not= \mbox{D}) \until (estop = 0)\big) \oor
    \always(state \not= \mbox{D})
  \end{equation}
\item Actuation Interface sends a full brake command to the
  Acceleration Module if the current state is DISABLED, PAUSED, RESUMING or SHIFTING.
  In addition, if the vehicle is disabled, then the gear is shifted to 0.
  \begin{equation}
    \label{chap:alice:Eq:gcdriveExFullBrake}
    \begin{array}{c}
      \always(state \in \{\mbox{D, P, Re, S}\}
      \implies acc = -1) \aand\\
      \always(state = \mbox{D} \implies gear = 0)
    \end{array}
  \end{equation}
\item After receiving an estop pause command, the vehicle may resume the operation 5 seconds
  after an estop run command is received.
  \begin{equation}
    \label{chap:alice:Eq:gcdriveRun5Sec}
    \always(state = \mbox{Ru} \implies timer \geq 5)
  \end{equation}
\end{enumerate}

Model checking revealed that an early implementation of Gcdrive failed to satisfy the above
properties.
In particular, the counterexample showed that the variable $state$ did not change as required when
an estop command was sent.
The counterexample suggested that we incorporate the following assumptions.
\begin{enumerate}[(a)]
\item Actuation Interface gets executed at least once each time an estop command is sent.
\item Actuation Interface reads the current estop status at the beginning of each iteration. It then performs
  a computation based on this estop status for the rest of the iteration.
\item All the estop commands are eventually received.
\end{enumerate}
We introduce a global variable $\mathit{enableEstop}$ %
to incorporate assumption (a).
Assumption (b) is enforced using atomic sequences (See \citep{holzmannSPIN}).
Lastly, assumption (c) is enforced by letting the variable $\mathit{estop}$ represent the estop command
received by Gcdrive as well.
With these assumptions, Spin can verify the correctness of the system with respect to the
desired properties.
The \acrshort{promela} models of the components involved in this example can be found in
\citep{WongpiromsarnThesis}.

Realizing that these assumptions needed to be enforced, we then modified
the implementation of Alice by having Gcdrive store all the estop commands in a queue and
process all these commands one by one.
If an estop command is not stored but only sampled at the beginning of each iteration,
an estop disable or pause command may not be handled appropriately.
Consider, for example, the case where an estop pause command is sent
while Actuation Interface is in the middle of an iteration
and an estop run command is sent immediately after.
In this case, Alice will not stop because the estop pause command is not processed,
leading to an incorrect, unsafe behavior.

\chapter{Closed-System Synthesis}
\glsresetall

This section provides an overview of correct-by-construction synthesis of control protocols for autonomous systems and discusses approaches that merge concepts from formal methods and controls. These concepts include but are not limited to formal specification languages, discrete protocol synthesis, and optimization-based control.
It also points to approaches that deal with settings where the set of desired specifications is not realizable as a whole.

We consider a discrete system modeled as an action-deterministic finite transition system, i.e.,
at every state, the actions uniquely determine the next state.
We refer to these systems as deterministic systems.
In particular, we consider closed systems by referring to systems whose outputs are generated purely by themselves without any exogenous input.
We assume that at any time instance, the state of the system is fully observable.

\section{Control Protocol Synthesis}
In this section, we are interested in synthesizing a control protocol for a transition system
to ensure that a given \gls{ltl} specification is satisfied.
We define a control protocol for a transition system as follows:

\begin{definition}
  Let $TS = (S, Act, \to, I, AP, L)$ be a transition system.
  A \emph{control protocol} for $TS$ is a function $\control : S^+ \to Act$
  such that $\control(s_0 s_1 \ldots s_n) \in Act(s_n)$ for all
  $s_0 s_1 \ldots s_n \in S^+$, where $S^{+}$ denotes the set of nonempty finite strings of $S$.
\end{definition}

A control protocol $\control$ for a transition system $TS$ essentially restricts the non-deterministic choices in $TS$ by picking an action based on the path fragment that leads to the system's current state.
Hence, $\control$ induces a transition system $TS^\control$ that formalizes the behavior of $TS$ under the control protocol $\control$.

In general, $TS^\control$ contains all of the states in $S^+$; therefore, the induced transition system may not be finite even though $TS$ is finite.
However, for special cases where $\control$ is a memoryless or a finite-memory control protocol,
it can be shown that $TS^\control$ can be identified with a finite transition system.
Roughly, a memoryless control protocol picks an action based on the current state of $TS$,
irrespective of the path fragment that led to that state. For example, a memoryless control protocol is sufficient for a car in a controlled intersection; the car proceeds if the light is green and stops if the light is red. 
A finite memory control protocol, on the other hand, maintains a ``mode'', picks an action based on the current mode and the current state of $TS$, and modifies the mode according to the next state. For example, an uncontrolled intersection environment with stop signs requires cars to have finite memory control protocols. The ``mode'' of the controller keeps the number of cars that arrived at the intersection before, and the car proceeds if and only if there are no cars that arrived before.

\begin{figure}
  \centering
  \input{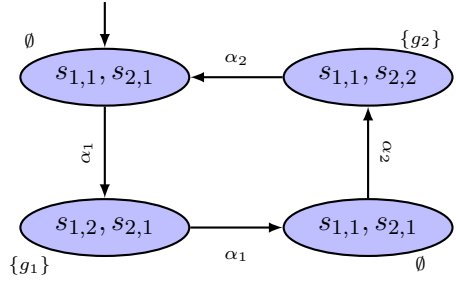}
  \caption{$(T_1 || T_2)^\control$, the transition systems induced by applying $\control$ defined in Example \ref{ex:lights_contr} on
    the traffic light system $T_1 || T_2$.}
  \label{fig:lights_model_contr}
\end{figure}

\begin{example}
  \label{ex:lights_contr}
  Consider the transition system that represents the complete traffic light system in Example
  \ref{ex:lights_model} (see \figurename~\ref{fig:lights_model}).
  Define a control protocol $\control : S^+ \to Act$ such that
  \begin{itemize}
  \item $\control(\langle s_{1,1}, s_{2,1} \rangle) = \alpha_1$,
  \item $\control(\pi \langle s_{1,1}, s_{2,1} \rangle \langle s_{1,2}, s_{2,1} \rangle) = \alpha_1$,
  \item $\control(\pi \langle s_{1,2}, s_{2,1} \rangle \langle s_{1,1}, s_{2,1} \rangle) = \alpha_2$,
  \item $\control(\pi \langle s_{1,1}, s_{2,1} \rangle \langle s_{1,1}, s_{2,2} \rangle) =
    \alpha_2$, and
  \item $\control(\pi \langle s_{1,1}, s_{2,2} \rangle \langle s_{1,1}, s_{2,1} \rangle) = \alpha_1$,
  \end{itemize}
  for any $\pi \in S^*$.
  According to the transition system $T_{1} || T_{2}$ in \figurename~\ref{fig:lights_model}, the
  initial state is $\langle s_{1,1}, s_{2,1} \rangle$.
  At this state, the control protocol picks the action $\control(\langle s_{1,1}, s_{2,1} \rangle) =
  \alpha_1$, causing the system to transition to the state $\langle s_{1,2}, s_{2,1} \rangle$.
  The path fragment so far is then given by $\langle s_{1,1}, s_{2,1} \rangle \langle s_{1,2}, s_{2,1}
  \rangle$.
  Thus, the control protocol picks the action
  $\control(\pi \langle s_{1,1}, s_{2,1} \rangle \langle s_{1,2}, s_{2,1} \rangle) = \alpha_1$,
  with $\pi = \emptyset$, causing the system to transition back to the initial state $\langle
  s_{1,1}, s_{2,1} \rangle$.
  Following this procedure, the next action is
  $\control(\pi \langle s_{1,2}, s_{2,1} \rangle \langle s_{1,1}, s_{2,1} \rangle) = \alpha_2$,
  with $\pi = \langle s_{1,1}, s_{2,1} \rangle$.
  The system then transition to $\langle s_{1,1}, s_{2,2} \rangle$, at which the control action
  $\control(\pi \langle s_{1,1}, s_{2,1} \rangle \langle s_{1,1}, s_{2,2} \rangle) =
  \alpha_2$ is applied, causing the system to transition back to the initial state $\langle s_{1,1},
  s_{2,1} \rangle$ once again.
  The transition system induced by $\control$ is shown in \figurename~\ref{fig:lights_model_contr}.
  This transition system satisfies $\always(\neg g_1 \oor \neg g_2)$ since no state in the induced system has a label including $g_1$ and $g_2$. It also satisfies $\always \eventually g_1$ since state $\langle s_{1,2}, s_{2,1} \rangle$ with label $\lbrace g_1 \rbrace$ is visited infinitely often. However, it violates $\always (g_1 \oor g_2)$ since states $\langle s_{1,1},s_{2,1} \rangle$ and $\langle s_{1,2},s_{2,2} \rangle$ in the induced system do not have labels including $g_1$ or $g_2$.
\end{example}

\noindent \textbf{Control Protocol Synthesis Problem:}
Given a finite transition system $TS$ and a specification $\varphi$ expressed as an \gls{ltl} formula,
automatically synthesize a control protocol $u$ such that the induced transition system $TS^{\control}$ satisfies $\varphi$.
\\

\begin{example}
  \label{ex:robot}
  Consider the robot motion planning problem where the robot navigates in an area that is partitioned
  into cells as shown in \figurename~\ref{fig:robot}.
  The dynamics of the robot is abstracted to a finite transition system $TS$ shown on the right of \figurename~\ref{fig:robot}.
  The state of $TS$ represents the cell occupied by the robot.
  If $TS$ is action-deterministic,
  then the system is deterministic.
  Otherwise, the system is non-deterministic.
  In this case, non-determinism potentially arises due to disturbances that affect the dynamics of the robot,
  leading to multiple possible next states when an action is taken.
  The desired property of the system is for a robot to visit cell $C_8$, then $C_1$
  and, subsequently, cover $C_{10}$, $C_{17}$, and $C_{25}$ in any order while always avoiding cells $C_2$, $C_{14}$, and $C_{18}$.
  This property can be expressed in \gls{ltl} as
  $$\eventually \big(C_8 \aand \eventually(C_1 \aand \eventually C_{10} \aand \eventually C_{17} \aand \eventually C_{25}) \big)
  \aand \always \neg(C_2 \oor C_{14} \oor C_{18}).$$
\end{example}

\begin{figure}
  \centering
  \input{./figures/figure4_2.tikz}
  \caption{The robot motion planning problem in Example \ref{ex:robot}. Each cell represents a state of $TS$. The possible transitions are between the adjacent cells. }
  \label{fig:robot}
\end{figure}

\begin{example}
  \label{ex:robot2}
  Consider the simplified autonomous driving problem described in Example \ref{ex:autonomous}.
  The system consists of the autonomous vehicle, obstacles (i.e., $Obs$ in Traffic rule 1),
  and other vehicles (i.e., $Veh$ in Traffic rule 1).
  If the obstacles and other vehicles are not stationary, and their motion is not known exactly,
  then the system is non-deterministic since the system does not have control over the motion of the obstacles and other vehicles.
  In this case, a control protocol for this system needs to ensure that the desired properties described in Example \ref{ex:autonomous}
  are satisfied for all the possible motion (i.e., behavior) of the obstacles and other vehicles.
\end{example}

\section{Model Checking-Based Synthesis}
\label{sec:logic}
Consider a closed system that is modeled as an action-deterministic finite transition system $TS$.
The control-protocol synthesis problem can be formulated as finding a path in $TS$ that satisfies a given specification $\varphi$, which is essentially a model checking problem as described in Section \ref{sec:model-checking}.
Using model checking methods, we can start with the hypothesis that $\varphi$ is not satisfiable; that is, we investigate whether or not there exist a path $\pi$ of $TS$ that satisfies $\varphi$. In case the hypothesis can be refuted with a counterexample, the counterexample can be used as a synthesized path $\pi$ of $TS$ that satisfies $\varphi$.

Let $\mathcal{A}$ be a non-deterministic B\"{u}chi automaton over $2^{\props}$ that accepts all and only
words over $\props$ that satisfy $\varphi$.
We construct the product $TS_p = TS \otimes \mathcal{A}$ as described in Definition \ref{def:product}.
Let $\pi_p = \langle s_0, q_0 \rangle \langle s_1, q_1 \rangle \ldots$ be an accepting path on
$TS_p$ and $\pi = s_0 s_1 \ldots$ be the path on $TS$ corresponding to $\pi_p$.
We define a control protocol $\control$ for $TS$ by
\begin{equation*}
  \control(s_0' s_1' \ldots s_i') =
  \left\{
    \begin{array}{ll}
      \alpha_i &\hbox{if } s_0' s_1' \ldots s_i' = s_0 s_1 \ldots s_i,\\
      \alpha_i' &\hbox{otherwise},
    \end{array}\right.
\end{equation*}
where $\alpha_{i}$ satisfies $s_i \stackrel{\alpha_i}{\to} s_{i+1}$ and
$\alpha_i' \in Act(s_i')$ is any arbitrary action.
Under the control protocol $\control$, the system simply picks the next state according to the path
$\pi$, which ensures that the resulting path satisfies $\varphi$.
Note that this construction of control protocol works because we consider a closed system, which has a
full control over the non-deterministic choices in $TS$ and is not affected by exogenous inputs
(e.g., from the environment).

\begin{figure}
  \centering
  \input{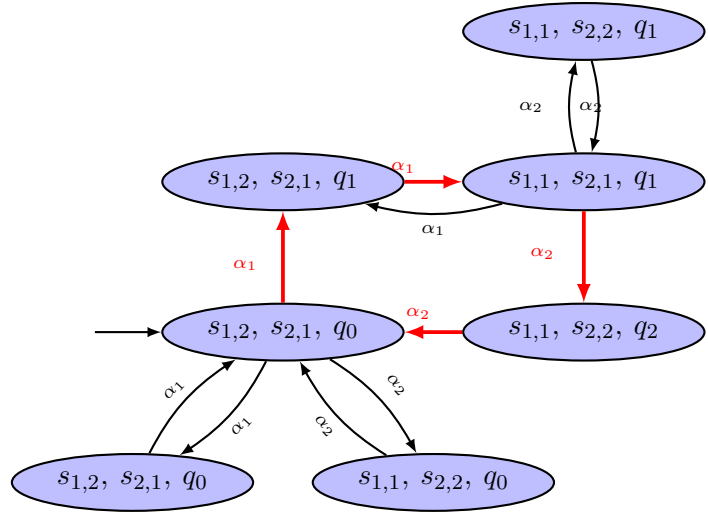}

  \caption{The product transition system $TS_p = (T_1 || T_2) \otimes \mathcal{A}$, showing only reachable states.
    An accepting path is highlighted by double (red) arrows.}
  \label{fig:lights_product}
\end{figure}

\begin{example}
  Consider the traffic light system $TS = T_1 || T_2$ shown in \figurename~\ref{fig:lights_model}
  and the desired property $\varphi = \always(\neg g_1 \oor \neg g_2) \aand \always\eventually g_1 \aand \always\eventually g_2$. In words, the property makes sure that the two lights are never green at the same time and each light turns green infinitely often.
  A non-deterministic B\"{u}chi automaton $\mathcal{A}$ that recognizes $\varphi$ and
  the product transition system $TS_p = TS \otimes \mathcal{A}$ are shown in \figurename~\ref{fig:lights_aut_syn} and
  \figurename~\ref{fig:lights_product}, respectively.
  In relation to \figurename~\ref{fig:lights_model_contr}, projecting the path that is shown in \figurename~\ref{fig:lights_product} onto the state of $TS$ yields the same transition system that applying the control protocol in Example \ref{ex:lights_contr} induces.
\end{example}

\begin{figure}
  \centering
  \includegraphics{figures/figure4_3.tex}

  \caption{Non-deterministic B\"{u}chi automaton $\mathcal{A}$ that recognizes
    $\varphi = \always(\neg g_1 \oor \neg g_2) \aand \always\eventually g_1 \aand \always\eventually g_2$.
    Accepted states are drawn with a double (red) circle.}
  \label{fig:lights_aut_syn}
\end{figure}

\section{A Case Study}
\label{sec:case_study_left_turn}
\begin{figure}
  \centering
  \includegraphics[width=0.3\textwidth]{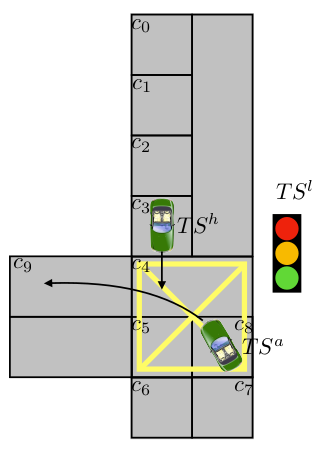}
  \caption{An unprotected left turn scenario.}
  \label{fig:unprotected_left}
\end{figure}

We now take a closer look into a case study motivated by autonomous driving. Consider a scenario where an autonomous vehicle $TS^{a}$ needs to make an unprotected left turn
with an oncoming vehicle $TS^{h}$ as shown in Figure \ref{fig:unprotected_left}.
The complete system consists of the autonomous vehicle, the oncoming vehicle, and the traffic light.

\begin{figure}
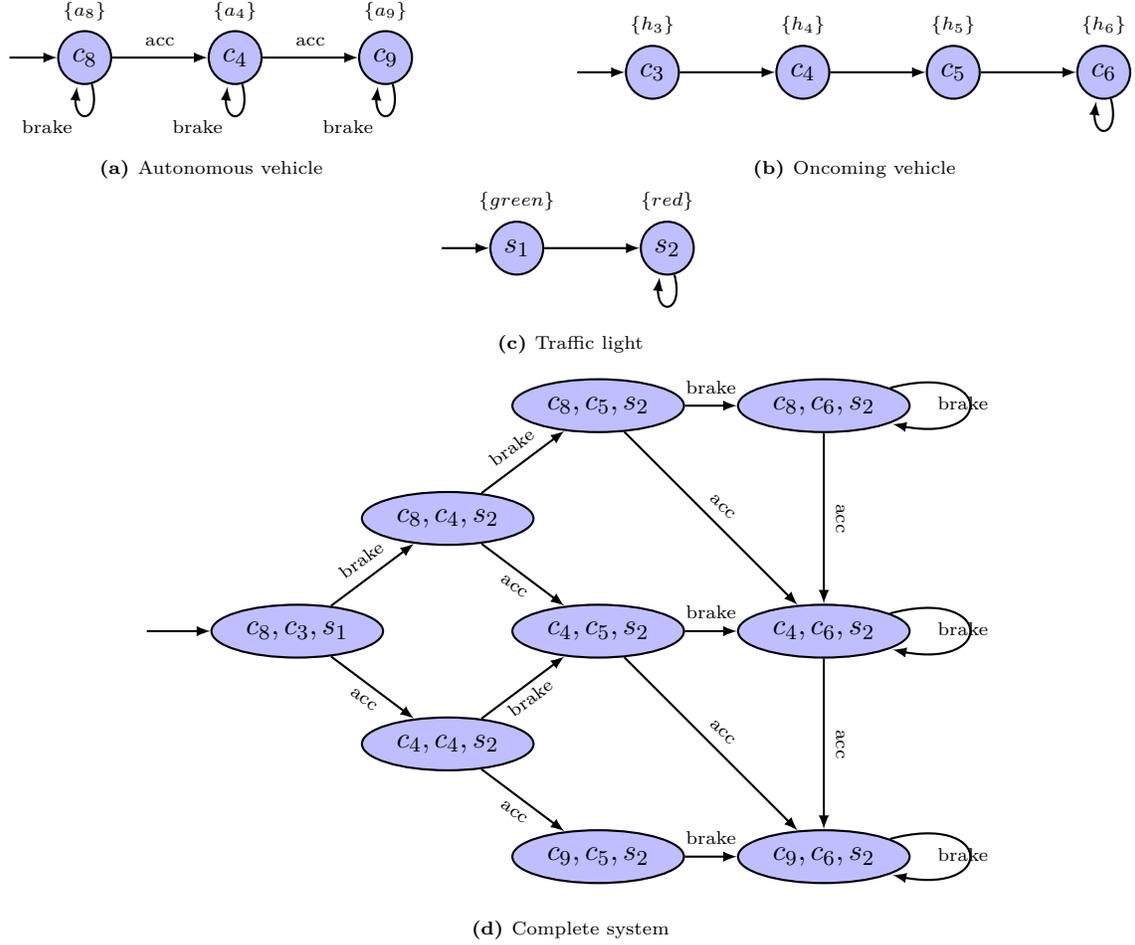

  \centering
  \subfloat[Autonomous vehicle]%
  {\input{./figures/figure4_6a.tikz}}
  \hfill
  \subfloat[Oncoming vehicle]%
  {\input{./figures/figure4_6b.tikz}}
  \hfill
  \subfloat[Traffic light]%
  {\input{./figures/figure4_6c.tikz}}\\
  \subfloat[Complete system]%
  {\input{./figures/figure4_6d.tikz}}
  \caption{The finite transition systems that describe the behavior of the autonomous vehicle,
    the oncoming vehicle, the traffic light, and the complete system.}
  \label{fig:unprotected_left_det_model}
\end{figure}

First, consider the case where the complete system is deterministic, i.e., starting from any state,
the actions of the autonomous vehicle lead to a unique state of the complete system.
This implies that the behavior of the oncoming vehicle and the traffic light is deterministic.
Figure \ref{fig:unprotected_left_det_model} shows the finite transition systems
that describe the behavior of each element as well as the complete system $TS$.
Here, a state $s \in S$ is of the form $s = (c_{i}, c_{j}, s_{k})$,
where $i$ and $j$ are the labels of the cells occupied by the autonomous vehicle and the oncoming
vehicle, respectively, and $k$ indicates whether or not the light is green; $k = 1$ if the light is green and $k = 2$ if the light is red. $Act = \{acc, brake\}$ where $acc$ and $brake$ represent accelerating and braking, respectively.
$AP = \{a_{0}, \ldots, a_{9}, h_{0}, \ldots, h_{9}, g\}$ and the labeling function $L$ is defined
such that for any state $s = (c_{i}, c_{j}, s_{k})$, $a_{i}, h_{j} \in L(s)$ and $g \in L(s)$ if and
only if $k = 1$.

\begin{figure}
  \centering
  {\input{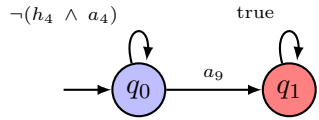}}
  \caption{The nondeterministic B\"uchi automaton corresponding to \gls{ltl} formula $\neg(h_{4} \aand
    a_{4}) \until a_{9}$.}
  \label{fig:unprotected_left_det_nba}
\end{figure}

\begin{figure}
  \centering
  {\input{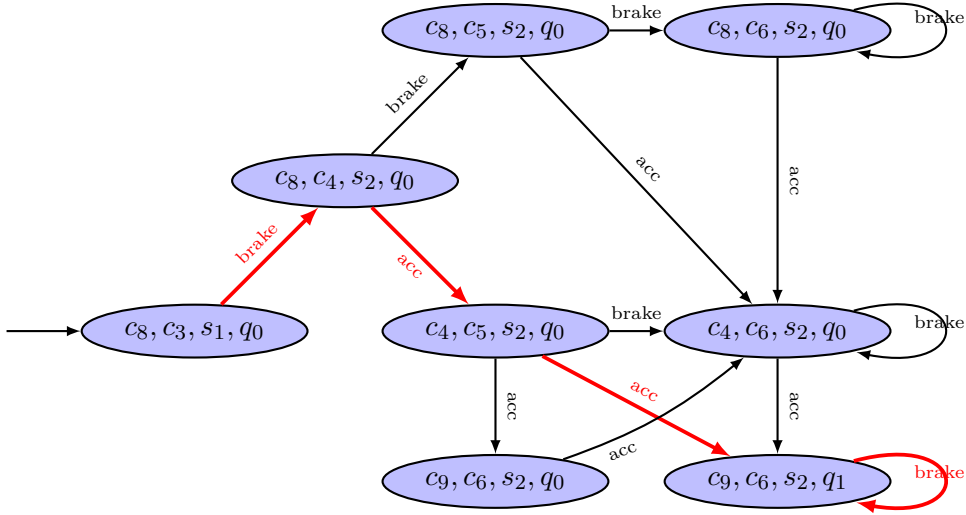}}
  \caption{The product $TS \otimes \mathcal{A}$ as well as a run on $TS$ that satisfies $\varphi$
    shown in green.}
  \label{fig:unprotected_left_det_sol}
\end{figure}

An \gls{ltl} formula $\varphi = \neg (h_{4} \aand a_{4}) \until a_{9}$ specifies that the
autonomous vehicle and the oncoming vehicle do not simultaneously occupy cell $c_{4}$ until the
autonomous vehicle reaches cell $c_{9}$.
Figure \ref{fig:unprotected_left_det_nba} shows the corresponding nondeterministic B\"uchi automaton
$\mathcal{A}$.
The product $TS \otimes \mathcal{A}$ as well as a solution is shown in Figure \ref{fig:unprotected_left_det_sol}.

\section{Minimum-Violation Synthesis}
\label{sec:mvp}
Autonomous systems are often subject to multiple regulatory requirements or safety rules
that may not be equally important.
In general, it is infeasible to guarantee the satisfaction of all the rules under all conditions.
Thus, there is a need to allow for formally justifiable violation of these rules.
In particular, we assume that each rule has a certain penalty associated with its
violation.
The goal of \emph{minimum-violation synthesis} is to minimize such penalties.

The formulation of the minimum-violation synthesis requires some extension to the models and
specifications considered in Section \ref{chap:models_and_specifications}.
First, we consider a system that can be modeled as a weighted, finite, and action-deterministic
transition system.

\begin{definition}
  A \gls{wts} is a tuple
  $TS = (S, Act, \to, I, \AP, L, W)$ where
  $S$, $Act$, $\to$, $I$, $AP$, and $L$ are the same as in Definition \ref{def:transition_system} and, for some fixed $m \in \naturals$, $W : S \times S \to \reals^{m}_{\geq 0}$ is a weight function.
\end{definition}

For states $s_{1}, s_{2} \in S$, the weight $W(s_{1}, s_{2})$ typically represents the time or
distance between $s_{1}$ and $s_{2}$.
Similar to conventional transition systems, a \gls{wts} is \emph{finite}
if $S$, $Act$ and $\AP$ are all finite, and is \emph{action-deterministic}
if $|I| \leq 1$ and, for all $s \in S$ and $\alpha \in Act$, $|Post(s, \alpha)| \leq 1$.
The weight of a finite path fragment $\pi = s_{0}s_{1} \ldots s_{n}$ is $W(\pi) = \sum_{i = 0}^{n-1}W(s_{i}, s_{i+1})$.

We will use \gls{fltl} to formalize each rule.
As opposed to \gls{ltl}, an \gls{fltl} formula $\spec$ is interpreted over a finite word
$\sigma = \sigma_{0} \sigma_{1} \ldots \sigma_{n} \in (2^{\props})^{n+1}$. We write $\sigma \models \spec$ if and only if $\sigma$ satisfies $\spec$.
For example, for some $p \in \props$, $\sigma \models p$ if and only if $p \in \sigma_{0}$. Additionally,
$\sigma \models \always p$ if and only if, for all $i \in \{0, \ldots, n\}$, we have $p \in \sigma_{i}$. As a more complicated example, with $p,p' \in \props$, let $\spec = \always(p \implies (\nextop p \oor
\nextop p'))$. Then, $\sigma \models \spec$ if and only if, for all $i \in \{0, \ldots, n-1\}$ such that
$p \in \sigma_{i}$, we have $p \in \sigma_{i+1}$ or $p' \in \sigma_{i+1}$.

Given an \gls{fltl} formula $\spec$, a finite automaton $\automaton = (Q,Q_0,\delta,F)$ that accepts all and only finite
words that satisfy $\spec$ can be automatically constructed \citep{TACAS-2002-GunterP}.
Here, the definitions of $Q$, $Q_0$, $\delta$ and $F$ are similar to those in the definition of a
B\"uchi automaton (see Section \ref{sec:spec}).
However, a finite automaton is evaluated over a finite word $\sigma = \sigma_{0} \ldots \sigma_{n}$.
A run of $\automaton$ over $\sigma$ is defined as a finite sequence $q_{0} \ldots q_{n+1}$ such that $q_{0} \in
Q_{0}$ and $(q_{i}, \sigma_{i}, q_{i+1}) \in \delta$ for all $i \in \{0, \ldots, n\}$.
We say that $\automaton$ accepts $\sigma$ if and only if there exists a run
$q_{0} \ldots q_{n+1}$ of $\automaton$ over $\sigma$ such that $q_{n+1} \in F$.

We now introduce the concept of prioritized safety specification to formalize a set of rules
with unequal importance.

\begin{definition}
  A \emph{prioritized safety specification} is a tuple
  $\pspec = (\props, \Phi,\allowbreak \Psi, \allowbreak \priority)$ where
  $\props$ is a set of atomic propositions,
  $\Phi$ is a set of \gls{fltl} formulas over $\props$,
  $\Psi = (\Psi_0, \Psi_1, \ldots, \Psi_N)$
  organizes the formulas in $\Phi$ into a hierarchy based on their priorities
  such that $\Psi_i \subseteq \Phi$, for all $i \in \naturalsle{N}$,
  and
  $\priority : \Phi \to \naturals$
  is a function that assigns the weight to each
  $\spec \in \Phi$.
  Throughout the article, we refer to each $\spec \in \Phi$
  as an \emph{atomic safety rule}.
  \label{def:prioritized-safety-specification}
\end{definition}

\begin{example}
  \label{ex:prioritized-safety-specification}
  Consider an autonomous vehicle that is required to
  (1) avoid collision,
  (2) stay on the road,
  (3) keep sufficient clearance from other vehicles, and
  (4) stay within a correct lane.
  To capture these rules, we define the following atomic propositions.
  \begin{itemize}
  \item $\mathsf{collision}$ represents a state at which the autonomous vehicle
    collides with another vehicle or obstacle.
  \item $\mathsf{onroad}$ represents a state at which the autonomous vehicle is fully on the road.
  \item $\mathsf{close}$ represents a state at which the autonomous vehicle overlaps with the
    clearance zone around another vehicle.
  \item $\mathsf{inlane}$ represents a state at which the autonomous vehicle is fully within a
    correct lane.
  \end{itemize}
  Following Example \ref{ex:autonomous}, we let $(x, \theta, v) \in \reals^{3}$ represent the state
  of the autonomous vehicle, $\mathsf{FP}(x,\theta) \subset \reals^{2}$ represent its footprint, and
  $\mathsf{Obs} \subset \reals^{2}$ represent the union of the footprints of all the vehicles and
  obstacles in the environment.
  Additionally, we define the following environment states.
  \begin{itemize}
  \item $\mathsf{RD} \subset \reals^{2}$ is the road, i.e., the area where a vehicle is allowed to
    drive, and
  \item $\mathsf{CZ} \subset \reals^{2}$ is the union of the clearance zone around other vehicles constructed
    from their footprints and the required lateral and longitudinal clearance,
  \item $\mathsf{LN} \subset \reals^{2}$ is the right lane, i.e., the lane with the correct travel direction
    for the autonomous vehicle.
  \end{itemize}
  The state of the complete system (autonomous vehicle and the environment) is then given by $(x,
  \theta, v, \mathsf{Obs}, \mathsf{RD}, \mathsf{CZ}, \mathsf{LN})$.
  The labeling function $L$ is defined such that for any
  $x, \theta, v, \mathsf{Obs}, \mathsf{RD}, \mathsf{CZ}, \mathsf{LN}$,
  \begin{itemize}
  \item $\mathsf{collision} \in L(x, \theta, v, \mathsf{Obs}, \mathsf{RD}, \mathsf{CZ}, \mathsf{LN})$ iff
    $\mathsf{FP}(x, \theta) \intersect \mathsf{Obs} \not= \emptyset$,
  \item $\mathsf{onroad} \in L(x, \theta, v, \mathsf{Obs}, \mathsf{RD}, \mathsf{CZ}, \mathsf{LN})$ iff
    $\mathsf{FP}(x, y, \theta) \subseteq \mathsf{RD}$,
  \item $\mathsf{close} \in L(x, \theta, v, \mathsf{Obs}, \mathsf{RD}, \mathsf{CZ}, \mathsf{LN})$ iff
    $\mathsf{FP}(x, y, \theta) \intersect \mathsf{CZ} \not= \emptyset$, and
  \item $\mathsf{inlane} \in L(x, \theta, v, \mathsf{Obs}, \mathsf{RD}, \mathsf{CZ}, \mathsf{LN})$ iff
    $\mathsf{FP}(x, y, \theta) \subseteq \mathsf{LN}$.
  \end{itemize}

  We consider the following atomic safety rules, each of which
  can be expressed by an \gls{fltl} formula.
  \begin{enumerate}[(i)]
  \item Avoiding collision: $\spec_{0} = \always \neg \mathsf{collision}$.
  \item Staying on the road: $\spec_{1} = \always \mathsf{onroad}$.
  \item Keeping sufficient clearance from other vehicles: $\spec_{2} = \always \neg\mathsf{close}$.
  \item Staying within a correct lane: $\spec_{3} = \always \mathsf{inlane}$.
  \end{enumerate}

  Define the prioritized safety specification $\pspec = (\props, \Phi,\allowbreak \Psi, \allowbreak \priority)$
  as
  \begin{align*}
    \props &= \{\mathsf{collision}, \mathsf{close}, \mathsf{onroad}, \mathsf{inlane}\},\\
    \Phi &= \{\spec_{0}, \ldots, \spec_{3}\},\\
    \Psi &= \{\Psi_{0}, \Psi_{1}, \Psi_{2}\}, \text{ and}\\
    \priority(\spec_{0}) &= 1, \forall i,
  \end{align*}
  where, $\Psi_{0} = \{\spec_{0}\}$, $\Psi_{1} = \{\spec_{1}\}$, and $\Psi_{2} = \{\spec_{2}, \spec_{3}\}$.
  In this definition of $\pspec$, $\spec_{0}$ is the only rule at the top level in the hierarchy.
  As we will see later, this means that $\spec_{0}$ is the most important rule, and the system
  should minimize the violation of this rule, even at the cost of infinitely violating the other rules.
  Next, $\spec_{1}$ is the only rule at the second level.
  This means that after minimizing $\varphi_{0}$, the system will minimize the violation of
  $\varphi_{1}$, even at the cost of infinitely violating the other rules at the lower levels.
  Finally, $\spec_{2}$ and $\spec_{3}$ are at the lowest level.
  This means that after minimizing the violation of $\varphi_{0}$ and $\varphi_{1}$, the system then
  tries to minimize the weighted violation of $\spec_{2}$ and $\spec_{3}$, where the weights are
  given by $\priority$.
  Note that the weights only affect the rules at the same level.
\end{example}

We use the \emph{level of unsafety} $\lambda_{\spec}(\pi)$ to measure the violation of a
finite path fragment $\pi$ with respect to an atomic safety rule $\spec$.
The dual concept, including robustness, has been proposed for some variants of temporal logics such
as Signal Temporal Logic \citep{Donze:2010:Robust,Mehdipour:2019:Average}.
However, when applying to a system that is subject to multiple requirements,
this measure  introduces the additional complexity of having to ensure that the system cannot
take advantage of robustly satisfying one rule, in order to slightly violate another rule.
Additionally, when combining with the concept of rule hierarchy, this measure may cause the system
to compromise lower-level rules by robustly satisfying the higher-level rules.
For example, consider the prioritized safety specification in Example
\ref{ex:prioritized-safety-specification}.
Using the robustness as the performance measure may incentivize the system to violate staying on the road
rule $\varphi_{2}$ in order to robustly satisfy the collision avoidance rule $\varphi_{1}$ since
this behavior allows the system to be as far away from other vehicles as possible, and hence,
maximizing the robustness of the top-level rule.

Various definitions of the level of unsafety have been proposed.
For example, \citet{TumovaHKFR13} defines $\lambda_{\spec}(\pi)$ as the minimum number of states in
$\pi$ that needs to be removed so that the resulting path fragment satisfies $\spec$.
Formally, given a finite sequence $\pi = s_{0}s_{1} \ldots s_{n}$ and
a set $I \subseteq \{0, \ldots, n\}$,
let $\vanish(\pi, I)$ represent a subsequence of $\pi$ obtained by removing all
$s_{i}$ with $i \in I$.
Then, $\lambda_{\spec}(\pi) = \min_I \{ |I| \text{ s.t. } \vanish(\pi, I) \models \pi\}$.
In contrast, \citep{Castro:2013:CDC} defines the level of unsafety as
$\lambda_{\spec}(\pi) =  \min_I \{\sum_{i \in I \setminus \{n\}} W(s_{i}, s_{i+1}) \text{ s.t. }
\vanish(\pi,I) \models \pi\}$.
Following \citet{Wongpiromsarn:2021:Minimum}, we restrict atomic safety rules to be expressed using a
class of \gls{fltl} known as $\siFLTLX$ defined as follows:

\begin{definition}
    An $\siFLTLX$ formula over a set $\AP$ of atomic propositions is an \gls{fltl} formula
    that is stutter-invariant (see below) and is of the form
    \begin{equation*}
    \spec = \always \PX,
    \label{eq:siFLTLX}
  \end{equation*}
  where $\PX$ belongs to the smallest set defined inductively by the following rules:
  \begin{itemize}
  \item $p$ is a formula for all $p \in \props \cup \{\true, \false\}$;
  \item $\nextop p$ is a formula for all $p \in \props  \cup \{\true, \false\}$; and
  \item if $\PX^1$ and $\PX^2$ are formulas, then so are
    $\neg \PX^1$,
    $\PX^1 \oor \PX^2$,
    $\PX^1 \aand \PX^2$ and
    $\PX^1 \implies \PX^2$.
  \end{itemize}
  In other words, $\PX$ is a Boolean combination of propositions from $\props$ and expressions
  of the form $\nextop p$ where $p \in \props$.
  \label{def:siFLTLX}
\end{definition}

Roughly, a specification is stutter-invariant if its satisfaction
with respect to any word is not affected
by operations that duplicate some letters or remove some duplicate letters in that word.
For example, consider $\sigma = \sigma_0 \sigma_1 \ldots \sigma_n$ and
$\sigma' = \sigma_0 \sigma_1 \ldots \sigma_{i-1} \sigma_i \sigma_i \sigma_{i+1} \ldots \sigma_n$, which is constructed from
$\sigma$ by duplicating $\sigma_i$ for some $i \in \{0, \ldots, n\}$.
If $\spec$ is stutter-invariant,
then $\sigma \satisfies \spec$ if and only if $\sigma' \satisfies \spec$.
We refer the reader to \citep{Peled97} for the definition
of stutter-invariant properties.
See, e.g., \citep{Klein:2007:CIAA,Michaud:2015:PSC} for approaches to check
whether a specification is stutter-invariant.

\begin{example}
  All the rules $\spec_{0}, \ldots, \spec_{3}$ defined in Example \ref{ex:prioritized-safety-specification}
  are $\siFLTLX$ formulas.
  For example, consider $\spec_{0} = \always \neg \mathsf{collision}$.
  It can be written as $\spec_{0} = \always \PX$, where $\PX = \neg \mathsf{collision}$.
\end{example}

Regardless of their simplicity, $\siFLTLX$ formulas turn out to be sufficiently expressive
in many applications;
for example,
\citep{Wongpiromsarn:2011:Synthesis} shows
that all the rules enforced in the DARPA Urban Challenge 2007
can be expressed with $\siFLTLX$ formulas.
Moreover, all of the traffic rules in the examples presented in \citep{Castro:2013:CDC}
can be described using $\siFLTLX$ formulas.

\citet{Wongpiromsarn:2021:Minimum} show that
the violation of a $\siFLTLX$ formula is caused either by
visiting an unsafe state or by taking an unsafe transition.
As a result, this work defines the level of unsafety as the total
time spent in an unsafe state and the total number of unsafe transitions.

Let $\pspec = (\props, \Phi, \Psi, \priority)$ be a prioritized safety specification
where $\Psi = (\Psi_0, \Psi_1, \ldots, \Psi_N)$.
We define the level of unsafety of a finite-path fragment $\pi$ with respect to $\pspec$ as
\begin{equation*}
   \lambda_{\pspec}(\pi) = \left(\lambda_{\Psi_0}(\pi), \ldots, \lambda_{\Psi_N}(\pi)\right)
   \in \reals^{N+1},
   \label{eq:unsafety-prioritized}
\end{equation*}
where, for every $i \in \{0, 1, \ldots, N\}$,
\begin{equation*}
  \lambda_{\Psi_i}(\pi) =
  \sum_{\spec \in \Psi_i} \priority(\spec) \lambda_{\spec}(\pi).
  \label{eq:unsafety-set}
\end{equation*}

Let $Path_{G}(TS)$ be the set of finite path fragments that end in a goal state.
Given a weighted, finite, and action-deterministic transition system $TS = (S, Act, \to, I, \AP, L, W)$ and
a prioritized safety specification $\pspec = (\props, \Phi, \Psi, \priority)$,
the minimum-violation synthesis problem is to compute an optimal path fragment $\pi^{*} \in
Path_{G}(TS)$ that minimizes the weight $W(\pi^{*})$ among all the path fragments that minimize the
level of unsafety with respect to $\pspec$.
Formally, we define the cost function
$J : Path_{G}(TS) \to \reals^{N+2}$ as
\begin{equation}
  J(\pi) =
  \big(\lambda_{\pspec}(\pi), W(\pi)\big).
  \label{eq:cost-function}
\end{equation}

Using the cost function $J$, we now formally define the minimum-violation synthesis problem.

\noindent \textbf{Minimum-Violation Synthesis Problem:}
Based on the standard lexicographical ordering, compute an optimal finite path fragment $\pi^{*}$ such that
$$\pi^{*} = \arg\min_{\pi \in Path_{G}(TS)} J(\pi).$$

\begin{example}
  Consider the prioritized safety specification $\pspec$ in Example
  \ref{ex:prioritized-safety-specification}.
  In this case, we have $\lambda_{\pspec}(\pi) = \left(\lambda_{\Psi_0}(\pi), \lambda_{\Psi_{1}}(\pi),
    \lambda_{\Psi_2}(\pi)\right)$, where
  \begin{align*}
    \lambda_{\Psi_0}(\pi) &= \lambda_{\varphi_{0}}(\pi),\\
    \lambda_{\Psi_1}(\pi) &= \lambda_{\varphi_{1}}(\pi), \text{ and}\\
    \lambda_{\Psi_2}(\pi) &= \lambda_{\varphi_{2}}(\pi) + \lambda_{\varphi_{3}}(\pi).
  \end{align*}
  As a result, the cost function $J : Path_{G}(TS) \to \reals^{4}$ is defined as
  \begin{equation*}
    J(\pi) = \big(\lambda_{\Psi_0}(\pi), \lambda_{\Psi_1}(\pi), \lambda_{\Psi_2}(\pi), W(\pi)\big).
  \end{equation*}
  For convenience, we define the following subsets.
  \begin{itemize}
  \item $Path_{G,0}(TS)$ is the subset of $Path_{G}(TS)$ that
    minimize the level of unsafety with respect to $\spec_0$.
  \item $Path_{G,0,1}$ is the subset of
    $Path_{G,0}(TS)$ that minimize the level of unsafety with respect to $\spec_{1}$.
  \item $Path_{G,0,1,2}$ is the subset of
    $Path_{G,0,1}(TS)$ that minimize the level of unsafety with respect to $\spec_{2}$
    and $\spec_{3}$.
  \end{itemize}
  An optimal path $\pi^{*}$ is defined as a path in $Path_{G,0,1,2}$ that minimizes the weight
  $W(\pi^{*})$, which typically represents the time or distance on the path fragment.
\end{example}

\citet{TumovaHKFR13} and \citet{Castro:2013:CDC} solve the minimum-violation synthesis problem by
constructing a weighted finite automaton $\automaton$
that is the product of a collection of weighted finite automata,
each of which corresponds to an atomic safety rule
$\spec \in \Phi$.
A weighted finite automaton is essentially a finite automaton with
weight $W(q, \sigma, q') \in \reals^{m}_{\geq 0}$ for some $m \in \naturals$
assigned to each transition $(q, \sigma, q')$.
The weights of the transitions of $\automaton$ are defined such that
the weight of the shortest accepting run over any word $\sigma$
is the level of unsafety of $\sigma$.
It can be shown that the minimum-violation synthesis problem is equivalent to finding the shortest path
in $TS \otimes \automaton$.

As the size of $\automaton$ is exponential in the length of $\spec$ \citep{BK08},
the approach presented in \citep{Wongpiromsarn:2021:Minimum} avoids constructing the product
$TS \otimes \automaton$ to reduce computational complexity.
The main idea is to construct a weighted finite transition system $TS'$ with the same sets
$S$, $Act$, $I$, and $AP$ of states, actions, transitions, initial states, and atomic
propositions as well as the same transition relation $\to$ and labeling function $L$ as $TS$.
Instead, the weight $W'$ will be defined such that the minimum-violation synthesis problem can be
translated to computing the shortest path on $TS'$.

To enable such a construction of $TS'$, we first translate a $\siFLTLX$ formula over $\props$
into a $\siFLTL$ formula over $\props \times \props$.

\begin{definition}[$\siFLTL$]
  A $\siFLTL$ formula over $\props \times \props$ is a $\siFLTLX$ formula
  $\spec = \always P$ where $P$ is a propositional logic formula over
  $\props \times \props$.
  \label{def:siFLTL}
\end{definition}

A propositional logic formula $P$ over
$\props \times \props$ is interpreted over a pair
$(l, l') \in 2^\props \times 2^\props$
with the satisfaction relation $\satisfies$ defined as follows:
for $p, p' \in \props \union \{\true, \false\}$ and
$(l, l') \in 2^\props \times 2^\props$,
$(l, l') \satisfies (p, p')$ if and only if
$l \satisfies p$ and $l' \satisfies p'$.
Here, for any $l \in 2^\props$, we have
$l \satisfies \true$,
$l \not\satisfies \false$,
and for any $p \in \props$,
$l \satisfies p$ if and only if $p \in l$.
The logic connectives are defined as in the standard propositional logic.

Based on the semantics of \gls{fltl},
given a finite word $\sigma = \sigma_0 \sigma_1 \ldots \sigma_n \in (2^\props)^{n+1}$
and a $\siFLTL$ formula $\spec = \always P$ over $\props \times \props$,
we say that $\sigma$ satisfies $\spec$,
written $\sigma \satisfies_{\props \times \props} \spec$
if and only if $(\sigma_i, \sigma_{i+1}) \satisfies P$ for all $i \in \naturalsle{n-1}$
and $(\sigma_n, \sigma_n) \satisfies P$.
Note that the terminal condition $(\sigma_n, \sigma_n) \satisfies P$ results
from the assumption that $\spec$ is stutter-invariant, which ensures that
$\sigma \satisfies \spec$ if and only if
$\sigma' = \sigma_0 \sigma_1 \ldots \sigma_n \sigma_n \satisfies \spec$.

Given a $\siFLTLX$ formula $\spec$ over $\props$,
we define an operation called $\denext$ which
constructs a $\siFLTL$ formula over $\props \times \props$ from $\spec$ by
replacing each instance of $p$ in $\spec$ with $(p, \true)$
and replacing each instance of $\nextop p$ in $\spec$ with $(\true, p)$,
for all $p \in \props$.
For example, consider a $\siFLTLX$ formula $\spec = \always(p \implies (\nextop p \oor
\nextop p'))$.
The corresponding $\siFLTL$ formula over $\props \times \props$ is given by

\vspace{-5mm}
\begin{equation*}
  \denext(\spec) = \always
  \Big( (p, \true) \implies \big((\true, p) \oor (\true, p') \big) \Big).
  \label{ex:denext}
\end{equation*}

\begin{example}
  \label{ex:denext}
  Consider Example \ref{ex:prioritized-safety-specification}.
  $\siFLTLX$ formulas $\spec_{0}, \ldots, \spec_{4}$ over $\AP$ can be translated to
  the corresponding $\siFLTL$ formulas over $\props \times \props$ as follows.
  \begin{enumerate}[(i)]
  \item Avoiding collision: $\denext(\spec_{0}) = \always \neg (\mathsf{collision}, \true)$.
  \item Staying on the road: $\denext(\spec_{1}) = \always (\mathsf{onroad}, \true)$.
  \item Keeping sufficient clearance from other vehicles: $\denext(\spec_{2}) = \always \neg (\mathsf{close},
    \true)$.
  \item Staying within a correct lane: $\denext(\spec_{3}) = \always (\mathsf{inlane}, \true)$.
  \end{enumerate}
\end{example}

\citet{Wongpiromsarn:2021:Minimum} establish the equivalence of the level of unsafety
with respect to a $\siFLTLX$ formula over $\props$ and
the level of unsafety with respect to the corresponding $\siFLTL$ formula over $\props \times
\props$.

The translation of $\siFLTLX$ formula over $\props$
to a $\siFLTL$ formula over $\props \times \props$
allows us to convert the original prioritized safety specification
$\pspec = (\props, \Phi, \Psi, \priority)$ to
$\hat{\pspec} = (\props \times \props, \hat{\Phi}, \hat{\Psi}, \hat{\priority})$
with each atomic safety rule obtained from that of $\pspec$
by applying $\denext$ operation.
Formally,
$\hat{\Phi} = \big\{ \denext(\spec) \hspace{1mm}|\hspace{1mm} \spec \in \Phi \big\}$;
$\hat{\Psi} = (\hat{\Psi}_0; \hat{\Psi}_1, \ldots, \hat{\Psi}_N)$;
$\hat{\Psi}_i = \big\{ \denext(\spec) \hspace{1mm}|\hspace{1mm} \spec \in \Psi_i \big\}$, for all $i
\in \{0, \ldots, N\}$; and
$\hat{\priority}(\denext(\spec)) = \priority(\spec)$, for all $\spec \in \Phi$.

For each atomic safety rule $\spec \in \Phi$, define a propositional logic formula $P_{\spec}$ such that
$\denext(\spec) = \always P_{\spec}$.
We construct the weighted finite transition system
$TS' = (S, Act, \to, I, \AP, L, W')$ such that the weight $W'$ corresponds to the
cost function in (\ref{eq:cost-function}).
Formally, $W' : S \times S \to \reals^{N+2}$ is defined by
\begin{equation*}
  W'(s_{1}, s_{2}) = (\lambda_{\hat{\Psi}_0}(s_{1}, s_{2}), \ldots, \lambda_{\hat{\Psi}_N}(s_{1},
  s_{2}), W(s_{1}, s_{2})),
\end{equation*}
where $\lambda_{\hat{\Psi}_i}(s_{1}, s_{2}) = \sum_{\spec \in \hat{\Psi}_i} \hat{\priority}(\spec)\lambda_{\spec}(s_{1}, s_{2})$
and $\lambda_{\spec}(s_{1}, s_{2})$ is defined based on the satisfaction of the propositional logic
formula $P_{\spec}$ at $s_{1}$ and $s_{2}$ as follows:

\begin{equation}
  \lambda_{\spec}(s_{1}, s_{2}) = \left\{
    \begin{array}{ll}
      0 &\text{ if } (L(s_{1}), L(s_{2})) \models P_{\spec}\\
      W(s_{1}, s_{2}) &\text{ if } (L(s_{1}), l) \not\models P_{\spec} \text{ for all } l \in 2^{\AP}\\
      1 &\text{ otherwise}
    \end{array}
  \right\}
  \label{eq:mvp-transition-weight}
\end{equation}

Note that the first condition of (\ref{eq:mvp-transition-weight}) corresponds to the case where
$P_{\spec}$ is satisfied by the transition from $s_{1}$ to $s_{2}$, thereby no violation cost being incurred.
The second condition corresponds to visiting an unsafe state $s_{1}$ and
the third condition corresponds to taking an unsafe transition.

\begin{example}
  Let us revisit Example \ref{ex:prioritized-safety-specification} with the corresponding
  $\siFLTL$ formulas over $\props \times \props$ defined in Example \ref{ex:denext}.
  For any states $s_{1}, s_{2} \in S$, the weight $W'$ is defined as
  $$W'(s_{1}, s_{2}) = (\lambda_{\hat{\Psi}_0}(s_{1}, s_{2}), \lambda_{\hat{\Psi}_1}(s_{1}, s_{2}), \lambda_{\hat{\Psi}_2}(s_{1},
  s_{2}), W(s_{1}, s_{2})),$$ where
  \begin{enumerate}[(i)]
  \item $\lambda_{\hat{\Psi}_0}(s_{1}, s_{2})$ corresponds to the violation of the avoiding
    collision rule and is defined as $\lambda_{\hat{\Psi}_0}(s_{1}, s_{2}) = 0$ if
    $\mathsf{collision} \not\in L(s_{1})$ (i.e., the autonomous vehicle does not collide with
    another vehicle or obstacle at state $s_{1}$) and
    $\lambda_{\hat{\Psi}_0}(s_{1}, s_{2}) = W(s_{1}, s_{2})$ otherwise,
  \item $\lambda_{\hat{\Psi}_1}(s_{1}, s_{2})$ corresponds to the violation of the staying on the road
    rule and is defined as $\lambda_{\hat{\Psi}_1}(s_{1}, s_{2}) = 0$ if
    $\mathsf{road} \in L(s_{1})$ (i.e., the vehicle is fully on the road at state $s_{1}$) and
    $\lambda_{\hat{\Psi}_0}(s_{1}, s_{2}) = W(s_{1}, s_{2})$ otherwise, and
  \item $\lambda_{\hat{\Psi}_2}(s_{1}, s_{2}) = \lambda_{\spec_{2}}(s_{1}, s_{2}) +
    \lambda_{\spec_{3}}(s_{1}, s_{2})$, where
    \begin{itemize}
    \item $\lambda_{\spec_{2}}(s_{1}, s_{2})$ corresponds to the violation of the clearance rule and
      is defined as $\lambda_{\spec_{2}}(s_{1}, s_{2}) = 0$ if $\mathsf{close} \not\in L(s_{1})$
      (i.e., the autonomous vehicle does not overlap with the clearance zone around another vehicle
      at state $s_{1}$) and $\lambda_{\spec_{2}}(s_{1}, s_{2}) = W(s_{1}, s_{2})$ otherwise, and
    \item $\lambda_{\spec_{3}}(s_{1}, s_{2})$ corresponds to the violation of the lane rule and
      is defined as $\lambda_{\spec_{3}}(s_{1}, s_{2}) = 0$ if $\mathsf{inlane} \in L(s_{1})$
      (i.e., the autonomous vehicle is fully within a correct lane at state $s_{1}$) and
      $\lambda_{\spec_{3}}(s_{1}, s_{2}) = W(s_{1}, s_{2})$ otherwise.
    \end{itemize}
  \end{enumerate}
\end{example}

\citet{Wongpiromsarn:2021:Minimum} show that the minimum-violation synthesis problem is equivalent to
computing the shortest path (based on the standard lexicographical ordering) on $TS'$, which has the
same size as $TS$.
As a result, the construction of $TS'$ allows temporal logic specifications to be handled with
the same computational complexity as traditional graph-search algorithms
such as Dijkstra and A*.

\chapter{Reactive Synthesis}
\glsresetall

This chapter continues the discussion of correct-by-construction synthesis of control protocols focusing on non-deterministic systems. In particular, we consider open systems whose behaviors can be affected by exogenous inputs.
Non-determinism can be used to capture uncertainties in the system, and is particularly useful in capturing uncertainties arising from valid environment behaviors that the system cannot control.
Such a system is called reactive as it must react to the environment's behavior.
Similarly to the previous chapter, we focus on discrete systems modeled as finite transition systems and we assume that the system's state is observable at all times.
This chapter puts emphasis on methods that alleviate some of the difficulties -- such as computational complexity and conflicting specifications -- that naturally arise when constructing autonomous protocols.

\blfootnote{This section incorporates the results from the following publications~\citep*{Wongpiromsarn:2013:Synthesis,Kress-Gazit:2011:Correct,dimitrova2018maximum}.}

\section{Synthesis of Reactive Control Protocol}
\label{sec:reactive-synthesis}

\begin{figure}
  \centering
  \input{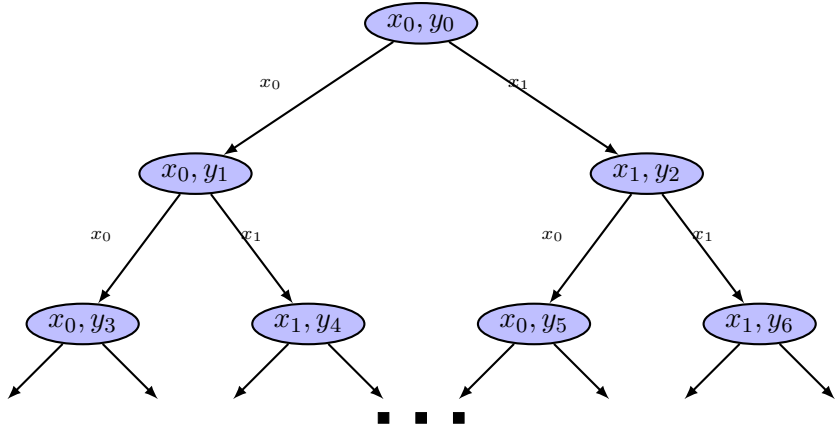}
  \caption{A tree representing a control protocol for non-deterministic systems.
    The state of the system is a tuple $(x, y)$ where $x \in \{x_0, x_1\}$
    represents the non-deterministic choice of the environment behavior whereas
    $y \in \{y_0, y_1, \ldots\}$ represents the state over which the system has control.}
  \label{fig:control_tree}
\end{figure}

Similarly to as for closed systems, we say that a reactive system is \emph{correct} with respect to specification $\varphi$ if it satisfies $\varphi$.
However, for a non-deterministic system, the correctness needs to be interpreted with respect to the non-deterministic choices over which the system does not have control.
In this case, we require that the control protocol ensures that specification $\varphi$ is satisfied for all possible non-deterministic choices, for example, for all of the environment's possible behaviors.
As discussed in \citet{Pnueli:1989:SRM,piterman06},
the control protocol synthesis in this case can be treated as a two-player game between the system and the environment, also called the adversary.
The system and the environment alternate in picking actions.
The environment then attempts to falsify $\varphi$ while the system attempts to satisfy $\varphi$.
A provably correct control protocol therefore needs to ensure that $\varphi$ is satisfied for any possible behavior from the environment.
The control protocol may thus be represented by a tree whose branches represent the possible environment actions and capture non-determinism
as shown in \figurename~\ref{fig:control_tree}.

Solving the above two-player game typically involves computing the winning set, which is defined as the set of initial states from which there exists a strategy for the system to satisfy the specification,
for all the possible environment behaviors.
Similar to model-checking-based policy synthesis, computing a winning set requires computing the product of the transition system and a finite automata, except for that a non-deterministic B\"{u}chi automaton $\mathcal{A}_\varphi$ recognizing $\varphi$ must be transformed into a deterministic Rabin automaton $\mathcal{R}_\varphi$ \citep{BK08}. We call such a transformation \textit{determinization}.
We then compute the product transition system $TS_p = TS \otimes \mathcal{R}_\varphi$ which is defined similarly to as in Definition \ref{def:product}.
A fixed-point strategy can be applied to $TS_p$ in order to derive the winning set.
Finally, a control protocol can be constructed using the intermediate values in the computation of the winning set.

The size of \(TS_p\) and the runtime of this synthesis algorithm are both at most double exponential in the length of \(\varphi\).
The first exponent results from the construction of the non-deterministic B\"{u}chi automaton $\mathcal{A}_\varphi$ from $\varphi$ and
the second exponent results from  the determinization of $\mathcal{A}_\varphi$ into a deterministic Rabin automaton $\mathcal{R}_\varphi$.
We refer the reader to \citep{Pnueli:1989:SRM,Kloetzer:HSCC2008} for more details.

For a class of specifications of the form $\always p$, $\eventually p$, $\always \eventually p$ and $\eventually \always p$, where
$p$ is a proposition, there exist efficient synthesis algorithms \citep{Asarin98:IFAC,Alur04:DGG}.
The main idea behind these algorithms is that they avoid translating the specification to a non-deterministic B\"{u}chi automaton
as well as avoiding the determinization of the non-deterministic B\"{u}chi automaton into a deterministic Rabin automaton.
For example, in a  reachability game with specification $\varphi = \eventually p$,
we define the set $W = \{s \in S : s \models p\}$ to be the set of states at which $p$ is satisfied,
and we define the predecessor operator
$Pre_{\exists\forall} \mid 2^S \to 2^S$ as follows: $Pre_{\exists\forall}(R)$ is the set of states
whose subsequent successors have at least one successor in $R$. In particular,
$Pre_{\exists\forall}(R) = \{s \in S \mid \forall s' \in S, s \to s' \hbox{ implies } \exists s'' \in R \hbox{ such that } s' \to s''\}$.
The set of states from which the controller can lead the system into $W$ can be computed efficiently by the iteration sequence
\begin{eqnarray*}
  R_0 &=& W,\\
  R_i &=& R_{i-1} \cup Pre_{\exists\forall}(R_{i-1}), \forall i > 0.
\end{eqnarray*}
Using the Tarski-Knaster Theorem, it can be shown that there exists a natural number $n$ such that $R_n = R_{n-1}$.
In addition, $R_n$ is the minimal solution of the fix-point equation $R = W \cup Pre_{\exists\forall}(R)$.

The methodology proposed by \citet{piterman06} allows for solving a broader class of games efficiently.
A summary of the algorithm is as follows:
First, a \emph{game structure} is defined as a tuple $\game = ( V, X, Y, \theta_e, \theta_s$, $\rho_e, \rho_s, AP, L, \varphi )$ where
\begin{itemize}
\item $V$ is a finite set of variables over finite domains,
\item $X \subseteq V$ is a set of environment variables,
\item $Y = V \setminus X$ is a set of controlled variables,
\item $\theta_e(X)$ is a proposition over $X$ characterizing the initial states of the environment,
\item $\theta_s(V)$ is a proposition over $V$ characterizing the initial states of the system,
\item $\rho_e(V,X')$ is a proposition that relates a state $s \in \dom(V)$ to a possible next input value $s_X \in \dom(X)$ and characterizes the transition relation of the environment,
\item $\rho_s (V , X' , Y' )$ is a proposition that relates a state $s \in \dom(V)$ and an input value $s_X \in \dom(X)$ to an output value $s_Y \in \dom(Y)$ and characterizes the transition relation of the system,
\item $AP$ is a set of atomic propositions,
\item $L : \dom(V) \to 2^{AP}$ is a labeling function, and
\item $\varphi$ is the winning condition, characterized by a \gls{ltl} formula.
\end{itemize}
We let $\dom(V)$, $\dom(X)$ and $\dom(Y)$ denote the set of all the possible assignments to variables in $V$, $X$ and $Y$,
respectively.
An environment state $s_X \in \dom(X)$ is a valid input in state $s \in \dom(V)$ if $(s, s_X ) \models \rho_e$. Analogously,
a controlled state $s_Y \in \dom(Y)$ is a valid system output at state $s \in \dom(V)$, after reading input $s_X$, if $(s,s_X,s_Y) \models \rho_s$.

\begin{example}
  \label{ex:unprotected_left_reactive}
  \begin{figure}
    \centering
    \includegraphics[width=0.3\textwidth]{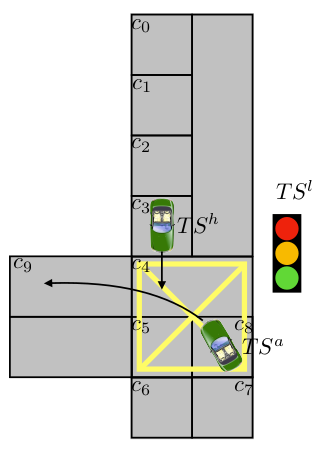}
    \caption{An unprotected left turn scenario.
      As described in Section \ref{sec:case_study_left_turn}, the autonomous vehicle $TS^{a}$
      needs to make an unprotected left turn with an oncoming vehicle $TS^{h}$. $TS^{l}$ denotes the
      traffic light.}
    \label{fig:unprotected_left_reactive}
  \end{figure}

  \begin{figure}
    \centering
    \input{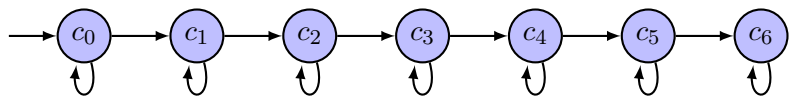}
    \caption{The finite transition system that represents the nondeterministic model of the oncoming vehicle.}
    \label{fig:unprotected_left_nondet_model}
  \end{figure}

  Let us revisit the unprotected left turn scenario described in Section
  \ref{sec:case_study_left_turn} and illustrated in Figure \ref{fig:unprotected_left_reactive}.
  In practice, the behavior of the oncoming vehicle and the traffic light may not be known exactly.
  For example, in one time step, the oncoming vehicle may stay in the same cell or move to the next
  cell.
  Figure \ref{fig:unprotected_left_nondet_model} shows a nondeterministic model corresponding to such
  behaviors.

  We consider the following requirements:
  \begin{enumerate}
  \item The autonomous vehicle should eventually go to cell $c_{9}$.
  \item The autonomous vehicle should not collide with the oncoming vehicle.
  \end{enumerate}
  The corresponding game structure $\game = ( V, X, Y, \theta_e, \theta_s, \rho_e,$ $\rho_s, AP, L, \varphi )$
  is defined as follows:
  \begin{itemize}
  \item $X = \{x_{h}\}$, where $x_{h} \in \{0, \ldots, 9\}$ is the label of the cell occupied by the oncoming vehicle.
  \item $Y = \{x_{a}\}$, where $x_{a} \in \{0, \ldots, 9\}$ is the label of the cell occupied by the autonomous vehicle.
  \item $\theta_e = (x_{h} = 0 \oor x_{h} = 1 \oor x_{h} = 2)$ indicates
    that the oncoming vehicle can start from cell $c_{0}$, $c_{1}$ or $c_{2}$.
  \item $\theta_s = (x_{a} = 4 \oor x_{a} = 7 \oor x_{a} = 8 \oor x_{a} = 9)$
    indicates that the autonomous vehicle can start from cell $c_{4}$, $c_{7}$, $c_{8}$ or $c_{9}$.
  \item $\rho_e = \bigwedge_{i=0}^{5}\big(x_{h} = i \implies (\nextop x_{h} = i \oor \nextop x_{h} =
    i+1)\big) \ \aand \ \big(x_{h} = 6 \implies (\nextop x_{h} = 6)\big)$ indicates that the
    oncoming vehicle may stay in the same cell or move to the next cell.
  \item $\rho_s = \rho_{s}^{trans} \ \aand \rho_{s}^{safe}$.
    Here, $\rho_{s}^{tran} =
    \big(x_{a} = 7 \implies (\nextop x_{a} = 7 \oor \nextop x_{a} = 8)\big) \ \aand \
    \big(x_{a} = 8 \implies (\nextop x_{a} = 8 \oor \nextop x_{a} = 4)\big) \ \aand \
    \big(x_{a} = 4 \implies (\nextop x_{a} = 4 \oor \nextop x_{a} = 9)\big) \ \aand \
    \big(x_{a} = 9 \implies (\nextop x_{a} = 9\big)$ indicates that the
    autonomous  vehicle may stay in the same cell or move to the next cell, and
    $\rho_{s}^{safe} = \neg (x_{a} = 4 \ \aand \ x_{h} = 4)$ indicates that the autonomous vehicle and
    the oncoming vehicle do not simultaneously occupy cell $c_{4}$.
  \item $\varphi = \varphi_{e} \implies \always\eventually (x_{a} = 9)$ indicating that the
    autonomous vehicle is at cell $c_{9}$ infinitely often.
    Here $\varphi_{e}$ represents the assumption on the environment, which we will revisit in
    Example \ref{ex:unprotected_left_reactive_synth}.
    Note that from the transition system representing the autonomous vehicle, once the vehicle reaches
    $c_{9}$, it will stay at $c_{9}$ forever.
    As a result, $\eventually (x_{a} = 9)$ and $\always\eventually (x_{a} = 9)$ are equivalent winning conditions.
  \end{itemize}

\end{example}

A game is played as follows:
The environment initially chooses an assignment $s_X \in \dom(X)$ such that $s_X \models \theta_e$, and the system chooses an assignment $s_Y \in \dom(Y)$ such that $(s_X , s_Y ) \models \theta_e \aand \theta_s$. %
From a state $s$, the environment chooses an input $s_X \in \dom(X)$ such that $(s, s_X ) \models \rho_e$ and the system chooses an output $s_Y \in \dom(Y)$ such that $(s,s_X,s_Y) \models \rho_s$.
Formally, we define a \emph{play} as a maximal sequence of states $\sigma = s_0 s_1 \ldots$ such that $s_0 \models \theta_e \aand \theta_s$ and, for every $j \geq 0$, $(s_j, s_{j+1}) \models \rho_e \aand \rho_s$.

A \emph{finite memory control protocol} for the system can be identified with a partial function
$f: M \times \dom(V) \times \dom(X) \to M \times \dom(Y)$ such that, for every $s \in \dom(V)$, every $s_X \in \dom(X)$, and every $m \in M$, if $(s, s_X) \models \rho_e$ and $f(m,s,s_X) = (m',s_Y)$, then $(s,s_X,s_Y) \models \rho_s$.
Here, M is some memory domain with a designated initial value $m_0 \in M$.

Protocol $f$ is winning for the system starting from state $s_0$ if any play $\sigma = s_0 s_1 \ldots$ such that, for all $i\ge 0$, $f(m_i, s_i, s_{i+1}|_X) = (m_{i+1}, s_{i+1}|_Y)$, either (i) is infinite and satisfies $\varphi$, or (ii) is finite and there is no assignment $s_X \in \dom(X)$ such that $(s_n, s_X ) \models \rho_e$, where $s_n$ is the last state in $\sigma$.
We let $Win_s$ denote a proposition characterizing the set of states starting from which there exists a winning strategy for the system.
A game structure is \emph{winning for the system} if, for all $s_X \in \dom(X)$ such that $s_X \models \theta_e$,
there exists $s_Y \in \dom(Y)$ such that $(s_X, s_Y) \models \theta_s$ and $(s_X, s_Y) \in Win_s$.

For certain \gls{ltl} specifications, $\mu$-calculus
over game structures can be employed to characterize the set of winning states of the system.
The description of $\mu$-calculus, however, is beyond the scope of this paper and
we refer the reader to \citep{Kozen83,piterman06}.
As an example, the $\mu$-calculus formula $\mu R (p \oor \cox R)$
characterizes the set of states from which the system can force the game to eventually visit
$p$-states, i.e., states that satisfy proposition $p$.
This formula thus provides the solution for the reachability game previously discussed.
Here, $\mu$ is the least fixpoint operator in $\mu$-calculus,
$R$ is known as a ``relational variable'' and
the operator $\cox$ is defined roughly similar to the predecessor operator $Pre_{\exists\forall}$.

\citet{piterman06} consider a broader class of \gls{ltl} formula known as
\emph{\gls{gr[1]}} which covers \gls{ltl} formulas of the form
\begin{equation}
  \label{eq:gr1}
  \varphi = (\always \eventually p_1 \aand \ldots \aand \always \eventually p_m)
  \implies (\always \eventually q_1 \aand \ldots \aand \always \eventually q_n).
\end{equation}
Roughly, the left hand side of $\implies$ specifies the assumption on the environment behavior
whereas the right hand side of $\implies$ specifies the desired property of the system.
\citet{piterman06} show that there exists a $\mu$-calculus formula that characterizes
the set of winning states of the system for \gls{gr[1]} winning conditions.
The formulation allows for
the synthesis problem to be solved based on fixpoint computation in time proportional to $nm|\dom(V)|^3$,
where $|\dom(V)|$ is the size of the state space.
The proposed synthesis procedure has been implemented in JTLV \citep{piterman06}
and in TuLiP \citep{Wongpiromsarn:2011:TuLiP}.
We refer the reader to \citep{piterman06} for more details,
including a discussion on the expressiveness of \gls{gr[1]} and
an extension to handle formulas of the form
$\varphi_e \implies \varphi_s$ where
$\varphi_e$ and $\varphi_s$ are any \gls{ltl} formulas that can be represented by a deterministic B\"{u}chi automaton.
A deterministic B\"{u}chi automaton is defined as a non-deterministic B\"{u}chi automaton with additional constraints that
$|Q_0| \leq 1$ and for any $q \in Q$ and $\sigma \in \Sigma$, $(q, \sigma, q') \in \delta$ and $(q, \sigma, q'') \in \delta$
imply that $q' = q''$.
\gls{ltl} formulas that can be represented by a deterministic Buchi automaton include
those of the form $\always (p_1 \implies \eventually p_2)$
where $p_1$ and $p_2$ are propositions.

\begin{example}
  \label{ex:unprotected_left_reactive_synth}
  Consider the unprotected left turn scenario described in Example \ref{ex:unprotected_left_reactive}.
  Without any assumption on the oncoming vehicle (i.e., $\varphi_{e} = \true$),
  the game structure $\game$ defined in Example \ref{ex:unprotected_left_reactive} is not winning
  for the system.
  To see this, consider the case where the oncoming vehicle starts at $c_{2}$ while the autonomous
  vehicle starts at $c_{7}$.
  In two steps, the oncoming vehicle can reach $c_{4}$ and stay there forever,
  blocking the autonomous vehicle from entering $c_{4}$ without violating the safety requirement.
  A sufficient assumption to ensure that the game structure will be winning is that the oncoming
  vehicle visits cell $c_{6}$ infinitely often, i.e., $\varphi_{e} = \always\eventually (x_{h} = 6)$.
  With this assumption, a possible control protocol for the autonomous vehicle is to wait until the
  oncoming vehicle reaches $c_{5}$ or $c_{6}$ before entering $c_{4}$.

\end{example}

\begin{remark}
  As demonstrated in Example \ref{ex:unprotected_left_reactive_synth}, the system (e.g., the
  autonomous vehicle) has no control over the environment (e.g., the oncoming vehicle).
  As a result, the synthesis algorithm needs to take into account all the possible values of the
  environment variables (e.g., $x_{h}$) and ensures that the resulting behavior of the system and
  the environment satisfies the specification $\varphi$.
  In many cases, in order to obtain a solution, one needs to limit the power of the environment,
  which is captured by $\theta_e$, $\rho_e$, and $\varphi_{e}$ in the proposed model.
\end{remark}

\section{Receding Horizon Temporal Logic Planning}
\label{sec:rhtlp}
The main limitation of the discrete synthesis described in Section \ref{sec:logic} and Section \ref{sec:reactive-synthesis} is the state explosion problem.
In the worst case, the entire system's state space has to be taken into account.
For example, if the system has $|V|$ variables, each can take any of the $P$ possible values.
Then, we must consider as many as $P^{|V|}$ states.
This type of computational complexity limits the application of systhesis to relatively small
problems.

Similar computational complexity is also encountered in the area of constrained optimal control.
In the controls domain, an effective and well-established technique to address this issue is to
design and implement control strategies in a receding horizon manner, i.e., optimize over a {\em
  shorter\/} horizon, starting from the currently observed state, implement the initial control
action, move the horizon one step ahead, and re-optimize.
This approach reduces the computational complexity by essentially solving a sequence of {\em
  smaller\/} optimization problems, each with a specific initial condition (as opposed to optimizing
with {\em any\/} initial condition in traditional optimal control).
Under certain conditions, receding horizon control strategies are known to lead to closed-loop stability
\citep{Murray02onlinecontrol, Mayne00MPC, Jadbabaie00thesis}.
See, e.g., \citep{goodwin04} for a detailed discussion on constrained optimal control, including
finite horizon optimal control and receding horizon control.

\begin{figure}
  \centering
  \includegraphics[width=0.48\textwidth]{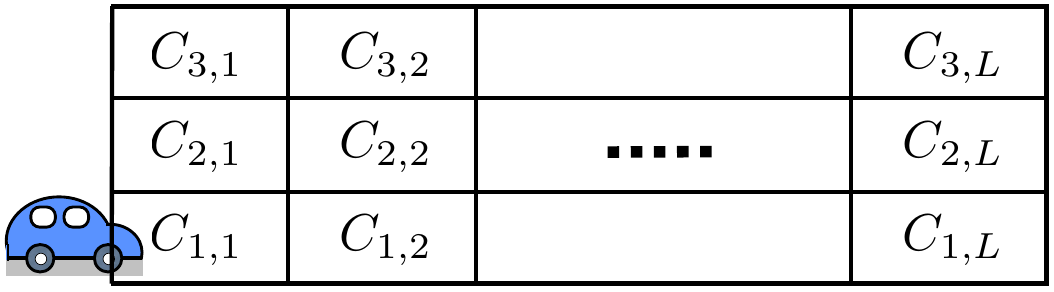}
  \caption{The autonomous driving example where the road is partitioned into $3L$ cells where $L$ is the length of the road.}
  \label{fig:aut_car_ex}
\end{figure}

To partially alleviate the state explosion problem in the synthesis of finite state automata,
\citet{wongpiromsarn10hscc} and \citet{wongpiromsarn12tac} consider reactive module synthesis with
\gls{gr[1]} specifications and show that for systems with a certain structure, the synthesis problem
can be solved in a receding horizon fashion, i.e., compute the plan or strategy over a ``shorter'' horizon,
starting from the current state, implement the initial portion of the plan, move the horizon one step ahead, and recompute.
This approach essentially reduces the discrete control protocol synthesis problem into a set of smaller problems.
The size of these smaller problems depends on the horizon length.
For example, consider the autonomous driving problem where an autonomous vehicle
needs to navigate the road shown in \figurename~\ref{fig:aut_car_ex} starting from cell $C_{1,1}$ and
with destination $C_{1,L} \cup C_{2,L} \cup C_{3,L}$.
Suppose the horizon length is $l$, i.e., the vehicle plans for $l$ cells ahead.
Then, the state space for each short-horizon problem contains at most $3l2^{3l}$ states (whereas the size of the original problem is $3L2^{3L}$).
Hence, the horizon length should be made as small as possible, subject to the realizability of the resulting short-horizon specifications; horizons that are too short typically render the specifications unrealizable.

Sufficient conditions that ensure that this receding horizon implementation preserves the desired system-level properties are presented
in \citep{wongpiromsarn10hscc,wongpiromsarn12tac}.
For the simplicity of the presentation, in this article, we consider the case where the specification is given by
\begin{equation}
  \varphi = (\varphi_{init} \aand \varphi_{env}) \implies (\varphi_{safety} \aand \varphi_{goal}),
\end{equation}
where $\varphi_{init}$ is a proposition characterizing the set of initial states,
$\varphi_{env}$ is an \gls{ltl} formula characterizing the assumption on the environment behavior and
can be written as the conjunction of a safety formula and
the progress formulas on the left hand side of $\implies$ in (\ref{eq:gr1}),
$\varphi_{safety}$ is a safety formula and
$\varphi_{goal}$ is of the form $\varphi_{goal} = \always\eventually q$ where $q$ is a proposition
characterizing the set of some goal states to be visited infinitely often.

The receding horizon approach works as follows.
First, we organize the discrete state space into a partially ordered set $(\{\mathcal{W}_0, \ldots, \mathcal{W}_M\}, \prec_\varphi)$
such that $\mathcal{W}_0$ only contains the goal states and
$\mathcal{W}_0 \prec_\varphi \mathcal{W}_i$ for all $i \not= 0$.
The partial order relation $\prec_{\varphi}$ can be defined based on the notion of ``distance'' to
the goal states.

Next, we define a map $\mathcal{F} : \{\mathcal{W}_0, \ldots, \mathcal{W}_M\} \to \{\mathcal{W}_0,
\ldots, \mathcal{W}_M\}$ that captures the horizon length and satisfies
$\mathcal{F}(\mathcal{W}_i) \prec_\varphi \mathcal{W}_i$ for all $i \not= 0$.
Finally, we specify a proposition $\Phi$ that characterizes the receding horizon invariant
such that any state that satisfies $\varphi_{init}$ also satisfies $\Phi$.
In other words, $\varphi_{init} \implies \Phi$ is a tautology.

With the partially ordered set $(\{\mathcal{W}_0, \ldots, \mathcal{W}_M\}, \prec_\varphi)$,
the map $\mathcal{F}$, and the receding horizon invariant $\Phi$,
we define a short-horizon specification $\Psi_i$ associated with
each $\mathcal{W}_i$, for $i \in \{0, \ldots, M\}$ as
\begin{equation}
  \label{eq:short-horizon-spec}
  \Psi_i = \big( (\nu \in \mathcal{W}_i) \aand \Phi \aand \varphi_{env} \big) \implies
  \big( \always \Phi \aand \varphi_{safety} \aand \eventually(\nu \in \mathcal{F}(\mathcal{W}_i)) \big)
\end{equation}
where $\nu$ is the state of the system.
The left-hand side of $\implies$ states that
(a) the initial state is assumed to be in $\mathcal{W}_i$ and satisfies $\Phi$, and
(b) the environment is assumed to satisfy the assumptions $\varphi_{env}$ stated in the original
specification.
The right-hand side of $\implies$ then specifies that
(a) $\Phi$ holds throughout an execution,
(b) the original safety properties $\varphi_{safety}$ are satisfied, and
(c) the system eventually reaches a state in $\mathcal{F}(\mathcal{W}_i)$.

According to (\ref{eq:short-horizon-spec}), $\mathcal{F}(\mathcal{W}_i)$ essentially defines an
intermediate goal for states in $\mathcal{W}_i$.
In addition, $\Phi$ is introduced to ensure that
a provably correct plan exists when the system reaches the end of the current horizon and needs to compute a new plan.
We refer the reader to \citep{wongpiromsarn12tac} for a detailed discussion on this receding horizon framework,
including an extension to the case where there are multiple goals that may be visited in an arbitrary order.

\begin{figure}
  \centering
  {\input{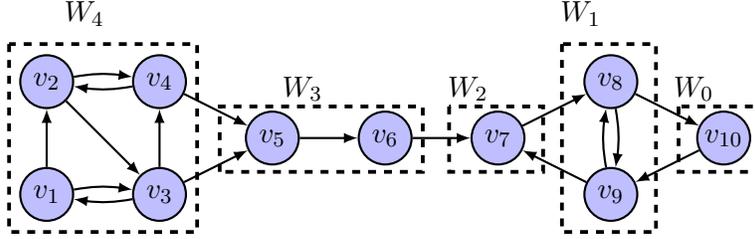}}
  \caption{A graphical description of the receding horizon framework for a special case where there is only one goal $\nu_{10}$. $\nu_1, \ldots, \nu_{10}$ are the discrete states.}
  \label{fig:RHTLP_visualization}
\end{figure}

Consider a simple example shown in \figurename~\ref{fig:RHTLP_visualization} where $\nu_{10}$ is the goal state.
The partial order may be defined as $\mathcal{W}_0 \prec_\varphi \mathcal{W}_1 \prec_\varphi \ldots \prec_\varphi \mathcal{W}_4$
and the map $\mathcal{F}$ may be defined as $\mathcal{F}(\mathcal{W}_j) = \mathcal{W}_{j-2}$, for all $j \geq 2$
and $\mathcal{F}(\mathcal{W}_1) = \mathcal{F}(\mathcal{W}_0) = \mathcal{W}_0$.
The key idea of the receding horizon framework is to synthesize a control protocol for short-horizon specification $\Psi_4$,
which corresponds to going from $\nu_1$ only to a state in $\mathcal{F}(\mathcal{W}_4) = \mathcal{W}_{2}$,
rather than synthesizing a control protocol for going from the initial state $\nu_1$ to the goal state $\nu_{10}$ in one shot,
taking into account all the possible behavior of the environment.
Once a state in $\mathcal{W}_3$, i.e., $\nu_5$ or $\nu_6$ is reached,
we then recompute a protocol for the short-horizon specification $\Psi_3$ for
going to a state in $\mathcal{F}(\mathcal{W}_3) = \mathcal{W}_{1}$.
This process is then continually repeated.
From the finiteness of the set $\{\mathcal{W}_0, \ldots, \mathcal{W}_M\}$ and its partial order,
it can be shown that this receding horizon implementation of the short-horizon strategies ensures the correctness of the global specification,
provided that all of the short horizon specifications $\Psi_i$, for $i \in \{0, \ldots M\}$ are realizable \citep{wongpiromsarn12tac}.
In this case, the invariant $\Phi$ is introduced to rule out the states that render the short horizon problems unrealizable.

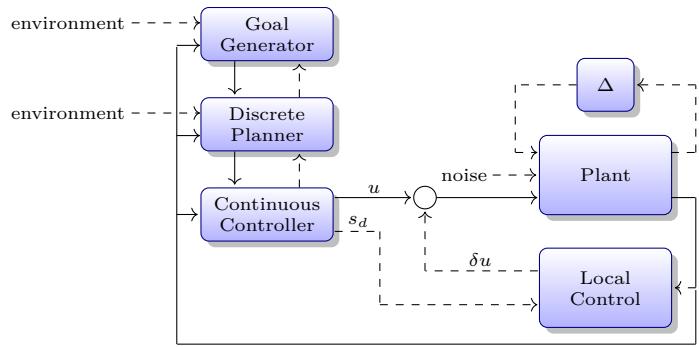
\begin{figure}
  \centering
  \begin{tikzpicture}[scale=0.3]
    \tikzstyle{every node}=[ text centered, font=\scriptsize, top color=white, bottom color=blue!30,
    draw=blue!50!black!100, drop shadow, minimum height = 20pt, rounded corners=3pt, text width=1.5cm]
    \draw(0, 7.25) node [rectangle] (goal) {Goal Generator};
    \draw(0,3.25) node [rectangle] (traj) {Discrete Planner};
    \draw(0,-0.75) node [rectangle] (planner) {Continuous Controller};
    \draw (15,-4) node [rectangle, minimum height = 30pt] (contr) {Local Control};
    \draw (15,1) node [rectangle, minimum height = 30pt] (plant) {Plant};
    \draw (15,5) node [rectangle, text width=0.5cm] (delta) {$\Delta$};
    \filldraw[fill=white] (7,0) circle (0.5);
    \tikzstyle{every node}=[font=\scriptsize,fill=none]
    \draw [ shorten >= 1pt, -, dashed ] (delta) to (10.9,5);
    \draw [ shorten >= 1pt, -, dashed ] (11,5) to (11,1.9);
    \draw [ shorten >= 1pt, ->, dashed ] (11,2) to (12.1,2);

    \draw [ shorten >= 1pt, -, dashed ] (17.9,2) to (19.1,2);
    \draw [ shorten >= 1pt, -, dashed ] (19,2) to (19,5.1);
    \draw [ shorten >= 1pt, ->, dashed ] (19,5) to (delta);

    \draw [ shorten >= 1pt, ->, dashed ] (10,1) to node[above=1pt, left=7pt] {noise} (plant);

    \draw [ shorten >= 1pt, -> ] (7.5,0) to (12.1,0);

    \draw [ shorten >= 1pt, - ] (17.9,0) to (19.1,0);
    \draw [ shorten >= 1pt, - ] (19,0) to (19,-4.1);
    \draw [ shorten >= 1pt, ->, dashed ] (19,-4) to (contr);
    \draw [ shorten >= 1pt, - ] (19,-4.1) to (19,-6.6);
    \draw [ shorten >= 1pt, - ] (19,-6.5) to node[above=4pt, left=-20pt] {} (-4.1,-6.5);
    \draw [ shorten >= 1pt, - ] (-4,-6.5) to (-4,6.8);
    \draw [ shorten >= 1pt, -> ] (-4,2.75) to (-2.85, 2.75);
    \draw [ shorten >= 1pt, -> ] (-4,-0.75) to (planner);
    \draw [ shorten >= 1pt, -> ] (-4,6.75) to (-2.85, 6.75);

    \draw [ shorten >= 1pt, ->, dashed ] (-6,3.75) to node[above=1pt, left=12pt] {environment} (-2.85, 3.75);
    \draw [ shorten >= 1pt, ->, dashed ] (-6,7.75) to node[above=1pt, left=12pt] {environment} (-2.85, 7.75);

    \draw [ shorten >= 1pt, -> ] (traj.220) to node[left=-2pt] {} (planner.140);
    \draw [ shorten >= 1pt, ->, dashed ] (planner.040) to node[right=-2pt] {} (traj.320);
    \draw [ shorten >= 1pt, -> ] (goal.220) to node[left=-2pt] {} (traj.140);
    \draw [ shorten >= 1pt, ->, dashed ] (traj.040) to node[right=-2pt] {} (goal.320);

    \draw [ shorten >= 1pt, -> ] (3,0) to node[above=-2pt] {$u$} (6.5,0);

    \draw [ shorten >= 1pt, -, dashed ] (12,-3.25) to node[above=-2pt] {$\delta u$} (6.9,-3.25);
    \draw [ shorten >= 1pt, ->, dashed ] (7,-3.25) to (7,-0.5);

    \draw [ shorten >= 1pt, -, dashed ] (3,-1.5) to node[above=-2pt] {$s_d$} (5.1,-1.5);
    \draw [ shorten >= 1pt, -, dashed ] (5,-1.5) to (5,-4.85);
    \draw [ shorten >= 1pt, ->, dashed ] (5,-4.75) to (12.1,-4.75);
  \end{tikzpicture}
  \caption{The hierarchical control structure with the goal generator.}
  \label{fig:hierarchical_goal}
\end{figure}

\begin{figure}
  \centering
  \input{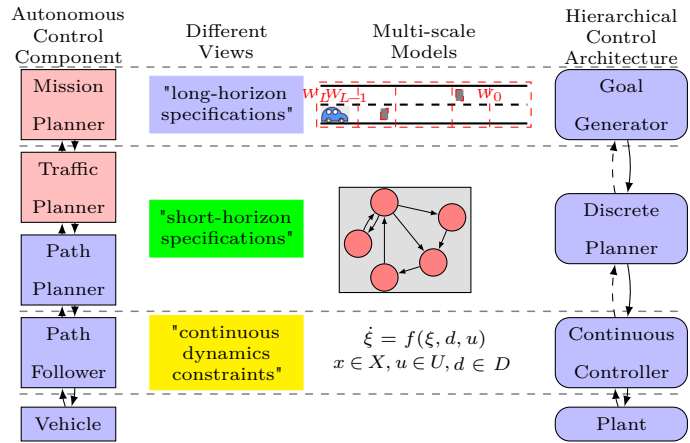}
  \caption{The hierarchical control structure with the receding horizon implementation, showing the similarity
    with the navigation protocol stack implemented on Alice.
    The goal generator has similar functionality as Mission Planner.
    It determines a sequence of intermediate goals for the discrete planner such that the original ``long-horizon'' specification is satisfied.
    Its computation relies on the graph $\mathbb{G}$ that encodes the partially ordered set
    $(\{\mathcal{W}_0, \ldots, \mathcal{W}_M\}, \prec_\varphi)$.
    The discrete planner has similar functionality as the composition of Traffic Planner and Path Planner.
    It computes a discrete plan for the system such that
    the short-horizon specification in (\ref{eq:short-horizon-spec}) with the next intermediate goal computed
    by the goal generator is satisfied based on a finite, abstract model of the physical system.
    Finally, the continuous controller deals with the continuous dynamics and constraints
    to ensure that the physical system follows the plan computed by the discrete planner.
    This functionality is similar to that of Path Follower in Alice.}
  \label{fig:receding_horizon_impl}
\end{figure}

Given a receding horizon invariant $\Phi$,
the partial order $\prec_\varphi$ as well as the horizon length defined by the map $\mathcal{F}$ can be automated
by adding an additional component, namely the \textit{goal generator} to the hierarchical control structure in \figurename~\ref{fig:hierarchical_goal}.
The goal generator works on a graph $\mathbb{G}$ with $\mathcal{W}_i, i \in \{0, \ldots M\}$ being its states.
For each $i\in \{0, \ldots M\}$ and $j \in \{0, \ldots M\}$, a transition from $\mathcal{W}_i$ to $\mathcal{W}_j$ in $\mathbb{G}$ is added if $i \not= j$ and
the short-horizon specification $\Psi_i$ is realizable with $\mathcal{F}(\mathcal{W}_i) = \mathcal{W}_j$.
After $\mathbb{G}$ is constructed, the goal generator then performs a graph search to find a path from
$\mathcal{W}_i$, to which the current state of the system belongs, to a goal state in $\mathcal{W}_0$.
This path essentially defines a sequence of intermediate goals for each short-horizon problem.
The resulting hierarchical control structure with this implementation of the receding-horizon framework is shown in
\figurename~\ref{fig:hierarchical_goal}.
\figurename~\ref{fig:receding_horizon_impl} shows the similarity of this hierarchical control structure with that implemented on Alice, the representative autonomous vehicle in Section \ref{sec:model_checking_for_Alice}, illustrating that the techniques presented in this article can be utilized to formalize and enable automatic design
of the navigation protocol stack of an autonomous system.

\begin{figure}
  \centering
  \includegraphics[width=0.3\textwidth]{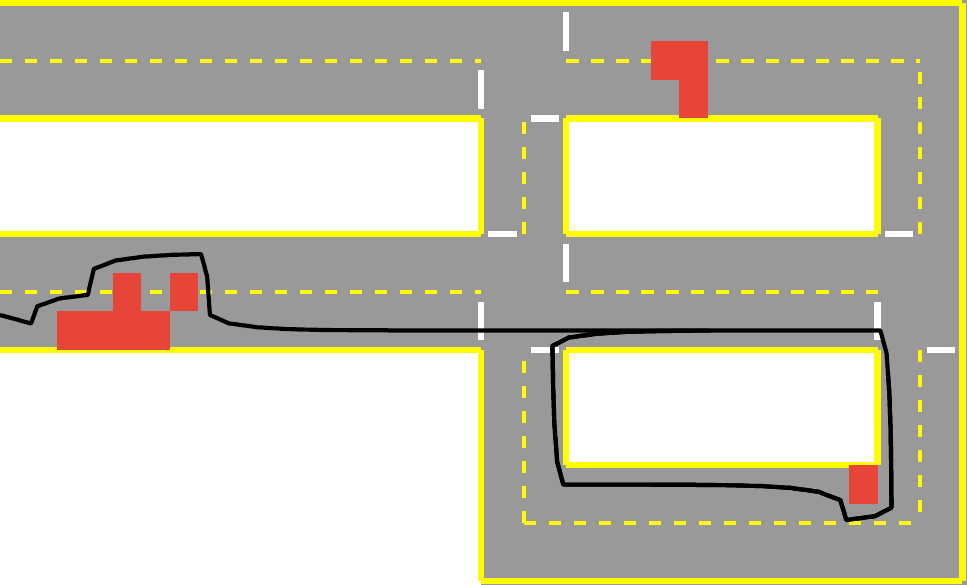}
  \\
  \vspace{2mm}
  \includegraphics[width=0.3\textwidth]{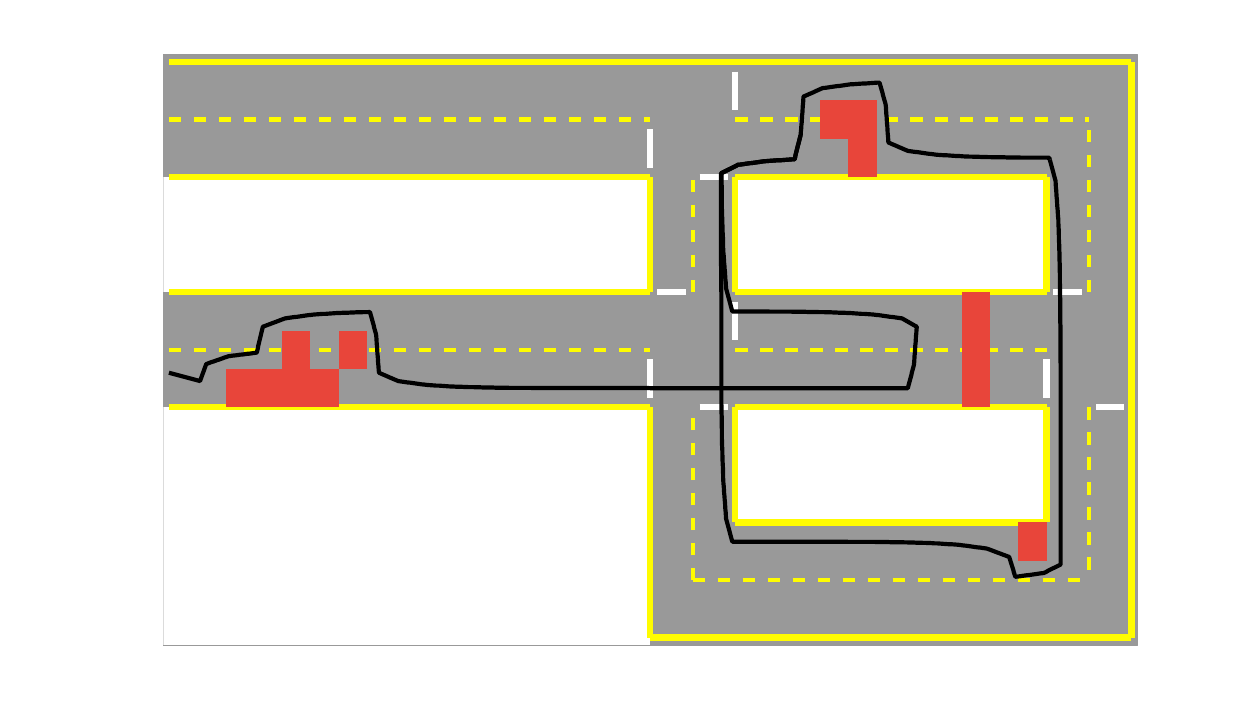}
  \caption{Simulation results with (top) no road blockage, (bottom) a road blockage on the middle road. The corresponding movies can be downloaded from \texttt{http://sourceforge.net/projects/tulip-control/}.}
  \label{pic:RHTLPSim}
\end{figure}

\noindent \textbf{Computational Complexity and Completeness:}
The receding-horizon implementation reduces the computational complexity by restricting the state space
considered in each subproblem;
however, it is not complete.
Even if the original specification is realizable, there may not exist a combination of horizon length, partial order relation, and receding horizon invariant that render all of the short horizon specifications realizable.
Nevertheless, its successful applications to autonomous driving problems have been illustrated in \citet{ wongpiromsarn10aaai,wongpiromsarn12tac}.
Examples of these applications are provided in Figure \ref{pic:RHTLPSim}.

\begin{remark}
  Computation of the horizon length, partial order relation, and receding horizon invariant requires insights for each problem domain. Automatic construction of these elements is subject to on-going research.
  \citet{wongpiromsarn12tac} describe automatic construction of certain elements, given other elements, e.g., automatic computation of the horizon length and partial order relation, given a receding horizon invariant, and automatic computation of the receding horizon invariant, given a horizon length and partial order relation.
\end{remark}

\begin{figure}
  \centering
  \includegraphics[width=0.4\textwidth]{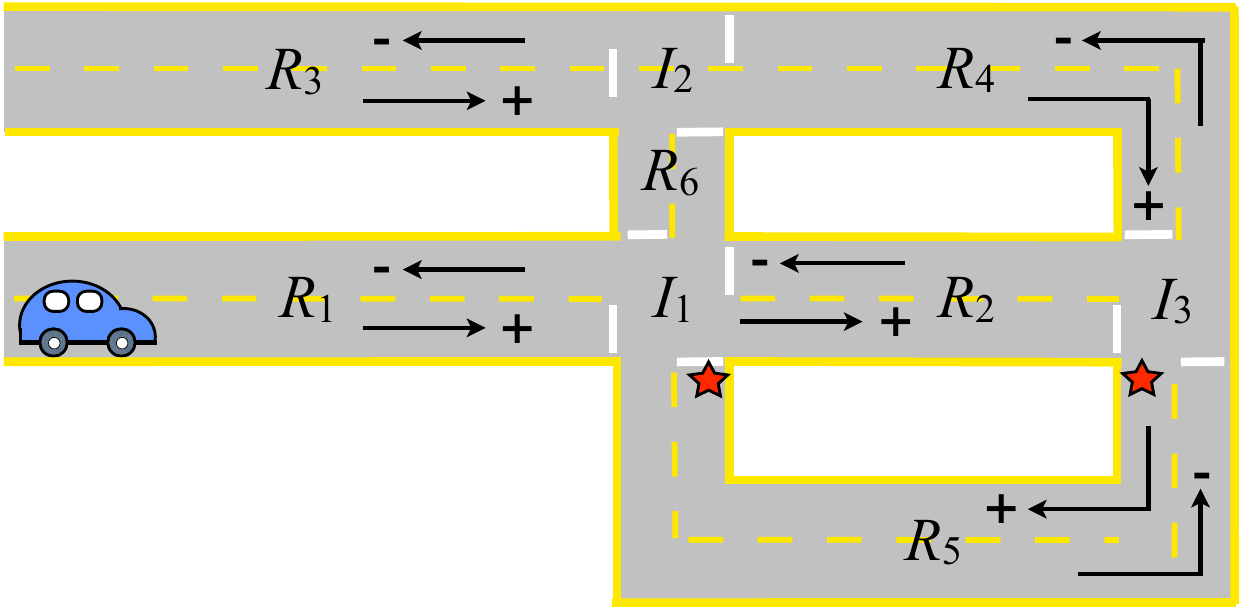}
  \\
  \vspace{2mm}
  \includegraphics[width=0.4\textwidth]{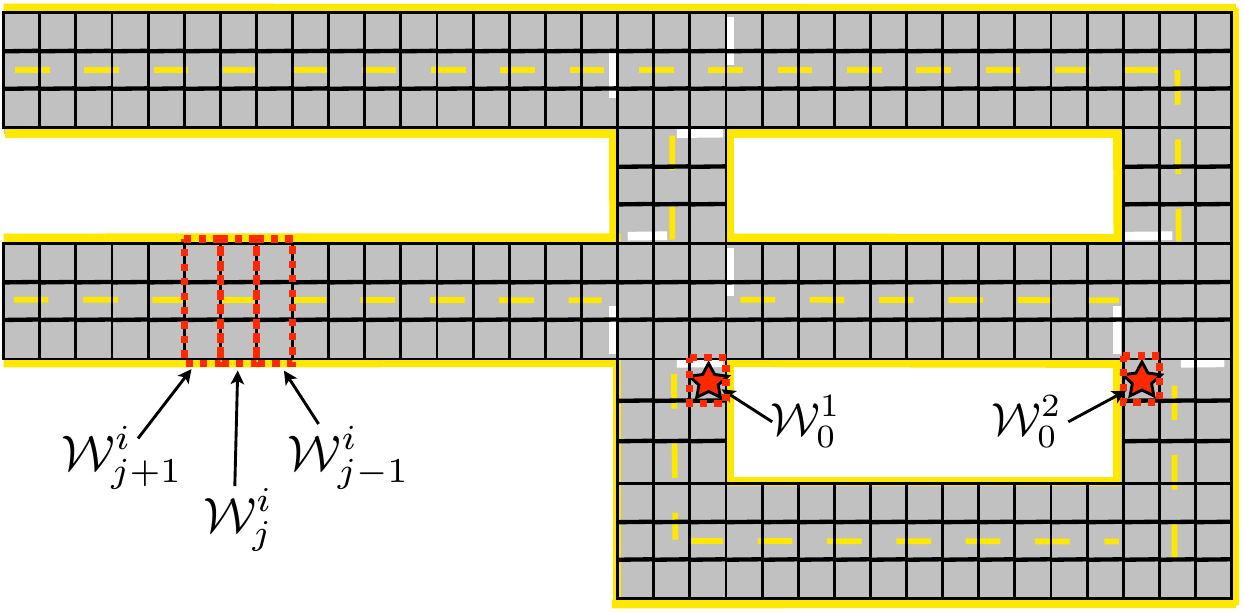}
  \caption{The road network and its partition for the autonomous vehicle example.
    The stars indicate the cells that need to be visited infinitely often.}
  \label{fig:road}
\end{figure}
\begin{example}
  Consider an autonomous driving problem in an urban-like environment.
  We consider the road network shown in \figurename~\ref{fig:road}, which is
  partitioned into $N = 282$ cells.
  Each of these cells may or may not be occupied by an obstacle.
  The desired properties include:

  \begin{itemize}
  \item Each of the two cells marked by star needs to be visited infinitely often.
  \item No collision is allowed, i.e., the vehicle cannot occupy the same cell as an obstacle.
  \item The vehicle stays in the right lane unless there is an obstacle blocking the lane.
  \item The vehicle can only proceed through an intersection when the intersection is clear.
  \end{itemize}
  \citet{wongpiromsarn12tac} show that with some mild assumptions on the environment behavior,
  there exists a receding-horizon invariant that ensures that all the short-horizon specifications are realizable with horizon length 2,
  i.e, $\mathcal{F}(\mathcal{W}_i) = \mathcal{F}(\mathcal{W}_{i-2})$.
  Hence, the size of the state space for each short-horizon problem is at most 4608
  whereas the size of the state space of the original problem is in the order of $10^{87}$.
  Roughly, the receding-horizon invariant requires that the vehicle is not surrounded by obstacles
  and if the vehicle is not in the travel lane, there must be an obstacle blocking the lane.
  Using JTLV, each short-horizon synthesis problem can be solved in approximately 1.5 seconds
  on a MacBook with a 2 GHz Intel Core 2 Duo processor and 4 Gb of memory.
  Simulation results when the receding-horizon approach is applied are shown in \figurename~\ref{pic:RHTLPSim}.
\end{example}

\section{Reactive Synthesis with Maximum Realizability}\label{sec:max}

\label{sec:max-intro}

In conventional synthesis, either an implementation is constructed for a given specification, or the specification is identified as unrealizable. Nevertheless, specifications may arise from different design perspectives, especially in large systems, and if they consist of a large number of individual requirements, it is easy to encounter specifications that are unrealizable. In other scenarios, the user may have several alternative requirements in mind, potentially with some preferences, and want to know the best realizable combination of them with respect to some metric. Such cases usually lead to alternating between specification modification and synthesis procedure and hence, defeating the purpose of facilitating the design process.

The possibility of conflict amongst the provided requirements calls for a more comprehensive synthesis procedure that, in the case of unrealizability, can generate an implementation that minimally violates the specifications. In order to define the notion of minimality, one requires a quantitative metric on the satisfaction of \gls{ltl} formulas.
The approach we pursue in this section relies on multiple levels of relaxations of an \gls{ltl} formula as well as relying on forming a value function that captures the levels of relaxations applied over the specifications. Then, the maximum realizability of a set of \gls{ltl} formulas turns into seeking an implementation that maximizes the corresponding value.

The backbone of this section's approach toward maximum realizability is bounded synthesis, originally introduced by Schewe and Finkbeiner~\citet{ScheweF07a}. Bounded synthesis tackles the computational complexity of reactive synthesis from \gls{ltl} properties by restricting the size of the search space and incrementing the bound on the size if necessary.
More specifically, it searches for a realizable implementation of the size up to a prespecified bound. If no such implementation exists, it increments the bound and repeats the search process.
Each instance of bounded search for an implementation can be encoded as a SAT (or QBF, or SMT) problem~\citep{FaymonvilleFRT17}. The algorithm is complete as a theoretical bound on the maximum size of the implementation exists.

In this section, we formulate maximum realizability as iterative \gls{maxsat} solving~\citep{biere2009handbook}. In each iteration, we construct a \gls{maxsat} instance that characterizes the existence of an implementation of size within the given bound that not only realizes the hard specification but is also optimal with respect to defined value function for soft specifications. We prove that, for any given finite set of soft specifications, there exists an optimal implementation with a bounded size. Consequently, the proposed algorithm that gradually increases the bound on the implementation size is complete.

\subsection{Related Work}\label{sec:max-rel-work}

Maximum realizability and several closely related problems have attracted significant attention in recent years. \citet{TumovaHKFR13} studied the problem of planning over a finite horizon with prioritized safety requirements, where the goal is to synthesize a least-violating control strategy. \citet{KimFS15} studied a similar problem for the case of infinite-horizon temporal logic planning. \citet{LahijanianAFKV15} describe a method for computing plans for co-safe \gls{ltl} specifications that minimize the cost of violating each atomic proposition.
These approaches are developed for the planning setting without an adversarial environment. \citet{LahijanianK16} considered the case of probabilistic environments and \citet{lahijanian2016iterative} studied the problem of partial satisfaction of guarantees in an unknown environment, maximizing the number of soft specifications that are satisfied.
\citet{Tomita2017} study a maximum realizability problem in which the specification is a conjunction of a must \gls{ltl} specification, and a number of weighted desirable \gls{ltl} specifications, formulated as a mean-payoff optimization.

Two other main research directions related to maximum realizability are \emph{quantitative synthesis} and \emph{specification debugging}.
Related to quantitative synthesis, the goal in \citet{BloemCHJ09} is to generate an implementation that maximizes the value of a mean-payoff objective, while possibly satisfying some $\omega$-regular specification, while in \citet{AlmagorBK16,TabuadaN16}, the system requirements are formalized in a multi-valued temporal logic.
\citet{AlurKW08} studied an optimal synthesis problem for an ordered sequence of prioritized $\omega$-regular properties, where the classical fixpoint-based game-solving algorithms are extended to a quantitative setting.
In specification debugging there is a lot of research dedicated to finding good explanations for the unsatisfiability or unrealizability of temporal logic specifications~\citep{CimattiRST07,Schuppan12,RamanK13}, and more generally to the analysis of specifications~\citep{CimattiRST08,EhlersR14}.

\label{sec:max-background}

We start by an overview of the required concepts to
formally state the synthesis problem.
Then, we proceed to go over definitions of run graph and annotations to describe the bounded synthesis method and its SAT encoding.
Lastly, we provide a brief description of the \gls{maxsat} problem, particularly a class of that called partial weighted \gls{maxsat}.

\subsubsection{Bounded Synthesis Approach}\label{sec:max-def-boundedsynth}

The \emph{run graph} of a universal automaton $\autA = (Q,\alphabet,\delta,Q_0,F)$ on a transition system $\trans = (S,s_0,\tau)$ is the unique graph $G = (V,E)$ with a set of nodes $V = S \times Q$ and a set of labeled edges $E \subseteq V \times \alphabet \times V$ such that $((s,q),\sigma,(s',q')) \in E$ if and only if $(q,\sigma,q') \in \delta$ and $\tau(s,\sigma\cap \inpv) = (s',\sigma\cap \outv)$.
That is, $G$ is the product of $\autA$ and $\trans$.

A run graph of a universal B\"uchi (resp.\ co-B\"uchi) automaton is accepting if every infinite path $(s_0,q_0),(s_1,q_1), \ldots$ contains infinitely (resp.\ finitely) many occurrences of states $q_i$ in $F$. A transition system $\trans$ is accepted by a universal automaton $\autA$ if the unique run graph of $\autA$ on $\trans$ is accepting. We denote with  $\lang\autA$ the set of transition systems accepted by $\autA$.

The bounded synthesis approach is based on the fact that for every \gls{ltl} formula $\varphi$ one can construct a universal co-B\"uchi automaton $\autA_\varphi$ with at most $2^{O(|\varphi|)}$ states such that $\trans \in \lang{\autA_\varphi}$ iff $\trans \models \varphi$, for every transition system $\trans$~\citep{KupfermanV05}.

An \emph{annotation} of a transition system $\trans = (S,s_0,\tau)$  with respect to a universal co-B\"uchi automaton $\autA = (Q,\alphabet,\delta,Q_0,F)$ is a function $\anot : S \times Q \to \natsbot$ that maps nodes of the run graph of $\autA$ on $\trans$ to the set $\natsbot$. Intuitively, such an annotation is valid if every node $(s,q)$ that is reachable from the node $(s_0,q_0)$ is annotated with a natural number, which is an upper bound on the number of rejecting states visited on any path from $(s_0,q_0)$ to $(s,q)$.
Valid annotations of finite-state transition systems correspond to accepting run graphs. An annotation $\anot$ is $c$-bounded if $\anot(s,q) \in \{0,\ldots,c\}\cup\{\bot\}$ for all $s \in S$ and $q \in Q$.

The synthesis method proposed in \citep{ScheweF07a,FinkbeinerS13} employs the following result in order to reduce the bounded synthesis problem to checking the satisfiability of propositional formulas: a transition system $\trans$ is accepted by a universal co-B\"uchi automaton $\autA = (Q,\alphabet,\delta,Q_0,F)$ iff there exists a $(|\trans|\cdot|F|)$-bounded valid annotation for $\trans$ and $\autA$. One can estimate a bound on the size of the transition system, which allows to reduce the synthesis problem to its bounded version. Namely, if there exists a transition system that satisfies an \gls{ltl} formula $\varphi$, then there exists a transition system satisfying $\varphi$ with at most $\big(2^{(|\subf\varphi| +\log |\varphi|)}\big)!^2$ states, where $|\varphi|$ denotes the size of the formula $\varphi$ and $\subf\varphi$ denotes the set of all subformulas of $\varphi$.

\subsubsection{MaxSAT}\label{sec:max-def-maxsat}

Consider a propositional logic formula in \gls{cnf}, i.e., a formula that is a conjunction of disjunction of literals, where a literal is a Boolean variable or its negation and a disjunction of literals is called a clause.
\emph{\gls{maxsat}} is the problem of assigning truth values to a set of Boolean variables such that the number of clauses of a propositional logic formula in \gls{cnf} that are made true, is maximized \citep{biere2009handbook}.
\emph{partial weighted \gls{maxsat}} is a variant of \gls{maxsat} problem where the clauses are categorized as hard and soft clauses and each of the soft clauses is associated with a positive numerical weight. The objective is to find a truth assignment to the variables that not only makes all the hard clauses true but also maximizes the sum of the weights of the soft clauses that become true.

We exploit the separation of the hard and soft clauses in partial weighted \gls{maxsat} to capture the hard and soft constraints that arise in the encoding of the maximum realizability problem. Furthermore, we design the weights of the soft clauses in a way to promote the quantitative objective associated with the conjunction of the given soft specifications.

The procedure proposed by Finkbeiner and Schewe~\citet{FinkbeinerS13} provides a SAT encoding of synthesis when the size of the implementation is bounded. Maximum realizability is an optimization variant of synthesis while \gls{maxsat} is an optimization variant of SAT. For a proposed value function, the maximum realizability problem under a bounded implementation size can be reduced to a \emph{partial weighted \gls{maxsat}} instance.

\subsection{Maximum Realizability}\label{sec:max-prob}

\label{sec:max-quantitative-semantics}
Let $\gphi_1,\ldots,\gphi_n$ be a set of \gls{ltl} specifications, where each $\softSpec_i$ is a safety \gls{ltl} formula. In order to formalize the maximal satisfaction of $\gphi_1\wedge\ldots\wedge\gphi_n$, we first give a quantitative semantics of formulas of the form $\gphi$.

\paragraph{Quantitative semantics of safety specifications.}
For an \gls{ltl} formula of the form $\gphi$ and a transition system $\trans$, we define \emph{the value $\val\trans{\gphi}$ of $\gphi$ in $\trans$} as
$$
\val\trans\gphi \defeq
\begin{cases}
(1,1,1) & \text{if } \trans\models\LTLglobally\varphi,\\
(1,1,0) & \text{if } \trans\not\models\LTLglobally\varphi \text{ and } \trans\models\LTLfinally\LTLglobally\varphi,\\
(1,0,0) & \text{if } \trans\not\models\LTLglobally\varphi \text{ and } \trans\not\models\LTLfinally\LTLglobally\varphi \text{ and }\trans\models\LTLglobally\LTLfinally\varphi,\\
(0,0,0) & \text{if } \trans\not\models\LTLglobally\varphi \text{ and } \trans\not\models\LTLfinally\LTLglobally\varphi \text{ and }\trans\not\models\LTLglobally\LTLfinally\varphi.\\
\end{cases}
$$
The value of $\gphi$ in a transition system $\trans$ is a vector $(v_1,v_2,v_3) \in \{0,1\}^3$, where the value $(1,1,1)$ corresponds to the $\true$ value in the classical semantics of \gls{ltl}. When $\trans\not\models\gphi$, the values $(1,1,0)$, $(1,0,0)$ and $(0,0,0)$ capture the extent to which $\varphi$ holds or not along the traces of $\trans$. For example, if $\val\trans\gphi =(1,0,0)$, then $\varphi$ holds infinitely often on each trace of $\trans$, but there exists a trace of $\trans$ on which $\varphi$ is violated infinitely often. If $\val\trans\gphi =(0,0,0)$, then, on some trace of $\trans$, $\varphi$ holds for at most finitely many positions.
Thus, the lexicographic ordering on $\{0,1\}^3$ captures the preference of one transition system over another with respect to the quantitative satisfaction of $\gphi$.

\begin{example}\label{ex:max-one-safety}
    Suppose that we want to synthesize a transition system representing a navigation strategy for a robot working at a restaurant. We require that the robot  serves the VIP area  infinitely often, formalized in \gls{ltl} as $\LTLglobally\LTLfinally \mathit{vip\_area}$. We also desire that the robot never enters the staff's office, formalized as $\LTLglobally\neg\office$. Now, suppose that initially the key to the VIP area is in the office. Thus, in order to satisfy $\LTLglobally\LTLfinally \mathit{vip\_area}$, the robot must violate $\LTLglobally\neg\office$. A strategy in which the office is entered only once, and satisfies $\fg \neg\office$, is preferable to one which enters the office over and over again, and only satisfies $\gf\neg\office$.
    Thus, we want to synthesize a strategy $\trans$  maximizing $\val\trans{\LTLglobally\neg\office}$.
\end{example}

In order to compare implementations with respect to their satisfaction of a conjunction of several safety specifications $\gphi_1 \wedge \ldots \wedge \gphi_n$, we will extend the above definition. We first consider the case where the specifier has not expressed any preference for the individual conjuncts and later on, extend that to the case with a given priority ordering. Consider the following example.

\begin{example}\label{ex:max-two-safety}
    We consider again the restaurant robot, now with two soft specifications. The soft specification $\LTLglobally (\mathit{req1}\rightarrow \LTLnext\mathit{table1})$ requires that each request by table  1 is served immediately at the next time instance. Similarly, $\LTLglobally (\mathit{req2}\rightarrow \LTLnext\mathit{table2})$, requires the same for table number 2. Since the robot cannot be at both tables simultaneously, formalized as the hard specification $\LTLglobally (\neg\mathit{table1} \vee\neg\mathit{table2})$, the conjunction of these requirements is unrealizable. Unless the two tables have priorities, it is preferable to satisfy each of $\mathit{req1}\rightarrow \LTLnext\mathit{table1}$ and $\mathit{req2}\rightarrow \LTLnext\mathit{table2}$ infinitely often, rather than serve one and the same table all the time.
\end{example}

\paragraph{Quantitative semantics of conjunctions.}
To capture the idea illustrated in Example~\ref{ex:max-two-safety}, we define a value function, which intuitively
gives higher values to transition systems in which a fewer number of soft specifications have low values. Formally, let \emph{the value of $\gphi_1 \wedge \ldots \wedge \gphi_n$ in $\trans$} be
\[\val\trans{\gphi_1 \wedge \ldots \wedge \gphi_n} \defeq \big(
\sum_{i=1}^n v_{i,1},
\sum_{i=1}^n v_{i,2},
\sum_{i=1}^n v_{i,3}
\big),\]
where
$\val\trans{\gphi_i} = (v_{i,1},v_{i,2},v_{i,3})$ for $i \in \{1,\ldots,n\}$. To compare transition systems according to these values, we use lexicographic ordering on $\{0,\ldots,n\}^3$.

\begin{example}\label{ex:max-two-safety-val}
    For the specifications in Example~\ref{ex:max-two-safety}, the defined value function assigns value $(2,0,0)$ to a  system satisfying  $\gf(\mathit{req1}\rightarrow \LTLnext\mathit{table1})$ and $\gf (\mathit{req2}\rightarrow \LTLnext\mathit{table2})$, but  neither of $\fg  (\mathit{req1}\rightarrow \LTLnext\mathit{table1})$ and
    $\fg (\mathit{req2}\rightarrow \LTLnext\mathit{table2})$. It assigns the smaller value
    $(1,1,1)$
    to an implementation that gives priority to table 1 and satisfies $\LTLglobally (\mathit{req1}\rightarrow \LTLnext\mathit{table1})$ but not $\gf(\mathit{req2}\rightarrow \LTLnext\mathit{table2})$.
\end{example}

According to the definition above, a transition system that satisfies all soft requirements to some extent is considered better in the lexicographic ordering than a transition system that satisfies one of them exactly and violates all the others. We could instead inverse the order of the sums in the triple, thus giving preference to satisfying some soft specification exactly, over having some lower level of satisfaction over all of them. The next example illustrates the differences between the two variations.

\begin{example}\label{ex:max-ordering}
    For the two soft specifications from Example~\ref{ex:max-two-safety}, reversing the order of the sums in the definition of $\val\trans{\gphi_1 \wedge \ldots \wedge \gphi_n}$ results in giving the higher value $(1,1,1)$ to a transition system that satisfies $\LTLglobally (\mathit{req1}\rightarrow \LTLnext\mathit{table1})$ but not $\gf(\mathit{req2}\rightarrow \LTLnext\mathit{table2})$, and the lower value $(0,0,2)$ to the one that only guarantees $\gf(\mathit{req1}\rightarrow \LTLnext\mathit{table1})$ and $\gf (\mathit{req2}\rightarrow \LTLnext\mathit{table2})$. The most suitable ordering usually depends on the specific application.
\end{example}

\subsection{Problem Formulation}\label{sec:max-prob-form}
Using the definition of quantitative satisfaction of soft safety specifications, we now define the maximum realizability problem, which asks to synthesize a transition system that satisfies a given \emph{hard} \gls{ltl} specification, and is optimal with respect to the satisfaction of a conjunction of \emph{soft} safety specifications.

{\bf Bounded maximum realizability problem:} Given an \gls{ltl} formula $\spec$ and formulas $\gphi_1,\ldots,\gphi_n$, where each $\softSpec_i$ is a safety \gls{ltl} formula, and a bound $b\in \nats_{>0}$, the bounded maximum realizability problem asks to determine if there exists a transition system $\trans$ with $|\trans| \leq b$ such that $\trans \models \spec$, and if the answer is positive, to synthesize a transition system $\trans$ such that $\trans \models \spec$, $|\trans| \leq b$ and for every transition system $\trans'$ with $\trans'\models \spec$ and $|\trans'| \leq b$, it holds that $\val\trans{\gphi_1 \wedge \ldots \wedge \gphi_n} \geq \val{\trans'}{\gphi_1 \wedge \ldots \wedge \gphi_n}$.

\subsection{Maximum Realizability as Iterative \gls{maxsat} Solving}\label{sec:max-maxsat-encoding}

We now describe the \gls{maxsat}-based approach to maximum realizability proposed by \citet{dimitrova2018maximum}. The approach first establishes an upper bound on the minimal size of an implementation that satisfies a given \gls{ltl} specification $\varphi$ and maximizes the satisfaction of a conjunction of the soft specifications $\gphi_1, \ldots,\gphi_n$, according to the value function defined in Section~\ref{sec:max-quantitative-semantics}.
This bound can be used to reduce the maximum realizability problem to its bounded version, which will be encoded as a \gls{maxsat} problem.

\begin{figure}[t]
    \centering
    \includegraphics[width=1\textwidth]{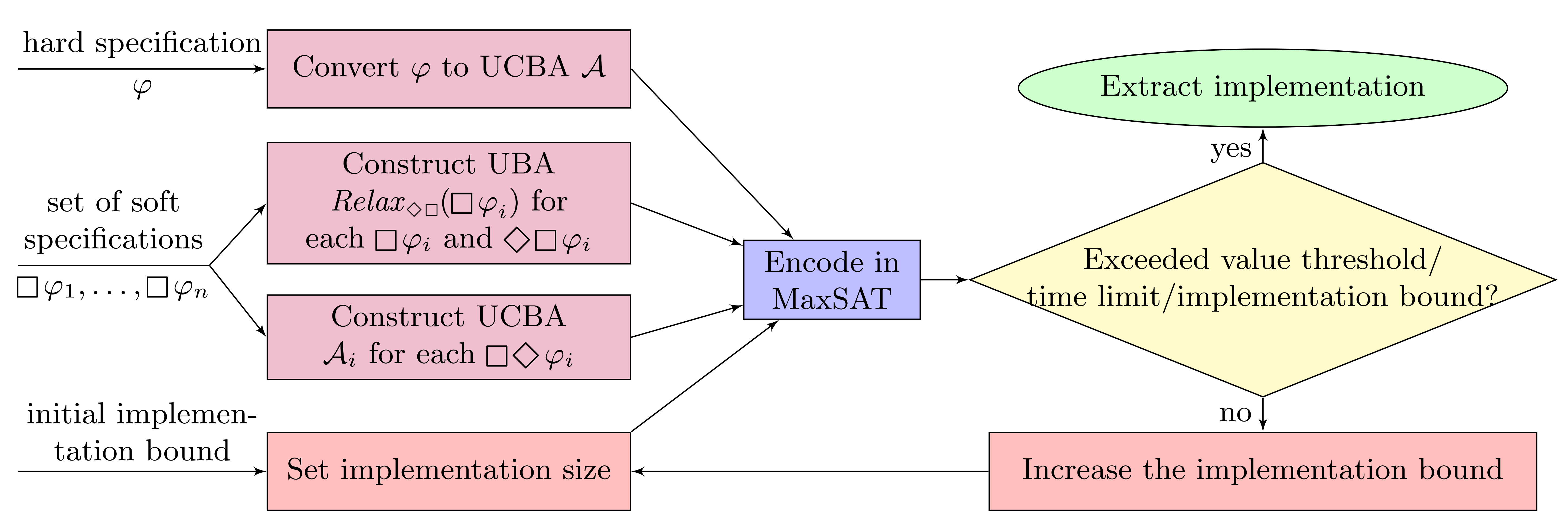}
    \caption{Outline of the maximum realizability procedure.}
    \label{fig:max-proc_diag}
\end{figure}

For each of the possible values of $\gphi_1 \wedge\ldots \wedge\gphi_n$ there is a corresponding \gls{ltl} formula that encodes this value in the classical \gls{ltl} semantics. This property can be utilized to establish an upper bound on the minimal optimal implementation.

\begin{theorem}\label{thm:max-optimal-bound-safety}
    Given an \gls{ltl} specification $\spec$ and soft safety specifications $\gphi_1,\ldots,$ $\gphi_n$,
    if there exists a transition system $\trans \models \spec$, then there exists  $\trans^*$ such that
    \begin{compactitem}
        \item [(1)] $\val{\trans^*}{\gphi_1 \wedge \ldots \wedge \gphi_n} \geq\val{\trans}{\gphi_1 \wedge \ldots \wedge \gphi_n}$ for all $\trans$ with $\trans \models \spec$,
        \item[(2)] $\trans^* \models \spec$ and $|\trans^*| \leq \left((2^{(b+\log b)})!\right)^2$,
    \end{compactitem}
    where $b = \max\{|\subf{\spec\wedge\softSpec_1'\wedge\ldots\wedge\softSpec_n'}| \mid \forall i:\ \softSpec_i' \in\{\gphi_i,\fgphi_i,\gfphi_i\}\}$.
\end{theorem}

The bound above is estimated based on the size of the specifications, using a worst-case bound on the size of the corresponding automata. Given the automata for all the specifications $\gphi_i,\fgphi_i$ and $\gfphi_i$, a potentially better bound can be estimated based on the sizes of these automata.

Figure~\ref{fig:max-proc_diag} gives an overview of the maximum realizability procedure and the automata constructions it involves. As in the bounded synthesis approach, we construct a universal co-B\"uchi automaton $\autA$ for the hard specification $\varphi$.  For each soft specification $\gphi_j$, we construct a pair of automata corresponding to the relaxations of $\gphi_j$. The relaxation $\gfphi_j$ is treated as in bounded synthesis. For $\gphi_i$ and $\fgphi_i$, we construct a single universal B\"uchi automaton and define a corresponding annotation function.

\subsubsection{\gls{maxsat} Encoding of Bounded Maximum Realizability}\label{sec:max-maxsat}

Let $\autA = (Q,\alphabet,\delta,Q_0,F)$ be a universal co-B\"uchi automaton for the \gls{ltl} formula $\spec$.
For each syntactically safe formula $\gphij$, $j \in\{1,\ldots,n\}$, we consider two universal automata:
the universal automaton $\autB_j = \relaxfg(\gphi_j)= (Q_j,\alphabet,\delta_j,Q_0^j,F_j)$
and a universal co-B\"uchi automaton $\autA_j = (\widehat Q_j,\alphabet,\widehat\delta_j,\widehat Q_0^j,\widehat F_j)$ for the formula $\gfphi_j$.
Given a bound $b$ on the size of the desired transition system, we encode the bounded maximum realizability problem as a \gls{maxsat} problem.

A transition system extracted from an optimal satisfying assignment for the \gls{maxsat} problem is optimal with respect to the value of $\gphi_1 \wedge \ldots \wedge \gphi_n$, as stated in the following theorem that establishes the correctness of the encoding.
\begin{theorem}\label{thm:max-encoding-correctness}
    Let $\autA$ be a given co-B\"uchi automaton for $\varphi$, and for each $j \in \{1,\ldots,n\}$, let $\autB_j = \relaxfg(\gphi_j)$ be the universal automaton for $\gphi_j$, and let $\autA_j$ be a universal co-B\"uchi automaton for $\gfphi_j$.
    The constraint system for bound $b \in \nats_{>0}$ is satisfiable if and only if there exists an implementation $\trans$ with $|\trans| \leq b$  such that $\trans \models \varphi$. Furthermore, from the optimal satisfying assignment to the variables $\tau_{s,\inpval,s'}$ and $o_{s,\inpval}$, one can extract a transition system $\trans^*$ such that for every transition system $\trans$ with $|\trans| \leq b$ and $\trans \models \varphi$ it holds that $\val{\trans^*}{\gphi_1 \wedge \ldots \wedge \gphi_n} \geq \val{\trans}{\gphi_1 \wedge \ldots \wedge \gphi_n}$.
\end{theorem}

\begin{figure}
    \centering
    \scalebox{1}{
        \input{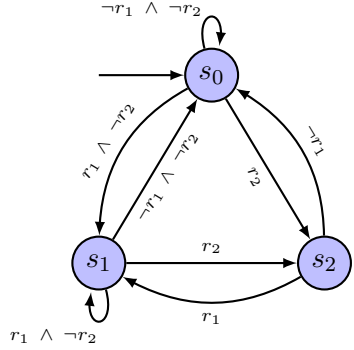}}
    \caption{An optimal implementation for Example~\ref{ex:max-two-safety}. }
    \label{fig:max-waiter-strategy}
\end{figure}
Figure~\ref{fig:max-waiter-strategy} shows a transition system extracted from an optimal satisfying assignment for Example~\ref{ex:max-two-safety} with bound $3$ on the implementation size. The  transitions depicted in the figure are defined by the values of the variables $\tau_{s,\inpval,s'}$. The outputs of the implementation (omitted from the figure) are defined by the values of $o_{s,\inpval}$. The output in state $s_1$ when $\mathit{r1}$ is $\trueval$ is $\mathit{table1}\wedge \neg\mathit{table2}$, and the output in $s_2$ when $\mathit{r2}$ is $\trueval$ is $\neg\mathit{table1}\wedge \mathit{table2}$. For all other combinations of state and input, the output is $\neg\mathit{table1}\wedge \neg\mathit{table2}$.

The next proposition establishes the size of the \gls{maxsat} encoding.
\begin{proposition}\label{prop:max-encoding-size}
    Let $\autA$ be a given co-B\"uchi automaton for $\varphi$, and for each $j \in \{1,\ldots,n\}$, let $\autB_j = \relaxfg(\gphi_j)$ be the universal B\"uchi automaton for $\gphi_j$, and let $\autA_j$ be a universal co-B\"uchi automaton for $\gfphi$.
    The constraint system for bound $b \in \nats$ has weights in $\mathcal{O}(n^2)$. It has
    \begin{equation*}
    \begin{aligned}
        \mathcal{O}\Big(
        &(b^2 + b \cdot |\outv|)\cdot 2^{|\inpv|} +
        b \cdot |Q|\cdot (1+ \log(b \cdot |Q|)) +\\
        &\phantom{\mathcal{O}(}\sum_{j=1}^n\big(b \cdot |Q_j| (1 + \log(b\cdot |Q_j|))\big)
        + \sum_{j=1}^n\big(b \cdot |\widehat Q_j| (1 + \log(b\cdot |\widehat Q_j|))\big)
        \Big)
    \end{aligned}
    \end{equation*}
    variables, and its size is
    \begin{equation*}
    \begin{aligned}
        \mathcal{O}\Big(
        &|Q|^2 \cdot b^2 \cdot 2^{|\inpv|} \cdot (d + \log(b\cdot |Q|)) +
        \mathcal{O}(\sum_{j=1}^n\big(|Q_j|^2 \cdot b^2 \cdot 2^{|\inpv|}
        \\&\cdot (d_j + r_j + \log(b\cdot |Q_j|))\big) +
        \mathcal{O}(\sum_{j=1}^n\big(|\widehat Q_j|^2 \cdot b^2 \cdot 2^{|\inpv|} \cdot (\widehat d_j + \log(b\cdot |\widehat Q_j|))\big)
        \Big),
    \end{aligned}
    \end{equation*}
    where
    \begin{equation*}
    \begin{aligned}
    d &= \max_{s,q,\inpval,q'}|\delta_{s,q,\inpval,q'}|,&
    d_j &= \max_{s,q,\inpval,q'}|\delta_{s,q,\inpval,q'}^j|,\\
    \widehat d_j &= \max_{s,q,\inpval,q'}|\widehat \delta_{s,q,\inpval,q'}^j|,&
    \text{ and }
    r_j &= \max_{s,q,\inpval,q'}|\rej^j(s,q,q',\inpval)|.
    \end{aligned}
    \end{equation*}
\end{proposition}

\subsection{A case study}\label{sec:max-experiments}

\begin{figure}
    \centering
    \includegraphics[width=0.75\linewidth]{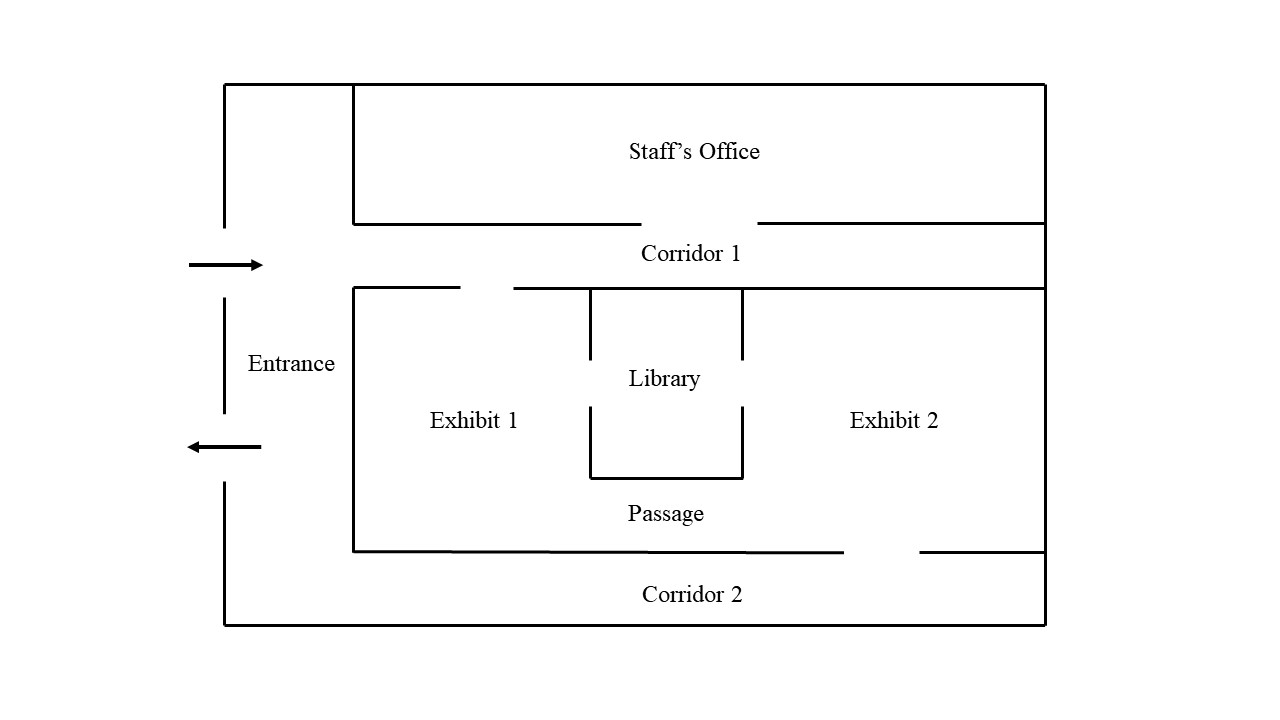}
    \caption{Map of the museum.}
    \label{fig:max-map}
\end{figure}

Consider a robotic museum guide in a museum shown in Figure~\ref{fig:max-map}. The robot has to give a tour of the exhibitions in a specific order, which constitutes the hard specification. The tour starts at the entrance of the museum where the robot picks up newly arrived visitors. The main objective is to take the group through the two exhibitions on that floor and then return to the entrance to pick up a new group of people.
Preferably, it also avoids certain locations, such as the library, or the passage when it is occupied. These preferences are encoded in the soft specifications.
In particular, on one hand, the robot can only gain access to Exhibition 2 by getting a key from the staff's office. On the other hand, the robot is asked not to disturb the employees in the office. There is a library between Exhibition 1 and Exhibition 2 which can be used to go from one to the other, but it is preferred that visitors do not enter the library. However, it is also desirable that when the other passage between these two exhibitions is occupied, the robot does not go through there.

These specifications cannot be realized in conjunction. Given their priorities, we categorize the requirements into hard and soft specifications, and synthesize a strategy which satisfies the hard specifications and maximizes the satisfaction of the soft specifications. We formalize the problem as follows.

\smallskip
\noindent
{\bf Propositions:}
The set $\inpv$ contains a single Boolean variable $\occupied$ that indicates whether the passage between the two exhibitions is occupied.
The set of output propositions $\outv$ consists of eight Boolean variables corresponding to the eight locations on the map: $\entr$, $\corr_1$, $\corr_2$, $\exh_1$, $\exh_2$, $\passage$, $\office$, $\library$.

\smallskip
\noindent
{\bf The hard specification} is the conjunction of the following formulas.

\begin{itemize}
    \item The robot starts at the entrance:
    \begin{equation*}
    \entr.
    \end{equation*}

    \item At each time step, the robot can occupy only one location:
    \begin{equation*}
    \LTLglobally \bigwedge_{o_1 \in \outv} \left( o_1 \rightarrow \bigwedge_{o_2 \in \outv \backslash \{o_1\}} \neg o_2 \right) .
    \end{equation*}

    \item The admissible actions of the robot are to stay in the current location or move to an adjacent one. This leads to eight requirements describing the map.
    For instance:
    \begin{equation*}
    \LTLglobally \left( \corr_1 \rightarrow \LTLnext \left(\corr_1 \lor \office \lor \exh_1 \right) \right) .
    \end{equation*}
\end{itemize}

{\it Remark: }Due to the requirements above,  the robot will always be in exactly one valid location, i.e., in a transition system that satisfies the specifications it is impossible to reach a state where all output variables are false.

\begin{itemize}

    \item The robot must infinitely often visit both exhibitions:
    \begin{equation*}
    \begin{aligned}
    & \LTLglobally \LTLfinally \exh_1 ,\\
    & \LTLglobally \LTLfinally \exh_2 .
    \end{aligned}
    \end{equation*}

    \item The robot has to respect the order of visits, by starting from Exhibition 1, going to Exhibition 2 and finishing at the entrance:
    \begin{equation*}
    \begin{aligned}
    & \LTLglobally \left(\exh_1 \rightarrow \LTLnext \left( \left( \neg \entr \land \neg \exh_1 \right) \LTLuntil \exh_2 \right) \right) ,\\
    & \LTLglobally \left(\exh_2 \rightarrow \LTLnext \left( \left( \neg \exh_1 \land \neg \exh_2 \right) \LTLuntil \entr \right) \right) ,\\
    & \LTLglobally \left(\entr \rightarrow \LTLnext \left( \left( \neg \exh_2 \land \neg \entr \right) \LTLuntil \exh_1 \right) \right) .
    \end{aligned}
    \end{equation*}

    \item The robot does not have access to Exhibition 2 before it visits the office:
    \begin{equation*}
    \neg \exh_2 \LTLuntil\office.
    \end{equation*}

\end{itemize}

\smallskip
\noindent
{\bf The set of soft specifications} describes the desirable requirements that the robot does not enter the office, the library, or a occupied passage. Formally:

\begin{itemize}
    \item The robot must not enter the office from corridor 1:
    \begin{equation*}
    \LTLglobally \left(\corr_1 \rightarrow \LTLnext \neg \office \right) .
    \end{equation*}

    \item The robot must not enter the library from the exhibitions:
    \begin{equation*}
    \LTLglobally \left( \exh_1 \lor \exh_2 \rightarrow \LTLnext \neg \library \right) .
    \end{equation*}

    \item The robot must not enter the passage from the exhibitions when it is occupied:
    \begin{equation*}
    \LTLglobally \left( \left( \exh_1 \lor \exh_2 \right) \land \LTLnext \occupied \rightarrow \LTLnext \neg \passage \right) .
    \end{equation*}

\end{itemize}

The maximum realizability as iterative MaxSAT approach can be applied to synthesize a policy for the robotic museum guide.
Table~\ref{tab:max-inst-robot} summarizes the results. With implementation bound of 8, the hard specification is realizable and a partial satisfaction of soft specifications is achieved. This strategy always selects the passage to transition from Exhibition 1 to Exhibition 2 and hence, avoids the library. It also violates the requirement of not entering the staff's office, to acquire access to Exhibition 2. For implementation bound 10 the solver times out. Notice that strategies with higher values exists, however, they require larger implementation size.

\begin{table*}[h!]\centering\scriptsize
    \caption{Results of applying synthesis with maximum realizability to the robotic navigation example, with different bounds on implementation size $|\trans|$. We report on the number of variables and clauses in the encoding, the satisfiability of hard constraints, the value (and bound) of the \gls{maxsat} objective function, the running times of Spot, Open-WBO, and the time of the solver plus the time for generating the encoding.}
    \begin{tabular*}{1\linewidth}{@{\extracolsep{4pt}}rrrrrrrr@{}}
        \toprule
        & \multicolumn{2}{c}{Encoding} & \multicolumn{2}{c}{Solution} & \multicolumn{3}{c}{Time (s)} \\
        \cline{2-3} \cline{4-5} \cline{6-8}
        $|\trans|$ & \# vars & \# clauses & sat. & $\Sigma weights$ & Spot & Open-WBO & enc.+solve \\
        \midrule
        2 & 4051 & 25366 & UNSAT & 0 (39) & 0.93 & 0.011 & 0.12 \\
        4 & 19965 & 125224 & UNSAT & 0 (39) & 0.93 & 0.079 & 0.57 \\
        6 & 45897 & 289798 & UNSAT & 0 (39) & 0.93 & 1.75 & 2.9 \\
        8 & 95617 & 596430 & SAT & 31 (39) & 0.93 & 956 & 959 \\
        10 & 152949 & 954532 & SAT & - (39) & 0.93 & time-out & time-out \\
        \bottomrule
    \end{tabular*}
    \label{tab:max-inst-robot}
\end{table*}

\chapter{Probabilistic Synthesis and Verification}
\glsresetall
\label{chap:probabilistic-synthesis}

\blfootnote{This section incorporates the results from the following publications~\citep*{atvaqcqp,cubuktepe2020scenario}.}
Autonomous systems operate in uncertain, dynamic environments and involve many sub-components such as perception, localization, planning, and control. The interaction between all of these components involves uncertainty. The sensors cannot entirely capture the environment around the autonomous system and are inherently noisy. Perception and localization techniques often rely on machine learning, and the outputs of these techniques involve uncertainty. Overall, the autonomous system needs to plan its decisions based on the uncertain output from perception and localization, which leads to uncertain outcomes.

To model decision-making under uncertainty, we start by describing the sequential decision-making model given in Figure~\ref{fig:intro}.
At any time step $t$, an agent observes the system's state $s_t$ and makes a decision $a_t$ based on this observation.
This action yields two results. 
First, the agent receives an immediate reward $r_{t}$ (or a cost.)
Second, the system transitions to a new state $s_{t+1}$ according to a probability distribution at time step $t+1$, which is determined by the choice of action $a_{t}$ in state $s_{t}$.
At the subsequent time steps, the agent faces a similar problem and needs to make a decision in a potentially different state in the system and from a potentially different set of actions.
We list the key ingredients in this sequential decision model:

\tikzstyle{block} = [rectangle, draw, 
text width=6em, text centered, rounded corners, minimum height=3em]

\tikzstyle{line} = [draw, -latex]

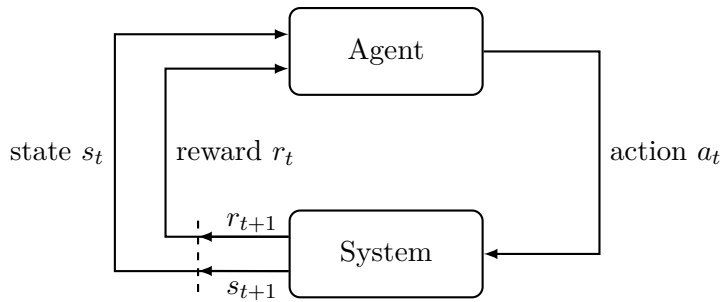
\begin{figure}
	\centering
\begin{tikzpicture}[node distance = 7em, auto, thick,every text node part/.style={align=center}]
\node [block] (Agent) {Agent};
\node [block, below of=Agent] (Environment) {System};

\path [line] (Agent.0) --++ (4em,0em) |- node [near start]{action $a_t$} (Environment.0);
\path [line] (Environment.190) --++ (-6em,0em) |- node [near start] {state  $s_{t}$} (Agent.170);
\draw [line] (Environment.190) -- (-2.5,-2.92);
\draw [line] (Environment.170) -- (-2.5,-2.465);
\path [line] (Environment.170) --++ (-4.25em,0em) |- node [near start, right] {reward $r_{t}$} (Agent.190);
\node[align=left,above left =-1.35cm and 0cm of Environment]  {$s_{t+1}$};
\node[align=left,above left =-0.45cm and 0cm of Environment]  {$r_{t+1}$};
\draw [dashed,]  (-2.5,-3.2) -- (-2.5,-2.2);

\end{tikzpicture}
\caption{A representation of a sequential decision-making problem.}
\label{fig:intro}
\end{figure}

\begin{enumerate}
		\item A set of states, which describes all possible configurations of the system.
	\item A set of available action in each state, which describes the set of decisions in the all states of the system.
	\item A set of rewards or costs for each state and action, which describes the objective or performance criterion of the system.
	\item A set of transition probabilities to the states in the system for each state and action, which describes the dynamics of the system.
		\item A set of (potentially infinite) decision horizon, which describes the planning period.
	\end{enumerate}

We focus on a particular sequential decision-making model, which is called \glspl{mdp}~\citep{puterman2014markov}. In MDPs, the set of available actions, the rewards or costs, and the transition probabilities depend only on the current state and implemented action on the system.
\begin{definition}[Markov Decision Process (MDP)]\label{atva2018def:mdp}
    A \emph{Markov decision process \gls{mdp}} is a tuple $\MdpInit$ with a finite set $S$ of \emph{states}, an \emph{initial state} $\sinit \in S$, a finite set $\Act$ of \emph{actions}, and a \emph{transition function} $\probmdp \colon S \times \Act \times S \rightarrow [0, 1]$ such that $\sum_{s'\in S}\probmdp(s,\act,s') = 1$ for all $s \in S$ and $\act \in \ActS$ where $\ActS$. denotes the set of available actions in the state $s$. A \emph{cost function} $c\colon S\times\Act\rightarrow\R_{\geq 0}$ associates cost to state-action pairs.
\end{definition}
MDPs have numerous applications in various domains thanks to their generality. These applications include but not limited to reinforcement learning~\citep{jansen2020safe,mason2017assured,lecarpentier2019non}, robotics~\citep{liu2018reinforcement,omidshafiei2017decentralized,ghasemi2019online}, human-robot interaction~\citep{chen2020trust,akash2020human,cubuktepe2018autonomy1},  aircraft collision avoidance~\citep{julian2019deep},  healthcare~\citep{hosseini2014coordinated}, disease management~\citep{radoszycki2015solving}, finance~\citep{borkar2014risk}, and digital marketing~\citep{thomas2017predictive}.

The central problem for MDPs is to find a control policy, which determines what action to take with the current knowledge of the system at a given time.
In other words, the policy is a mapping from the states to the actions.
The typical aim is to optimize a given objective, such as minimizing expected cost, for example, minimizing the fuel usage of the system, or maximizing the probability of the successful operation of a system for a (potentially infinite) horizon.
Given an MDP, the problem of finding an optimal policy can be cast as a dynamic programming problem, and numerous methods based on value or policy iteration and reinforcement learning exist to find such a policy.

A related problem of finding a policy in an MDP is called \emph{model checking}.
Given a model that represents the behavior of the system and a specification that determines the objective, model checking refers to a set of techniques that systematically check whether the system satisfies the given specification.
For MDPs, a related technique is \emph{probabilistic model checking}. 
Given an MDP and a specification, typically expressed as a formula in temporal logic, probabilistic model checking refers to determining whether there exists a policy that satisfies the specification.
Many related problems such as minimizing the expected cost of satisfying the specification can also be solved using the probabilistic model checking framework.

In Section \ref{section:mdp}, we first consider the probabilistic model checking problem for MDPs and explain different approaches to solve this problem.
We illustrate this problem with a case study on human-autonomy interactions in Section \ref{mdp-human-autonomy}.
Then, in Section~\ref{sec:parameter-synthesis}, we define an extension of MDPs called parametric MDPs, where functions now give the transition and reward function of an MDP over parameters.
We discuss a convex-optimization-based technique by \citet{atvaqcqp} to solve the so-called \emph{parameter synthesis problem} that scales to thousands of parameters as opposed to a handful of parameters for the existing methods.
Next, in Section~\ref{sec:uncertain-mdp}, we consider a setting where the parameters of the transition probabilities and rewards belong to an uncertainty set parameterized by a collection of random variables. 
The problem is to compute the satisfaction probability to satisfy a temporal logic specification within any MDP that corresponds to a sample from these unknown distributions. We give a technique by \citet{cubuktepe2020scenario} from so-called \emph{scenario optimization} to compute high confidence bounds on the satisfaction probabilities.

\section{Probabilistic Model Checking in Markov Decision Processes}
\label{section:mdp}
Probabilistic model checking is a rigorous technique that can precisely solve the aforementioned control problems, and this rigor provides guarantees on appropriate behavior for all possible events in the system.
Probabilistic model checking has been extensively studied for MDPs~\citep{BK08}, and mature tools exist for efficient model checking~\citep{KNP11,DBLP:conf/cav/DehnertJK017,iscasmc}.

In this section, we first give the formal definitions of concepts related to MDPs. Then, we give methods for verification and synthesis in MDPs subject to temporal logic specifications.

\subsection{Preliminaries for Markov Decision Processes}
A \emph{probability distribution} over a finite or countably infinite set $\distDom$ is a function $\distFunc\colon\distDom\rightarrow\Ireal$ with $\sum_{\distDomElem\in\distDom}\distFunc(\distDomElem)=1$.
The set of all distributions on $\distDom$ is denoted by $\Distr(\distDom)$.

\begin{figure}[t]
    \centering
        \begin{tikzpicture}
	
	\tikzstyle{dot}=[
	circle,
	radius=0.06,
	draw=black,
	fill = black,
	];
	\node[state] (v0) at (0,0) {$s_0$};
	\node[state] (v1) at (2,1) {$s_1$};
	\node[state] (v2) at (2,-1) {$s_2$};
	\node[state] (v3) at (4,2) {$s_3$};
	\node[state] (v4) at (4,0) {$s_4$};
	\node[state] (v5) at (4,-2) {$s_5$};
	\node[state, fill=red!50] (v61) at (6,0) {$s_6$};
	\node[state] (v7) at (6,-2) {$s_7$};
	
	\draw [-latex, thick] (-1,0) -- (v0);
	\node[dot](A) at (.9,0){};
	\draw [thick] (v0) -- (A);
	\draw [-latex, thick,postaction={decorate,decoration={raise=1ex,text along path,text align=center,text={|\scriptsize|{$0.7$}}}}] (A) -- (v1);
	\draw [-latex,  thick,postaction={decorate,decoration={raise=-2ex,text along path,text align=center,text={|\scriptsize|{$0.3$}}}}] (A) -- (v2);

	\node[dot](B) at (2.9,1){};
	\draw [thick] (v1) -- (B);
	\draw [-latex,  thick,postaction={decorate,decoration={raise=1ex,text along path,text align=center,text={|\scriptsize|{$0.5$}}}}] (B) -- (v3);
	\draw [-latex,  thick,postaction={decorate,decoration={raise=-2ex,text along path,text align=center,text={|\scriptsize|{$0.5$}}}}] (B) -- (v4);
	
	\node[dot](C) at (2.9,-1){};
	\draw [thick] (v2) -- (C);
	\draw [-latex, thick,postaction={decorate,decoration={raise=-2ex,text along path,text align=center,text={|\scriptsize|{$0.5$}}}}] (C) -- (v5);
	\draw [-latex, thick,postaction={decorate,decoration={raise=1ex,text along path,text align=center,text={|\scriptsize|{$0.5$}}}}] (C) -- (v4);
	
	\draw[-latex, thick,postaction={decorate,decoration={raise=1ex,text along path,text align=center,text={|\scriptsize|{$a$}}}}] (v4) -- (v61);
	\draw[-latex, thick,postaction={decorate,decoration={raise=1ex,text along path,text align=center,text={|\scriptsize|{$b$}}}}] (v3) -- (v61);
	\draw [-latex,thick,postaction={decorate,decoration={raise=-1.75ex,text along path,text align=center,text={|\scriptsize|{$b$}}}}] (v5) -- (v7);
	
	\node[dot](D) at (5,-1.2){};
	\draw [thick] (v5) -- (D);
	\draw [-latex, thick,postaction={decorate,decoration={raise=1ex,text along path,text align=center,text={|\scriptsize|{$0.3$}}}}] (D) -- (v61);
	\draw [-latex,thick,postaction={decorate,decoration={raise=-1.5ex,text along path,text align=center,text={|\scriptsize|{$0.7$}}}}] (D) -- (v7);

	\draw [thick] (v3) edge[>=latex, loop above] (v3) {};
	\draw [thick] (v4) edge[>=latex, loop above] (v4) {};
	\draw [ thick] (v61) edge[>=latex, loop above] (v61) {};
	\draw [thick] (v7) edge[>=latex, loop above] (v7) {};
	
	\tkzLabelPoint (5.2, -1) {\scriptsize{$a$}};
	\tkzLabelPoint (3.8,3.1) {\scriptsize{$a$}};
	\tkzLabelPoint (3.85,1.2) {\scriptsize{$b$}};
\end{tikzpicture}
    \caption{An MDP with the state space $S=\{s_0, s_1, s_2, s_3, s_4, s_5, s_6, s_7\}$, action space $\Act=\{a, b\}$ and the initial state $\sinit=s_0$. The transition function $\probmdp$ is given by (possibly branching edges) in the graph. To avoid clutter, we omit the transitions with probability $1$, and we merge action selections that induce the same transition probabilities, but differ only in action names.}
    \label{backgroundfig:MDP}
\end{figure}
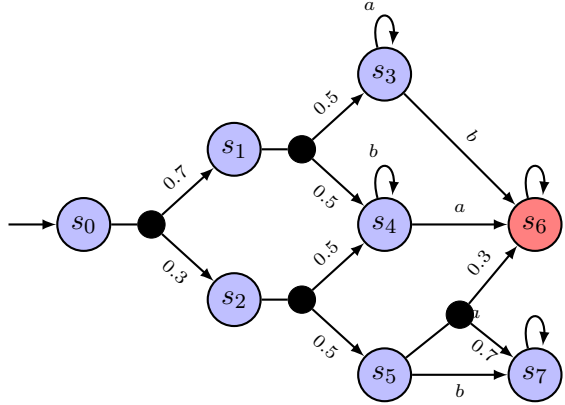

\begin{definition}[Policy]
To define measures on
MDPs, nondeterministic action choices are resolved by a so-called \emph{policy} $\sched\colon S\rightarrow\Act$ with $\sched(s) \in \ActS$.
The set of all policies over $\mdp$ is $\Sched^\mdp$.
\end{definition}

We note that the above definition of a policy is a deterministic mapping from states to actions. Such policies are called memoryless deterministic and suffice for MDPs for the performance criteria considered in this review~\citep{BK08}.

Applying a policy to an MDP yields an \emph{induced \gls{mc}} where all nondeterminism is resolved.
\begin{definition}[Induced Markov Chain (MC)]\label{atva2018def:induced_dtmc}
    For MDP $\MdpInit$ and policy $\sched\in\Sched^\mdp$, the \emph{MC induced by $\mdp$ and $\sched$} is  $\mdp^\sched=(S,\sinit,\Act,\pmdp^\sched)$ with
    \begin{align*}
         \pmdp^\sched(s,s')= \pmdp(s,\sched(s),s')\textit{ for all }s,s'\in S.
    \end{align*}
\end{definition}
Intuitively, the transition probabilities in $\mdp^\sched$ are obtained with respect to the action choices of the policy.

\begin{definition}[Occupancy Measure]
    The occupancy measure $x_\sched$ of a policy $\sched$ for an MDP $\mdp$ is defined as
    \begin{align}
        \displaystyle x_{\sched}(s,\act)=\sum\nolimits_{t=0}^{\infty}\Pr^{\sched}(s_t=s, \act_t=\act | s_0=\sinit),
    \end{align}
    where $\Pr^{\sched}$ denotes the probability measure induced by $\sched$, and $s_t$ and $\act_t$ denote the state and action in $\mdp$ at time $t$.
\end{definition}

The occupancy measure $x_{\sched}(s,\act)$ is the expected number of times to take action $\act$ at state $s$ under the policy $\sched$. 

For an MC $\dtmc$, the \emph{reachability specification} $\reachPropSymbol=\reachProplT$ asserts that a set $T \subseteq \states$ of \emph{target states}  is reached with probability at most $\lambda\in [0,1]$.
If $\reachPropSymbol$ holds for $\dtmc$, we write $\dtmc\models\reachPropSymbol$.
Accordingly, for an \emph{expected cost specification}, $\ereachPropSymbol=\expRewProp{\kappa}{G}$, $\dtmc\models\ereachPropSymbol$ holds if and only if the expected cost of reaching a set $G \subseteq \states$ is bounded by $\kappa \in \R$.
We use standard measures and definitions as in~\citep[Ch.\ 10]{BK08}.

\paragraph{Specifications.} We consider specifications that are combinations of \emph{reachability specifications} and \emph{expected cost specifications}. A reachability property $\reachPropSymbol=\reachProplT$  with upper probability bound $\lambda\in\Irat$ and target set $T\subseteq S$ constrains the probability to finally reach $T$ from $\sinit$ in $\mdp$ to be at most $\lambda$.
Analogously, expected cost specifications $\ereachPropSymbol=\expRewProp{\kappa}{G}$ impose an upper bound $\kappa\in\R$ on the expected cost to reach goal states $G\subseteq S$ with respect to the cost function $c$.
Combining both types of specifications, the intuition is that a set of bad states $T$ shall only be reached with a certain probability $\lambda$ (safety specification) while the expected cost for reaching a set of goal states $G$ has to be below $\kappa$ (performance specification).
We overload the notation $\finally T$ to denote both a reachability specifications and the set of all paths that finally reach $T$ from the initial state $\sinit$ of an MC. The probability and the expected cost for reaching $T$ from $\sinit$ are denoted by $\pr(\finally T)$ and $\er(\finally T)$, respectively. Hence, $\pr(\finally T)\leq\lambda$ and $\er(\finally G)\leq\kappa$ express that the properties $\reachProplT$ and $\expRewProp{\kappa}{G}$ respectively are satisfied by MC $\dtmc$. We note that \gls{ltl} specifications can be
reduced to reachability and expected cost specifications, and we refer the reader to~\citep{BK08} for a detailed introduction.

An MDP $\mdp$ satisfies both reachability specification $\reachPropSymbol$ and expected cost specification $\ereachPropSymbol$, if and only if \emph{for all} policies $\sched$ it holds that the induced MC $\mdp^\sched$ satisfies the properties $\varphi$ and $\psi$, i.e., $\mdp^\sched\models\varphi$ and $\mdp^\sched\models\psi$.  In our setting, we are also  interested in the so-called \emph{synthesis problem}, where the aim is to find a policy $\sched$ such that both specifications are satisfied (while this does not necessarily hold for all specifications).  If $\mdp^\sched\models\varphi$, policy $\sched$ is said to \emph{admit} the property $\varphi$; this is denoted by $\sched\models\varphi$.

\begin{example}
    Consider the MDP in Figure~\ref{backgroundfig:MDP} with the target set $T=\{s_6\}.$
    Given a reachability specification $\reachPropSymbol=\reachProplT$ with $\lambda=0.85$, an example policy $\sched_1$ that satisfies the specification is given by $\sched_1(s_3)=b, \sched_1(s_4)=a$, and $\sched_1(s_5)=a$, which induces a reachability probability of $0.895$, and is greater than the threshold $\lambda$.
    In this example, the policy $\sched_2$, which is defined by selecting the same actions as $\sched_1$ except $\sched_2(s_5)=b$, induces a MC that satisfies the specification with probability $0.85$, and also satisfies the specification, even though it may not maximize the reachability probability.
\end{example}

\subsection{Verification and Synthesis in Markov Decision Processes Subject to Temporal Logic Specifications}
In this section, we describe the primal and dual \gls{lp} formulations for verification and synthesis problems in MDPs subject to temporal logic specifications~\citep{puterman2014markov,forejt2011quantitative}.
We first start with the primal LP formulation, which is mainly used in verification problems.
The variables in the primal LP formulation specify the probability of satisfying the reachability specification $\reachPropSymbol$ induced by a maximizing policy, hence it is used in verification problems. We then describe the dual LP formulation, which is mainly used in policy synthesis problems subject to (multiple) temporal logic specifications. The variables in the dual LP are the expected number of times of taking action in the states of the MDP, and an optimal policy can also be obtained from an optimal solution of the dual LP.

\paragraph{Primal LP.} The primal LP formulation  computes the maximum probability $p_{\sinit}$ of reaching the target set $T$ from the initial state $\sinit$ using Bellman's principle of optimality. The result of the LP serves as a certificate for the verification problem~\citep[Ch.\ 10]{BK08}.
The LP reads as follows:
\begin{align}
    \text{minimize} &\quad p_{\sinit}\label{eq:min_mdp}\\
    \text{subject to} &\nonumber \\
    p_s=1,\label{eq:targetprob_mdp}&\quad   \forall s\in T,  & \\
    \probmdp(s,\act,s')\geq 0,&\quad    \forall s,s'\in S\setminus T,\, \forall\act\in\ActS, &\label{eq:well-defined_probs_mdp}\\
    \sum_{s'\in S}\probmdp(s,\act,s')=1,\label{eq:well-defined_probs_mdp1}  &\quad\forall s\in S\setminus T,\,  \forall\act\in\ActS,  \\
    p_s \geq \sum_{s'\in S} \probmdp(s,\act,s')\cdot p_{s'}&\quad   \forall s\in S\setminus T,\,\forall \act\in\ActS,    \label{eq:probcomputation_mdp}\\
        \lambda \geq p_{\sinit},&  \label{eq:probthreshold_mdp}
\end{align}
where $p_{s}$ are the variables.
For $s \in S$, the \emph{probability variable} $p_s\in[0,1]$ represents an upper bound of the probability of reaching target set $T\subseteq S$. We minimize $p_{\sinit}$ to compute the maximum probability of reaching set $T$.
The constraint~\eqref{eq:probthreshold_mdp} ensures that the probability of reaching $T$ is above the threshold $\lambda$.
This constraint is optional for stating the verification problem, but with that constraint, the LP is feasible if all policies in the MDP satisfy the reachability specification $\varphi$.

The constraints of the primal LP have the following meanings. The probability of reaching a state in $T$ from $T$ is set to one~\eqref{eq:targetprob_mdp}. The constraints \eqref{eq:well-defined_probs_mdp} and  \eqref{eq:probthreshold_mdp} ensure the validity of the transition probabilities and trivially hold for a valid MDP.
For each state $s\in S\setminus T$ and action $\act\in\ActS$, the probability induced by the \emph{maximizing policy} is a lower bound to the probability variables $p_s$~\eqref{eq:probcomputation_mdp}.
The constraint~\eqref{eq:probthreshold_mdp} ensures that the probability of reaching $T$ is below the threshold $\lambda$.

\paragraph{Expected cost specifications.}
The LP in \eqref{eq:min_mdp} -- \eqref{eq:probcomputation_mdp} considers reachability probabilities.
If we have instead an expected cost specification, we can similarly define the LP as follows:
\begin{align}
    \text{minimize} &\quad p_{\sinit}\label{eq:min_rew_mdp}\\
\text{subject to} &\nonumber \\
    p_s=0,&\quad\forall s\in G,      \label{eq:targetrew}\\
    \probmdp(s,\act,s')\geq 0,&\quad    \forall s,s'\in S\setminus G,\, \forall\act\in\ActS, &\label{eq:well-defined_rew_mdp}\\
    \sum_{s'\in S}\probmdp(s,\act,s')=1,\label{eq:well-defined_rew_mdp1}  &\quad\forall s\in S\setminus G,\,  \forall\act\in\ActS,  \\
    p_s\geq  c(s,\act) + \sum_{s'\in S} \probmdp (s,\act,s')\cdot p_{s'},&\quad    \forall s\in S\setminus G,\, \forall\act\in\ActS,
    \label{eq:rewcomputation}\raisetag{10pt}\\
    \kappa \geq p_{\sinit}&\label{eq:strategyah:lambda}.
\end{align}
We have $p_s\in\R$, as these variables represent the expected cost to reach $G$.
At $G$, the expected cost is set to zero \eqref{eq:targetrew};
The constraints \eqref{eq:well-defined_rew_mdp} and \eqref{eq:well-defined_rew_mdp1} are analogous to \eqref{eq:well-defined_probs_mdp} and \eqref{eq:well-defined_probs_mdp1}. The actual expected cost for other states is a lower bound to $p_s$ \eqref{eq:rewcomputation}.
Finally, %
$p_{\sinit}$ is bounded by the threshold $\kappa$.

\paragraph{Dual LP.} In this section, we recall the dual LP formulation to compute a policy that maximizes the probability of satisfying a reachability specification $\reachPropSymbol$ in an MDP~\citep{puterman2014markov,forejt2011quantitative}.

The variables of the dual LP formulation are following:
\begin{itemize}
    \item $x_{\sched}(s,\act)\in [0,\infty)$ for each state $s\in S \setminus T$ and action $\act\in\Act$ defines the occupancy measure of a state-action pair for the policy $\sched$, i.e., the expected number of times for taking action $\act$ in state $s$.
    \item $x_{\sched}(s) \in [0, 1]$ for each state $s \in T$ defines the probability of reaching a state $s\in T$.
\end{itemize}
\vspace{-0.07cm}
\begin{align}
    \displaystyle   &\text{maximize} \quad  \sum\nolimits_{s\in T}x_{\sched}(s)\label{eq:strategylp:obj}\\
    &\text{subject to} \nonumber \\&
     \label{eq:strategylp:welldefined_sched}     \hspace{-0.15cm} \sum_{\act\in\Act}x_{\sched}(s,\act) = \hspace{-0.05cm} \sum_{s'\in S \setminus T}\sum_{\act\in\Act}\probmdp(s',\act,s)x_{\sched}(s',\act)+\alpha_s, \quad \forall s\in S\setminus T,\\&
      x_{\sched}(s)= \sum_{s' \in S\setminus T}\sum_{\act\in\Act}\probmdp(s',\act,s)x_{\sched}(s',\act)+\alpha_s, \quad \displaystyle      \forall s\in T,  \displaystyle \label{eq:strategylp:mec_sched}\\ &
    \displaystyle       \sum\nolimits_{s \in T} x_{\sched}(s) \geq \beta\label{eq:strategylp:probthresh}
\end{align}
where $\alpha_s=1$ if $s=\sinit$ and $\alpha_s=0$ if $s\neq \sinit$.
The constraints in~\eqref{eq:strategylp:welldefined_sched} and~\eqref{eq:strategylp:mec_sched} ensure that the expected number of times transitioning to a state $s \in S$ is equal to the expected number of times to take action $\act$ that transitions to a different state $s' \in S$.
The constraint in~\eqref{eq:strategylp:probthresh} ensures that the specification $\reachPropSymbol$ is satisfied with a probability of at least $\beta$.
We determine the states with probability $0$ to reach $T$ by a preprocessing step on the underlying graph of the MDP.
To ensure that the variables $x_{\sched}(s)$ encode the actual probability of reaching a state $s \in T$, we then set the variables of the states with probability $0$ to reach $T$ to zero.

For any optimal solution $x_{\sched}$ to the LP in~\eqref{eq:strategylp:obj}--\eqref{eq:strategylp:probthresh},

\begin{align}
    \displaystyle {\sched}(s,\act)= \dfrac{x_{\sched}(s,\act)}{ \displaystyle\sum\nolimits_{\act'\in\Act}x_{\sched}(s,\act')}\label{eq:occupmeasure}
\end{align}
is an optimal policy, and $x_{\sched}$ is the occupancy measure of $\sched$, see~\citep{puterman2014markov} and~\citep{forejt2011quantitative} for details.

\section{A Case Study}

\label{mdp-human-autonomy}

\begin{figure}[t]
\centering
\includegraphics[width=.7\columnwidth]{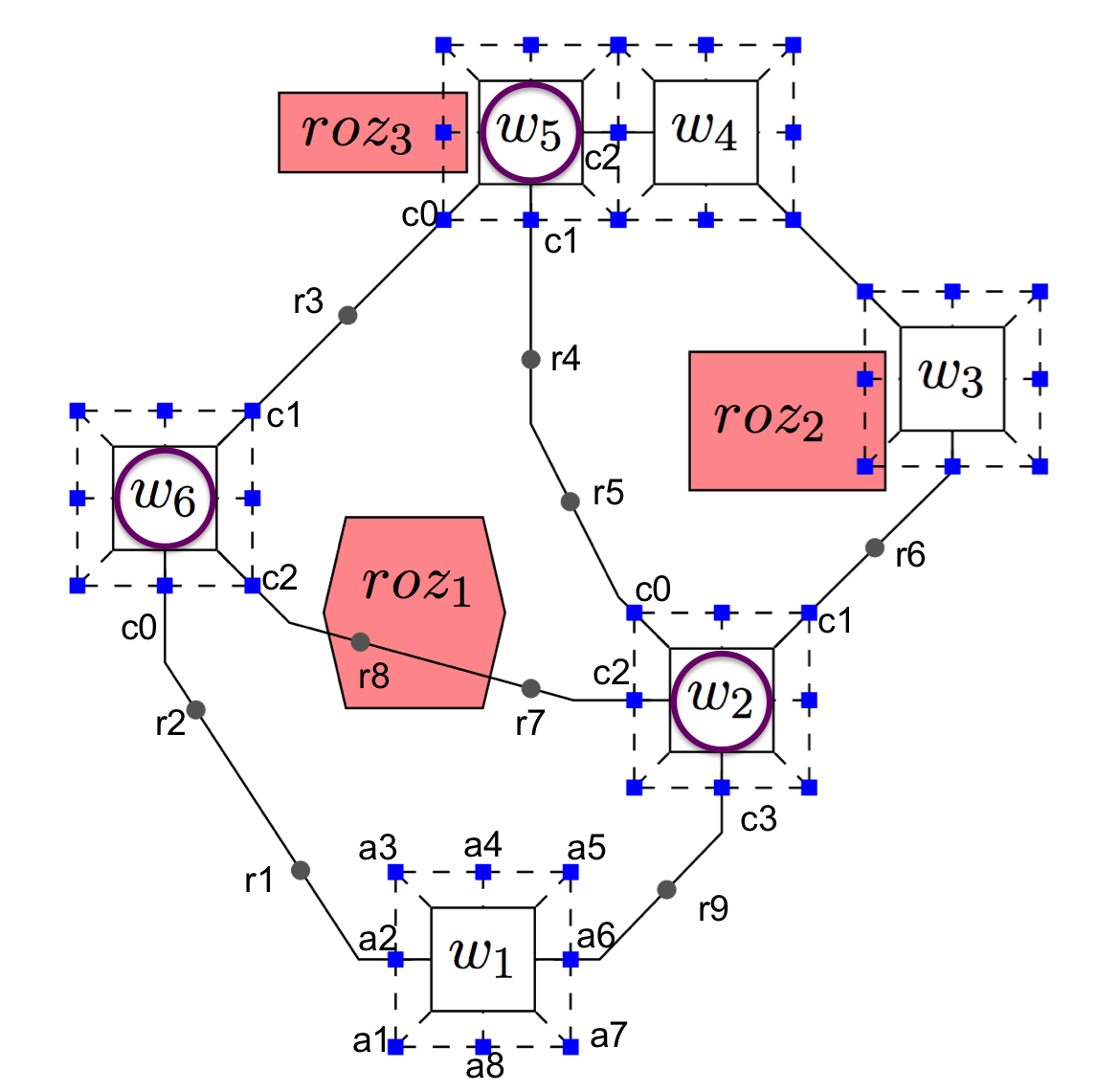}
\vspace{-0.2cm}
\caption{A road network for UAV ISR missions (adapted from \citep{HWT14}).}
\vspace{-0.5cm}
\label{fig:map}
\end{figure}

We now visit an example from \citep{feng2015controller} as an illustration of synthesis in MDPs for autonomous systems that interact with human operators, where a remotely controlled unmanned air vehicle (UAV) is used to perform
intelligence, surveillance, and reconnaissance (ISR) missions
over a road network. 
Figure \ref{fig:map} shows a map of the road network,
which has six surveillance \emph{waypoints} labeled $w_1, w_2, \dots, w_6$.
Approaching a waypoint from certain angles may be better than others,
\eg in order to obtain desired look angles on a waypoint target using an ellipsoidal loiter pattern.
Angles of approach are thus discretized in increments of 45$^\circ$ around each waypoint, 
resulting in eight \emph{angle points} $a_1, a_2, \dots, a_8$ around each waypoint.
Roads connecting waypoints are
discretized into \emph{road points} $r_1, r_2, \dots, r_9$. Red polygons represent ``restricted operating zones'' (ROZs), areas in which flying the UAV may be dangerous or lead to a higher chance of being detected by an adversary.

In current practice \citep{cooke_2009a}, at least two human operators are required for a  
UAV ISR mission: one to pilot the UAV, and the other to steer the onboard sensor 
and interpret the sensor imagery.
Here, we assume the UAV has a certain degree of autonomy that is used to
fulfill most of the piloting functions, \eg, maintaining loiter patterns around waypoints,  
selecting most of the points that comprise the route, and flying the route.
The human operator primarily performs sensor tasks, \eg
steering the onboard sensor to capture imagery of targets at waypoints.
However, the operator also retains the ability to affect some of the piloting functions of the UAV.
The operator decides how many loiters to perform at each waypoint, 
since more loiters may be needed if the operator is not satisfied 
with the sensor imagery obtained on previous loiters.
Additionally, waypoints $w_2$, $w_5$, and $w_6$ in Figure \ref{fig:map} 
will be designated as \emph{checkpoints}. At checkpoints, 
the operator can directly impact the choices made by the protocol we synthesize
by selecting different roads to be taken between waypoints.

The optimal piloting plan for the UAV varies depending on mission objectives.
Specification patterns for a variety of UAV missions are presented in \citep{HWT14},
including safety, reachability, coverage, sequencing of waypoints, \etc. 
Through a few concrete examples,
\eg surveillance of the road network with minimum fuel consumption, 
or flying to certain waypoints while trying to avoid ROZs, the goal is to synthesize the optimal UAV piloting plan for a specific mission objective,
which would be implemented by the UAV's onboard automation interface to control the route.
In particular, the results shed light on how the uncertainties and imperfections 
of a human operator's behavior affect the optimal UAV piloting plan.
Specifically, what is the influence of 
an operator's proficiency, workload, and fatigue level on UAV mission performance? 
Can we synthesize individualized optimal UAV piloting plans for different operators?
Can the automation provide informative feedback to operators to assist them in decision-making?

We now introduce the models for the operator, the UAV and the interactions between the two.

\paragraph{The operator model.}

\begin{figure}[t]
\centering
\includegraphics[width=.6\columnwidth]{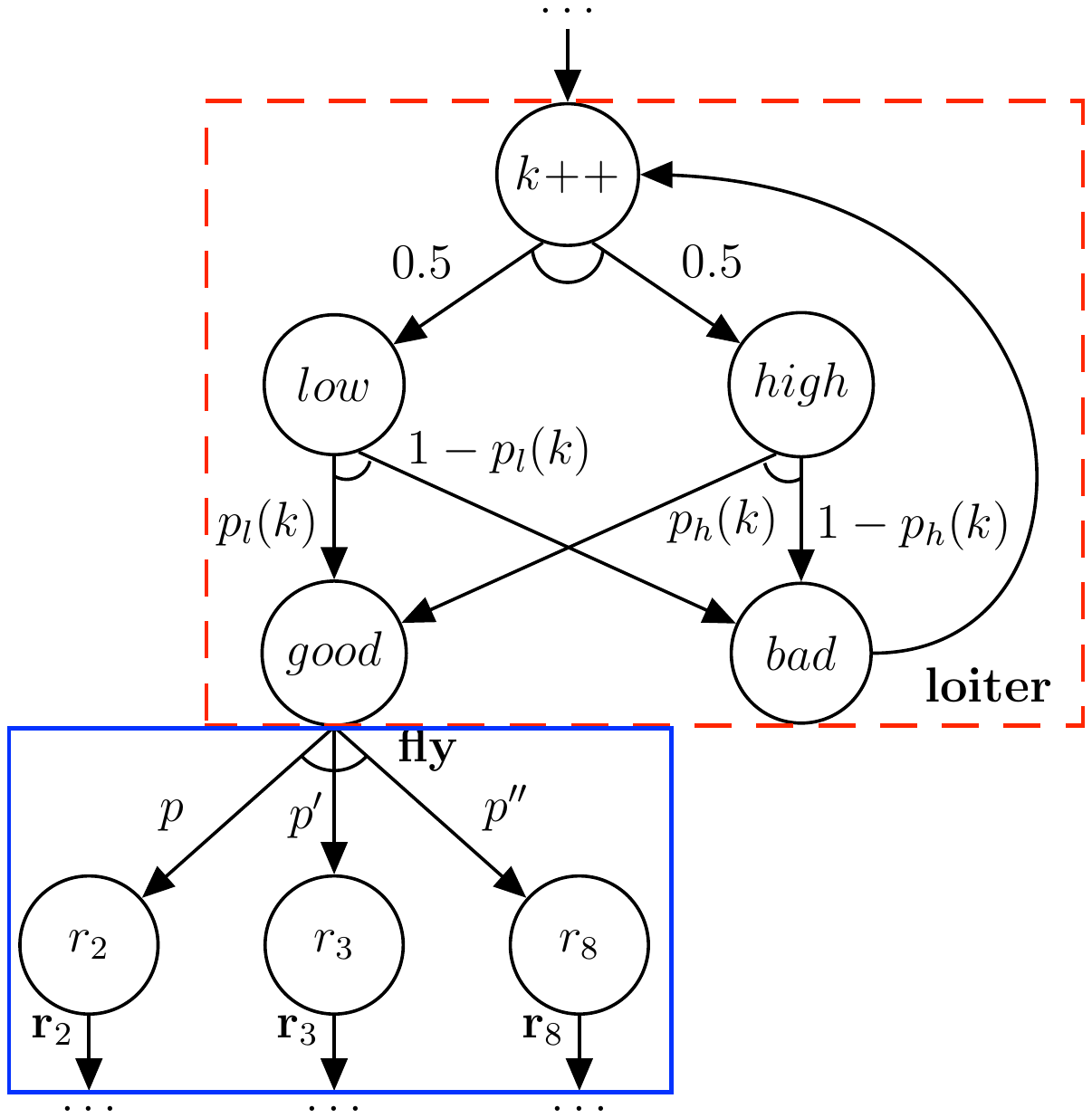}
\vspace{-0.3cm}
\caption{A fragment of the operator model $M_{\mathsf{OP}}$ 
representing the operator's behavior.
The red dashed square highlights the common behavior that is repeated at all waypoints,
while the blue solid square indicates specific choices of roads at waypoint $w_6$.
The operator's behavior at other waypoints is omitted from the figure, indicated by $\cdots$
}
\vspace{-0.5cm}
\label{fig:operator}
\end{figure}

We build abstractions of the operator's possible behavior 
as a probabilistic model $M_{\mathsf{OP}}$.
Figure \ref{fig:operator} shows a fragment of the model, 
representing the possible behavior 
at waypoint $w_6$.
There is a non-negative integer variable $k$ 
counting the number of sensor tasks 
performed by the operator
since the beginning of the mission.
The updates ``$k$++'' represent
increasing the value of $k$ by one. 
The purpose of using $k$ in the model is to measure the operator's fatigue level. 
To obtain a finite state model, let the value of $k$ stop increasing once it 
reaches a certain threshold $T$
(a constant that will be used later in modeling fatigue).

In general, operators' workload levels are driven by a number of factors 
including mission characteristics, 
\eg how many UAVs the operator supervises simultaneously 
and the phase of the mission. 
For simplicity and to reduce the complexity of the models 
(so that the results discussed later are easier to interpret), 
we model the operator's workload 
as a uniform distribution over two levels:  \emph{low} and \emph{high}.
Operators' accuracy on vigilance tasks tends to decline 
with lower levels of proficiency and higher levels of workload \citep{boff_1988a}.
We model an operator's accuracy in steering sensors to 
capture high resolution imagery of targets as probability distributions 
that correlate with proficiency, workload, and fatigue. 
Specifically, when the operator's workload level is \emph{low}, the probabilities of capturing 
\emph{good} and \emph{bad} quality imagery are $p_l(k)$ and $1-p_l(k)$, respectively.
Here $p_l(k)$ is a function over the variable $k$ such that 
$p_l(k)=p_l(0)$ if $k<T$ and $p_l(k)=f \cdot p_l(0)$ if $k\ge T$,
where $p_l(0)$ is the initial parameter value of the accuracy function, 
$T$ is the fatigue threshold mentioned earlier, and $f$ is a fatigue discount factor.  We define the accuracy function $p_h(k)$ for \emph{high} workload analogously.
Note that $p_l(k) \ge p_h(k)$ for any $k$, modeling the fact that
an operator tends to make more errors under higher levels of workload and stress.
Furthermore, more proficient operators have higher values for the accuracy parameters 
$p_l(0)$ and $p_h(0)$.
If the quality of the captured imagery is \emph{bad}, the operator would ask the UAV to continue to \emph{loiter}
at the current waypoint in order to collect more sensor imagery;
otherwise, the operator allows the UAV to \emph{fly} to another waypoint. 
At each waypoint, the operator repeats the aforementioned behavior 
(the red dashed square in Figure \ref{fig:operator}).

The operator selects the next road for the UAV at waypoints that are checkpoints 
($w_6$ for example),
while the UAV controller chooses the road at any non-checkpoint waypoint.
As illustrated in the blue solid square in Figure \ref{fig:operator},
we model the operator's choices at $w_6$ as following a certain probability distribution,
i.e.,  picking the roads that connect to 
neighboring road points $r_2$, $r_3$, and $r_8$ with probabilities $p$, $p'$, and $p''$, 
respectively (note that $p + p' + p''=1$).

Suppose the operator chooses $r_3$. According to the map shown in Figure \ref{fig:map},
the next waypoint is $w_5$. 
For simplicity, the operator's behavior at $w_5$ is omitted from Figure \ref{fig:operator}.

\begin{figure}[t]
\centering
\includegraphics[width=.95\columnwidth]{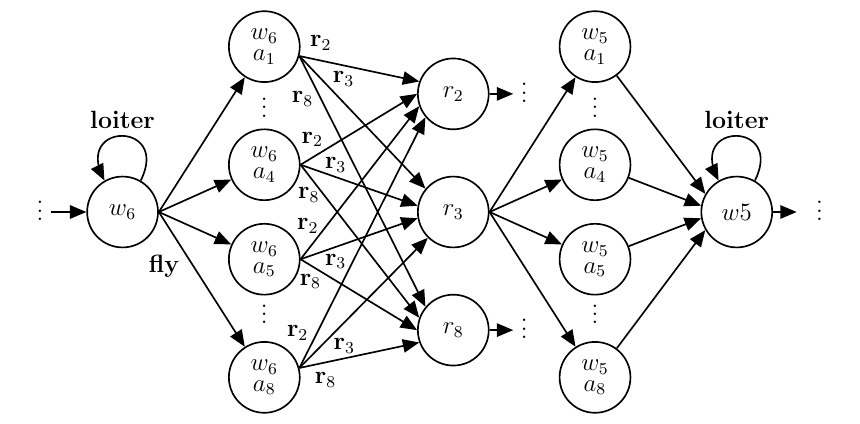}
\vspace{-0.3cm}
\caption{A fragment of the UAV model $M_{\mathsf{UAV}}$,
representing the UAV loitering and flying over the waypoints $w_6$ and $w_5$.}
\vspace{-0.5cm}
\label{fig:uav}
\end{figure}

\paragraph{The UAV model.}
We model the UAV's piloting behavior as an MDP $M_\mathsf{UAV}$, 
which contains 63 states (6 waypoints, 6$\times$8 angle points, and 9 road points).
At any waypoint or road point, the UAV can nondeterministically fly to 
a neighboring angle point or road point. 
These nondeterministic choices need to be resolved by a strategy.
Figure \ref{fig:uav} shows a fragment of the UAV model\footnote{Our models
are shown with several distributions associated with an action name
but after composition this can easily be resolved through renaming,
and we obtain MDPs.}, illustrating how the UAV loiters and flies 
over waypoints $w_6$ and $w_5$.
If the UAV receives a loiter instruction from the operator, it loiters 
at the current waypoint, allowing the operator to capture more sensor imagery;
otherwise, the UAV randomly picks one of the eight angle points $a_1,\cdots, a_8$ to exit $w_6$.
Then, a nondeterministic choice between three roads 
$r_2$, $r_3$, and $r_8$
needs to be resolved. 
Suppose $r_3$ is chosen by the operator;
then the UAV can fly to the waypoint $w_5$ and approach it via
one of the eight angles, or the UAV can also fly back to the waypoint $w_6$ 
(for clarity, this choice is not drawn in Figure \ref{fig:uav}).

\paragraph{The operator-UAV interactions.}
We model interactions between the operator and the UAV by composing 
$M_{\mathsf{OP}}$ and $M_{\mathsf{UAV}}$, 
which synchronize over common actions \emph{loiter} and \emph{fly},
and obtain a product MDP $M_{\mathsf{OP}} \| M_{\mathsf{UAV}}$, see~\citep{BK08} for a formal product model definition.
Synchronization between actions in our models
abstracts the concrete process of
interchanging information between the operator and the UAV,
which is assumed via a reliable communication protocol.
The model does not distinguish between
``sender'' and ``receiver'',
or outputs and inputs, of a protocol.
We can think of the operator initiating the \emph{loiter} action,
which the UAV can receive at any waypoint
(see the self-loops at $w_6$ and $w_5$ in Figure \ref{fig:uav}).
The \emph{fly} action is initiated as well by the operator,
but we can think of the UAV deciding which flight direction to take
(see the \emph{fly} transitions at $w_6$ in Figure \ref{fig:uav}).
Note that synchronization between actions
in the model also assumes that the 
operator and UAV synchronize their behaviors temporally.
 
Since $M_{\mathsf{OP}}$ is a discrete time \gls{mc} and has no nondeterminism,
synthesizing a strategy $\sigma$ for the MDP $M_{\mathsf{OP}} \| M_{\mathsf{UAV}}$
yields a strategy $\sigma'$ for $M_{\mathsf{UAV}}$ such that
$(M_{\mathsf{OP}} \| M_{\mathsf{UAV}})^\sigma = M_{\mathsf{OP}} \| M_{\mathsf{UAV}}^{\sigma'}$.
The strategy $\sigma'$
operates on the MDP model $M_{\mathsf{UAV}}$ of the UAV.
For $\sigma'$ to be implemented as a flight controller for the UAV,
it must know which state of $M_{\mathsf{UAV}}$ the UAV is in,
and hence, from the point of view of the strategy,
the model transitions are triggered by the UAV's behavior.

\paragraph{Representative analysis results.}

We present representative results obtained using the above MDP model.

\begin{figure}[t]
\centering
\includegraphics[width=.95\columnwidth]{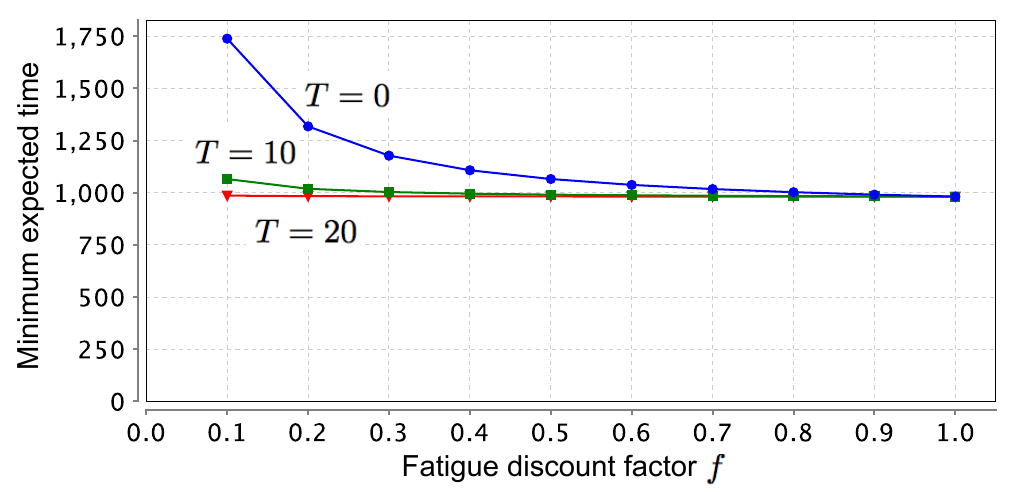}
\vspace{-0.3cm}
\caption{The effect of operator fatigue 
on minimum expected mission completion time,
for different values of the fatigue threshold $T$ and discount factor $f$
(with fixed parameters $p_l(0)=0.9$ and $p_h(0)=0.8$).}
\vspace{-0.5cm}
\label{fig:fatigue}
\end{figure}

Consider a UAV surveillance mission that requires covering all six waypoints in Figure \ref{fig:map}, 
where the objective is to complete the mission as fast as possible. 
Assume that each loiter takes 10 time units
and flying between any neighboring waypoint and/or road point takes 60 time units. 
Figure \ref{fig:fatigue} illustrates the influence of the operator's fatigue threshold $T$ and 
discount factor $f$ on the minimum expected time to complete the mission.
The general trend is that the UAV completes the mission faster 
if the operator has a higher fatigue threshold $T$ (i.e.,  less likely to get tired)
or a larger value of $f$ (i.e.,   the accuracy is less discounted).
The best UAV performance (i.e.,  the smallest expected mission completion time)
is achieved when $f=1$, that is, there is no accuracy discount due to fatigue.

\begin{figure}[t]
\centering
\includegraphics[width=.95\columnwidth]{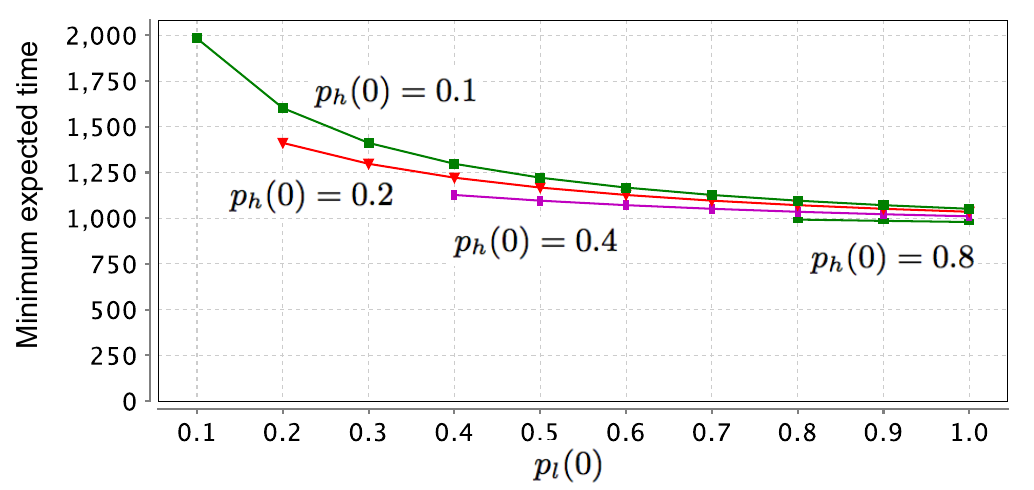}
\vspace{-0.3cm}
\caption{The effect of operator proficiency and workload 
on minimum expected mission completion time,
for different initial values of the accuracy functions $p_l(0)$ and $p_h(0)$
(with fixed parameters $T=10$ and $f=0.7$).}
\vspace{-0.5cm}
\label{fig:accu}
\end{figure}

The operator's accuracy in steering sensors and capturing good quality imagery are 
affected by proficiency and workload. 
Figure \ref{fig:accu} illustrates the influence of accuracy parameters $p_l(0)$ and $p_h(0)$
on the minimum expected time of finishing the mission (i.e.,  covering all six waypoints).
The trends show that a more proficient operator who has higher values for $p_l(0)$ and $p_h(0)$
can complete the mission faster.
In addition, the more accuracy declines due to high workload,
i.e.,  the larger the gap between $p_l(0)$ and $p_h(0)$,
the longer the time needed to complete the mission. 

\section{Parameter Synthesis for Markov Decision Processes}
\label{sec:parameter-synthesis}

In this section, we consider parametric Markov decision processes (parametric MDPs) whose transitions probabilities are affine functions of a finite set of parameters.
Every fixed set of parameters induce an MDP, and the goal is to find a good set of parameters such that the induced MDP satisfies the specifications.
We develop an approach that utilizes a \gls{scp} method.
The techniques improve the runtime and scalability by multiple orders of magnitude compared to black-box \gls{scp} by merging ideas from convex optimization and probabilistic model checking.
We demonstrate the approaches on a satellite collision avoidance problem with hundreds of thousands of states and tens of thousands of parameters and their scalability on a wide range of commonly used benchmarks.

\subsection{Parametric Markov Decision Processes}
A more general description of expressing the transition and reward function of an MDP is to define them as functions over parameters whose values are left unspecified~\citep{Daw04, lanotte, param_sttt}. 
Such \emph{parametric MDPs} describe uncountable sets of MDPs. %
A well-defined instantiation of the parameters yields an \emph{instantiated, parameter-free} MDP.
For a given finite-state parametric MDP, the \emph{parameter synthesis problem} is to compute a parameter instantiation such that the instantiated MDP satisfies a specification.
We can also think the parameter synthesis problem as a \emph{design problem}, where the objective is to compute an optimal design as a function of the parameters of the MDP.
This design problem can also include computing an optimal policy for the parametric MDP.

Traditionally, approaches for solving the parameter synthesis problems have been built around the notion of abstracting the parametric model into a solution function.
The solution function is the probability of satisfying the temporal logic specification as a function of the model parameters~\citep{Daw04,param_sttt,dehnert-et-al-cav-2015,DBLP:journals/tse/FilieriTG16}. 
The solution function can be exploited by the probabilistic model checking tools \tool{PARAM}~\citep{param_sttt}, \tool{PRISM}~\citep{KNP11} and \tool{Storm}~\citep{DBLP:conf/cav/DehnertJK017} to solve the parameter synthesis problem. 
However, this function is exponentially large in the parameters, and solving the problem is again exponential in the number of parameters, making the whole approach doubly exponential~\citep{DBLP:journals/iandc/BaierHHJKK20}. 
Consequently, these approaches typically can handle millions of states but only a \emph{handful of parameters}. 
Moreover, these approaches require a fixed policy or has to introduce a parameter for every state/action-pair in the MDP.

Orthogonally, 
~\citet{quatmann2016parameter} address an alternative parameter synthesis problem which focuses on proving the absence of parameter instantiations. 
The method iteratively solves simple stochastic games. 
~\citet{DBLP:conf/atva/SpelJK19} consider proving that the parameters behave monotonically, allowing for faster sampling-based approaches. 
However, this method is limited to a few parameters.
A recent survey on parameter synthesis in Markov models can be found in~\citep{DBLP:journals/corr/abs-1903-07993}.

Further variations of parameter synthesis, e.g., consider statistical guarantees for parameter synthesis, often with some prior on the parameter values~\citep{DBLP:conf/tacas/BortolussiS18,calinescu2016fact}.
These approaches cannot provide the absolute guarantees on an answer that the methods in this dissertation provide.

Parametric MDPs generalize interval models~\citep{DBLP:conf/tacas/SenVA06,chen2013complexity}. 
Such interval models have also been considered with convex uncertainties~\citep{seshia_et_al_cav_13,hahn2017multi,wu2008reachability,chen2013complexity}.
However, the resulting problems with interval models are easier to solve due to the lack of dependencies (or couplings) between parameters in different states.

\subsection{Preliminaries for Parametric Markov Decision Processes}
\label{atva2018sec:preliminaries}
Let $V=\{x_1,\ldots,x_n\}$ be a finite set of \emph{variables} over the real numbers $\R$. The set of multivariate polynomials over $V$ is $\mathbb{Q}[V]$. An \emph{instantiation} for $V$ is a function $\mathbf{v}\colon V \rightarrow \R$.

\begin{definition}[(Affine) \gls{pmdp}]\label{atva2018def:pmdp}
	A \emph{\gls{pmdp}} is a tuple $\pMdpInit$ with a finite set $S$ of \emph{states}, an \emph{initial state} $\sinit \in S$, a finite set $\Act$ of \emph{actions}, a finite set $\Paramvar$ of real-valued variables \emph{(parameters)} and a \emph{transition function} $\probpmdp \colon S \times \Act \times S \rightarrow \mathbb{Q}[V]$.
	A \gls{pmdp} is \emph{affine} if $\probpmdp(s,\act,s')$ is an affine function of $\Paramvar$ for every $s,s'\in S$ and $\act\in\Act$.
\end{definition}
For $s \in S$,  $\ActS = \{\act \in \Act \mid \exists s'\in S.\,\probpmdp(s,\,\act,\,s') \neq 0\}$ is the set of \emph{enabled} actions at $s$.
Without loss of generality, we require $\ActS \neq \emptyset$ for $s\in S$.
If $|\ActS| = 1$ for all $s \in S$, $\mdp$ is a \emph{parametric discrete-time Markov chain (pMC)}. 
MDPs can be equipped with a state--action \emph{cost function} $\rewFunction \colon S \times \Act \rightarrow \R_{\geq 0}$.

A \gls{pmdp} $\mdp$ is a \emph{Markov decision process (MDP)} if the transition function yields \emph{well-defined} probability distributions, \ie, $\probpmdp \colon S \times \Act \times S \rightarrow [0,1]$ and $\sum_{s'\in S}\probpmdp(s,\act,s') = 1$ for all $s \in S$ and $\act \in \ActS$. 
Applying an \emph{instantiation} $\mathbf{v}\colon V \rightarrow \R$ to a \gls{pmdp} $\mdp$ yields an \emph{instantiated MDP} $\mdp[\mathbf{v}]$ by replacing each $f\in\mathbb{Q}[V]$ in $\mdp$ by $f[\mathbf{v}]$.
An instantiation $\mathbf{v}$ is \emph{well-defined} for $\mdp$ if 
the resulting model $\mdp[\mathbf{v}]$ is an MDP.

\subsection{Formal Problem Statement for Parameter Synthesis}
\label{atva2018sec:problem}
In this section, we state the parameter synthesis problem, which is to compute a parameter instantiation such that the instantiated MDP satisfies the given temporal logic specification.
We then discuss the nonlinear program formulation of the parameter synthesis problem, which forms the basis of the considered solution method.
	\begin{problem}[Parameter Synthesis Problem]\label{atva2018prob:pmdpsyn}
		Given a parametric MDP $\pMdpInit$, and a reachability specification $\reachPropSymbol=\reachProplT$, 
		compute a well-defined
		instantiation $\instantiation\colon V \rightarrow \R$ for $\parammdp$ such that $\parammdp[\instantiation]\models\reachPropSymbol$.
	\end{problem}
In words, we seek for an instantiation of the parameters that satisfies $\reachPropSymbol$ for all possible strategies in the instantiated MDP.
We show necessary adaptions for an expected cost specification $\ereachPropSymbol=\expRewPropkT$ later.

For a given well-defined instantiation $\instantiation$, Problem~\ref{atva2018prob:pmdpsyn} can be solved by verifying whether $\parammdp[\instantiation]\models\reachPropSymbol$. 
The standard formulation uses a linear program (LP) to minimize the probability $p_{\sinit}$ of reaching the target set $T$ from the initial state $\sinit$ while ensuring 
that this probability is realizable under any strategy~\citep[Ch.\ 10]{BK08}.
The straightforward extension of this approach to pMDPs to \emph{compute} a satisfiable instantiation $\mathbf{v}$ yields the following nonlinear program (NLP)~\citep{cubuktepe2017sequential,atvaqcqp}: with the variables $p_s$ for $s \in S$, and the \emph{parameter variables} in $V$ in the transition function $\probpmdp(s,\act,s')$ for $s,s' \in S$ and $\act \in \ActS$:
\begin{align}
\text{minimize} &\quad p_{\sinit}\label{atva2018eq:min_mdp}\\
\text{subject to} &\nonumber \\
\forall s\in T,	 &\quad p_s=1,\label{atva2018eq:targetprob_mdp}\\
\forall s,s'\in S\setminus T,\, \forall\act\in\ActS,	 &\quad \probpmdp(s,\act,s')\geq 0,\label{atva2018eq:well-defined_probs_mdp}\\
\forall s\in S\setminus T,\, \forall\act\in\ActS,	 &\quad \sum_{s'\in S}\probpmdp(s,\act,s')=1,\label{atva2018eq:well-defined_probs_mdp1}\\
&\quad  \lambda \geq p_{\sinit},\label{atva2018eq:probthreshold_mdp}\\
\forall s\in S\setminus T,\,\forall \act\in\ActS,	&\quad p_s \geq \sum_{s'\in S}	\probpmdp(s,\act,s')\cdot p_{s'}\label{atva2018eq:probcomputation_mdp}.
\end{align}

For $s \in S$, the \emph{probability variable} $p_s\in[0,1]$ represents an upper bound of the probability of reaching target set $T\subseteq S$.
The \emph{parameters} in the set $V$ enter the NLP as part of the functions from $\mathbb{Q}[V]$ in the transition function $\probpmdp$.
The constraint~\eqref{atva2018eq:probthreshold_mdp} ensures that the probability of reaching $T$ is below the threshold $\lambda$.
This constraint is optional for stating the problem, but we use the constraint for finding a parameter instantiation that satisfies the specification $\varphi$.
We minimize $p_{\sinit}$ to assign probability variables their minimal values with respect to the parameters $V$.

The probability of reaching a state in $T$ from $T$ is set to one~\eqref{atva2018eq:targetprob_mdp}.
The constraints~\eqref{atva2018eq:well-defined_probs_mdp} and~\eqref{atva2018eq:well-defined_probs_mdp1} ensure \emph{well-defined} transition probabilities.
Recall that $\probpmdp(s,\act,s')$ is an affine function in $V$.
Therefore, the constraints~\eqref{atva2018eq:well-defined_probs_mdp} and~\eqref{atva2018eq:well-defined_probs_mdp1} only depend on the parameters in $V$, and they are affine in the parameters.
Constraint~\eqref{atva2018eq:probthreshold_mdp} is optional but necessary later, and ensures that the probability of reaching $T$ is below the threshold $\lambda$.
For each state $s\in S\setminus T$ and action $\act\in\ActS$, the probability induced by the \emph{maximizing scheduler} is a lower bound to the probability variables $p_s$~\eqref{atva2018eq:probcomputation_mdp}.
To assign probability variables to their minimal values with respect to the parameters in $V$, $p_{\sinit}$ is minimized in the objective~\eqref{atva2018eq:min_mdp}.
We state the correctness of the NLP in Proposition~\ref{atva2018prop:1}.

\begin{proposition}\label{atva2018prop:1}
	The NLP in \eqref{atva2018eq:min_mdp} -- \eqref{atva2018eq:probcomputation_mdp} computes the \emph{minimal probability} of reaching $T$ under a \emph{maximizing} strategy,  and an instantiation $\mathbf{v}$ is feasible to the NLP if and only if $\parammdp[\mathbf{v}]\models\reachPropSymbol$.
\end{proposition}
\begin{proof}
	The NLP in \eqref{atva2018eq:min_mdp} -- \eqref{atva2018eq:probcomputation_mdp} is an extension of the LP in~\citep[Theorem 10.105]{BK08}. We refer to~\citep[Theorem 4.20]{Jun20} for a formal proof.
\end{proof}

\begin{figure}[t]
	\centering
	\input{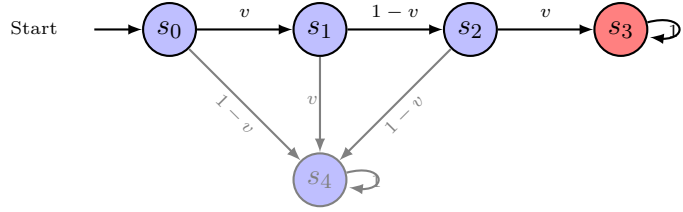}	
	\caption{A parametric MC with a single parameter $v$.}
	\label{atva2018fig:pmc_reform}
\end{figure}

\begin{remark}[Graph-Preserving Instantiations]
	In the LP formulation for MDPs, states with probability $0$ to reach $T$ are determined via a preprocessing on the underlying graph, and their probability variables are set to zero to ensure that the variables encode the actual reachability probabilities.
	This preprocessing requires the underlying graph of the parametric MDP to be preserved under any valuation of the parameters. 
	Thus, as in~\citep{param_sttt,dehnert-et-al-cav-2015}, we consider only graph-preserving valuations. 
	Concretely, we exclude valuations $\instantiation$ with $f[\instantiation]=0$ for $f\in\probpmdp(s,\act,s')$ for all $s,s'\in S$ and $\act\in\Act$.
	We replace the set of constraints \eqref{atva2018eq:well-defined_probs_mdp} by%
	\begin{align}
	\forall s,s'\in S.\, \forall\act\in\ActS,	 &\quad \probpmdp(s,\act,s')\geq \epsgraph,\label{atva2018eq:well-defined_eps}
	\end{align}
	where $\epsgraph>0$ is a small constant. 
	
\end{remark}
We demonstrate the constraints for the NLP in~\eqref{atva2018eq:min_mdp} -- \eqref{atva2018eq:well-defined_eps} for a parametric MC by Example~\ref{atva2018ex:nlp}.
\begin{example}\label{atva2018ex:nlp}
	Consider the parametric Markov chain in Fig.~\ref{atva2018fig:pmc_reform} with parameter set $V=\{v\}$, initial state $s_0$, and target set $T = \{s_3\}$. Let $\lambda$ be an arbitrary constant.
	The NLP in~\eqref{atva2018eq:nlp-ex1} -- \eqref{atva2018eq:nlp-ex2} minimizes the probability of reaching $s_3$ from the initial state:
	\begin{align}
	\textnormal{minimize} & \quad p_{s_0} \label{atva2018eq:nlp-ex1} \\
	\textnormal{subject to} &\quad p_{s_3}=1,\\
	&\quad \lambda \geq p_{s_0}, \\
	&\quad  p_{s_0} \geq v\cdot p_{s_1},\\
	&\quad p_{s_1} \geq (1-v)\cdot p_{s_2},\\
	&\quad  p_{s_2} \geq v\cdot p_{s_3},  \\
	&\quad 1-\epsgraph \geq v\geq \epsgraph. \label{atva2018eq:nlp-ex2}
	\end{align}
\end{example}
\paragraph{Expected Cost Specifications.}
The NLP in \eqref{atva2018eq:min_mdp} -- \eqref{atva2018eq:well-defined_eps} considers reachability probabilities.
If we have instead an expected cost specification $\ereachPropSymbol=\expRewProp{\kappa}{G}$, we replace \eqref{atva2018eq:targetprob_mdp}, \eqref{atva2018eq:probthreshold_mdp}, and \eqref{atva2018eq:probcomputation_mdp} in the NLP by the following constraints:
\begin{align}
\forall s\in G,	 &\; p_s=0,\label{atva2018eq:targetrew}\\
\forall s\in S\setminus G,\, \forall\act\in\ActS,	&\; p_s\geq  c(s,\act) + \sum_{s'\in S}	\probpmdp (s,\act,s')\cdot p_{s'},
\label{atva2018eq:rewcomputation}\\
&\; \kappa \geq p_{\sinit}\label{atva2018eq:strategyah:lambda}.
\end{align}
We have $p_s\in\R$, as these variables represent the expected cost to reach $G$. 
At $G$, the expected cost is set to zero \eqref{atva2018eq:targetrew}, and the actual expected cost for other states is a lower bound to $p_s$ \eqref{atva2018eq:rewcomputation}.
Finally, %
$p_{\sinit}$ is bounded by the threshold $\kappa$. 

For an affine parametric MDP $\parammdp$, the functions in the resulting NLP ~\eqref{atva2018eq:min_mdp} -- \eqref{atva2018eq:probthreshold_mdp} for parametric MDP synthesis are affine in $V$.
However, the functions in the constraints~\eqref{atva2018eq:probcomputation_mdp} are \emph{quadratic}, as a result of multiplying affine functions occurring in $\probpmdp$ with the probability variables $p_{s'}$.
The problem in~\eqref{atva2018eq:min_mdp} -- \eqref{atva2018eq:probcomputation_mdp} is a quadratically constrained quadratic program (QCQP)~\citep{boyd_convex_optimization} and is generally nonconvex~\citep{atvaqcqp}, and computationally hard to solve. In the rest of the chapter, we discuss two methods to obtain a \emph{locally optimal solution} to the problem in ~\eqref{atva2018eq:min_mdp} -- \eqref{atva2018eq:probcomputation_mdp}.

\subsection{Sequential Convex Programming}
\label{atva2018sec:scp}
In this section, we discuss a method by \citet{cubuktepe2021convex} for solving the parameter synthesis problem, which is a \gls{scp} approach with trust region constraints~\citep{yuan2015recent,mao2018successive,chen2013optimality}.
The \gls{scp} method computes a locally optimal solution by iteratively approximating a nonconvex optimization problem.
The resulting approximate convex problem is an LP, and the convexified functions are no longer upper bounds of the original functions. 
Therefore, approximation may generate optimal solutions in the convexified problem that are infeasible in the original problem.
Therefore, we include \emph{trust regions} and an additional model checking step to ensure that the new solution improves the objective.
The trust regions ensure that the resulting LP accurately approximates the nonconvex QCQP.
If the new solution indeed improves the objective, we accept and update the assignment of the variables and enlarge the trust region.
Otherwise, we contract the trust region, and do not update the assignment of the variables.
\subsubsection{Constructing the Affine Approximation}
We now explain in detail how we linearize the bilinear functions in the constraints in~\eqref{atva2018eq:probcomputation_mdp}.
Recall that this constraint appears as
\begin{align*}
			\forall s\in S\setminus T,\,\forall \act\in\ActS,	&\quad p_s \geq \sum_{s'\in S}	\probpmdp(s,\act,s')\cdot p_{s'}.
\end{align*}

Similar to the previous seciton, consider the bilinear function in the above constraint
\begin{equation}
h(s,\act,s')=\probpmdp(s,\act,s')\cdot p_{s'}
\end{equation}
and let 
\begin{equation}
\probpmdp(s,\act,s')=2d\cdot y+c, \text{ and } p_{s'}=z,
\end{equation} where $y$ is the parameter variable, $z$ is the probability, and $c, d$ are constants, similar to the previous chapter. We then convexify $h(s,\act,s')$ as
\begin{equation}
h_{\textrm{a}}(s,\act,s') \colonequals
2 d\cdot((\hat{y}+\hat{z})+\hat{y}\cdot(z-\hat{z})+\hat{z}\cdot(y-\hat{y}))+c\cdot z,
\end{equation} where $\langle \hat{y}, \hat{z}\rangle$ are any assignments to $y$ and $z$.
Note that the function $h_{\textrm{a}}(s,\act,s')$ is  affine in the parameter variable $y$ and the probability variable $z$.

After the linearization, the set of constraints~\eqref{atva2018eq:probcomputation_mdp} is replaced by the convex constraints
\begin{align}
&\forall s \in S \setminus T.\, \forall \act \in \Act(s),
\quad p_s \geq  \sum_{s' \in S} h_{\textrm{a}}(s,\act,s').\nonumber
\end{align}
\begin{remark}
	If the parametric MDP is not affine, i.e., $\probpmdp(s,\act,s')$ is not affine in $\Paramvar$ for every $s,s'\in S$ and $\act\in\Act(s)$, then $h(s,\act,s')$ will not be a quadratic function in $V$ and probability variables $p_s'$.
	In this case, we can compute $h_a(s,\act,s')$ by computing a first order approximation with respect to $\Paramvar$ and $p_{s'}$ around the previous assignment.
\end{remark}

We use \emph{penalty variables} $k_{s}$ for all $s\in S \setminus T$ to all linearized constraints, ensuring that they are always feasible.
We minimize the sum of penalty variables to minimize the violation of the constraints in~\eqref{atva2018eq:slackvariable_scp}
However, the functions in these constraints do not over-approximate the functions in the original constraints. 
Therefore, a feasible solution to the linearized problem is potentially infeasible to the parameter synthesis problem.
To make sure that the linearized problem accurately approximates the parameter synthesis problem, we use a trust region contraint around the previous parameter instantiations.
The resulting LP is:%
\begin{align}
			\textnormal{minimize} &\quad p_{\sinit}+ \tau\sum_{\forall s\in S\setminus T}k_{s}\label{atva2018eq:min_qcqp_scp}\\
			\textnormal{subject to}\nonumber\\
			\forall s\in  T,	 &\quad p_s=1,\label{atva2018eq:targetprob_scp}\\
			\forall s,s'\in S\setminus T, \forall\act\in\ActS,	 &\quad \probpmdp(s,\act,s')\geq \epsgraph, \label{atva2018eq:well-defined_probs_scp}\\
			\forall s\in S,\forall \act \in \ActS,	 &\quad \sum_{s'\in  S}\probpmdp(s,\act,s')=1,\label{atva2018eq:well-defined_probs_scp1}\\
			&\quad \lambda \geq p_{\sinit},
								 \label{atva2018eq:probthreshold_scp}\\
	\forall s\in S\setminus T, \forall \act \in \ActS,	&\quad k_{s} + p_s \geq  \sum_{s' \in \setminus T}   h_{\textrm{a}}(s,\act,s'),\label{atva2018eq:probcomputation_scp}\\
			\forall s\in S\setminus T, &\quad k_{s} \geq 0,
			 \label{atva2018eq:slackvariable_scp}\\
			  \forall s \in S\setminus T, & \quad \nicefrac{\hat{p}_{s}}{\delta'}\leq p_s\leq \hat{p}_s \cdot \delta'\label{atva2018eq:trust_region_scp},\\
				\forall s,s'\in S\setminus T,\forall \act \in \ActS,& \nonumber	\\
			 	\quad\nicefrac{\hat{\probpmdp}(s,\act,s')}{\delta'} \leq &\probpmdp(s,\act,s')\leq \hat{\probpmdp}(s,\act,s')\cdot \delta',\label{atva2018eq:trust_region_scp2}
		\end{align}
where $\tau > 0$ is a constant, and $\hat{\probpmdp}(s,\act,s')$ and $\hat{p}_s$ denotes the previous assignment for the parameter and probability variables. 
The constraints~\eqref{atva2018eq:trust_region_scp}--\eqref{atva2018eq:trust_region_scp2} are the trust region constraints. $\delta > 0$ is the size of the trust region, and $\delta'=\delta+1$. 
	We demonstrate the linearization in Example~\ref{atva2018ex:scp}.

\begin{figure}[h!]
	\centering
	\input{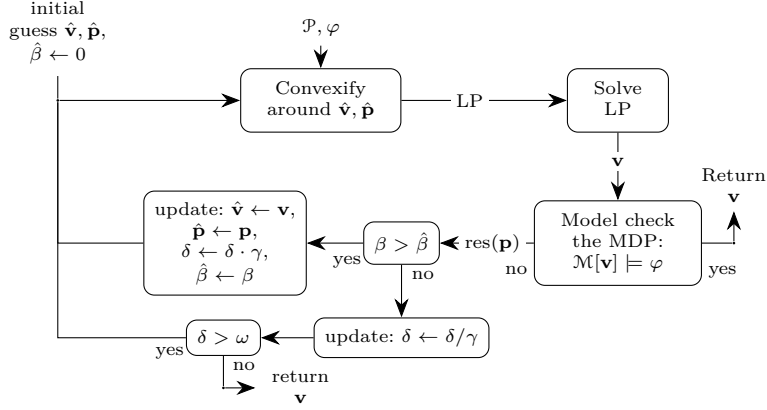}
	\caption{\gls{scp} with model checking in the loop. 
		The NLP~\eqref{atva2018eq:min_mdp}--\eqref{atva2018eq:probcomputation_mdp} is linearized around $\hat{\instantiation}, \hat{\mathbf{p}}$.
		Then, we solve the LP~\eqref{atva2018eq:min_qcqp_scp}--\eqref{atva2018eq:trust_region_scp} an optimal solution to the parameter values, denoted by $\hat{\instantiation}$.
		After each iteration, we model check the instantiated MDP $\mdp[\instantiation]$ to determine whether the specification is satisfied. 
		If the instantiated MDP satisfies the specification, we return the parameter instantiation. 
		Otherwise, we check whether the reachability probability, denoted by $\beta$, is improved compared to the previous iteration, denoted by $\hat{\beta}$.
	If the probability is improved, we accept this step, update the assignment for the parameters and the probability variables, and increase the size of the trust region $\delta$ by $\gamma$.
Otherwise, we do not update the assignment, and decrease the size of the trust region.}
	\label{atva2018fig:scpwithmc}
\end{figure}

\begin{example}
	\label{atva2018ex:scp}
	Recall the parametric Markov Chain (MC) in Fig.~\ref{atva2018fig:pmc_reform} and the QCQP from Example~\ref{atva2018ex:nlp}.  After linearizing around an assignment for $\hat{v}, \hat{p}_{s_0}, \hat{p}_{s_1},$ and $\hat{p}_{s_2}$, the resulting LP with a trust region radius $\delta>0$ is given by
	\begin{align*}
		&\textnormal{minimize } \quad p_{s_0}+\tau\sum_{i=0}^2 k_{s_i} \\
&	\textnormal{subject to}  \nonumber\\
&\; p_{s_3}=1,\; \lambda \geq p_{s_0},\\
&\;  k_{s_0}+p_{s_0} \geq \hat{v}\cdot\hat{p}_{s_1}+\hat{p}_{s_1}\cdot(v-\hat{v})+\hat{v}\cdot(p_{s_1}-\hat{p}_{s_1}),\\
&\; k_{s_1}+p_{s_1} \geq p_{s_2}-\hat{v}\cdot\hat{p}_{s_2}-\hat{p}_{s_2}\cdot(v-\hat{v})-\hat{v}\cdot(p_{s_2}-\hat{p}_{s_2}),\\
&\;  k_{s_2}+p_{s_2} \geq\hat{v}\cdot\hat{p}_{s_3}+\hat{p}_{s_3}\cdot(v-\hat{v})+\hat{v}\cdot(p_{s_3}-\hat{p}_{s_3}),\\
& \; k_{s_0}\geq 0, k_{s_1}\geq 0, k_{s_2} \geq 0, \\
& \; \nicefrac{\hat{p}_{s_0}}{\delta'}\geq  p_{s_0}\geq \hat{p}_{s_0}\cdot\delta',
\; \nicefrac{\hat{p}_{s_1}}{\delta'}\geq  p_{s_1}\geq \hat{p}_{s_1}\cdot\delta',
\\&\; \nicefrac{\hat{p}_{s_2}}{\delta'}\geq  p_{s_2}\geq \hat{p}_{s_2}\cdot\delta', \; \nicefrac{\hat{v}}{\delta'}\geq  v\geq \hat{v}\cdot\delta'.
	\end{align*}
\end{example}
We detail our \gls{scp} method in Fig.~\ref{atva2018fig:scpwithmc}. %
We initialize the method with a guess for the parameters $\hat{\instantiation}$, for the probability variables $\hat{\mathbf{p}}$, and the trust region $\delta>0$.
Then, we solve the LP~\eqref{atva2018eq:min_qcqp_scp}--\eqref{atva2018eq:trust_region_scp} that is linearized around $\hat{\instantiation}$ and probability variables $\hat{\mathbf{o}}$.

After obtaining an instantiation to the parameters $\instantiation$, we model check the instantiated MDP $\parammdp[\instantiation]$ to obtain the values of probability variables $\textrm{res}(\mathbf{p})$ for the instantiation $\instantiation$.
If the instantiated MDP indeed satisfies the specification, we return the instantiation $\instantiation$.
Otherwise, we check whether the probability of reaching the target set  $\beta$ is larger than the previous best value $\hat{\beta}$.
If $\beta$ is larger than $\hat{\beta}$, we update the values for the probability and the parameter variables, and enlarge the trust region.
Else, we reduce the size of the trust region, and resolve the problem that is linearized around $\instantiation$ and $\mathbf{p}$.
This procedure is repeated until a parameter instantiation that satisfies the specification is found, or the value of $\delta$ is smaller than $\omega >0$.
The intuition behind enlarging the trust region is as follows: If the instantiation to the parameters $\instantiation$ increases the probability of reaching the target set  $\beta$ over the previous solution, then we conclude that the linearization is accurate. 
	Consequently, the \gls{scp} method may take a larger step in the next iteration for faster convergence in practice.

For expected cost specifications, the resulting algorithm is similar, except, we accept the parameter instantiation if the expected cost is reduced compared to the previous iteration, and initialize $\hat{\beta}$ with a large constant.
For various numerical results about the SCP Method, we refer the reader to~\citep{atvaqcqp}.

\section{Scenario-Based Verification in Uncertain Markov Decision Processes}\label{Chapter:TACAS2020}
\label{sec:uncertain-mdp}

In this chapter, we consider Markov decision processes (MDPs) in which the transition probabilities and rewards belong to an uncertainty set parametrized by a collection of random variables.
The probability distributions for these random parameters are unknown.
The problem is to compute the probability to satisfy a temporal logic specification within any MDP that corresponds to a sample from these unknown distributions.
In general, this problem is undecidable, and we resort to techniques from so-called scenario optimization. 
Based on a finite number of samples of the uncertain parameters, each of which induces an MDP, the proposed method estimates the probability of satisfying the specification by solving a finite-dimensional convex optimization problem. 
The number of samples required to obtain a high confidence on this estimate is independent from the number of states and the number of random parameters. 
Experiments on a large set of benchmarks show that a few thousand samples suffice to obtain high-quality confidence bounds with a high probability.

There are several approaches for verification of uncertain MDPs. \citet{bacci2019model} consider the analysis of Markov models in the presence of uncertain rewards, utilizing statistical methods to reason about the probability of a parametric MDP satisfying an expected cost specification.
This approach is restricted to reward parameters and does not explicitly compute confidence bounds.
\citet{DBLP:conf/sigsoft/LlerenaBBSR18}~compute bounds on the long-run probability of satisfying a specification with probabilistic uncertainty for Markov chains.
Other related techniques include multi-objective model checking to maximize the average performance with probabilistic uncertainty sets~\citep{DBLP:conf/valuetools/Scheftelowitsch17}, sampling-based methods which minimize the \emph{regret} with uncertainty sets~\citep{DBLP:journals/jair/AhmedVLAJ17}, and Bayesian reasoning to compute parameter values that satisfy a metric temporal logic specification on a continuous-time Markov chain~\citep{DBLP:conf/tacas/BortolussiS18}.
\citet{arming2018parameter} consider a variant of the problem in this dissertation where the probability distribution of the uncertainty sets is assumed to be known.
This work formulates the policy synthesis problem as a partially observable Markov decision process (POMDP) synthesis problem and use off-the-shelf point-based POMDP methods~\citep{pineau2003point,cassandra1997incremental}.
The work in~\citep{seshia_et_al_cav_13,DBLP:conf/cdc/WolffTM12} consider the verification of MDPs with convex uncertainties. 
However, the uncertainty sets for different states in an MDP are restricted to be independent, which does not hold in the considered problem setting where we have parameter dependencies.

Uncertainties in MDPs have received quite some attention in the artificial intelligence and planning literatures. Interval MDPs~\citep{seshia_et_al_cav_13,givan2000bounded} use probability intervals in the transition probabilities. Dynamic programming, robust value iteration and robust policy iteration have been developed for MDPs with uncertain transition probabilities whose parameters are statistically independent, also referred to as rectangular, to find a policy ensuring the highest expected total reward at a given confidence level~\citep{nilim2005robust,DBLP:conf/cdc/WolffTM12}. The work in~\citep{wiesemann2013robust} relaxes this independence assumption a bit and determines a policy that satisfies a given performance with a pre-defined confidence provided an observation history of the MDP is given by using conic programming. State-of-the art exact methods can handle models of up to a few hundred of states~\citep{ho2018fast}. Multi-model MDPs~\citep{steimle2018multi} treat distributions over probability and cost parameters and aim at finding a single policy maximizing a weighted value function. For deterministic policies this problem is NP-hard, and it is PSPACE-hard for history-dependent policies.

\subsection{Uncertain Markov Decision Processes}

We now introduce the setting that we study in this chapter. 
Specifically, we use parameters to define the uncertainty in the transition probabilities and cost functions of an MDP. 
Each random parameter follows an unknown probability distribution from which we can sample the parameter values. 

\begin{definition}[\Gls{umdp}]
	An \emph{\gls{umdp}} $\vmdp$ (\gls{umdp}) is a tuple $\vmdp = \left(\pmdp,   \probdist \right)$ where $\pmdp$ is a parametric MDP (\gls{pmdp}), and  $\probdist$ is a probability distribution over the parameter space $\paramspace[\pmdp]$.
	If $\pmdp$ is a pMC, then we call $\vmdp$ a uMC.
\end{definition}
Intuitively, a \gls{umdp} is a \gls{pmdp} with an associated distribution over possible (graph-preserving) parameter instantiations.
That is, a realization of $\probdist$ yields a concrete MDP $\pmdp[u]$ with the respective instantiation $u\in\paramspace[\pmdp]$ (and $\probdist(u)>0$). 
\begin{remark}
In a \gls{umdp}, we distinguish \emph{controllable} and \emph{uncontrollable} parameters.
The uncontrollable parameters follow the probability distribution $\probdist$. 
In contrast, we can actively \emph{instantiate} the controllable parameters. In the following, we specifically allow cost parameters to be controllable. 
\end{remark}
\begin{definition}[Satisfaction Probability]
Let $\vmdp = \left(\pmdp, \probdist\right)$ be a \gls{umdp} and $\varphi$ a specification.
	The (weighted) \emph{satisfaction probability of $\varphi$} is  
	\[ \satprob= \int_{\paramspace[\pmdp]}   I_\varphi(u)\; d\,\probdist(u)\]
	with $u\in \paramspace[\pmdp]$ and $I_\varphi \colon \paramspace[\pmdp] \rightarrow \{0,1\}$ is the indicator for $\varphi$, i.e.\ $I_\varphi(u) = 1$ iff %
$\pmdp[u] \models \varphi$. 
\end{definition} 
Note that $I_\varphi$ is measurable, as $\paramspace[\pmdp]$ is the finite union of semi-algebraic sets~\citep{Book_Basu_RAAlgorithms}. %
Moreover, we have that $\satprob \in [0,1]$ and $\satprob + \falseprob= 1$.

\begin{example}
Consider the uMC in the left figure of Figure~\ref{tacas-2020fig:exampleparam} with the uncontrollable parameter set $V=\{v\}$, initial state $s_0$, target set $T = \{s_3\}$ and an uniform distribution for the parameter $v$ over the interval $[0,1]$. We plot the probability of satisfying the specification $\reachPropSymbol=\reachPropgT$ as a function of $v$ in the right figure of Figure~\ref{tacas-2020fig:exampleparam}. We also show the satisfying region and its complementary as green and red regions. The satisfying region is given by the union of the intervals $\left[0.13, 0.525\right]$ and $\left[0.89,1.0\right]$, and the satisfaction probability $\satreachprob$ is $0.395+0.11=0.505$.
\end{example} 

\begin{figure}[t]
\centering
\begin{tikzpicture}
\node[state] (v0) at (0.5,-2) {$s_0$};
\node[state] (v1) at (2,-2) {$s_1$};
\node[state] (v2) at (4.5,-2) {$s_2$};
\node[state, fill=red!50] (v31) at (6,-2) {$s_3$};
\node[state] (v4) at (2,-3) {$s_4$};
\node[state] (v5) at (4.5,-3) {$s_5$};
\node[state,draw=gray, text=gray] (v6) at (3.25,-5) {$s_6$};
\node[state,draw=gray, text=gray] (v7) at (3.25,1) {$s_7$};

\draw[-latex, thick] (-.5,-2) -- (v0);
\draw[-latex, thick,postaction={decorate,decoration={raise=1ex,text along path,text align=center,text={|\scriptsize|{$1-v$}}}}] (v0) -- (v1);
\draw[-latex, thick,postaction={decorate,decoration={raise=1ex,text along path,text align=center,text={|\scriptsize|{$0.1 \cdot (1-v)$}}}}] (v1) -- (v2);
\draw[-latex, thick,postaction={decorate,decoration={raise=1ex,text along path,text align=center,text={|\scriptsize|{$v$}}}}] (v2) -- (v31);
\draw[-latex, thick,postaction={decorate,decoration={raise=-1.5ex,text along path,text align=center,text={|\scriptsize|{$v$}}}}] (v0) -- (v4);
\draw[-latex, thick,postaction={decorate,decoration={raise=1ex,text along path,text align=center,text={|\scriptsize|{$0.5 \cdot v^2$}}}}] (v4) -- (v5);
\draw[-latex, thick,postaction={decorate,decoration={raise=-1.5ex,text along path,text align=center,text={|\scriptsize|{$1-v$}}}}] (v5) -- (v31);
\draw[-latex, thick,color=gray,postaction={decorate,decoration={raise=-1.75ex,text along path,text align=center,text color=gray,text={|\scriptsize|{$1-0.5 \cdot v^2$}}}}] (v4) -- (v6);
\draw[-latex, thick,color=gray,postaction={decorate,decoration={raise=-1.5ex,reverse path,text along path,text align=center,text color=gray,text={|\scriptsize|{$v$}}}}] (v5) -- (v6);
\draw [thick, color=gray] (v6) edge[>=latex, loop right] (v6) {};
\draw[-latex, thick,color=gray,postaction={decorate,decoration={raise=1.5ex,text along path,text align=center,text color=gray,text={|\scriptsize|{$v+0.9 \cdot (1-v)$}}}}] (v1) -- (v7);
\draw[-latex, thick,color=gray,postaction={decorate,decoration={raise=1ex,reverse path,text along path,text align=center,text color=gray,text={|\scriptsize|{$1-v$}}}}] (v2) -- (v7);
\draw [thick, color=gray] (v7) edge[>=latex, loop right] (v7) {};
\tkzLabelPoint [color = gray](4, 1.25) {\scriptsize{$1$}};
\tkzLabelPoint [color = gray](4, -4.75) {\scriptsize{$1$}};
\draw [thick] (v31) edge[>=latex, loop right] (v31) {};
\tkzLabelPoint (6.75, -1.8) {\scriptsize{$1$}};

\end{tikzpicture}%
\scalebox{.8}{
\begin{tikzpicture}[baseline]
\definecolor{color1}{rgb}{0.1,0.498039215686275,0.9549019607843137}
\begin{axis}[width=2.2in,height=2.2in,ymin=-0.0001,xmin=-0.01,xmax=1.01,xmajorgrids,ymajorgrids,xlabel={$v$},ylabel={$\reachPropgT$},ytick={0,0.05,0.1,0.13,0.15},yticklabels={0,0.05,0.1,$\lambda$,0.15},clip=false]

\addplot [color=color1,line width=0.5pt,dashed,domain=0:1, samples=101,unbounded coords=jump, name path=C]{0.5*x^3*(1-x) + 0.1*x^6*(1-x) +  0.1*x*(1-x)^3 +  0.1*(1-x)^2 + 0.04 + 0.05*x};
\addplot [color=color1,line width=1.5pt,domain=0:0.15, samples=101,unbounded coords=jump, name path=CC]{0.5*x^3*(1-x) + 0.1*x^6*(1-x) +  0.1*x*(1-x)^3 +  0.1*(1-x)^2 + 0.04 + 0.05*x};
\addplot [color=gray!50!white,line width=0pt,domain=0:0.15, samples=10,unbounded coords=jump, name path=DD]{0};
\addplot [color=gray!50!white,line width=0pt,domain=0.15:0.525, samples=10,unbounded coords=jump, name path=EE]{0};
\addplot [color=color1,line width=1.5pt,domain=0.525:0.89, samples=101,unbounded coords=jump, name path=CCC]{0.5*x^3*(1-x) + 0.1*x^6*(1-x) +  0.1*x*(1-x)^3 +  0.1*(1-x)^2 + 0.04 + 0.05*x};
\addplot [color=gray!50!white,line width=0pt,domain=0.525:0.89, samples=10,unbounded coords=jump, name path=DDD]{0};
\addplot [color=gray!50!white,line width=0pt,domain=0.89:1, samples=10,unbounded coords=jump, name path=EEE]{0};
\addplot [color=gray!50!white,line width=0pt,domain=0:0.15, samples=10,unbounded coords=jump, name path=DDA]{0.13};
\addplot [color=gray!50!white,line width=0pt,domain=0.15:0.525, samples=10,unbounded coords=jump, name path=EEA]{0.13};
\addplot [color=gray!50!white,line width=0pt,domain=0.525:0.89, samples=10,unbounded coords=jump, name path=DDDA]{0.13};
\addplot [color=gray!50!white,line width=0pt,domain=0.89:1, samples=10,unbounded coords=jump, name path=EEEA]{0.13};
     \addplot[red!20!white,domain=0:0.15] fill between[of=DD and DDA];
    \addplot[red!20!white,domain=0.525:0.89]  fill between[of=DDD and DDDA];
         \addplot[green!20!white,domain=0.15:0.525] fill between[of=EE and EEA];
    \addplot[green!20!white,domain=0.89:1]  fill between[of=EEE and EEEA];
    \draw [color=black,dashed,line width=1.5pt](axis cs:0,0.0) -- node[left]{} (axis cs:0,0.14);
    \draw [color=black,dashed,line width=1.5pt](axis cs:0.15,0.0) -- node[left]{} (axis cs:0.15,0.13);
    \draw [color=black,dashed,line width=1.5pt](axis cs:0.525,0.0) -- node[left]{} (axis cs:0.525,0.13);
    \draw [color=black,dashed,line width=1.5pt](axis cs:0.89,0.0) -- node[left]{} (axis cs:0.89,0.13);
    \draw [color=black,dashed,line width=1.5pt](axis cs:0,0.13) -- node[left]{} (axis cs:1.0,0.13);

\end{axis}
\end{tikzpicture}}%
\caption{Left: A uMC with parameter $v$. Right: The probability of satisfying the reachability specification $\reachPropSymbol=\reachPropgT$ versus the value of the parameter $v$. Intervals that satisfy $\reachPropSymbol$ are green, intervals that violate $\reachPropSymbol$ are red.}
\label{tacas-2020fig:exampleparam}
\end{figure}

\subsection{Formal Problem Statement}

In this section, we state the problem that we study in this chapter.
We seek to compute the satisfaction probability of the parameter space for a reachability or an expected cost specification $\varphi$ on an \gls{umdp}.
Intuitively, we seek the probability that a randomly sampled instantiation from the parameter space induces an MDP which satisfies $\varphi$.
Formally: Given an uncertain MDP $\vmdp=\left(\pmdp,  \probdist\right)$, and a specification $\varphi$, compute the satisfaction probability $\satprob$. 
However, as mentioned, the problem is in general undecidable~\citep{arming2018parameter}.
Therefore, we consider an approximation of computing the satisfaction probability:
\begin{problem}
\label{tacas-2020prob:primal_with_confidence}
Given an uncertain MDP $\vmdp=\left(\pmdp,  \probdist\right)$, a reachability specification $\varphi_r = \reachPropgT$, and a \emph{tolerance probability} $\nu$, compute a confidence probability $\alpha_{\nu}$ such that $\satreachprob \geq 1 - \nu$ holds with a probability of at least $1-\alpha_{\nu}$.
  \end{problem}
We illustrate the problem statement with the following example.
 \begin{example}
 For a UAV motion planning example, consider the question \textquotedblleft What is the probability on a given day such that there exists a policy for the UAV to successfully finish the mission.\textquotedblright~   %
 A possible result is, e.g.,  0.78 (confidence probability: 0.99) and 0.81 (confidence probability: 0.95).
 Then, with a confidence probability of 0.99, the actual satisfaction probability is indeed greater than 0.78, and with a (slightly lower) confidence probability of 0.95 it is greater than 0.81. 
Such a result shows that it is quite likely that the UAV will finish the mission successfully with a probability that is at least 81\%.
 \end{example}

\subsection{Scenario-Based Verification}
\label{tacas-2020sec:robust}
In this section, we present an approach by \citet{cubuktepe2020scenario} to solve Problem~\ref{tacas-2020prob:primal_with_confidence}, that is, to approximate the satisfaction probability with respect to a specification.
We first consider the robust \textcolor{red}{verification} problem that accounts \emph{for all possible values} in the uncertainty set, potentially leading to a very pessimistic result.
This problem can be formulated as a semi-infinite convex optimization problem, which is NP-hard~\citep{wiesemann2013robust}.
Here, we exploit the structure of this problem, which includes finitely many variables but infinitely many constraints. 
The presented approach is based on \emph{scenario optimization}~\citep{DBLP:journals/tac/CalafioreC06,DBLP:journals/siamjo/CampiG08}: We sample a finite number of parameter values and restrict the semi-infinite problem to these samples.
The resulting \emph{finite-dimensional} convex optimization problem can be solved efficiently~\citep{boyd_convex_optimization}.  
Based on the solution of the optimization problem, we compute high confidence in the estimate of the satisfaction probability. 
The estimate also generalizes to the samples from the probability distribution that are not in the sample set.

\begin{remark}
For ease of presentation, we focus on uncertain Markov chains (uncertain MCs).
The results and methods generalize to uncertain MDPs (uncertain MDPs). 
\end{remark}
We first develop the main results for the simple setting where \emph{all sampled} instantiated MCs from the parameter space $\paramspace[\dtmc]$ satisfy the reachability specification $\varphi_r$. 
This assumption does not imply that \emph{all} instantiated MCs satisfy $\varphi_r$: The sample set does not contain an MC that violates $\varphi_r$ even though there exists such an MC in the parameter space. %
In Chapter~\ref{tacas-2020sec:scenario:generalized}, we drop this assumption and allow sampled points in $\paramspace[\dtmc]$ to violate $\varphi_r$.
This completes our treatment of Problem~\ref{tacas-2020prob:primal_with_confidence}.

\subsection{Restriction to Satisfying Samples}
\label{tacas-2020sec:scenario:closetoone}
In this section, we assume that all instantiated MCs satisfy $\varphi_r$.
We then generalize the presented method to any values of $\nu$.
We want to check if an uncertain MC $\dtmc$ satisfies a reachability specification $\reachPropSymbol=\reachPropgT$ for all instantiations in the sample set $\mathcal{U}$. 
For each instantiation, we can formulate a linear program (LP) that is feasible if and only if $\varphi_r$ is satisfied~\citep{puterman2014markov}. 
For a subset $\mathcal{U} \subseteq \paramspace[\dtmc]$ of the parameter space $\paramspace[\dtmc]$ of the uncertain MC $\dtmc$, we can then write the conjunction of these LPs. 
We assume that $\vert \mathcal{U}\vert$ is finite and sampled from the probability distribution $\PP$ over the parameter space $\paramspace[\dtmc]$.

For each instantiation $u \in \mathcal{U}$, we introduce a set of linear constraints that are parametrized by $u$.
We assume that each sample has a unique index.
We use the following variables.
For $s \in S$ and $u \in \mathcal{U}$, the variable $p^u_s \in [0,1]$ represents the probability of reaching the target set $T \subseteq S$ from state $s$. 
The variable $\tau$ represents an upper bound on the probability of satisfying $\varphi_r$ for all instantiations in $\mathcal{U}$. Note that $\tau$ is a variable in our formulation, whereas $\lambda$ is the threshold of the reachability specification, and thus constant. 
The set $\neg\exists\finally T$ represents the set of states which cannot reach the target set $T$. 
The probability of reaching $T$ from these states is zero, and the set $\neg\exists\finally T$ does not change for different graph-preserving instantiations~\citep{param_sttt}.  
The set $\neg\exists\finally T$ can be found in polynomial time in the size of an uncertain MC by using standard graph-based search algorithms~\citep{BK08}. 
We solve the following LP $\mathcal{L}_r(\mathcal{U})$, which is parametrized by each instantiation $u$ in $\mathcal{U}$,

\begin{align}
	\displaystyle \minimize  &\quad 	\tau 	\label{tacas-2020eq:robustmc_spec_obj}\\
		\subjectto & \quad \forall  u \in \mathcal{U}, \nonumber\\
		&\quad p^u_{\sinit} \leq \tau,\label{tacas-2020eq:robustmc_spec_cons1}\\
	& \quad p^u_{\sinit} \leq \lambda,\label{tacas-2020eq:robustmc_spec_cons2}\\
\quad\forall s \in T,	&\quad	 \displaystyle p^u_s = 1,\label{tacas-2020eq:robustmc_spec_cons3}\\
	\quad\forall s \in \neg\exists\finally T,	&\quad \displaystyle  p^u_s =0,\label{tacas-2020eq:robustmc_spec_cons4}\\
\quad\forall s \in S \setminus \left( T \cup \neg\exists\finally T \right),&\quad	 \displaystyle  p^u_s = \sum_{s' \in S}\mathcal{P}(s,s')[u] \cdot p^u_{s'}.\label{tacas-2020eq:robustmc_spec_cons5}
\end{align}
The objective~\eqref{tacas-2020eq:robustmc_spec_obj} minimizes the maximal probability that can be achieved by all MCs induced by $\mathcal{U}$. 
The constraint~\eqref{tacas-2020eq:robustmc_spec_cons1} represents an upper bound on the reachability probability for all instantiations. We minimize the upper bound to compute the maximal probability of satisfying $\varphi_r$ for all instantiated MCs.
The constraint~\eqref{tacas-2020eq:robustmc_spec_cons2} ensures that the probability of reaching $T$ from the initial state $\sinit$ is below the threshold $\lambda$. 
The constraint~\eqref{tacas-2020eq:robustmc_spec_cons3} sets the probability to reach a state in $T$ from $T$ to 1. 
The constraint~\eqref{tacas-2020eq:robustmc_spec_cons4} sets the reachability probabilities from the states in $ \neg\exists\finally T$ to zero.
The constraint~\eqref{tacas-2020eq:robustmc_spec_cons5} computes the probability of satisfying the specification for each non-target state $s \in S$ in the standard way.  

There are infinitely many constraints in the semi-infinite LP $\mathcal{L}_r(\paramspace[\dtmc])$ as the cardinality of $(\paramspace[\dtmc])$ is infinite.
To obtain a LP with finitely many constraints, we instantiate the parameters $u \in \paramspace[\dtmc]$ by sampling the probability distribution $\PP$. %
Then, for a given violation probability $\nu \in (0, 1)$, we compute a solution that violates the constraints in the LP $\mathcal{L}_r(\paramspace[\dtmc])$ with a \emph{tolerance probability} that is not larger than $\nu$. 
We first give some properties of the LP $\mathcal{L}_r(\mathcal{U})$. For proofs in this section, we refer the reader to~\citep{cubuktepe2020scenario}.

\begin{thm}
	Let uncertain MC $\dtmc$ and the sample sets $\mathcal{U} \subseteq \paramspace[\dtmc]$ with $K = \vert \mathcal{U} \vert \geq  2$. 
	Assume for all $u \in \mathcal{U}$, $\dtmc[u] \models \reachPropSymbol$.
	\label{tacas-2020thm:probabilistic_bound_1}
  For a given \emph{tolerance probability} $\nu\in[0,1)$, let the associated \emph{confidence probability}
    \begin{equation}
    \label{tacas-2020eq:tolerance_level}
\alpha_\nu=\sum_{i=0}^{1}\dbinom{K}{i}(1-\nu)^{K-i}\nu^i.
\end{equation}
	Then, with a probability of at least $1-\alpha_\nu$, we have 
	\begin{equation}
	\label{tacas-2020eq:probabilistic_bound_1} \satreachprobmc \geq 1-\nu.
	\end{equation}
\end{thm}

\begin{remark}[Independence to Model Size]\label{tacas-2020remark:independent}
	The confidence probability in Theorem~\ref{tacas-2020thm:probabilistic_bound_1} is in fact independent from the number of states, transitions, or random parameters of the uncertain MC. 
	From a practical perspective, the number of samples that are needed for a certain confidence does not depend on the model size. 	
\end{remark}
	Finally, Theorem~\ref{tacas-2020thm:probabilistic_bound_1} asserts that with a probability of at least $1-\alpha_\nu$, the next sampled point from $\paramspace[\dtmc]$ will satisfy the specification with a probability of at least $1-\nu$. Note that $\alpha_\nu$ is the tail probability of a binomial distribution. 
	It converges exponentially rapidly to $0$ in $\vert  \mathcal{U} \vert$~\citep{DBLP:journals/siamjo/CampiG08}. %

\subsection{Satisfaction Probability by Treating Violating Samples}\label{tacas-2020sec:frac}
\label{tacas-2020sec:scenario:generalized}

Theorem~\ref{tacas-2020thm:probabilistic_bound_1} assumes that all sampled points, that is, the induced MCs, satisfy the specification $\varphi_r$. 
This is a severe assumption in general.
To lift this assumption, we consider the \emph{discarding approach} from~\citep{campi2011sampling}.
Specifically, after sampling a set of instantiations $\mathcal{U}$ from $\paramspace[\dtmc]$ according to the probability distribution $\PP$, we remove the constraints for the MCs that violate the specification $\varphi_r$ from the LP. 
We construct the set $\mathcal{R}=\mathcal{U}\setminus\mathcal{Q}$, where $\mathcal{Q}$ denotes the set of samples that induce MCs violating the specification $\varphi_r$.
Therefore, the set $\mathcal{R}$ denotes the set of sampled MCs that satisfy the specification $\varphi_r$. %
We then solve the LP $\mathcal{L}_r(\mathcal{R})$
\begin{equation}
	\label{tacas-2020eq:robustmc_spec_sample_discard}
	\begin{array}{llllllll}
	&\displaystyle \minimize \;\;  \tau \\
	&\subjectto \quad \forall u \in \mathcal{R},\\
	&\eqref{tacas-2020eq:robustmc_spec_cons1}-\eqref{tacas-2020eq:robustmc_spec_cons5},
	\end{array}
\end{equation}
where for $u \in \mathcal{R}$ and $s \in S$, $p^u_s$ gives the probability of satisfying the reachability specification of the instantiated MC $\dtmc[u]$ at state $s$.
The other constraints in the optimization problem in LP $\mathcal{L}_r(\mathcal{R})$ are identical to the LP $\mathcal{L}_r(\mathcal{U})$.
We give the main result of this section.

\begin{thm}
Let uncertain MC $\dtmc$ and the sample sets $\mathcal{U},\mathcal{Q}\subseteq \paramspace[\dtmc]$,  with $K = \vert \mathcal{U} \vert \geq  2$ and $L = \vert \mathcal{Q}\vert $. 
	\label{tacas-2020thm:probabilistic_bound_discard}
  For a given \emph{tolerance probability} $\nu\in[0,1)$, the associated \emph{confidence probability} is
    \begin{equation}
    \label{tacas-2020eq:tolerance_level_disard}
\alpha_\nu=\dbinom{L+1}{L}\sum_{i=0}^{L+1}\dbinom{K}{i}(1-\nu)^{K-i}\nu^i.
\end{equation}
	Then, with a probability of at least $1-\alpha_\nu$, we have	\begin{equation}
	\label{tacas-2020eq:probabilistic_bound_discard} \satreachprobmc \geq 1-\nu.
	\end{equation}
\end{thm}

\subsection{Building Scenario-Based Algorithms}%

The question remains how we leverage the theoretical results to compute an estimate on the satisfaction probability to solve the problem in this section.
For instance, let $\nu$  be a violation probability and $\mathcal{U}$ the sample set. 
Then, we can use Theorem~\ref{tacas-2020thm:probabilistic_bound_discard} to compute the confidence probability $\alpha_\nu$ by using the discarding approach from~\citep{campi2011sampling}.
Similarly, for a the sample set $\mathcal{U}$ and a threshold on the confidence probability $\alpha_\nu$ we do a \emph{bisection} on $\nu$. 
Specifically, we repeatedly apply Theorem~\ref{tacas-2020thm:probabilistic_bound_discard} for different values of $\nu \in (0, 1)$, to see if the corresponding confidence probability $\alpha_\nu$ is below the threshold. 
We then approximate the lower and upper bounds on $\nu$.

The correctness of the approach is based on scenario-based optimization. 
However, it also applies to an obtained solution by any procedure~\citep{campi2018general}.
For instance, for any obtained value for the controlled parameters, we can construct a scenario program by sampling from random parameters.
We can then apply Theorem~\ref{tacas-2020thm:probabilistic_bound_discard} to compute the confidence probability $\alpha_\nu$ or the violation probability~$\nu$.

\paragraph{Generalization to Uncertain MDPs.} %
Recall that we want to compute the satisfaction probability for an uncertain MDP, which is the probability that for any sampled MDP, we are able to synthesize a policy that satisfies the specification $\varphi_r$.
To generalize the presented results to uncertain MDPs, we can modify the constraint~\eqref{tacas-2020eq:robustmc_spec_cons5} in the LP $\mathcal{L}_r(\mathcal{U})$ as
\begin{align}
	 \forall s \in S \setminus \left( T \cup \neg\exists\finally T \right),\;\forall \act \in \ActS,\displaystyle \quad  p^u_s \leq  \sum_{s' \in S}\mathcal{P}(s,\act,s')[u] \cdot p^u_{s'}.\label{tacas-2020eq:MDP_spec}
\end{align}
The constraints~\eqref{tacas-2020eq:robustmc_spec_cons1}--\eqref{tacas-2020eq:robustmc_spec_cons4} and~\eqref{tacas-2020eq:MDP_spec} assert that, for each non-target state $s \in S$ and action $\act \in \ActS$, the probability induced by the \emph{minimizing policy} is an upper bound to the probability variables $p^u_s$. 
Recall that, the reachability specification $\varphi_r$ is satisfied if and only if the reachability probability at the initial state induced by the minimizing policy is less than $\lambda$.
Then, our theoretical results apply to the uncertain MDPs.

\subsection{Case Study: UAV Motion Planning}
We implemented the approach from Chapter~\ref{tacas-2020sec:robust} using the model checker Storm~\citep{DBLP:conf/cav/DehnertJK017} to construct and  analyze samples of MDPs.
To solve the scenario optimization problems with cost parameters, we used the SCS solver~\citep{scs}. All computations ran on a computer with 8 2.2 GHz cores, and 32 GB of RAM. We note that further benchmarks evaluating scenario-based algorithms with a varying number of samples in~\citep{DBLP:conf/tacas/Cubuktepe0JKT20, badings2022scenario}. Specifically, the obtained confidence probabilities decrease exponentially rapidly with an increasing number of samples.

\begin{figure}[t]
\centering
\scalebox{1.2}{
\input{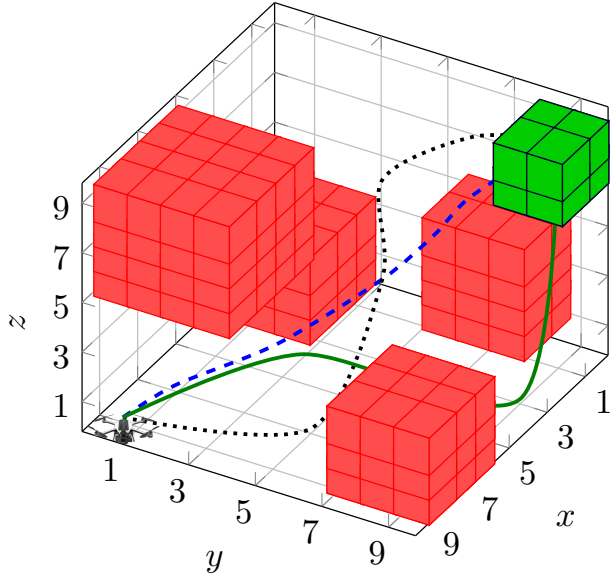}
}
\caption{An example of a 3D UAV benchmark with obstacles and a target area.}
\label{tacas-2020fig:drone}
\end{figure}

In the benchmark, we consider a UAV motion planning example to model a realistic problem with a high number of random parameters.  
We model the problem as an uncertain MDP, where the parameters represent how the weather conditions affect the movement of the UAV, and how the weather may change.
In particular, different wind conditions induce specific satisfaction probabilities.
We assume that the planning area is a certain valley where we have historic weather data which provide distributions over parameter values. 
The mission of the UAV is to transport a payload to a specific location and return safely to its initial position. 
The problem is to compute the satisfaction probability, that is, the probability that for any sampled MDP for this scenario we are able to synthesize a UAV policy that satisfies the specification.

We model the problem as follows:
States encode the position of the UAV, the current weather situation, and the general wind direction in the valley. 
Parameters describe how the weather affects the position of the UAV for different zones in the valley, and how the weather/wind may change during the day. 
Fig.~\ref{tacas-2020fig:drone} shows an example environment with zones to avoid (red) and a target zone (green).
We define four different weather conditions that each induce certain probability distributions over the eight different wind directions. 
The parameters of the model determine the probabilities of transitioning between different weather and wind conditions at each time step.
The specification is to reach the target zone safely with a probability of at least $0.9$. 
The number of states in our example is 266\,880, and the number of parameters is 2\,500.

For the distributions over parameter values, that is, over weather conditions, we consider the following cases.
First, we assume a uniform distribution over the different weather conditions in each zone. 
Second, the probability for a weather condition inducing a wind direction that pushes the UAV into the positive $y$-direction is five times more likely than others. 
Similarly, in the third case, it is five times more likely to push the UAV into the negative $x$-direction.
We depict some example trajectories of the UAV for three different conditions in Fig.~\ref{tacas-2020fig:drone}.
The trajectory given by the blue dashed line represents the expected trajectory for the first case, taking a  direct route to reach the target area.
Similarly, the trajectories given by the black dotted and solid green lines represent the expected trajectories for the second and third cases.
For the second case, we observe that the UAV tries to avoid to get closer to the obstacles in $x$ direction as the wind may push the UAV to the obstacles.
For the third case, the UAV avoids the obstacle at the bottom and then reaches the target area.

We sample $1\,000$ parameters for each case and approximate the maximal satisfaction probability with a confidence probability of at least $1-\alpha_\nu$, with $\alpha_\nu=10^{-6}$. 
The highest satisfaction probability is given by the first weather condition with $0.86$, and the other conditions have a satisfaction probability of $0.78$ and $0.75$, showing that it may be harder to navigate around the obstacles with non-uniform probability distributions.
The average time to compute the satisfaction probabilities is $1\,341$ seconds.

Finally, we introduce costs to a 2-dimensional example, where hitting an obstacle causes (1) a cost of $100$ and (2) the UAV to return to the initial position.
Specifically, we introduce cost parameters for transitions that steer the UAV towards $x$ or $y$-directions.
We minimize the maximal possible expected cost (under all parameter values) to reach the target location.
The specification asserts that the resulting expected cost should be less than $20$.

We uniformly sample $1\,000$ parameter values for weather conditions and note that the UAV policies favor on average transitioning to $y$-direction more compared to the $x$-direction to minimize the cost while ensuring that the probability of hitting an obstacle is minimized.
The average expected cost of the induced MDPs is $7.41$ and the satisfaction probability is $0.71$.
The solving time for this example is $2\,274$ seconds.

\chapter{Dealing with Information Limitations}
\label{sec:information_limitations}
\glsresetall

Methods for the synthesis and verification of policies in \glspl{mdp}, \glspl{pmdp}, and \glspl{umdp} assume that the agent is able to observe the underlying state of the system.
However, many sensor or communication limitations may lead to imperfect or limited observations of the system's state in practice.
In this section, we study the problem of sequential decision-making under uncertainty when the decision-making agent has only limited observational capabilities.

There exist several formalisms for modeling decision-making under imperfect perception.
\citet{bai2014integrated} model an integrated perception and planning problem using \gls{pomdp}s.
In~\citet{ghasemi2018perception}, the authors propose a perception-aware point-based \gls{pomdp} solver.
In~\citet{benenson2006integrating}, the authors integrate \gls{slam} with a partial motion planner for autonomous navigation.
\citet{fu2016optimal} consider a robot with a temporal logic task in a probabilistic map obtained from a semantic \gls{slam}. 

We begin this section by presenting the \gls{pomdp}, a commonly used model for decision-making under partial observability.
We then present \glspl{upomdp}---in addition to having imperfect information about the current state, the transition and observation functions belong to uncertainty sets.
Following the exposition of \citet{DBLP:conf/aaai/Cubuktepe0JMST21}, we present an algorithm for the synthesis of policies that robustly satisfy specifications in uncertain \glspl{pomdp}.
We then present an application of \glspl{upomdp} through a spacecraft motion planning case study.
Finally, as an example of policy synthesis under an alternate model of partial observability we present an algorithm, and corresponding case studies, for the setting in which an \gls{mdp} model of the system is available, but the semantic labeling of the environment is only partially known. 

\section{Partially Observable Markov Decision Processes}

\Glspl{pomdp} generalize \glspl{mdp} by introducing an \emph{observation function}, which defines a probability distribution over a set of possible observations given the current state of the system.
\begin{definition}[\gls{pomdp}]
	\label{def:pomdp}
	A \gls{pomdp} is a tuple $\PomdpInit$, with $\mdp$ the \emph{underlying MDP of $\pomdp$}, a finite set of observations $\ObsSym$, and \emph{observation function} $\ObsFun\colon\states\rightarrow\ObsSym$.
\end{definition}
For brevity, we use so-called deterministic observation functions which may be derived from the more standard stochastic observation functions $\ObsFun \colon S \to \Distr(\ObsSym)$ via a (polynomial) reduction~\citep{ChatterjeeCGK16}.
For \glspl{pomdp}, observation-action sequences are based on a finite path $\path\in\pathsfin^{\mdp}$ of the underlying MDP $\mdp$ and have the form: 
$\path_o=\ObsFun(\path)=\ObsFun(s_0)\xrightarrow{\act_0} \ObsFun(s_1)\xrightarrow{\act_1}\cdots\ObsFun(s_n)$.
The set of all finite observation-action sequences for a POMDP $\pomdp$ is $\obsSeqFin^{\pomdp}$. 

While the agent acts within the environment, it encounters certain observations, according to which it can infer the probability of the system being in a certain state. 
Technically, this \emph{belief} $b$ is a distribution $b\in\Distr(S)$, such that $b(s)$ describes the probability of being in state $s\in S$.

Recall that a policy in an \gls{mdp} is a mapping from states to actions.
A policy in a \gls{pomdp} is a mapping from the observations (or a history of observations) to the actions.

The problem of computing an optimal policy for \glspl{pomdp} is undecidable~\citep{MadaniHC99}.
To achieve computational tractability, policies are often restricted to finite memory by computing a \gls{fsc}~\citep{meuleau1999solving,amato2010optimizing}, which is an NP-hard problem~\citep{VlassisLB12}.

\begin{definition}[Finite-Memory Policies.]
\label{def:fsc}
	An \emph{observation-based policy} $\osched\colon (Z \times \Act)^*\times Z\rightarrow\Distr(\Act)$ for a \gls{pomdp}  maps a \emph{trace}, i.e., a sequence of observations and actions, to a distribution over actions. 
	An \gls{fsc} consists of a finite set of memory states and two functions. 
	The \emph{action mapping} $\gamma(n,\obs)$ takes an \gls{fsc} memory state $n$ and an observation $\obs$, and returns a distribution over \gls{pomdp} actions.
	To change a memory state, the \emph{memory update} $\eta(n,\obs,\act)$ returns a distribution over memory states and depends on the action $\act$ selected by $\gamma$.
	An \gls{fsc} induces an observation-based policy by following a joint execution of these functions upon a trace of the \gls{pomdp}.
	An \gls{fsc} is \emph{memoryless} if there is a single memory state; memoryless \glspl{fsc} encode policies $\osched\colon Z \rightarrow\Distr(\Act)$.   
\end{definition}

We start with a simple 5-state reachability example to highlight the utility of \glspl{fsc} as finite-memory \gls{pomdp} policies.
	\begin{figure}[!t]
	\centering
	\input{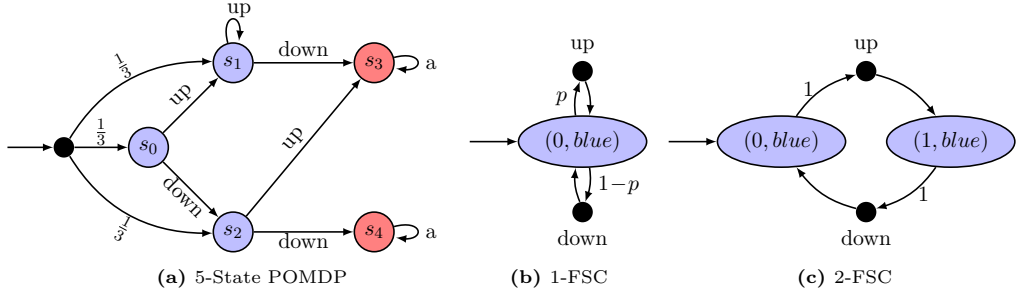}
	\caption{(a) \gls{pomdp} for Example~\ref{ex:motivating} with three observations $\lbrace blue, s_3,s_4\rbrace$ with (b) $1$-\gls{fsc} and (c) $2$-\gls{fsc}. Both \glspl{fsc} are defined for observing ``blue'' and subsequent action choices that may result in a change of memory node for the $2$-\gls{fsc}.}
	\label{fig:motivating}
\end{figure}

\begin{example}
	\label{ex:motivating}
	Consider the \gls{pomdp} in Figure~\ref{fig:motivating}.
	The \gls{pomdp} has three observations (\emph{blue}, $s_3$ and $s_4$), where observation \emph{blue} is received upon visiting $s_0$, $s_1$, and $s_2$. 
	That is, the agent is unable to distinguish between these states.
	The specification is $\varphi = \Pr_{\geq 0.9}(\Ever s_3)$;the agent should reach state $s_3$ with at least probability $0.9$.
	We define the $1$-\gls{fsc} $\fsc_1$, illustrated in Figure \ref{fig:1FSC}, with one memory node $0$:
	\begin{align*}
	\alpha(0,blue) &= \begin{cases} up & \textrm{with probability} \quad p,	\\ down & \textrm{with probability} \quad 1-p,\end{cases}\\
	\delta(0,\obs,\act) &= 0 \quad \forall \obs \in \ObsSym, \act \in \Act.
	\end{align*}
	A $2$-\gls{fsc} with two memory nodes ($0$ and $1$), see Figure~\ref{fig:2FSC}, allows for greater expressivity, i.e. the policy can base its decision on larger observation sequences.
	
	 With this memory structure, we can create an \gls{fsc} $\fsc_2$ that ensures the satisfaction of $\varphi$:
	\begin{align*}
	\alpha(0,blue) &= \begin{cases} up & \textrm{with probability} \quad 1,	\\ down & \textrm{with probability} \quad 0,\end{cases}\\
	\alpha(1,blue) &= \begin{cases} up & \textrm{with probability} \quad 0,	\\ down & \textrm{with probability} \quad 1,\end{cases}\\
	\delta(0,blue,up) &= 1, \\
	\delta(1,blue,down) &= 0. 
	\end{align*}
\end{example}

While \glspl{pomdp} provide an expressive modeling framework for partially observable settings, in many cases the transition and observation functions of the model may not be known exactly.
In these cases, it is necessary to also model our uncertainty in the \gls{pomdp} model.
We discuss such \glspl{upomdp} models in the following section.

\section{Uncertain Partially Observable Markov Decision Processes}
\label{sec:upomdps}

By combining \glspl{umdp} and \glspl{pomdp}, we obtain \glspl{upomdp}---in addition to imperfect information pertaining to the current state, the transition and observation functions belong to uncertainty sets.

\begin{definition}[\Gls{upomdp}]
	\label{ijcai-2020def:umdp}
	An \emph{\gls{upomdp}} is a tuple $\upomdp=(\states,$ $\sinit,$ $\Act,$ $\intervals,$ $\probumdp,$ $\ObsSym,$ $\ObsFun,$ $\rew)$ with a finite set $\states$ of states, an initial state $\sinit \in \states$, a finite set $\Act$ of actions,
	a set  $\intervals$ of probability intervals, an \emph{uncertain transition function} $\probumdp \colon \states \times \Act \times \states \to \intervals$, a finite set $\ObsSym$ of \emph{observations}, an \emph{uncertain observation function} $\ObsFun\colon\states\times \ObsSym \to\intervals$, and a reward function $R \colon \states \times \Act \to \R_{\geq 0}$.
\end{definition}

\noindent \emph{Nominal} probabilities are point intervals where the upper and lower bounds coincide.
If the probabilities of all transitions and observations are nominal, the model is a (nominal) \gls{pomdp}. 
Without a loss of generality, we may express any \gls{upomdp} as a set of nominal \glspl{pomdp} that vary only in their transition functions.
For a transition  function $P \colon \states \times \Act \times S \to \mathbb{R}$, we write $P \in \probumdp$ if for all $s,s' \in \states$ and $\alpha \in \Act$ we have $P(s,\act,s') \in \probumdp(s,\act,s')$ and $P(s, \act, \cdot)$ is a probability distribution over $S$.
Finally, we note that a fully observable \gls{upomdp} where each state has unique observations is an \gls{umdp}.

Existing approaches for policy synthesis in \glspl{upomdp} rely on dynamic programming~\citep{DBLP:conf/cdc/WolffTM12}, convex optimization~\citep{DBLP:conf/cdc/WolffTM12}, or value iteration~\citep{DBLP:conf/cdc/WolffTM12}.
While the complexity of solving a standard \gls{mdp} is polynomial in the number of states and actions, solving an \gls{umdp} is NP-hard in general~\citep{wiesemann2013robust}.
The existing approaches for \glspl{upomdp} rely on sampling~\citep{burns2007sampling} or robust value iteration~\citep{DBLP:conf/icml/Osogami15} on the belief space of the \gls{upomdp}.
The policy synthesis algorithm presented by \citet{suilen2020robust} is based on convex optimization and searches over memoryless policies.
However, their resulting optimization problems are exponentially larger than ours, and they only consider memoryless policies. %
Meanwhile, \cite{burns2007sampling} utilize sampling-based methods and~\cite{DBLP:conf/icml/Osogami15} employ a robust value iteration on the belief space of the \gls{upomdp}. 
\cite{ahmadi2018verification} use sum-of-squares optimization to verify uncertain POMDPs against temporal logic specifications.
\cite{itoh2007} assume distributions over the probability values of the uncertainty set.
Finally,~\cite{DBLP:conf/cdc/ChamieM18} consider a convexified belief space and computes a policy that is robust over this space.

\subsection{Synthesizing Robust Finite-State Controllers for Uncertain Partially Observable Markov Decision Processes}
Following the presentation of \cite{DBLP:conf/aaai/Cubuktepe0JMST21}, we now present an algorithm for the synthesis of finite-state controllers that robustly satisfy specifications in \glspl{upomdp}.

\subsubsection{Problem Formulation}

We begin by introducing observation-based policies, which are similar to memoryless \glspl{fsc} for \glspl{pomdp}.
We then introduce specifications for \glspl{pomdp}, followed by the notion of robustly satisfying a specification in a \gls{upomdp}.

\begin{definition}[Observation-based policy]
	An \emph{observation-based policy} $\osched\colon Z\rightarrow\Distr(\Act)$ for an \gls{upomdp} maps observations to distributions over actions. 
	Note that such a policy is referred to as memoryless and randomized. 
	More general types of policies take an (in)finite sequence of observations and actions into account.
	We use $\oSched^{\upomdp}$ to denote the set of observation-based strategies for $\upomdp$.
	Applying $\osched\in\oSched^{\upomdp}$ to $\upomdp$ resolves all choices and partial observability and results in an induced (uncertain) Markov chain $\upomdp^{\osched}$.
\end{definition}

\paragraph{Specifications in \glspl{pomdp}.}   
We constrain the undiscounted expected reward (the value) of a policy for a \gls{pomdp} using \emph{specifications}:
For a \gls{pomdp} $\pomdp$ and a set of goal states $G$
the specification $\expRewProp{\kappa}{G}$ states that the expected reward before reaching $G$ is at least $\kappa$.
For brevity, we require that the \gls{pomdp} has no dead-ends, i.e., that under every policy, we eventually reach $G$.
Reachability specifications to a subset of $G$ and discounted rewards are special cases~\citep{puterman2014markov}.

\paragraph{Robustly Satisfying Specifications in \glspl{upomdp}.}
A policy $\osched$ satisfies a specification $\varphi=\expRewProp{\kappa}{G}$, if the expected reward to reach $G$ induced by $\osched$ is at least $\kappa$.
\gls{pomdp} $\upomdp[P]$ denotes the \emph{instantiated} \gls{upomdp} $\upomdp$ with a fixed transition function $P \in \mathcal{P}$.
A policy \emph{robustly satisfies} $\varphi$ for the \gls{upomdp} $\upomdp$, if it does so for all $\upomdp[P]$ with $P\in\probumdp$. Thus, a (robust) policy for u\gls{pomdp}s accounts for all possible instantiations $P \in \probumdp$.

Intuitively, to robustly satisfy a specification in a \gls{upomdp}, we require a policy that satisfies the specification for all \gls{pomdp} instantiations from $\upomdp[P]$.
If we have several (expected cost or reachability) specifications $\varphi_1,\ldots,\varphi_m$, we write $\osched\models\varphi_1\land\ldots\land\varphi_n$ where $\osched$ robustly satisfies all specifications.
Note that general temporal logic constraints can be reduced to reachability specifications~\citep{BK08,bouton2020point}, therefore we omit a detailed introduction to the underlying logic.

\begin{assumption}
	For the correctness of the presented method, we require the lower bounds of the intervals to be strictly larger than zero, that is, an instantiation cannot ``eliminate'' transitions.
	Put differently, either a transition exists in \emph{all} instantiations of the \gls{umdp}, or in none.
	That assumption is standard and, for instance, also employed in~\citep{wiesemann2013robust}.
	Moreover, the problem statement would be different and theoretically harder to solve, see~\citep{winkler2019complexity}.
	We allow the upper and lower bound of an interval to be the same, resulting in \emph{nominal} transition probabilities.
\end{assumption}

\begin{problem}[Robust Synthesis for Uncertain \gls{pomdp}s]
	Given an uncertain \gls{upomdp} $\upomdp$ and an expected reward specification $\varphi=\expRewProp{\kappa}{G}$,
	compute an FSC that yields an observation-based policy $\osched$ which robustly satisfies $\varphi$ for $\upomdp$.\smallskip
	\label{prob:RobustSynth}
\end{problem}

\subsubsection{Optimization Problem for Uncertain Partially Observable Markov Decision Processes}
We now reformulate the above problem statement as a semi-infinite nonconvex optimization problem, with finitely many variables but infinitely many constraints.

\begin{figure}[t]
	\centering
	\input{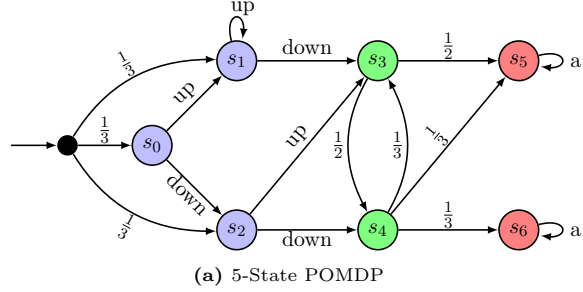}
	\caption{A simple and binary \gls{pomdp} with four observations $\lbrace blue, green, s_3,s_4\rbrace$. This \gls{pomdp} is simple as the states $\lbrace s_0, s_1, s_2 \rbrace$ only have the action choices in $\lbrace up, down \rbrace$, and the actions have deterministic outcomes. On the other hand, the states $\lbrace s_3, s_4 \rbrace$ only have \emph{uncertain} outcomes with probabilistic transitions belonging to intervals.}
	\label{fig:simple}
	
\end{figure}

To do so, we begin by adopting a small extension of (PO)MDPs in which only a subset of the actions are available in any given state, i.e., the transition function should be interpreted as a partial function.
We denote the set of actions at state $s$ by $\Act(s)$. 
We ensure that any states that share an observation also share the set of available actions.
Moreover, we translate the observation function to be deterministic without uncertainty, i.e., of the form $\ObsFun\colon\states\rightarrow \ObsSym$, by expanding the state space~\citep{ChatterjeeCGK16}.
\begin{definition}[Binary/Simple uncertain \gls{pomdp}]
	An uncertain \gls{pomdp} is \emph{binary}, if $|\Act(s)|\leq 2$ for all $s \in S.$ A binary uncertain \gls{pomdp} is \emph{simple} if for all $s \in S$, the following is true:
	\begin{align*}
	|\Act(s)|=2 \;\text{implies}\; \forall \act \in \Act(s). \,\exists s' \in S, \;\probumdp(s,\act,s') =1,
	\end{align*}
\end{definition}%
Simple \glspl{upomdp} differentiate the states with \emph{action} choices and \emph{uncertain} outcomes. 
All \glspl{upomdp} can be transformed into simple \glspl{upomdp}. 
We refer to~\citep{junges2018finite} for a transformation.
This transformation preserves the optimal expected reward of an \gls{upomdp}.
We denote $S_{\mathrm{a}}$ by the states with action choices, and $S_{\mathrm{u}}$ the states with uncertain outcomes.
We now give an example to a simple and binary POMDP in Figure~\ref{fig:simple}.

We now introduce the optimization problem with the nonnegative reward variables $\{r_s \geq 0\; | \; s \in S\}$ denoting the expected reward before reaching goal set $G$ from state s, and positive variables $\{\osched_{s,\act}> 0 \; | \; s \in S, \act \in \Act(s)\}$ denoting the probability of taking an action $\act$ in a state $s$ for the policy.
Note that we only consider policies where for all states $s$ and actions $\act$ it holds that $\osched_{s,\act} > 0$, such that applying the policy to the uncertain \gls{pomdp} does not change the underlying graph.%
\begin{align}
\text{maximize}& \quad    r_{s_I}
\label{aaai-2020NLP:obj}\\
\text{subject to} &\quad
\,\,  r_{s_I} \geq \kappa,\\
\forall s \in G,&\quad r_s = 0,  \label{aaai-2020NLP:constraint:2}\\
\forall s \in S,& \quad   \sum\limits_{\act \in \ActS} \osched_{s,\act} = 1, \label{aaai-2020NLP:constraint:3}\\
\forall s, s' \in S, \forall \act \in \ActS,&\quad \ObsFun(s) = \ObsFun(s')\implies\osched_{s,\act} = \osched_{s',\act},\label{aaai-2020NLP:constraint:4} \\
\forall s \in S_{\mathrm{u}}.\, \forall P \in \mathcal{P},&\quad   r_s \leq R(s)+\sum_{s' \in S} P(s,s')\cdot r_{s'}.
\label{aaai-2020NLP:simple-uncertain} \\
\forall s \in S_{\mathrm{a}}, \quad r_s \leq &\sum_{\act \in \Act(s)} \osched_{s,\act} \cdot \big(R(s,\act) +  \sum_{s' \in S}  \probumdp(s,\act,s')\cdot r_{s'}\big),
\label{aaai-2020NLP:simple-nondet}
\end{align}%
The objective is to maximize the expected reward $r_{s_I}$ at the initial state.
The constraint~\eqref{aaai-2020NLP:constraint:2} encodes the specification requirement and assigns the expected reward to $0$ in the states of goal set $G$.
We ensure that the policy is a valid probability distribution in each state by~\eqref{aaai-2020NLP:constraint:3}.
Next,~\eqref{aaai-2020NLP:constraint:4} ensures that the policy is observation-based.
We encode the computation of expected rewards for states with uncertain outcomes by~\eqref{aaai-2020NLP:simple-uncertain} and with action choices by~\eqref{aaai-2020NLP:simple-nondet}.
We omit denoting the unique actions in the transition function $P(s,s')$ and reward function $R(s)$ in~\eqref{aaai-2020NLP:simple-uncertain} for states with uncertain outcomes.

Let us consider some properties of the optimization problem.
First, the functions in~\eqref{aaai-2020NLP:simple-nondet} are \emph{quadratic}. 
Essentially, the policy variables $\osched_{s,\act}$ are multiplied with the reward variables $r_{s}$. In general, these constraints are \emph{nonconvex}, and we later \emph{linearize} them.
Second, the values of the transition probabilities $P(s,s')$ for $s, s' \in S_{\textrm{u}}$ in~\eqref{aaai-2020NLP:simple-uncertain}  belong to continuous intervals.
Therefore, there are infinitely many constraints over a finite set of reward variables. These constraints are similar to the LP formulation for MDPs~\citep{puterman2014markov}, and are affine; there are no policy variables.
We refer the reader to~\citep{DBLP:conf/aaai/Cubuktepe0JMST21} for a solution approach of the optimization problem.

\subsection{Spacecraft Motion Planning Case Study}

We evaluate the \gls{scp}-based approach from~\citep{DBLP:conf/aaai/Cubuktepe0JMST21} for solving the \gls{upomdp} problem on a case study based on a satellite motion planning problem.

This case study considers the robust spacecraft motion planning system presented by~\citet{frey2017constrained,hobbs2020taxonomy}.
The spacecraft orbits the earth along a set of predefined natural motion trajectories (NMTs)~\citep{kim2007mission}. 
While the spacecraft follows its current NMT, it does not consume fuel.
Upon an imminent close encounter with other objects in space, the spacecraft may be directed to switch into a nearby NMT at the cost of a certain fuel usage.
We consider two objectives: (1) To maximize the probability of avoiding a close encounter with other objects and (2) to minimize the fuel consumption, both within successfully finishing an orbiting cycle.
Uncertainty enters the problem in the form of potential sensing and actuating errors. 
In particular, there is uncertainty about the spacecraft position, the location of other objects in the current orbit, and the failure rate of switching to a nearby NMT.

\begin{figure*}[t]
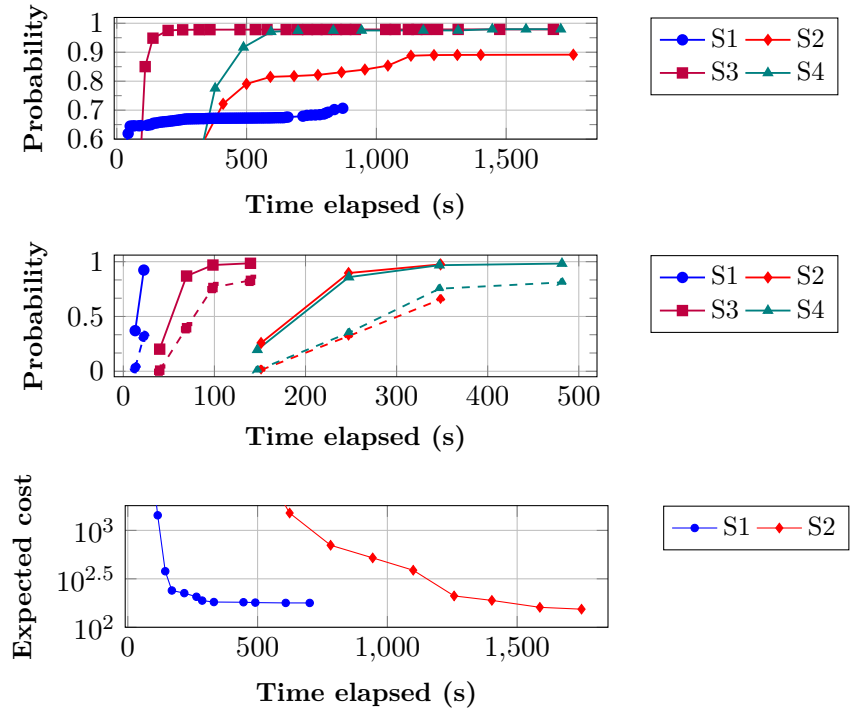

	\centering
	\begin{subfigure}[b]{0.99\textwidth}
	\centering
	\input{./07-part/figures/spacecraft_a}
	\end{subfigure}%
	\hfill
	\begin{subfigure}[b]{0.99\textwidth}
	\centering
	\input{./07-part/figures/spacecraft_b}
	\end{subfigure}%
	\hfill
	\begin{subfigure}[t]{0.99\textwidth}
		\centering
	\input{./07-part/figures/spacecraft_c}
	\end{subfigure}
	\caption{Computational effort versus the performance of the different policies for the spacecraft motion planning case study. Top: Obtained probability of avoiding close encounters between the spacecraft and other objects in the orbit. Middle: The performance of policies synthesizes using a nominal \gls{pomdp} model, applied to the nominal model (solid lines) and to the \gls{upomdp} model (dashed lines). Bottom: The obtained expected cost of successfully finishing an orbit.}
	\label{aaai-2020fig:conv}
\end{figure*}

\paragraph{Modeling Spacecraft Motion Planning Using \glspl{upomdp}.}
We encode the problem as an \gls{upomdp} with two-dimensional state variables for the NMT $n~\in~\{1,\ldots,36\}$, and the (discretized) time index $i~\in~\{1,\ldots,I\} $ for a fixed NMT. We use different values of resolution $I$ in the examples. 
Every combination of $\langle n, i\rangle$ defines an associated point in the 3-dimensional space.
The transition model is built as follows.
In each state, there are two types of actions, (1) staying in the current NMT, which increments the time index by $1$, and (2) switching to a different NMT if two locations are close to each other.
More precisely, we allow a transfer between two points in space defined by $\langle n,i\rangle$ and $\langle n',i'\rangle$ if the distance between the two associated points in space is less than 250km. 
A switching action may fail. 
In particular, the spacecraft may transfer to an unintended nearby orbit.
The observation model contains $1440$ different observations of the current locations of the orbit, which give partial information about the current NMT and time index in orbit.
Specifically, for each NMT, we can only observe the location up to an accuracy of $40$ different locations in each orbit. 
The points that are close to each other have the same observation.

\paragraph{Experimental Setup and Algorithm Variants.}
We consider three benchmarks. 
S1 is our standard model with a discretized trajectory of $I=200$ time indices.
S2 denotes an extension of S1 where the considered policy is an \gls{fsc} with $5$ memory states.
S3 uses a higher resolution ($I=600$). 
Finally, S4 is an extension of S3, where the policy is an \gls{fsc} with $2$ memory states.
In all models, we use an uncertain probability of switching into an intended orbit and correctly locating the objects, given by the intervals $[0.50, 0.95]$. %
The four benchmarks have approximately $3.6\mathrm{e}4, 3.5\mathrm{e}5, 1.1\mathrm{e}5$ and $3.4\mathrm{e}5$ states as well as $6.5\mathrm{e}4, 7.0\mathrm{e}5, 2.0\mathrm{e}5$ and $6.7\mathrm{e}5$ transitions, respectively.
In this example, the objective is to maximize the probability of avoiding a close encounter with objects in the orbit while successfully finishing a cycle.
\begin{figure*}[t]
\centering
\begin{subfigure}[t]{0.499\textwidth}
\centering
\includegraphics[trim =0 0 0 0, clip, width=0.999\linewidth]{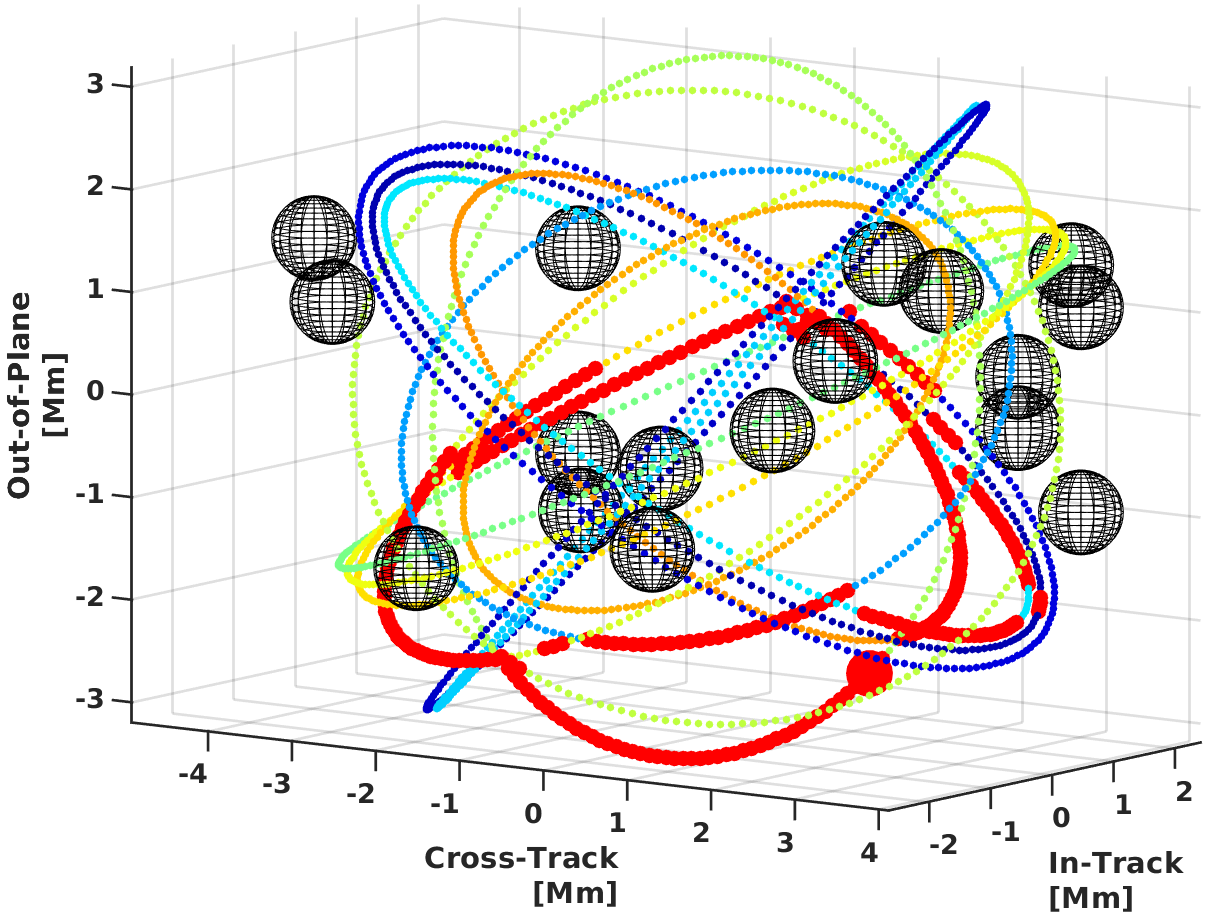}	
	\caption{}
	\label{aaai-2020fig:sat1}	
\end{subfigure}%
\hfill
	\begin{subfigure}[t]{0.499\textwidth}
	\centering
	\includegraphics[trim = 0 0 0 0, clip, width=0.999\linewidth]{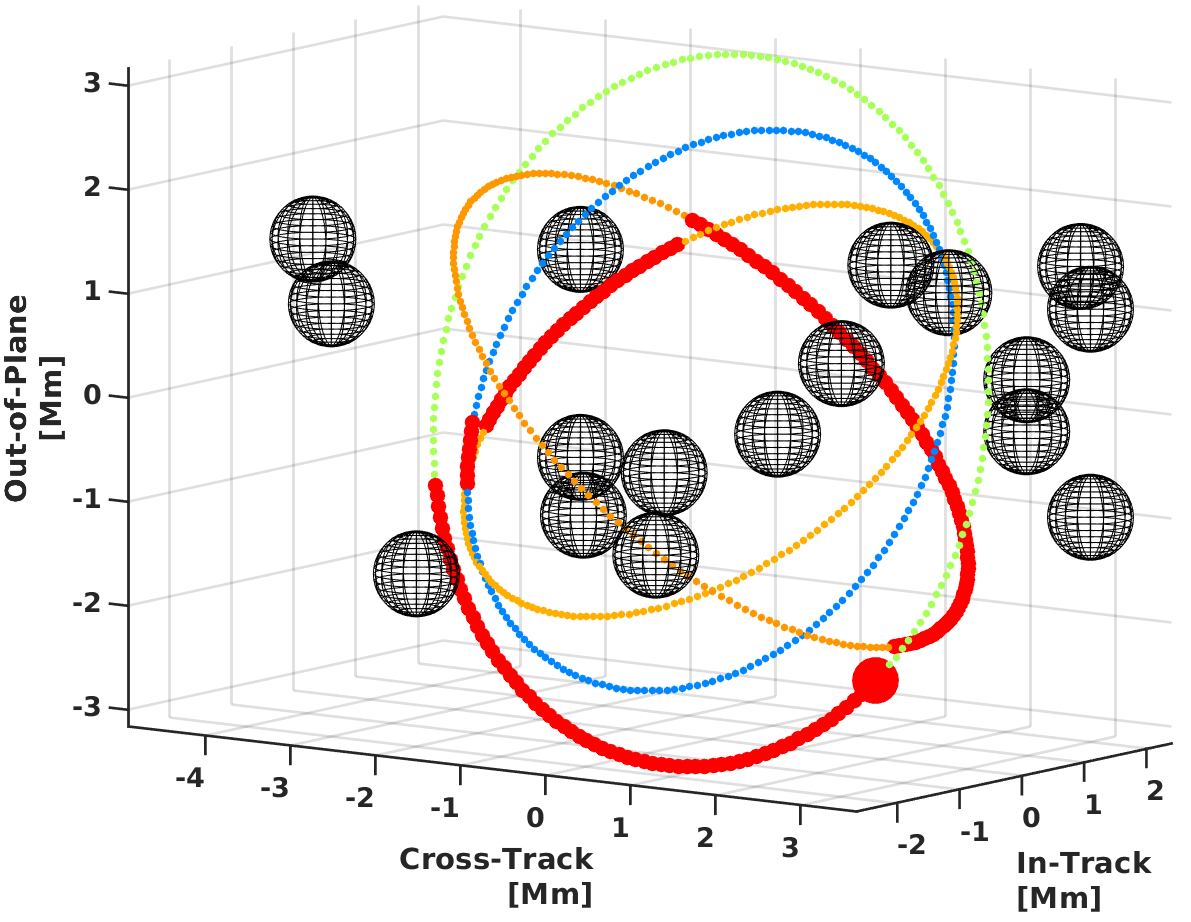}	
	\caption{}
	\label{aaai-2020fig:sat5}
	\end{subfigure}
	\caption{We show the obtained trajectory from a  policy in red that finishes an orbit around the origin. 
    (a) Trajectory from a memoryless policy.
    (b) Trajectory from policy with 5 memory states.
    We highlight the initial location by a big red circle.  
    We depict the other objects by black spheres, and all NMTs that were used as a part of the trajectory in blue, green, or yellow.}
	\label{aaai-2020fig:sat}
\end{figure*}

\paragraph{Memory Yields the Best Policies.}
Fig.~\ref{aaai-2020fig:conv} (top) shows the convergence of the reachability probabilities for each model, specifically the probability versus the runtime in seconds. 
First, we observe that after 20 minutes of computation, using larger \gls{pomdp}s that have a higher resolution or memory in the policy yields superior policies. Second, the policy with memory is superior to the policy without memory.
Finally, we observe that larger models indeed require more time per iteration, which means that on the smaller uncertain \gls{pomdp} S1, the algorithm converges faster to its optimum. 

\paragraph{Comparing the Policy Performances.}
Fig.~\ref{aaai-2020fig:sat} shows a comparison of policies and depicts the resulting spacecraft trajectories.
In particular, we show the trajectories from two different policies, a memoryless policy (one memory state) in Fig.~\ref{aaai-2020fig:sat1}---computed on S1---and from a policy with finite memory (five memory states) in Fig.~\ref{aaai-2020fig:sat5}--- computed on S2. 
The trajectory from the memoryless policy switches its orbit $17$ times, whereas the trajectory from the finite-memory policy switches its orbit only $4$ times.
Additionally, the average length of the trajectory with the finite-memory policy is $188$, compared to $375$ for the memoryless one.
These results demonstrate the utility of finite-memory to improve the reachability probability and to minimize the number of switches.

\paragraph{Robust Policies are Indeed More Robust.}
We demonstrate the utility of computing robust policies against uncertainty in Fig.~\ref{aaai-2020fig:conv} (middle).
Intuitively, we compute policies on nominal models and use them on uncertain models.
We give results on the nominal transition probabilities of the four considered models.
The performance of the policies on the nominal models has  solid lines, and the performance of the policies on the uncertain models has dashed lines.
Note that when we apply the policies synthesized using nominal models on the \gls{upomdp}, they perform significantly worse and fail to satisfy the objective in each case.
The results clearly demonstrate the trade-offs between computing policies for nominal and uncertain problems.
In each case, the computation time for the problem with uncertainty is roughly an order of magnitude larger. 
Yet, the resulting policies are better: we observe that the probability of a close encounter with another object increases up to $60$ percentage points, if we do not consider the uncertainty in the model.

\paragraph{Expected Energy Consumption.}
Finally, we consider an example where there is a cost for switching orbits.
In this example we only consider models S1 and S2, which both have discretized trajectories with $I=200$ time indices.
The objective is to minimize the cost of successfully finishing a cycle in orbit.
We obtain the cost of each switching action according to the parameters in~\citep{frey2017constrained}.
Additionally, we define the cost of a close encounter to be $10000 N{\cdot}s$, a much higher cost than all possible switching actions.
We reduce the uncertainty in the model by setting the previously mentioned intervals for these models to $[0.90, 0.95]$.
In particular, the worst-case probability to correctly detect objects is now higher than before, reducing the probability of close encounters with those objects.
Fig.~\ref{aaai-2020fig:conv} shows the convergence of the costs for each model.
The costs of the resulting policies for models S1 and S2 are $178N{ \cdot }s$ and $153 N{ \cdot }s,$ respectively.
Similar to the previous example, the results demonstrate the utility of finite-memory policies in reducing fuel costs in spacecraft motion planning problems.

\section{Task-Oriented Active Perception and Planning}\label{sec:task}
In the previous subsections we presented \glspl{fsc} for \glspl{pomdp}, as well as an approach to compute robust policies for \glspl{upomdp}.
In general, \glspl{pomdp} provide a useful modeling framework for policy synthesis when perception and planning are closely coupled.
However, in some cases the uncertainty stemming from perception-based information limitations must be decoupled from the the uncertainty that arises from the stochasticity of the underlying system dynamics. For instance, in the case of probabilistic knowledge over atomic propositions, a \glspl{pomdp} model would require an exponential expansion in its state space, resulting in a computationally intractable problem.  

Following the presentation of \citet{ghasemi2020task}, we now present a model and an accompanying algorithm that is able to separately reason about these two sources of uncertainty.
More specifically, we consider an agent that is assigned a temporal logic task in an environment that is only partially known.
We represent the environment using semantic labeling, i.e., with a set of state properties (labels) captured by atomic propositions over which the agent holds a probabilistic belief that is updated as new sensory measurements arrive. 
The goal is to design a joint perception and planning strategy for the agent that realizes the task with high probability. 
We present a planning strategy that takes the semantic uncertainties into account and by doing so provides probabilistic guarantees on the task success. 
Furthermore, as new data arrives, the belief over the atomic propositions evolves and the planning strategy adapts accordingly.

Temporal logic planning under imperfect perception has been studied from different perspectives.
\citet{jones2013distribution} proposed a new type of logic, called distribution temporal logic, to enable expressions over belief-based predicates.
In~\citet{ding2011ltl}, the authors considered an agent moving over a graph where the truth values of the predicates over the nodes depend on known probabilities.
There is also a family of solutions that rely on sampling methods~\citep{vasile2013sampling}.
In another related work, \citet{da2019active} propose a synthesis algorithm for probabilistic temporal logic over reals specifications in the belief space.
\citet{montana2017sampling} propose a sampling-based solution to temporal logic planning under imperfect state information, relying on constructing a transition system by sampling from a feedback-based information roadmap.
The work of \citet{kress2009temporal} considers uncertainty in the environment propositions and proposed to design a reactive controller offline such that it can satisfy the task for all admissible environments. 

Many solutions resort to replanning techniques, such as the iterative receding-horizon planning algorithm by \citet{wongpiromsarn12tac} mentioned in Section \ref{sec:rhtlp}.
For a subclass of temporal logic formulas, \citet{livingston2012backtracking} introduce a method to locally patch a nominal controller once a change in the environment is detected.
\citet{lahijanian2016iterative} propose an iterative replanning strategy that can relax the constraints imposed by the task if the discovery of a new obstacle deems the task unrealizable.
\citet{fu2015integrating,fu2016synthesis} design an alternating active sensing and planning strategy for temporal logic tasks. 
In the work of \citet{guo2013revising} the agent generates a plan according to its initial knowledge over a deterministic model and after new real-time information is gathered, it revises the plan.

\subsection{Problem Formulation}\label{sec:task-pro}

Returning to the presentation of \citet{ghasemi2020task}, we now introduce the agent model, the environment model, the observation model and the task specification language that we use in the formal problem statement.

\subsubsection{Agent and Environment Models}

We model the interaction between the agent and the environment by a \gls{mdp}. In particular, we use $\mdp = (\st, \sinit, \act, \tr)$ to represent the \gls{mdp} modeling the agent's decision-making.

The agent perceives its environment through atomic propositions. Different from the approaches discussed above, we use a time-varying labeling function to capture the belief of the agent about the truth values of the atomic propositions.
\begin{definition}
	An environment model is a tuple $\env = (\st, \AP, \Lb)$ where $\st$ is a finite, discrete state space, $\AP$ is a set of atomic propositions, and $\Lb : \st \to 2^{\AP}$ is a deterministic labeling function that captures the true state of the environment. The agent's belief at time $t$ is a probabilistic labeling function $\L_t : \st \times 2^{\AP} \to [0,1]$. For a state $s$ and a subset of atomic propositions $P \subseteq \AP$, $\L_t(s,P)$ assigns the probability of the event that $P$ holds true at $s$, i.e., $\P(\bigcap_{p \in P} s \models p)$.
\end{definition}
We denote by $\L_0$ the agent's prior belief. 
This prior belief may, for example, be an uninformative prior distribution. 
We assume that the truth values of the propositions are mutually independent in each state,
facilitating the update of the labeling function over time. Nevertheless, so long as the joint distribution model is known, the updates can be computed.

\subsubsection{Observation Model}
At each time step, the agent's perception module processes a set of sensory measurements regarding the atomic propositions. While the measurements may be from multiple sensing units, for ease of notation, we consider their joint model by a general observation function.
\begin{definition}
	Let $\z(s_1,s_2,p) \in \{\true,\false\}$ denote the perception output of the agent at state $s_1$ for the atomic proposition $p$ at state $s_2$.
	The joint observation model of the agent is $$\obs : \st \times \st \times \AP \times \{\true,\false\} \to [0,1],$$ where $\obs(s_1,s_2,p,b)$ represents the probability that $\z(s_1,s_2,p) = \true$ if the truth value of $p$ is given by the Boolean variable $b$.
\end{definition}
More specifically, $\z(s_1,s_2,p)$ represents a measurement of the property captured by $p$ at state $s_2$, i.e., whether $p$ holds at $s_2$ or not, when the agent takes this measurement from state $s_1$, i.e., the current state of the agent is $s_1$.
Hence, $\z(s_1,s_2,p)$ is a Bernoulli random variable and its distribution is dictated by the observation model $\obs(s_1,s_2,p,b)$, which depends on the true value of $p$ at state $s_2$ captured by $b$, i.e., $b$ indicates whether $p$ in reality holds at $s_2$ or not. An accurate observation model is one for which the output probability of $\obs(s_1,s_2,p,b)$ is one for $b = \true$ and zero for $b = \false$.

In the Bayesian framework, the observation model is used to update the agent's belief. Nevertheless, in the absence of such an observation model, one can perform the update in a frequentist way.

\subsubsection{Task Specification Language}

We use \gls{scltl}~\citep{kupferman2001model} to specify finite-horizon tasks for the agent.
Notice that \gls{scltl} is a variant of \gls{ltl}, introduced in Section \ref{sec:spec}, that deals only with finite-horizon tasks. Hence, one would need to investigate finite traces of a system for verification purposes. 
The language defined by an \gls{scltl} formula can equivalently be represented through a \gls{dfa} $\dfa = (\Q,\qinit,\Sigma,\del,\Facc)$ \citep{kupferman2001model}.

\subsubsection{Problem Statement}

We consider an agent whose interaction with the environment is captured by an \gls{mdp}. The agent is tasked with an \gls{scltl} specification that can be successfully completed within finite time. The agent is unaware of the environment state. However, it starts with a prior knowledge and over time, gathers observations that can be used to further revise its belief. 
The formal definition of the problem is stated next.
\begin{problem}
	Given an \gls{mdp} $\mdp = (\st, \sinit, \act, \tr)$, an environment model with unknown labeling function $\env = (\st, \AP,\_)$, an observation model $\obs$, and an \gls{scltl} task $\varphi$, find a policy $\pi$ that maximizes the probability of satisfying the task conditioned on the true labeling function, i.e.,
	\begin{equation*}
	\pi^* = \argmax_{\pi} \P(\mdp^\pi \models \varphi \,|\, \Lb).
	\end{equation*}
\end{problem}

\subsection{Joint Active Perception and Planning}\label{sec:task-alg}

\begin{figure}[t]
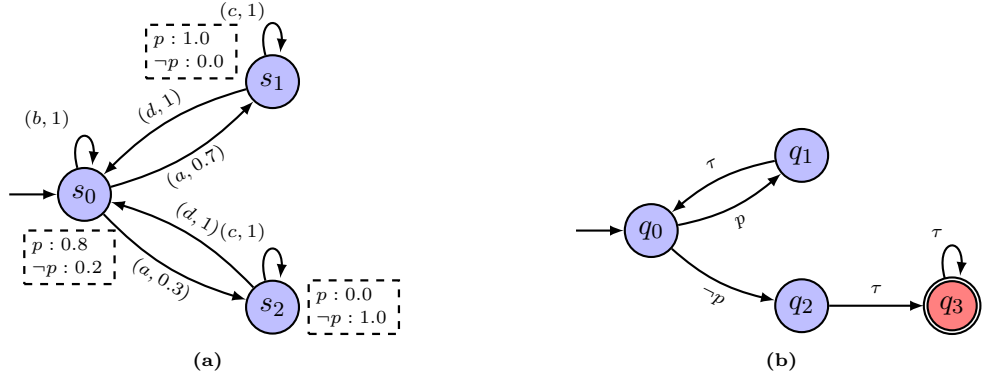

	\centering
	\begin{subfigure}{0.495\textwidth}
		\centering
		\input{figures/task/figure7_10a.tikz}
		\caption{}
	\end{subfigure}
	\hfill
	\begin{subfigure}{0.495\textwidth}
		\centering
		\input{figures/task/figure7_10b.tikz}
		\caption{}
	\end{subfigure}
	\caption{The \gls{mdp} transitions behave in a stochastic way while the uncertainty in knowledge does not. (a) An \gls{mdp}. The edge labels represent an action and a transition probability, respectively, while the node labels capture agent's knowledge about the value of property $p$ at each state. (b) A \gls{dfa} with knowledge uncertainty. The edge labels represent properties' valuations that lead to a transition where $\tau$ denotes any valuation (tautology). Node $q_3$ is the accepting state.}
	\label{fig:task-prob-nat}
\end{figure}

\begin{figure*}[t]
	\centering
	\includegraphics[width=1\textwidth]{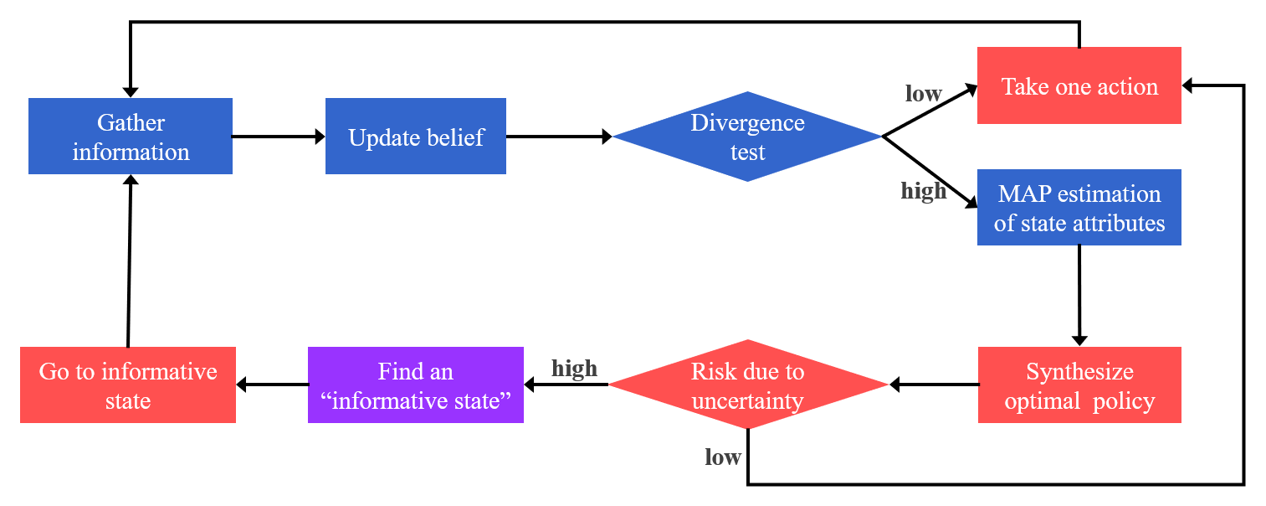}
	\caption{The schematic of the perception-planning loop. The blue blocks refer to pure perception modules and the red blocks refer to pure planning modules, and the purple block indicates a combined perception and planning module of the algorithm.}
	\label{fig:task-alg}
\end{figure*}

Before introducing the algorithm, it is necessary to first describe the challenges of having a probabilistic view of the environment and, in particular, atomic propositions. 
In a setting where the agent is uncertain about the valuations of all atomic propositions over the whole environment, there may be up to $2^{|\st||\AP|}$ possibilities for how the environment is configured. In this case, computing policies that can account for all possible configurations, as offline reactive synthesis does~\citep{BK08}, is indeed computationally intractable. 
Additionally, if the environment is not dynamically changing, then such a comprehensive policy that accounts for all possible configurations, is not necessary. 
Another fundamental challenge is the fact that the nature of the probabilistic perception differs from the stochasticity of the agent model. Therefore, as seen in Example~\ref{ex:task-prob-nat}, one cannot combine the belief probabilities on the perception side with the transition probabilities of the \gls{mdp}.
\begin{example}[Stochastic transition versus knowledge uncertainty]\label{ex:task-prob-nat}
	Consider the simple \gls{mdp} and \gls{dfa} in Figure~\ref{fig:task-prob-nat}. The \gls{dfa} accepts any path on the \gls{mdp} whose induced word is in the form of $(p \tau)^* \neg p \tau$. For instance, if the true labels are $s_0 \models p$, $s_1 \models p$, and $s_2 \models \neg p$, then the path 
	$$s_0 \stackrel{a}{\longrightarrow} s_2 \stackrel{c}{\longrightarrow} s_2 \stackrel{d}{\longrightarrow} s_0$$ 
	on the \gls{mdp} generates the run 
	$$q_0 \stackrel{p}{\longrightarrow} q_1 \stackrel{\neg p}{\longrightarrow} q_0 \stackrel{\neg p}{\longrightarrow} q_2 \stackrel{p}{\longrightarrow} q_3$$ 
	on the \gls{dfa} which is accepting and satisfies the task.
	
	A key observation is that the nature of the probabilities of the \gls{mdp} transitions are distinct from that of the agent's belief. A stochastic transition means that if the agent takes the same action at the same state multiple times, every time the next state is determined by the given probabilities. Therefore, a path like
	$$s_0 \stackrel{a}{\longrightarrow} s_1 \stackrel{d}{\longrightarrow} s_0 \stackrel{a}{\longrightarrow} s_2$$
	is possible on the \gls{mdp}.
	On the other hand, the true labels of the states are fixed and the distribution of the agent's belief does not translate into similar behavior. For instance, the path
	$$s_0 \stackrel{b}{\longrightarrow} s_0 \stackrel{b}{\longrightarrow} s_0 \stackrel{b}{\longrightarrow} s_0$$
	on the \gls{mdp} cannot generate the run
	$$q_0 \stackrel{p}{\longrightarrow} q_1 \stackrel{p \lor \neg p}{\longrightarrow} q_0 \stackrel{\neg p}{\longrightarrow} q_2 \stackrel{p \lor \neg p}{\longrightarrow} q_3$$
	on the \gls{dfa}. Since in this run, the truth value assignment of property $p$ at state $s_0$ is inconsistent.
\end{example}

The outline of the proposed algorithm is illustrated in Figure~\ref{fig:task-alg}. At each state, the agent gathers some perception outputs, e.g., sensing measurements, and uses them to update its belief about the environment. If the updated belief, called the posterior belief, is significantly different from the previous one, called the prior belief, the agent must replan. Otherwise, it will continue with its previous policy to take a step. To check the significance of the added information from the new perception data, we compute the difference between the prior and posterior beliefs using statistical distance measured by a divergence metric. If the threshold (a hyperparameter given to the algorithm) on this difference is exceeded, the agent first estimates the most probable configuration of the environment. Then, the agent applies a synthesis algorithm with the estimated environment model. This synthesis algorithm outputs a strategy that maximizes the probability of satisfying the given temporal logic task. The given strategy induces a Markov chain that is used to calculate the risk due to perception uncertainty. If the risk is lower than a threshold (another hyperparameter given to the algorithm), the agent uses the computed policy to take a step. Otherwise, it will find an active perception strategy to reduce its perception uncertainty. 
We now proceed to explain different stages of the proposed algorithm.

\subsubsection{Information Processing}

Consider the agent's state to be $s_t$ at time $t$. The agent will receive new perception outputs according to the observation model $\obs(s_t,.,.,.)$ for all states and atomic propositions.

The agent employs the observations to update its learned model of the environment in a Bayesian approach. For ease of notation, let $\L_t \equiv \L_t(s,p)$ and $\obs(b) \equiv \obs(s_t,s,p,b)$. Given the prior belief of the agent $\L_{t-1}$ and the received observations $\z$, the posterior (updated) belief follows
\begin{equation*}
\P(s \models p | \z(s_t,s,p) = \true) =
\frac{\L_{t-1} \obs(\true)}
{\L_{t-1} \obs(\true) + (1-\L_{t-1}) \obs(\false)},
\end{equation*}
\begin{align*}
    \P(s \models p | \z(s_t,s,p) &= \false) = \\
    &\frac{\L_{t-1} (1-\obs(\true))}
    {\L_{t-1} (1-\obs(\true)) + (1-\L_{t-1}) (1-\obs(\false))},
\end{align*}
for all $s \in \st$ and $p \in \AP$. Depending on the truth value observed for $p$,  $\L_t(s,p)$ will be updated according to one of the above expressions. Besides, for any $P \subseteq \AP$,
$$\L_t(s,P) = \prod_{p \in \AP} \L_t(s,p).$$

\subsubsection{Divergence Test on the Belief}

If the agent's knowledge about the environment configuration has not significantly changed from its knowledge in the previous state, then the agent will continue with its previous policy. Nevertheless, if its knowledge has significantly changed, the agent will synthesize a new policy. We use the Jensen-Shannon divergence to quantify the change in the belief distribution between two consecutive time steps. The cumulative Jensen-Shannon divergence over the states and the propositions can be expressed as
\begin{equation*}
\begin{aligned}
\jsd(\L_{t-1} \| \L_t) = &\frac{1}{2} \dkl(\L_{t-1} \| \L_m^t) + 
\frac{1}{2} \dkl(\L_t \| \L_m^t),
\end{aligned}
\end{equation*}
where $\dkl(. \| .)$ is the Kullback–Leibler divergence between two probability distributions and $\L_m^t = 1/2 \left(\L_{t-1} + \L_t\right)$ is the average probability distribution. One of the input parameters to the algorithm is a threshold $\gamma_d$ on the above divergence. If $\gamma_d$ is not exceeded, the agent uses its previous policy $\pi_{t-1}$ to pick an action and transitions according to its outcome. Otherwise, it has to synthesize a new policy.

For the policy synthesis step, the agent first estimates the most probable environment configuration from the distribution dictated by its updated belief. Let $\Lh_t : \st \to 2^{\AP}$ indicate the agent's inference of the environment configuration at time $t$. The maximum a posteriori estimation is fairly simple as it decomposes into finding the mode of the posterior distribution for each property at each state. For binary-valued atomic propositions that follow Bernoulli distributions, the inference turns into picking the more probable outcome for each property at each state, i.e.,
$$\Lh_t(s) = \{p \in \AP \,|\, \L_t(s,p) \ge 0.5\}.$$

\subsubsection{Policy Synthesis}

Finding an optimal policy, i.e., a policy that maximizes the probability of realizing a temporal logic task, translates into a reachability criterion on the product \gls{mdp}. Let $\Facc^\mdp_\dfa = \st \times \Facc$ denote the equivalent accepting states on the product \gls{mdp}. The agent must find a policy that with high probability reaches to $\Facc^\mdp_\dfa$.

Given that there exists an optimal memoryless deterministic policy on the product \gls{mdp}~\citep{BK08}, one can restrict the search space to that of memoryless deterministic policies and formulate
\begin{equation}\label{eq:task-opt-app}
\pi_t = \argmax_{\pi \in \Pi_{nm,d}} \P(\mdp_\dfa^\pi \models \Eventually \Facc^\mdp_\dfa \,|\, \Lh_t),
\end{equation}
where $\Pi_{nm,d}$ is the set of memoryless and deterministic policies.
To find $\pi_t$ in~\eqref{eq:task-opt-app}, we use a \gls{lp} approach~\citep{BK08}. 
The optimal value of the linear program is the maximum probability of reaching the set of accepting states. In order to find the corresponding optimal policy, it suffices to find the actions for which the corresponding \gls{lp} constraints are active. If there are more than one action with active constraint for a state, any of those actions can be chosen arbitrarily.

\subsubsection{Risk Assessment of Imperfect Perception}

To assess the risk of the computed policy due to perception uncertainties, we now factor in the probabilistic belief of the agent over the environment properties. In particular, we first generate the induced Markov chain $\mdp_\dfa^{\pi_t}$ by applying the policy $\pi_t$ over the product \gls{mdp} $\mdp_\dfa$.
Next, we verify the induced Markov chain with the uncertain labels against the task specification. 
We develop an algorithm via a computation graph that yields the exact probability of the task realization. 
However, due to the complexities explained in Example~\ref{ex:task-prob-nat}, such quantitative analysis has exponential complexity, as formalized in the next theorem.
\begin{theorem}
	Let $\mdp_\dfa^{\pi_t}$ to be a Markov chain with $n$ states and $\L_t$ to be a fully probabilistic labeling function (i.e., all labels are uncertain) over $m$ atomic propositions. Quantitative verification of $\mdp_\dfa^{\pi_t}$ against a reachability specification has a complexity of $O(n2^{nm})$.
\end{theorem}

Since the complexity of an exact quantitative analysis is prohibitive, we instead propose a statistical verification. More specifically, we approximate the expected value of the probability of the task realization over all possible instances of the environment
\begin{equation*}
\E_{\L \sim \Dist(\L)} \left[\P\left(\mdp_\dfa^{\pi_t} \models \varphi\right)\right]
\end{equation*}
by an empirical expectation with $N$ samples
\begin{equation*}
\hat{\E}_{\L \sim \Dist(\L)} \left[\P\left(\mdp_\dfa^{\pi_t} \models \varphi\right)\right] = 
\frac{1}{N} \sum_{i=1}^{N} \P\left(\mdp_\dfa^{\pi_t} \models \varphi | \L_i\right),
\end{equation*}
where $\L_i$ are samples drawn from $\Dist(\L) = \L_t$.
By the application of Hoeffding's inequality~\citep{hoeffding1994probability}, we establish the following concentration result for this approximation.
\begin{theorem}
	Let $f(\L_i) = \P\left(\mdp_\dfa^{\pi_t} \models \varphi | \L_i\right)$ denote the output of verification for an environment modeled by $\L_i$. The empirical expectation of $\E_{\L \sim \Dist(\L)} \left[f(\L)\right]$ with $N$ samples has the following concentration bound
	\begin{equation*}
	\P\left(\abs{\E_{\L \sim \Dist(\L)} \left[f(\L)\right] - \frac{1}{N} \sum_{i=1}^{N} f(\L_i)} \geq \epsilon\right)
	\leq 2\exp(-2N\epsilon^2).
	\end{equation*}
\end{theorem}
It is worth noting that $\P\left(\mdp_\dfa^{\pi_t} \models \varphi | \L_i\right)$ itself is the output of a system of linear equations~\citep{BK08} that depend on the sampled labeling function $\L_i$. Therefore, further characterization of $f(\L_i)$ such as bounding its higher moments is very difficult. $\P\left(\mdp_\dfa^{\pi_t} \models \varphi | \L_i\right)$ can also be computed via statistical verification techniques by sampling paths over the Markov chain~\citep{agha2018survey}.

For a policy $\pi_t$, we define a risk parameter
\begin{equation}
\risk(\mdp_\dfa,\pi_t,\L_t,\varphi) = \Big\lvert\P(\mdp_\dfa^{\pi_t} \models \varphi \,|\, \Lh_t)
- \E_{\L \sim \Dist(\L)} \left[\P\left(\mdp_\dfa^{\pi_t} \models \varphi\right)\right]\Big\rvert,
\end{equation}
which accounts for the variation in the task realization guarantee of the policy with respect to the perception uncertainty. Another input parameter to the proposed perception and planning algorithm is a threshold $\gamma_r$ on the risk due to perception uncertainty. If $\gamma_r$ is not exceeded, the agent acts according to the computed policy $\pi_t$. Otherwise, it takes an active perception strategy as explained next.

\subsubsection{Active Perception Strategy}

\begin{figure}[t]
	\centering
	\input{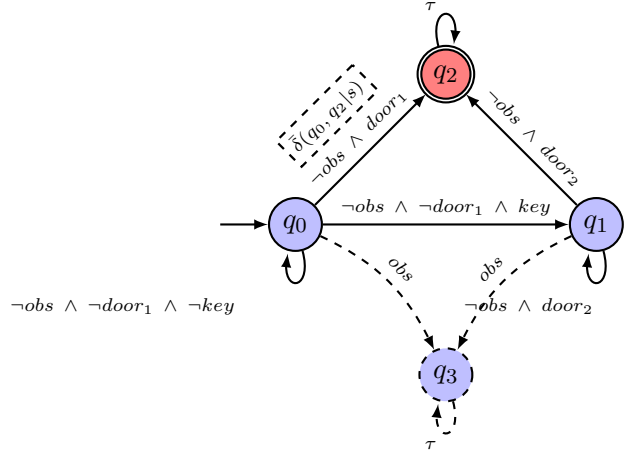}
	\caption{Total probabilistic finite automaton for the formula $\varphi = (\neg {obs} \,\Until\, {door}_1) \lor ((\neg {door}_2 \,\Until\, {key}) \land (\neg {obs} \,\Until\, {door}_2))$. State $q_3$, a sink state, and the transitions to it have been added to make the automaton \textit{total}. The transitions between the states of the automaton are probabilistic where the probabilities depend on the belief over the states' propositions. For example, $\delp(q_0,q_2|s)$ indicates how $\L_t(s,.)$ determines the probability of the transition between $q_0$ and $q_2$.}
	\label{fig:task-prob-aut}
\end{figure}

We develop an algorithm to compute an active perception strategy in the form of a sequence of actions that the agent follows to reduce its perception uncertainty. 
We consider three criteria for computing an active perception strategy. First, such a strategy should enable the agent to reduce its uncertainty about the value of the propositions that affect the task progress. These propositions are the ones that enable the transitions from the current stage of the task, i.e., state of the automaton, to the next ones. For example in Figure~\ref{fig:task-prob-aut}, if the agent is at state $q_1$, the propositions that matter are $obs$ and $door_2$. To measure the uncertainty reduction, we use expected entropy of the said propositions over the whole state space. Second, an active perception strategy must not affect the stage of the task and so, the agent has to remain in the same state of the automaton. Third, after the agent completes the sequence of actions, it should be able to return to the point from which it started the active perception strategy. 

Based on these criteria, we propose an algorithm
to construct an active perception strategy. The algorithm takes a bound $C_\act$ on the number of actions and uses that to construct a tree of depth $C_\act$. Each node of the tree has four parameters: distribution over the states of the \gls{mdp}, distribution over the states of the \gls{dfa}, expected entropy reduction, and reachability probability back to the root node. The distribution over the states of the \gls{mdp} depends on the \gls{mdp}'s transitions while the distribution over the states of the \gls{dfa} depend on the agent's belief over the propositions, as shown in Figure~\ref{fig:task-prob-aut}.
Once the tree is generated, the algorithm picks the best node using a hyperparameter $\beta$ that weighs safety versus information quality. Safety refers to the ability of the agent to remain in the same state of the automaton as well as its ability to return to the root node while information quality refers to the amount of entropy reduction. The sequence of actions leading to the optimum node with respect to a combination of safety and information quality results in the active perception strategy. 
After following this sequence of actions, the perception-planning loop starts over.

\subsection{Task-Oriented Active Perception and Planning Case Studies}
\subsubsection{Simulating Planar Navigation with Finite-Horizon Tasks}
We consider an agent that navigates in a discretized 2D environment and has a finite-horizon task. For instance, the task encoded as a \gls{dfa} in Figure~\ref{fig:task-prob-aut} asks the agent to either go to the state where $door_1$ is located or find a $key$ and go to the state where $door_2$ is located, while avoiding the obstacles. We implemented different versions of the task-oriented perception and planning algorithm to evaluate the effect of each module on the performance. 

\begin{figure}[!t]
	\centering
	\includegraphics[width=0.5\textwidth]{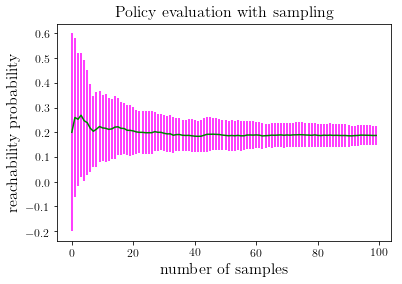}
	\caption{Statistical verification of an induced Markov chain with uncertain atomic propositions.}
	\label{fig:task-statver}
\end{figure}

Table~\ref{tab:task-res} reports the results for a reach-avoid task in an environment with 64 states and with randomly generated obstacles and target. In the table, \textit{No perc.} refers to a baseline scenario where the agent estimates the environment configuration with its prior knowledge and plans according to that. \textit{Perc. w/ no update} is a perception strategy that incorporates only the most recent perception output. \textit{Perc. w/ update} is perception with a Bayesian update, as described in Section~\ref{sec:task-alg}. 
With the exception of the first algorithm, all algorithms have a replanning module, however, the ones with \textit{div.} replan only if the divergence threshold over the change in the belief is exceeded. 
\textit{info.} means that active perception is enabled and hence the agent will perform active perception strategies when the risk due to perception uncertainty is high.
The results show that adding the divergence test reduces the number of policy synthesis steps. 
Furthermore, the divergence test reduces the success rate. 
On the other hand, adding the active perception module increases the success rate.
\begin{table}[!t]\centering\footnotesize
	\caption{Results of planar navigation under a reach-avoid task using different versions of the proposed algorithm. Success is the percentage of runs that complete the task. \#Step is the average length of the runs and \#Plan is the number of times that the agent synthesizes a new policy.}
	\begin{tabular*}{0.8\linewidth}{c|c|c|c}
		\toprule
		Algorithm & Success & \#Step & \#Plan\\
		\midrule
		No perc. & 0\% & 50 & 1\\
		Perc. w/ no update + replan & 0\% & 38.4 & 38.4\\
		Perc. w/ update + replan & 84\% & 21.8 & 21.8\\
		Perc. w/ update + div. & 80\% & 22.8 & 14.8\\
		Perc. w/ update + replan + info. & 92\% & 19.4 & 19.4\\
		Perc. w/ update + div. + info.  & 86\% & 22.6 & 14.6\\
		\bottomrule
	\end{tabular*}
	\label{tab:task-res}
\end{table}
Figure~\ref{fig:task-statver} depicts the results from a risk assessment step for the \gls{mdp} with 64 states and 2 atomic propositions. Even though the size of the sampling space is large ($2^{64}$), it can be seen that the empirical expectation of the reachability probability quickly converges with about 20 samples.

\subsubsection{Drone Navigation in Simulated Urban Environment}

\begin{figure}[!t]
	\centering
	\includegraphics[width=0.5\textwidth]{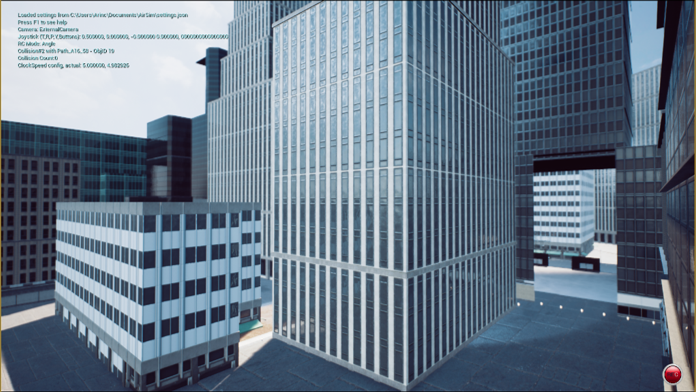}
	\caption{A scene from the created urban environment.}
	\label{fig:task-airsim}
\end{figure}

In the AirSim~\cite{airsim2017fsr} simulator, we designed an urban environment as shown in Figure~\ref{fig:task-airsim} and tasked a drone to fly from an initial state to a specific flagged building while avoiding collision with other entities of the environment. The drone is equipped with 4 cameras and 4 depth sensors. The perception module processes the cameras' readings as well as the depth measurements to map the semantic labels to a discretized model of the environment. We applied the proposed perception and planning scheme. However, in contrast to the previous simulation scenario, an observation model does not exist here. Therefore, we used a frequentist update rule for the agent's belief. Details of the simulation setting as well as recordings of the resulting behavior of the drone are available in \cite{ghasemi2020task}.

\chapter{Runtime Assurance via Shielding}

\newcommand{\buchi}{B\"uchi\xspace}
\newcommand{\win}{\mathsf{win}}
\newcommand{\B}{\mathbb{B}}
\newcommand{\NI}{\mathbb{N}^{\infty}}
\newcommand{\Rp}{\mathbb{R}^+}
\newcommand{\design}{\mathcal{D}}
\newcommand{\shield}{\mathcal{S}}
\newcommand{\obj}{\mathsf{obj}}
\newcommand{\gstates}{G}
\newcommand{\ginit}{g_0}
\newcommand{\init}{q_0}
\newcommand{\din}{I}
\newcommand{\dinalph}{\Sigma_I}
\newcommand{\dinletter}{{\sigma_I}}
\newcommand{\dintrace}{{\overline{\sigma_I}}}
\newcommand{\dout}{O}
\newcommand{\sout}{{O'}}
\newcommand{\doutalph}{\Sigma_O}
\newcommand{\doutletter}{{\sigma_O}}
\newcommand{\douttrace}{{\overline{\sigma_O}}}
\newcommand{\dalph}{\Sigma}
\newcommand{\dletter}{\sigma}
\newcommand{\dtrace}{\overline{\dletter}}
\newcommand{\langset}{\mathcal{L}}
\newcommand{\specv}{\varphi^v}
\newcommand{\distt}{d^\sigma}
\newcommand{\distl}{d^L}
\newcommand{\distta}{\mathcal{M}}
\newcommand{\distla}{\mathcal{N}}
\newcommand{\ds}{\mathsf{d2s}}
\newcommand{\sd}{\mathsf{s2d}}
\newcommand{\prop}{\mathsf{prop}}
\newcommand{\kin}{\!\in\!}
\newcommand{\err}{\text{\color{red}\Lightning}}
\newcommand{\comp}{\circ}

\glsresetall
\blfootnote{This section follows an adapted exposition from~\citet*{humphrey:2016} and~\citet*{konighofer:2017}.}
\emph{Runtime verification} detects violations
of a set of specified properties while a system is executing \citep{leucker:2009}.
An extension of this idea is to perform \emph{runtime enforcement} of specified properties,
in which violations are not only detected but also overwritten
in a way that specified properties are maintained.

A general approach for runtime enforcement of specified properties is
\emph{shield synthesis}, in which a shield monitors the system and
instantaneously overwrites incorrect outputs.  A shield must ensure
both \emph{correctness}, i.e., it corrects system outputs such that
all properties are always satisfied, as well as \emph{minimum
  deviation}, i.e., it deviates from system outputs only if necessary
and as rarely as possible.  The latter requirement is important
because the system may satisfy additional noncritical properties that
are not considered by the shield but should be retained as much as
possible.

Merging formal methods with shielding we require the notion of $k$-stabilizing
shields~\citep{bloem:2015}. 
Given a safety specification, we can identify
wrong outputs, i.e., outputs after which the specification is
violated or, more precisely, after which the environment can force the
specification to be violated. A ``wrong'' trace is then a trace that ends
in a wrong output. The idea of shields is that they may modify the
outputs so that the specification always holds, but that such
deviations last for at most $k$ consecutive steps after a wrong
output.  If a second violation happens during the $k$-step recovery
phase, the shield enters a mode where it only enforces correctness,
but no longer minimizes the deviation. 

\subsection{Overview of Shielding}
We apply shielding to \emph{reactive systems}, which model well with Mealy machines---finite-state machines that depend both on state and current inputs. We can view a Mealy machine formally as the tuple $\mathcal{M} = (\states, \init, \dinalph, \doutalph, \delta, \lambda)$ composed of the following data.
\begin{itemize}
    \item A finite set of states, $\states$.
    \item A starting state, $\init$.
    \item An input alphabet, $\dinalph$.
    \item An output alphabet, $\doutalph$.
    \item A transition function, $\delta\colon \states \times \dinalph \to \states$.
    \item An output function, $\lambda \colon \states \times \dinalph \to \doutalph$.
\end{itemize}

This tuple is the fundamental unit for the formalism of a system model, termed the \emph{design}, and its \emph{shield}.

\begin{definition}[Shield, \citet{bloem:2015}]
Let $\design = (\states, \init, \dinalph, \doutalph, \delta, \lambda)$ 
be a design, $\spec$ be a set of properties, and $\specv\subseteq \spec$ 
be a valid subset such that $\design \models \specv$.  A reactive system 
$\shield = (\states', \init', \dalph, \doutalph, \delta', \lambda')$ is 
a \emph{shield} of $\design$ with respect to $(\spec\setminus\specv)$ 
if and only if $(\design \comp \shield) \models \spec$.
We also require that for $\shield$ to be a it must be 
shield of \emph{any} design $\design$ such that $\design \models 
\specv$.
\end{definition}

\subsubsection{Correctness property}  With correctness we refer to the
property that the shield corrects any design's output such that a
given safety specification is satisfied.

Since a shield must work for any design, the synthesis procedure
does not need to consider the design's implementation. This property
is crucial because the design may be unknown or too complex to
analyze. On the other hand, the design may satisfy additional
(noncritical) specifications that are not specified in $\spec$ but
should be retained as much as possible.

\subsubsection{Minimum deviation property}  Minimum deviation requires a
shield to deviate only if necessary, and as infrequently as possible.
To ensure minimum deviation, a shield can only deviate from the design
if a property violation becomes unavoidable.  Given a safety
specification $\spec$, a shield $\shield$ \emph{does not
  deviate unnecessarily} if for any design $\design$ and any trace
 that is not wrong and
$\design$ does not violate $\spec$, $\shield$ keeps the output of
$\design$ intact.

\subsubsection{Admissibility property} To address shortcoming with $k$-stabilizing shields (1),
we guarantee the following: (a) Admissible shields are subgame
optimal. That is, for any wrong trace, if there is a finite number $k$
of steps within which the recovery phase can be guaranteed to end, the
shield will always achieve this. (b) The shield is \emph{admissible},
that is, if there is no such number $k$, it always picks a deviation
that is optimal in that it ends the recovery phase as soon as possible for
some possible future inputs. As a result, admissible shields work well in settings in which
finite recovery can not be guaranteed, because they guarantee
correctness and may well end the recovery period if the system does
not pick adversarial outputs. To address shortcoming (2), admissible
shields allow arbitrary failure frequencies and in particular failures
that arrive during recovery, without losing the ability to recover.

To create an admissible shield we must first define what $k$-stabilizing a trace means.
Given an unavoidable violation occurring in design $\design$,
the shield can deviate from the expected outputs of the design
for at most $k$ consecutive time 
steps. 
This means that violations cannot be consecutive and can only be tolerated after $k$ steps.
In the event that the design violates the specification within $k$ steps,
then the shield transitions the system to a fail-safe mode

\begin{definition}[Admissible Shield]
A shield $\shield$ is admissible if for any trace,
whenever there exists a $k$ and a shield $\shield'$ such that
$\shield'$ adversely $k$-stabilizes the trace, then $\shield$
adversely $k$-stabilizes the trace. If such a $k$ does not exist for
trace, then $\shield$  collaboratively $k$-stabilizes
the trace for a minimal $k$.
\end{definition}

\section{Shielding a UAV mission}
\label{sec:ex}

In this section, we apply shields on a scenario in which a UAV must
maintain certain properties while performing a surveillance mission in a dynamic environment.
We show how a shield can be used to enforce the desired properties,
where a human operator in conjunction with a lower-level autonomous planner
is considered as the reactive system that sends commands to the UAV's
autopilot.  We discuss how we would intuitively want a shield to
behave in such a situation. %

A common UAV control architecture consists of a ground control station
that communicates with an autopilot onboard the UAV~\citep{chao:2010}.
The ground control station receives and displays updates from the autopilot on the UAV's state,
including position, heading, airspeed, battery level, and sensor imagery.
It can also send commands to the UAV's autopilot,
such as waypoints to fly to.
A human operator can then use the ground control station to plan waypoint-based
routes for the UAV, possibly making modifications during mission execution to respond to events observed through the UAV's sensors.
However, mission planning and execution can be very workload intensive,
especially when operators are expected to control multiple UAVs simultaneously \citep{donmez_2010a}.
To address this issue, methods for UAV command and control have been explored
in which operators issue high-level commands, and automation carries out
low-level execution details.%

Several errors can occur in this type of human-automation paradigm \citep{chen_2012}.
For instance, in issuing high-level commands to the low-level planner,
a human operator might neglect required safety properties due to
high workload, fatigue, or an incomplete understanding
of exactly how the autonomous planner might execute the command.
The planner might also neglect these safety properties either because of software errors or by design. Waypoint commands issued by the operator or planner
could also be corrupted by software that translates
waypoint messages between ground station and autopilot specific formats or
during transmission over the communication link.

As the mission unfolds, waypoint commands will be sent periodically to the autopilot.
If a waypoint that violates the properties is received,
a shield that monitors the system inputs and can overwrite the waypoint
outputs to the autopilot would be able to make
corrections to ensure the satisfaction of the desired properties.

\begin{figure}[tb]
\centering
\includegraphics[width=3in]{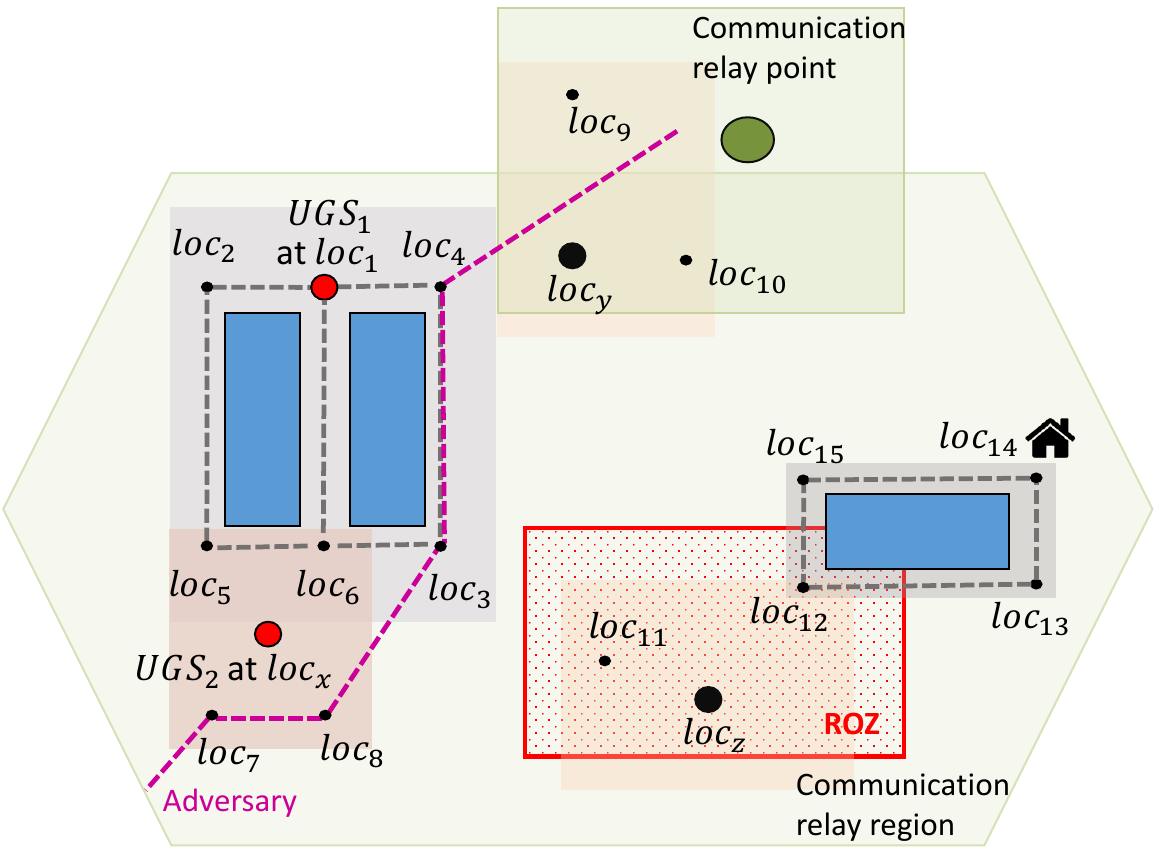}
\caption{A map for UAV mission planning.}
\label{fig:map1}
\end{figure}

\begin{figure}[!t]
  \centering
\includegraphics[width=2in]{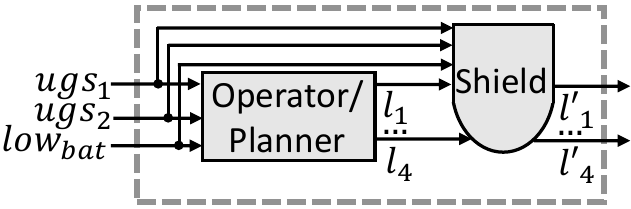}
\caption{The interaction between the operator/planner (acting as a reactive system) and the shield.}
\label{fig:attach_shield}
\end{figure}

Consider a mission map (Figure~\ref{fig:map1}) \citep{Feng16}, which contains three tall buildings (illustrated as blue blocks),
over which a UAV should not attempt to fly.
It also includes two unattended ground sensors (UGS) that provide data
on possible nearby targets, one at location $loc_1$ and one at $loc_x$,
as well as two locations of interest, $loc_y$ and $loc_z$.
The UAV can monitor $loc_x$, $loc_y$, and $loc_z$ from several nearby vantage points.
The map also contains a restricted operating zone
(ROZ), illustrated with a red box, in which flight might be dangerous,
and the path of a possible adversary that should be avoided
(the pink dashed line).
Inside the communication relay region (large green area), communication links
are highly reliable. Outside this region,
communication relies on relay points with lower reliability.
Given this scenario, properties of interest include: %

\begin{itemize}
  \item \textbf{Connected waypoints.} \label{connected}
  The UAV is only allowed to fly to directly connected waypoints.
  \item \textbf{No communication.} The UAV is not allowed to stay in
  a location with reduced communication reliability.
  \item \textbf{Restricted operating zones.} \label{ROZ}
  The UAV has to leave a ROZ within 2 time steps.
  \item \textbf{Detected by an adversary.} Locations on the
  adversary's path cannot be visited more than once over any window of 3 time steps.\label{adversary}
  \item \textbf{UGS.} If a UGS reports a possible nearby target, the UAV
  should visit a respective waypoint within 7 steps
  (for $UGS_1$ visit $loc_1$, for $UGS_2$ visit $loc_5$, $loc_6$, $loc_7$, or $loc_8$).\label{ugs}
  \item \textbf{Go home.} Once the UAV's battery is low, it
  should return to a designated landing site at $loc_{14}$ within 10 time steps.\label{home}
\end{itemize}
The task of the shield is to ensure these properties during operation.
In this setting, the operator in conjunction with a lower-level planner
acts as a reactive system that responds to mission-relevant inputs;
in this case data from the UGSs and a signal indicating whether the battery is low.
In each step, the next waypoint is sent to the autopilot, which is
encoded in a bit representation via outputs
$l_4$, $l_3$, $l_2$, and $l_1$.
The shield monitors (Figure~\ref{fig:attach_shield}) mission inputs and waypoint outputs,
correcting outputs immediately if a violation of the safety properties becomes unavoidable.

\begin{figure}[!t]
      \centering
\includegraphics[width=2.3in]{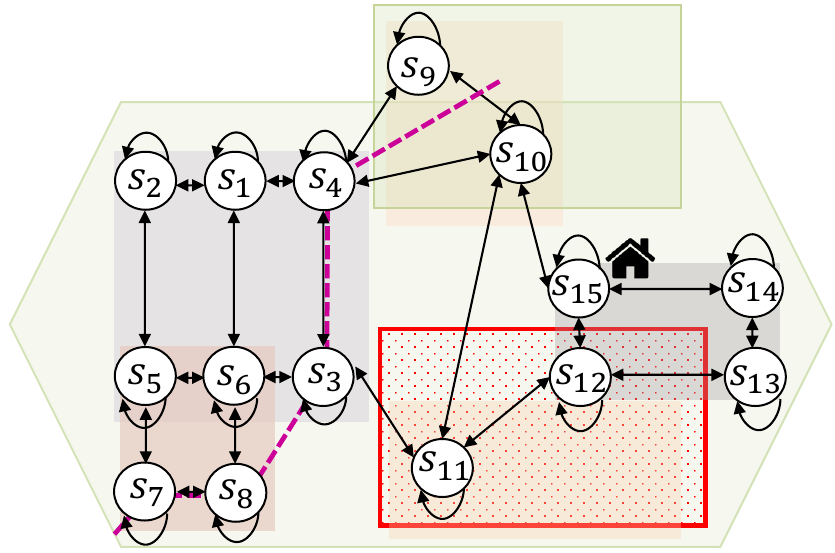}
\caption{Safety automaton of Property~\ref{connected} over the map in Figure~\ref{fig:map1}.}
\label{fig:model_map}
\end{figure}

We represent each of the properties by a safety automaton, the
product of which serves as the shield specification.
Figure~\ref{fig:model_map} models the ``connected waypoints'' property,
where each state represents a waypoint with the same number.
Edges are labeled by the values of the variables $l_4\dots l_1$.
For example, the edge leading from state $s_5$
to state $s_6$ is labeled by
$\neg l_4 l_3 l_2 \neg l_1$.
For clarity, we drop the labels of edges in Figure~\ref{fig:model_map}.
The automaton also includes an error state, which is not shown.
Missing edges lead to this error state, denoting forbidden situations.

How should a shield behave in this scenario?
If the human operator wants to monitor
a location in a ROZ, he or she would like to simply command the UAV
to ``monitor the location in the ROZ and stay there'',
with the planner handling the execution details.
If the planner cannot do this while
meeting all the safety properties, it is appropriate for the shield to revise its outputs.
Yet, the operator would still expect his or her commands to be followed
to the maximum extent possible, leaving the ROZ when necessary and returning whenever possible. Thus,
the shield should minimize deviations from the operator's
directives as executed by the planner.


\subsubsection{Using a $k$-stabilizing shield}
As a concrete example, assume the UAV is currently at $loc_3$,
and the operator commands it to monitor $loc_{12}$.
The planner then sends commands to fly to $loc_{11}$ then $loc_{12}$,
which are accepted by the shield.
The planner then sends a command to loiter at $loc_{12}$,
but the shield must overwrite it to maintain Property~\ref{ROZ},
which requires the UAV to leave the ROZ within two time steps.
The shield instead commands the UAV to go to $loc_{15}$.
Suppose the operator then commands the UAV to fly to $loc_{13}$, while
the planner is still issuing commands as if the UAV is at $loc_{12}$.
The planner then commands the UAV to fly to $loc_{13}$, but
since the actual UAV cannot fly from $loc_{15}$ to $loc_{13}$ directly, the
shield directs the UAV to $loc_{14}$ on its way to $loc_{15}$.
The operator might then respond to a change in the mission and
command the UAV fly back to $loc_{12}$, and the shield again deviates
from the route assumed by the planner, and directs the UAV back to
$loc_{15}$, and so on.
Therefore, a single specification violation can lead to an infinitely long deviation between
the UAV's actual position and the UAV's assumed position.
A $k$-stabilizing shield is allowed to deviate from the planner's commands
for at most $k$ consecutive time steps.
Hence, no $k$-stabilizing shield exists.

\subsubsection{Using an admissible shield}
Recall the situation in which the shield
caused the actual position of the UAV to ``fall behind'' the position assumed by the planner,
so that the next waypoint the planner issues is two or more steps away from the UAV's current
waypoint position.
The shield should then implement a best-effort strategy to ``synchronize'' the UAV's
actual position with that assumed by the planner.
Though this cannot be guaranteed,
the operator and planner are not adversarial towards the shield,
so it will likely be possible to achieve this re-synchronization,
for instance when the UAV goes back to a previous waypoint or remains at the current waypoint
for several steps.
This possibility motivates the concept of an admissible shield.
Assume that the actual position of the UAV is $loc_{14}$ and the its assumed position
is $loc_{13}$. If the operator commands the
UAV to loiter at $loc_{13}$, the shield will be able to catch up
with the state assumed by the planner and to end the deviation by the next
specification violation.

\section{Synthesizing Admissible Shields}
\label{sec:sol}

\begin{figure}[tb]
  \begin{center}
    \includegraphics[width=0.85\textwidth]{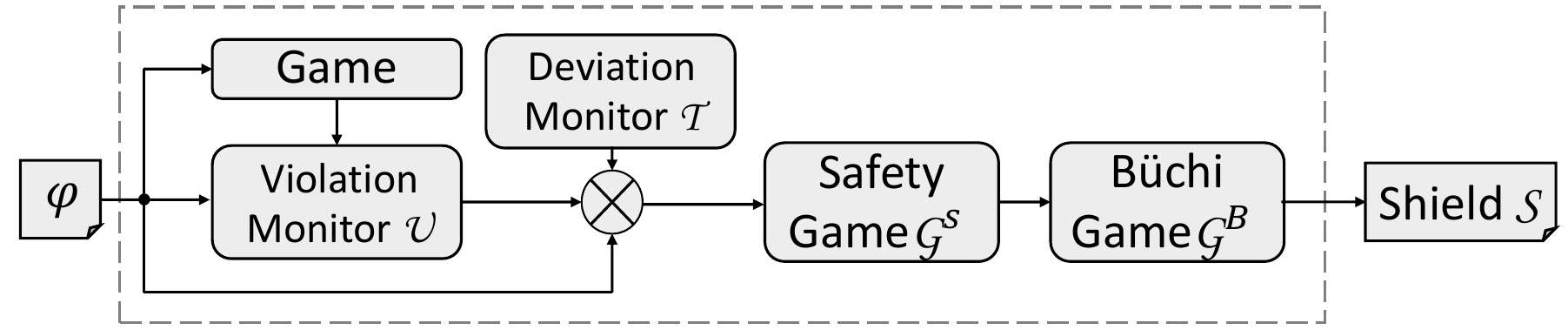}
    \caption{Synthesizing admissible shields.}
    \label{fig:c-stab}
  \end{center}
\end{figure}

We will illustrate shield synthesis using the admissible shielding definition,
but the process is similar for other types of shields.
Starting from a safety specification $\spec = (Q, q_{0}, \dalph, \delta,F)$
with $\dalph=\dinalph\times\doutalph$, the admissible shield synthesis procedure
consists of five steps (Figure~\ref{fig:c-stab}).

\subsubsection{Step 1. Constructing the Violation Monitor $\mathcal{U}$}
\label{sec:violation_monitor}
From $\spec$ we build the automaton $\mathcal{U} = (U, u_0, \dalph, \delta^u)$ to monitor property
violations by the design.  The goal is to identify the latest point in
time from which a specification violation can still be corrected with a
deviation by the shield. This constitutes the start of the \emph{recovery}
period, in which the shield is allowed to deviate from the design.
In this phase the shield monitors the design from all states that the design could reach
under the current input and a correct output.
A second violation occurs only if the next design's output is inconsistent with all
states that are currently monitored. In case of a second violation,
the shield monitors the set of all input-enabled states that are reachable from the current set of monitored states.

The first phase of the construction of the violation monitor $\mathcal{U}$ considers
$\spec = (Q, q_{0}, \dalph, \delta,F)$  as a
\emph{safety game} and computes its winning region $W\subseteq F$ so that every reactive system
$\design\models\spec$ must produce outputs such that the next state of $\spec$ stays in $W$.  Only
in cases in which the next state of $\spec$ is outside of $W$ the shield is allowed to interfere.

The second phase expands the state space $Q$ to
$2^{Q}$ via a subset construction, with the following rationale.
If the design makes a mistake (i.e., picks outputs such that
$\spec$ enters a state $q\not \in W$), we have to ``guess'' what the design
actually meant to do and we consider all output letters that would
have avoided leaving $W$ and continue monitoring the design
from all the corresponding successor states in parallel.  Thus,
$\mathcal{U}$ is essentially a subset construction of $\spec$,
where a state $u\in U$ of $\mathcal{U}$ represents a set of states in
$\spec$.

The third phase expands the state space of $\mathcal{U}$
by adding a counter $d\in\{0,1,2\}$ and a
output variable $z$.
Initially $d$ is 0.
Whenever a property is violated%
$d$ is set to 2.
If $d>0$, the shield is in the recovery phase and can deviate.
If $d=1$ and there is no other violation, $d$ is decremented to 0.
In order to decide when to decrement $d$ from 2 to 1,
we add an output $z$ to the shield. If this output is set to
$\true$ and $d = 2$, then $d$ is set to 1.

The final violation monitor is $\mathcal{U} = (U, u_0, \dalph^u,
\delta^u)$, with the set of states $U = (2^{Q} \times \{0,1,2\})$,
the initial state $u_0 = (\{q_0\}, 0)$,
the input/output alphabet $\dalph^u = \dinalph \times \doutalph^u$
with $\doutalph^u = \doutalph \cup z$, and the next-state function $\delta^u$ , which obeys the following rules:
\begin{enumerate}
\item $\delta^u((u,d), (\dinletter, \doutletter)) =
      \bigl(\{q' \kin W \mid \exists q\in u, \doutletter' \in \doutalph^u
         \scope \delta(q,(\dinletter,\doutletter')) = q'\}, 2\bigr)$
if $\forall q \in u \scope
    \delta(q,(\dinletter, \doutletter)) \not\in W$,
and
\label{eq:subset_m}
\item $\delta^u((u,d), \dletter) \!=\!
     \bigl(\{q'\kin W \mid \exists q\kin u \scope\delta(q,\dletter) = q'\},
       \textsf{dec}(d)\bigr)$
if $\exists q \kin u \scope \delta(q,\dletter) \kin W$,
and $\textsf{dec}(0) = \textsf{dec}(1) = 0$, and if
$z$ is $\true$ then $\textsf{dec}(2) = 1$,
else $\textsf{dec}(2) = 2$.
\label{eq:subset_n}
\end{enumerate}
Our construction sets $d=2$ whenever the design leaves the winning
region, and not when it enters an unsafe state.  Hence, the shield
$\shield$ can take a remedial action as soon as ``the crime is
committed'', before the damage is detected, which would have been too
late to correct the erroneous outputs of the design.

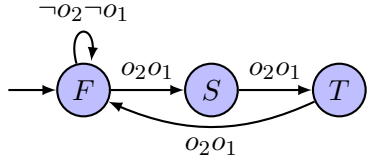
\begin{figure}[tb]
\vspace{-18pt}
\begin{minipage}{\linewidth}
  \centering
  \begin{minipage}{0.46\linewidth}
    \begin{figure}[H]
        \centering
          \centering\begin{tikzpicture}
  \node[state,draw,inner sep=1pt,minimum size=0.7cm] (s0) at (0,0) {$F$};
  \node[state,draw,inner sep=1pt,minimum size=0.7cm] (s1) at (1.7,0) {$S$};
  \node[state,draw,inner sep=1pt,minimum size=0.7cm] (s2) at (3.4,0) {$T$};
  \draw [thick, -latex](s0) -- node[above] {$o_2o_1$} (s1);
  \draw [thick, -latex](s1) -- node[above] {$o_2o_1$} (s2);
  \draw [thick, -latex](s2) to [bend left=25] node[below] {$o_2o_1$} (s0);
  \draw [thick, -latex] (-1,0) -- (s0);
  \draw [thick, -latex](s0) edge[>=latex,loop above] node[above] {$\neg o_2 \neg o_1$} (s0);
  \end{tikzpicture}
        \caption{Safety automaton of Example \ref{ex:monitor_U}.}
        \label{fig:ex_spec}
    \end{figure}
  \end{minipage}
  ~
  \begin{minipage}{0.49\linewidth}
    \begin{figure}[H]
        \centering
        \centering\begin{tikzpicture}
  \node[state,draw,inner sep=1pt,minimum size=0.7cm] (s0) at (0,0) {$t_0$};
  \node[state,draw,inner sep=1pt,minimum size=0.7cm] (s1) at (2.7,0) {$t_1$};
  \draw [thick, -latex](s0) to [bend right=15] node[below] {$\sigma_o\neq \sigma_o'$} (s1);
  \draw [thick, -latex](s1) to [bend right=15] node[above] {$\sigma_o=\sigma_o'$} (s0);
  \draw [thick, -latex] (-1,0) -- (s0);

  \draw [thick, -latex](s0) edge[>=latex,loop above] node[above] {$\sigma_o=\sigma_o'$} (s0);
  \draw [thick, -latex](s1) edge[>=latex,loop above] node[above] {$\sigma_o\neq\sigma_o'$} (s1);
  \end{tikzpicture}
        \caption{The deviation monitor $\mathcal{T}$.}
        \label{fig:dev_monitor}
    \end{figure}
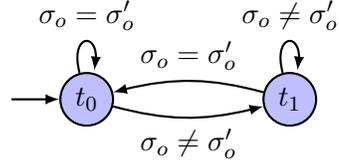
  \end{minipage}
\end{minipage}
\end{figure}

\begin{example}
\label{ex:monitor_U}
We illustrate the construction of $\mathcal{U}$ using the specification
$\spec$ from Figure~\ref{fig:ex_spec} over the outputs $o_1$ and $o_2$.
(Figure~\ref{fig:ex_spec} represents a safety automaton if we make all missing edges point to an
(additional) unsafe state.)
 The winning region consists of all safe
states, i.e., $W = \{F,S,T\}$.  The resulting violation monitor is
$\mathcal{U}=
(\{\text{F},\text{S},\text{T},\text{FS},\text{ST},\text{FT},\text{FST}\}
\times\{0,1,2\}, (\text{F},0), \dalph^u, \delta^u)$.
The transition relation $\delta^u$
is illustrated in Table~\ref{fig:ex1_table}
and lists the next states for all possible
present states and outputs.
Lightning bolts denote specification
violations.
The update of counter $d$, which is not included in
Table~\ref{fig:ex1_table}, is as follows: Whenever the design
commits a violation %
$d$ is set to $2$. If no violation exists, $d$ is decremented in the following way:
if $d=1$ or $d=0$, $d$ is set to 0. If $d=2$ and $z$ is $\true$, $d$ is set to 1, else $d$ remains 2. In this example, $z$ is set to $\true$,
whenever we are positive about the current state of the design (i.e. in $(\{F\},d)$,
$(\{S\},d)$, and $(\{T\},d)$).

Let us take a closer look at some entries of Table~\ref{fig:ex1_table}.
If the current state is $(\{F\},0)$ and we observe
the output $\neg o_2 o_1$, a specification violation occurs.
We assume that $\design$  meant to give an allowed output, either
$o_2 o_1$ or $\neg o_2 \neg o_1$. The shield continues
to monitor both $F$ and $S$; thus, $\mathcal{U}$ enters the state $(\{F,S\},2)$.
If the next observation is $o_2 o_1$, which is allowed from both
possible current states, the possible next states are $S$ and $T$, therefore
$\mathcal{U}$ traverses to state $(\{S,T\},2)$.
However, if the next observation is again $\neg o_2 o_1$,
which is neither allowed in $F$ nor in $S$, we know that a second violation occurs.
Therefore, the shield monitors the design from all three states
and $\mathcal{U}$ enters the state $(\{F,S,T\},2)$.

\begin{table}[tb]
\caption{Transition relation $\delta^u$ of monitor $\mathcal{U}$ for example \ref{ex:monitor_U}.}
\label{fig:ex1_table}
\centering
\begin{tabular}{l|ccc}
    &$\neg o_1 \neg o_2$   &$\neg o_1 o_2$ or $o_1\neg o_2$  &$o_1o_2$ \\
\hline
\{F\}   &\{F\}      &\{F,S\}$^\err$   &\{S\}   \\
\{S\}   &\{T\}$^\err$  &\{T\}$^\err$    &\{T\}   \\
\{T\}   &\{F\}$^\err$  &\{F\}$^\err$    &\{F\}   \\
\{F,S\}  &\{F\}      &\{F,S,T\}$^\err$   &\{S,T\}   \\
\{S,T\}  &\{F,T\}$^\err$ &\{F,T\}$^\err$   &\{F,T\}   \\
\{F,T\}  &\{F\}      &\{F,S,T\}$^\err$  &\{F,S\}   \\
\{F,S,T\} &\{F\}      &\{F,S,T\}$^\err$  &\{F,S,T\}   \\
\hline
\end{tabular}
\end{table}

\end{example}

\subsubsection{Step 2. Constructing the Deviation Monitor $\mathcal{T}$}
\label{sec:deviation_monitor}
We build $\mathcal{T} = (T, t_0, \doutalph \times \doutalph, \delta^t)$
to monitor deviations between the shield and design outputs.
Here, $T = \{t_0, t_1\}$ and $\delta^t(t, (\doutletter,
\doutletter')) = t_0$ if and only if $\doutletter = \doutletter'$. That is, if there is a deviation in the current time step, then $\mathcal{T}$ will be in $t_1$ in the next time step. Otherwise, it will be in $t_0$.  This deviation monitor is shown in
Figure~\ref{fig:dev_monitor}.

\subsubsection{Step 3. Constructing and Solving the Safety Game $\mathcal{G}^s$}
Given the automata $\mathcal{U}$ and $\mathcal{T}$ and the
safety automaton $\spec$, we construct a safety
game $\mathcal{G}^s = (G^s, g_0^s, \dinalph^s, \doutalph^s$
$\delta^s, F^s)$, which is the synchronous product of $\mathcal{U}$,
$\mathcal{T}$, and $\spec$, such that $G^s= U \times
T \times Q$ is the state space, $g_0^s = (u_0, t_0, q_0)$
is the initial state, $\dinalph^s=\dinalph\times\doutalph$ is the input of the
shield, $\doutalph^s=\doutalph\cup \{z\}$ is the output of the shield,
$\delta^s$ is the next-state function, and $F^s$ is the set of safe states such that
$\delta^s\bigl((u, t, q), (\dinletter, \doutletter),
(\doutletter',z)\bigr) = $
\[
\bigl( \delta^u[u,(\dinletter, (\doutletter,z))],
       \delta^t[t,(\doutletter, \doutletter')],
       \delta[q, (\dinletter, \doutletter')]
\bigr),
\]

\noindent and $F^s = \{(u, t, q)\in G^s \mid
             q \in F \wedge u=(w,0) \rightarrow t=t_0\}$.

We require $q \in F$, which ensures that the
output of the shield satisfies $\spec$, and that the shield can
only deviate in the recovery period (i.e., if $d=0$, no deviation is allowed).
We use standard algorithms for safety games (cf.
\citet{Faella09}) to compute the winning region $W^s$ and the most permissive non-deterministic winning strategy $\rho_s: \gstates \times \dinalph \rightarrow 2^{\doutalph}$ that is not only winning for the system, but also contains all deterministic winning strategies.

\subsubsection{Step 4. Constructing the \buchi Game $\mathcal{G}^b$}
Implementing the safety game ensures correctness ($\design \comp \shield \models \spec$)
and that the shield $\shield$ keeps the output of the design $\design$ intact,
if $\design$ does not violate $\spec$.
The shield still has to keep
the number of deviations per violation to a minimum.
Therefore, we would like the recovery period to be over infinitely often.
This can be formalized as a \buchi winning condition.
We construct the \buchi game $\mathcal{G}^b$
by applying the non-deterministic safety strategy $\rho^s$
to the game graph $\mathcal{G}^s$.

Given the safety game $\mathcal{G}^s=(G^s, g_0^s, \dinalph^s, \doutalph^s, \delta^s, F^s)$ with the
non-deterministic winning strategy $\rho^s$ and the winning region $W^s$, we construct a \buchi game
$\mathcal{G}^b=(G^b, g_0^b, \dinalph^s, \doutalph^s, \delta^b, F^b)$
such that $G^b=W^s$ is the state space, the initial state $g_0^b=g_0^s$ and
the input/output alphabet $\dinalph^b=\dinalph^s$ and $\doutalph^b=\doutalph^s$
remain unchanged, $\delta^b=\delta^s\cap\rho^s$ is the transition
function, and $F^b = \{(u, t, q)\in W^s \mid
             (u=(w,0) \vee u=(w,1))\}$ is the set of accepting states.
A play is winning if $d\leq1$ infinitely often.

\noindent
\subsubsection{Step 5. Solving the \buchi Game $\mathcal{G}^b$}
\label{sec:solving_buchi}
Most likely, the \buchi game $\mathcal{G}^b$
contains reachable states, for which $d\leq1$ cannot be enforced infinitely often.
We implement an admissible strategy
that enforces to visit $d\leq1$ infinitely often whenever possible.
This criterion essentially asks for a strategy
that is winning with the help of the design.

The admissible strategy $\rho^b$ for a \buchi game $\mathcal{G}^b=(G^b, g_0^b, \dinalph^b, \doutalph^b, \delta^b, F^b)$ can be computed as follows~\citet{Faella09}:
\begin{enumerate}
  \item Compute the winning region $W^b$ and a winning strategy $\rho_w^b$ for $\mathcal{G}^b$ (cf.~\citet{Mazala01}).

  \item Remove all transitions that start in $W^b$ and do not belong to $\rho_w^b$ from  $\mathcal{G}^b$.
      This results in a new \buchi game
      $\mathcal{G}_1^b=(G^b, g_0^b, \dinalph^b, \doutalph^b, \delta_1^b, F^b)$
      with \[(g,(\dinletter,\doutletter),g')\in\delta_1^b\]
      \[\text{if } (g,\dinletter, \doutletter)\in\rho_w^b\] 
      \[\text{or if }\forall \doutletter' \in \doutalph^b \scope (g,\dinletter, \doutletter')\notin\rho_w^b \wedge (g,(\dinletter,\doutletter),g')\in\delta^b.\]

  \item In the resulting game $\mathcal{G}_1^b$, compute a cooperatively winning strategy $\rho^b$.
        In order to compute $\rho^b$, one first has to transform all input variables to output variables. This results in the \buchi game
        $\mathcal{G}_2^b=(G^b, g_0^b, \emptyset, \dinalph^b\times\doutalph^b, \delta_1^b, F^b)$. Afterwards, $\rho^b$ can be computed with the standard algorithm for the winning strategy
        on $\mathcal{G}_2^b$.
\end{enumerate}

The strategy $\rho^b$ is an admissible strategy of the game $\mathcal{G}^b$,
since it is winning and cooperatively winning~\citep{Faella09}.
Whenever the game $\mathcal{G}^b$ starts in a state of the winning region $W^b$,
any play created by $\rho_w^b$ is winning.
Since $\rho^b$ coincides with $\rho_w^b$ in all states of the winning region $W^b$,
$\rho^b$ is winning.
We know that $\rho^b$ is cooperatively winning in the game $\mathcal{G}_1^b$.
A proof that $\rho^b$ is also cooperatively winning in the original game $\mathcal{G}^b$
can be found in \citet{Faella09}.

\begin{theorem}
A shield that implements the admissible strategy $\rho^b$
in the \buchi game $\mathcal{G}^b=(G^b, g_0^b, \dinalph^b, \doutalph^b, \delta^b, F^b)$ in a new reactive system
$\shield = (G^b, g^b_0, \dinalph^b, \doutalph^b,
\delta', \rho^b)$ with $\delta'(g,\dinletter) = \empty
\delta^b(g,\dinletter,\rho^b(g,\dinletter))$ is an admissible shield.
\end{theorem}
\begin{proof}
  First, the admissible strategy $\rho^b$ is winning for all winning states of the \buchi game $\mathcal{G}^b$. Since winning strategies for \buchi games are subgame optimal,
  a shield that implements $\rho^b$ ends deviations after the smallest number of steps possible,
  for all states of the design in which a finite number of deviations can be guaranteed.
 Second, $\rho^b$ is cooperatively winning in the \buchi game $\mathcal{G}^b$.
  Therefore, in all states in which a finite number of deviation cannot be guaranteed, a shield that implements the strategy $\rho^b$ recovers with the help of the design as soon as possible.
\end{proof}

The standard algorithm for solving \buchi games
contains the computation of attractors; the $i$-th attractor for the system contains all states  from which the system can ``force'' a visit of an accepting
state in $i$ steps. For all states $g\in G^b$ of the game $\mathcal{G}^b$, the attractor
number of $g$ corresponds to the smallest number of steps within
which the recovery phase can be guaranteed to end, or can end with the help of the design
if a finite number of deviation cannot be guaranteed.

\begin{theorem}
  Let $\spec=\{Q, q_{0}, \dalph, \delta, F\}$ be a safety specification and
  $|Q|$ be the cardinality of the state space of $\spec$.
  An admissible shield with respect to $\spec$ can be synthesized in $\mathcal{O}(2^{|Q|})$ time, if it exists.
\end{theorem}
\begin{proof}
  The safety game $\mathcal{G}^s$  and \buchi game $\mathcal{G}^b$ have at most
  $m=(2 \cdot 2^{|Q|}+|Q|)\cdot 2  \cdot |Q|$ states and at most $n=m^2$ edges.
  
  Safety games can be solved in $\mathcal{O}(m+n)$ time
  and \buchi games in $\mathcal{O}(m\cdot n)$ time~\citep{Mazala01}.
\end{proof}

\chapter{Verifiable Learning-Based Synthesis}
\glsresetall
\label{chapter:learning}

This section provides an overview of incorporating learning techniques in policy synthesis and discusses an approach that merges concepts from formal methods and machine learning. We consider the challenging task of computing a policy for a \gls{pomdp} that satisfies certain classes of specifications, first introduced in Section~\ref{sec:information_limitations}. The approach discussed in this section represents a policy using \glspl{rnn}. We then show how such a representation can be integrated with formal methods by extracting a policy that is compatible with state-of-the-art verification tools.

\blfootnote{This section incorporates the results from the following publications~\citep*{carr2019counterexample,CarrJT20,Carr0T21}.}

\Glspl{rnn} offer an effective policy representation for \glspl{pomdp} due to their ability to effectively process sequential data.
As opposed to the conventional feed-forward architectures present in artificial neural networks, in which the nodes in each layer are only connected to nodes in subsequent layers, recurrent architectures allow for backward connections between nodes.
Such backward connections allow \glspl{rnn} to create internal memory states, such as those in \gls{lstm} architectures~\citep{hochreiter1997long}, which infer temporal behavior from sequences of data~\citep{DBLP:journals/corr/PascanuGCB13}.
Reinforcement learning research has shown that \glspl{rnn} used in environments modeled by \glspl{pomdp} perform well as black-box functions for either state or value estimators \citep{wierstra2007solving,bakker2002reinforcement} or as control policies \citep{hausknecht2015deep,heess2015learning}.

In \glspl{pomdp} that model agents in safety-critical environments, policies that are guaranteed to prevent unsafe behavior are necessary.
The agent's behavior may have to obey more complicated specifications than maximizing an expected reward, such as reachability, liveness or, more generally, specifications expressed in temporal logic, \eg~\gls{ltl}, see Section~\ref{sec:logic}. 

Verifying whether an agent following an \gls{rnn}-based policy in a \gls{pomdp} satisfies temporal logic specifications is, in general, hard.
\gls{rnn}s are complex structures that capture non-linear input-output relations~\citep{DBLP:journals/csl/MulderBM15}.
To formally analyze how \gls{rnn}s interpret sequences of data, \citet{sherstinsky2020fundamentals} suggest fixing a defined sequence length for analysis and performing an \emph{unrolling} procedure, which converts the \gls{rnn} to a feed-forward neural network with the same number of layers as that defined length~\citep{goodfellow2016deep}.
Checking whether the agent's behavior satisfies the specification for the set of all possible sequences of data with a given length in the \gls{pomdp} is intractable~\citep{meuleau1999solving}.

The approach discussed in this section combines the representation power of \glspl{rnn} from machine learning with the provable guarantees that are at the heart of formal verification.
The latter can efficiently verify whether an agent following a given policy, typically in the form of an \gls{fsc}~\citep{poupart2004bounded,junges2018finite}, adheres to a temporal logic specification~\citep{BK08}.
However, directly synthesizing a policy requires---in general---memory of exponential size in the number of \gls{pomdp} states \citep{DBLP:journals/jacm/BaierGB12}.
Machine learning, on the other hand, provides an efficient approach, in the form of training \gls{rnn}-based policy representations from sequences of data, to find candidate policies that might ensure an agent in a \gls{pomdp} satisfies a temporal logic specification~\citep{DBLP:journals/corr/HeessHLS15}.

\begin{figure}
	\centering
	\hspace{-1.80cm}\scalebox{0.75}{\input{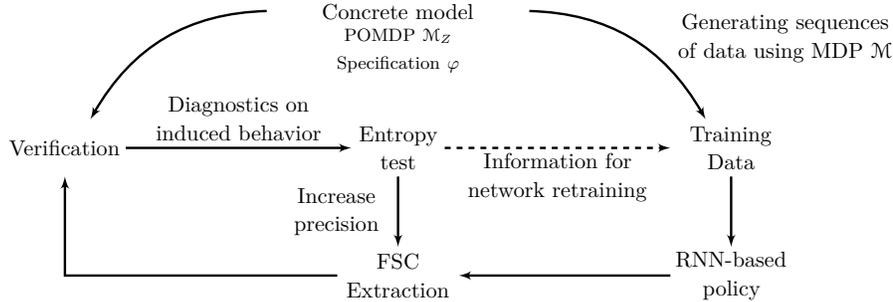}}
	\caption{High-level iterative policy improvement process.}
	\label{fig:high-level}
\end{figure}

There remains a central gap: How to close the loop between training an \gls{rnn}-based policy and efficiently verifying for a candidate policy? 
The approach closes this gap by tightly integrating formal verification and machine learning towards three key steps: (1) extracting an FSC from an \gls{rnn}-based policy, (2) verifying this candidate FSC for the \gls{pomdp} against a temporal logic specification, and (3) if needed, either refining the FSC or generating more training data for the \gls{rnn}, see Figure~\ref{fig:high-level}.

\section{Verifiable Recurrent Neural Network-Based Policies for Partially Observable Markov Decision Processes}

\label{sec:verifiable}

In this section, we formulate a problem that is similar to that introduced in Problem~\ref{prob:RobustSynth}. We restate it here without the robustness considerations:

\begin{problem}[\Gls{pomdp} synthesis]
 For a \gls{pomdp} $\pomdp$ and a specification $\reachPropSymbol$, where either $\reachPropSymbol=\p_{\sim \lambda}(\ltlformula)$ for ${\sim}\in\{ {<}, {\leq},{\geq},{>}\}$ and $\lambda\in[0,1]$ with $\ltlformula$ an LTL specification, or $\reachPropSymbol=\Ex_{\sim\lambda}(\Finally a)$,
	the problem is to determine a (finite-memory) policy $\osched\in\oSched^\pomdp$ such that $\pomdp^{\osched}\models\reachPropSymbol$.
	\label{prob:POMDPSynth}
\end{problem}

In Problem~\ref{prob:POMDPSynth}, if no (finite-memory) policy exists, the problem is infeasible.
Note that, in general, Problem~\ref{prob:POMDPSynth} is undecidable~\citep{MadaniHC99} and each method is necessarily incomplete.

Figure~\ref{fig:high-level} outlines the learning-based overall approach to Problem~\ref{prob:POMDPSynth}. 
A trained \gls{rnn} serves as an efficient representation of a POMDP policy. 
As safety-critical scenarios necessitate a \emph{sound} notion of correctness, the approach evaluates such an RNN-based policy using formal verification against LTL specifications. %
There are four main building blocks towards that approach: (1) \emph{Training} the \gls{rnn}, (2) \emph{extracting} \glspl{fsc} as a tractable representation of the \gls{rnn}-based policy, (3) \emph{evaluating} the policy, and (4) \emph{improving} the policy.

\subsection{Training a Recurrent Neural Network-Based Policy}
\label{ssec:DRNN}
We first define how \glspl{rnn} can be used as a representation of POMDP policies.
\begin{definition}[\gls{rnn}-based policy] \label{def:policy_network}
An \gls{rnn}-based policy for a \gls{pomdp} is a function $\RNNfun\colon \obsSeqFin^{\pomdp} \mapsto \Distr(\Act)$, where $\obsSeqFin^{\pomdp}$ is the set of all sequential observation-action inputs and $\Distr(\Act)$ is the set of all distributions over actions. 
To be more precise, we identify the main components of such a network.
	An \gls{rnn}-based policy $\RNNfun$ is sufficiently described by a \emph{hidden-state update function} 
	$\hat{\delta} \colon \Reals \times \ObsSym \times \Act \rightarrow \Reals$ and an \emph{action-mapping} $\osched_h \colon \Reals \rightarrow \Distr(\Act)$.
\end{definition}
Consider the following observation-action sequence:
\begin{align}
\ObsFun(\path)=\ObsFun(s_0)\xrightarrow{\act_0} \ObsFun(s_1) \xrightarrow{\act_1}\cdots \ObsFun(s_i)
\label{eq:obs_seq}
\end{align}

The \gls{rnn}-based policy receives an observation-action sequence and returns a distribution over the action choices.
Throughout the execution of the sequence, the \gls{rnn} holds a continuous hidden state $h\in \Reals$, typically described as an internal memory state, which captures previous information.
On each transition, this hidden state is updated to include the information of the current state and the last action taken under the hidden-state update function $\hat{\delta}$.
From the prior observation sequence in (\ref{eq:obs_seq}), the corresponding hidden state sequence would be defined as:
\begin{align*}
\hat{\delta}(\path) = h_0 \xrightarrow{\act_0,~\ObsFun(s_1)} h_1  \xrightarrow{\act_1,~\ObsFun(s_2)} \cdots h_i
\end{align*}
Additionally, the output of the \gls{rnn}-based policy is expressed by the action-distribution function $\osched_h$, which maps the value of hidden state to an action mapping~$\osched_h$.~
At an internal memory state $h_i$, we have $\hat{\delta}(h_i,\ObsFun(s_i),a_i) = h_{i+1}$ and $\osched_h(h_{i+1})=\mu(\Act)$ for state $s_i$ on path $\path$.

\begin{figure}[t!]
	\centering
    \scalebox{1.00}{%
			\input{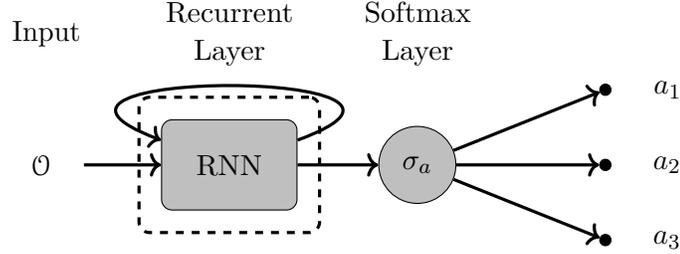}}
	\caption{\gls{rnn}-based Policy $\hat{\osched}$ with softmax layer activation $\sigma_a$. \label{fig:RNN_policy}}
\end{figure}

The approach constructs an \gls{rnn}-based policy using a three-layer network, shown in Figure~\ref{fig:RNN_policy}. The policy network uses an \gls{lstm} architecture~\citep{hochreiter1997long} for the recurrent layer $\hat{\delta}$ and then a softmax layer for the output action mapping $\osched_h$ (see Definition~\ref{def:policy_network}).
To fit the \gls{rnn} model to the sequences of training data, we use the Adam optimizer~\cite{kingma2014adam} with a categorical cross-entropy error function~\cite{goodfellow2016deep}.

\begin{figure}[t!]
	\centering
	\scalebox{1.00}{%
			\input{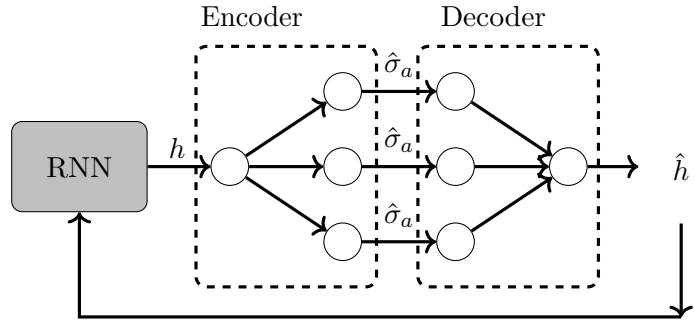}}
	\caption{\gls{rnn}-based policy structure with a QBN. \gls{rnn} block and associated QBN of $B_h=3$ with quantized activation $\hat{\sigma}_a \colon \Reals\rightarrow \lbrace-1,0,1\rbrace$.	\label{fig:QBN}}
	
\end{figure}

In practice, a machine learning-based approach requires a method of generating data. One such method involves first computing a policy $\policy\in \Policy^\mdp$ of the underlying \gls{mdp} $\mdp$ that satisfies $\varphi$ using the \tool{STORM} probabilistic model checker~\citep{DBLP:conf/cav/DehnertJK017}.~
Then it samples an initial state uniformly over the initial belief support $s_0\in\supp(b)$ and generate finite observation paths $\obspath$, thereby creating multiple trajectory trees \citep{kearns2000approximate}.
When generating sequences of data, selecting one of the trees and following it to a leaf, which forms either at a pre-defined maximum sequence length or a deadlock, gives a finite path $\path \in \pathsfin^{\mdp} $.
From this path $\path$, the method generates one possible observation-action sequence $\obspath \in \obsSeqFin^{\pomdp}$.

\begin{example}
	Consider the \gls{pomdp} in Example~\ref{ex:motivating} and Figure~\ref{fig:motivating}: a sample set of sequences of data would be:\\ $\mathcal{D} = \lbrace \path_o^0 = (\mathit{blue}, \textrm{up}, \mathit{blue}, \textrm{down},s_3),~\path_o^1 = (\mathit{blue},\textrm{down},s_3),~\path_o^2 = (\mathit{blue},\textrm{up},s_3)\rbrace$. An example \gls{rnn} policy $\osched$ trained on these sequences would yield a policy for observation-action sequence $\path_{o,0}=(\textrm{blue})$ as $\osched(\path_{o,0}) = \lbrace 0.67:\textrm{up}, 0.33:\textrm{down}\rbrace$, which has a categorical cross-entropy loss of approximately 0.585. Similarly, the same \gls{rnn} policy for a longer observation-action sequence such as $\path_{o,1} = (\textrm{blue},\textit{up},\textrm{blue})$ yields a policy $\osched(\path_{o,1}) = \lbrace 1.0:\textrm{down}\rbrace$, for a cross-entropy loss of 0.
\end{example}

\subsection{Finite-State Policy Extraction}
\label{ssec:extraction}

The discussed approach adapts a method called quantized bottleneck insertion~\citep{koul2018learning} to extract an \gls{fsc} from a given  \gls{rnn}-based policy.
Let us first explain the relationship between the main components of an  \gls{rnn}-based policy $\RNNfun$ (Definition~\ref{def:policy_network}) and an \gls{fsc} $\fsc$ (Definition~\ref{def:fsc}).
In particular, the hidden-state update function $\hat{\delta}$ takes as input a real-valued hidden state of the policy network, while the \gls{fsc}'s memory update function $\delta$ takes a memory node from the finite set $N$.
Figure~\ref{fig:RNN_policy} describes a simplified architecture for the former since its recurrent component acts as the hidden-state update function $\hat{\delta}$.
The key for linking the two is therefore a mechanism that encodes the continuous hidden state $h$ into a set $N$ of discrete memory nodes.
We outline such a mechanism in the sequel and in Figure~\ref{fig:QBN} in which we show the modified activation function (formed using an encoder and a decoder).

The approach leverages an autoencoder~\citep{goodfellow2016deep} in the form of a \emph{quantized bottleneck network} (QBN)~\citep{koul2018learning}.
This QBN, consisting of an encoder and a decoder, is inserted into the \gls{rnn}-based policy directly before the softmax layer, see Figure~\ref{fig:QBN}. 
In the encoder, the continuous hidden-state value $h\in\Reals$ is mapped to an intermediate real-valued vector $\Reals^{B_h}$ of pre-allocated size $B_h$.
The decoder then maps this intermediate vector into a discrete vector space defined by $\lbrace -1,0,1\rbrace^{B_h}$.
This process, illustrated in Figure~\ref{fig:QBN}, provides a mapping of the continuous hidden state $h$ into $3^{B_h}$ possible discrete values.
We denote the discrete state for $h$ by $\hat{h}$ and the set of all such discrete states by $\hat{H}$. 
Note, that $|\hat{H}|\leq 3^{B_h}$ since not all values of the hidden state may be reached in an observation sequence.

After the QBN insertion, we simulate a series of executions querying the modified \gls{rnn} for action choices on the \gls{pomdp}.
These executions form a dataset of consecutive pairs ($\hat{h}_{t}, \hat{h}_{t+1})$ of discrete hidden states, the action $\act_{t}$ and the observation $\obs_{t+1}$ that led to the transition $\lbrace \hat{h}_{t},\act_{t},\obs_{t+1},\hat{h}_{t+1}\rbrace$ at each time $t$ during the execution of the \gls{rnn}-based policy.
The number of accessed memory nodes $N\subseteq\hat{H}$ corresponds to the number of different discrete states $\hat{h} \in \hat{H}$ in this dataset.  
The deterministic memory update rule $\delta(n_{t},\act_{t},\obs_{t+1}) = n_{t+1}$ is obtained by constructing a $N\times (|\ObsSym|\times |\Act|)$ transaction table~\citep{koul2018learning}.
We can additionally construct the action-mapping $\alpha\colon N \times \ObsSym \rightarrow \Distr(\Act)$ with $\alpha(n_t,\obs_t) = \mu\in\Distr(\Act)$ by querying the softmax-output layer (see Figure~\ref{fig:RNN_policy}) for each memory node and observation.

\subsection{Evaluating the Extracted Policy}
\label{ssec:policy-evaluation}

We assume that for \gls{pomdp} $\PomdpInit$ and specification $\varphi$, we have an extracted \gls{fsc} $\fsc_{\hat{\osched}} \in  \oSched^{\pomdp}$ as in Definition~\ref{def:fsc}.
The approach applies the policy $\fsc_{\hat{\osched}}$ to obtain an induced \gls{dtmc} $\pomdp^{\fsc_{\hat{\osched}}}$. %
For this \gls{dtmc}, formal verification through model checking checks whether $\pomdp^{\fsc_{\hat{\osched}}}\models\varphi$ and thereby provides hard guarantees about the quality of the extracted \gls{fsc} $\fsc_{\hat{\osched}}$ regarding $\varphi$.
In particular, (probabilistic) model checking provides the probability -- or the expected reward -- to satisfy a specification for \emph{all states} $s\in S$ via solving linear equation systems, see Section~\ref{chap:probabilistic-synthesis} and \citep{BK08}.

\begin{example}
	Consider the case in the 1-FSC $\fsc_1$ from Example~\ref{ex:motivating} (Figure~\ref{fig:1FSC}) where the parameter $p=1$ and the probability of reaching the state $s_3$ in the induced \gls{dtmc} is $\Pr(\Ever s_3) = \frac{1}{3}$.
	The behavior induced by this 1-FSC violates the specification and we obtain two counterexamples of critical memory-state pairs for this policy $\fsc_{\hat{\osched}}\colon$~$(0,s_0)$ and $(0,s_1)$.
\end{example}

If the specification does not hold, the policy may require refinement.
On the one hand we can increase the number of memory nodes $B_h$ to extract a new \gls{fsc}. 
On the other hand, we may decide via a formal entropy check whether new data need to be generated.

\subsection{Improving the Policy with Counterexample Data}
\label{ssec:drnn_improve}

The approach attempts to determine whether an \gls{rnn}-based policy requires more training data $\mathcal{D}$ or not.
Existing approaches in supervised learning methods leverage benchmark comparisons between a train-test set using a loss function~\citep{baum1988supervised}.
Loss visualization, proposed by~\citet{goodfellow2014qualitatively} provides a set of analytical tools to show model convergence.
However, such approaches aim at continuous functions instead of the discrete representations as in the \gls{fsc}.
More importantly, the method leverages the information gained from a model-based approach. 

We first determine a set of states that are critical for the satisfaction of the specification under the current policy.
Consider a sequence of memory nodes and observations  $(n_0,\obs_0)\xrightarrow{a_0}\cdots \xrightarrow{a_{t-1}}(n_t,\obs_t)$ from the \gls{pomdp} $\pomdp$ under the \gls{fsc} $\fsc_{\hat{\osched}}$.
For each of these sequences, we collect the states $s\in \states$ underlying the observations, e.g., $\ObsFun(s)=\obs_i$ for $0\leq i\leq t$. 
As we know the probability or expected reward for these states to satisfy the specification from previous model checking, we can now directly assess their criticality regarding the specification.
We collect all pairs of memory nodes and states from $N\times S$ that contain critical states and build the set $\mathit{Crit^{\pomdp}_{\fsc_{\hat{\osched}}}}\subseteq N\times S$ that serves us as a counterexample. 
These pairs carry the joint information of critical states and memory nodes from the policy applied to the \gls{dtmc} $\pomdp^{\fsc_{\hat{\osched}}}$.

\paragraph{Entropy measure.}
The average entropy across the distributions over actions at the choices induced by the counterexample set $\mathit{Crit^{\pomdp}_{\fsc_{\hat{\osched}}}}$ is our measure of choice to determine the level of training for the \gls{rnn}-based policy. 
Specifically, for each pair $(n,s)\in\mathit{Crit^{\pomdp}_{\fsc_{\hat{\osched}}}}$, we collect the distribution $\distFunc\in\Distr(\Act)$ over actions that $\fsc_{\hat{\osched}}$ returns for the observation $\ObsFun(s)$ when it is in memory node $n$.
Then, we define the \emph{evaluation function} $H$ using the entropy $\mathcal{H}(\distFunc)$ of the distribution $\distFunc$:
\begin{align*}
H\colon Crit^{\pomdp}_{\fsc_{\hat{\osched}}}\rightarrow [0,1] \text{ with } H(n,s)=\mathcal{H}(\distFunc).
\end{align*}

For high values of $H$, the distribution is uniform across all actions and the associated \gls{rnn}-based policy is likely extrapolating from unseen inputs.

We observe that when there are fewer samples and higher memory nodes, the extracted \gls{fsc} tends to perform arbitrarily, see Section~\ref{ssec:policy_improve} and Figure~\ref{fig:entropy_plot} for a detailed empirical analysis.
We lift the function $H$ to the full set $\mathit{Crit^{\pomdp}_{\fsc_{\hat{\osched}}}}$:
\begin{align}
H(\mathit{Crit^{\pomdp}_{\fsc_{\hat{\osched}}}}) &= \frac{1}{|Crit^{\pomdp}_{\fsc_{\hat{\osched}}}|} \sum_{(n,s) \in Crit^{\pomdp}_{\fsc_{\hat{\osched}}}} H(n,s). \label{eq:ent}
\end{align}

We compare the average entropy over all decision-points of the counterexample against a constant threshold $\eta \in [0,1]$, that is, if $H(\mathit{Crit^{\pomdp}_{\fsc_{\hat{\osched}}}}) > \eta$, we will provide more training data.
Vice versa, if $H(\mathit{Crit^{\pomdp}_{\fsc_{\hat{\osched}}}}) \leq \eta$, we increase the upper bound on the number of memory nodes in the \gls{fsc}.

\begin{example}
Under the working Example~\ref{ex:motivating}, the policy $\fsc_1$ was the 1-\gls{fsc} with $p=1$ (Figure~\ref{fig:1FSC}), which produces two counterexample memory and state pairs: $Crit^{\pomdp}_{\fsc_1}=\{(0,s_0),(0,s_1)\}$.
The procedure would then examine the policy's average entropy at these critical components $(n,s) \in Crit^{\pomdp}_{\fsc_1}$, which in this trivial example is given by $H(Crit^{\pomdp}_{\fsc_1})=-p \log_2(p) - (1-p) \log_2(1-p)=0$ from (\ref{eq:ent}).
The average entropy is below a prescribed threshold, $\eta=0.5$, and thus we increase the number of memory nodes, which results in the satisfying FSC $\fsc_2$ in Figure~\ref{fig:2FSC}.
\end{example}

\subsection{Performance on Partially Observable Markov Decision Process Benchmarks}

\label{sec:experiments}

We evaluate the RNN-based synthesis on benchmark examples that are subject to either LTL specifications or expected reward specifications.
For the former, we compare to the tool \tool{PRISM-POMDP}~\citep{NPZ17}, and for the latter to \tool{PRISM-POMDP} and the point-based solver \tool{SolvePOMDP}~\citep{walraven2017accelerated}.

For a fair comparison, instead of terminating the synthesis procedure once the policy satisfies the specification, we always iterate 10 times, where one iteration encompasses the (re-)training of the RNN-based policy using counterexamples, the FSC extraction, the evaluations, and the policy improvement.
For instance, for a specification $\reachPropSymbol=\p_{\geq \lambda}(\ltlformula)$, we leave the $\lambda$ open and seek to compute $\p_{\max}(\ltlformula)$, that is, we compute the minimal probability of satisfying $\ltlformula$ to obtain a policy that satisfies $\reachPropSymbol$.
We cannot guarantee to reach that optimum, but we rather improve as far as possible within the predefined 10 iterations.

We created the following \tool{Python} toolchain to realize the full RNN-based procedure, combining the state-of-the-art tools from deep learning with those from formal verification.
First, we use the deep learning library \tool{Keras}~\citep{ketkar2017introduction} to train the RNN-based policy from sequences of data.
To evaluate policies, we employ the probabilistic model checkers \tool{PRISM}~\citep{NPZ17} and \tool{STORM}~\citep{DBLP:conf/cav/DehnertJK017} for LTL and undiscounted expected reward respectively.
We evaluated on a 2.3\,GHz machine with a 12\,GB memory limit and a specified maximum computation time of $10^5$~seconds. In Table~\ref{tab:LTL_Prop} TO/MO denote violations of the time/memory limit, respectively and Res. refers to the output value of the induced \gls{dtmc}.

\paragraph{Problem settings with temporal logic specifications.} 
We examined three problem settings involving motion planning with LTL specifications.
For each of the settings, we use a square gridworld of length $c$ with 4~action choices (cardinal directions of movement).
The motivation for gridworld examples is that they provide a minimal safety check: a policy that fails to behave safely in such simple environments is also unlikely to behave safely in the real world~\citep{DBLP:journals/corr/abs-1711-09883}.
Inside this environment, there are a set of static $(\hat{x})$ and moving $(\tilde{x})$ obstacles as well as possible target cells $G_1$ and $G_2$.

\begin{figure}[t!]
    \centering
    \subfloat[Gridworld for LTL examples  \label{fig:LTL_grid}]{
    \scalebox{0.85}{
    \input{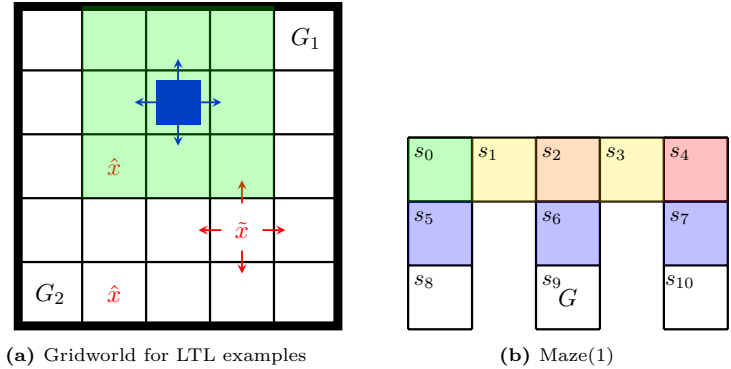}}} \quad
    \subfloat[Maze(1)\label{fig:maze}]{
	\centering
	\scalebox{0.85}{
\begin{tikzpicture}
\draw[black,line width=1pt] (0,0) grid[step=1] (1,3);
\draw[black,line width=1pt] (2,0) grid[step=1] (3,3);
\draw[black,line width=1pt] (4,0) grid[step=1] (5,3);
\draw[black,line width=1pt] (0,2) grid[step=1] (5,3);
\fill[draw=none,fill=green,opacity=0.25] (0,2) rectangle (1,3);
\fill[draw=none,fill=blue,opacity=0.25] (0,1) rectangle (1,2);
\fill[draw=none,fill=blue,opacity=0.25] (2,1) rectangle (3,2);
\fill[draw=none,fill=blue,opacity=0.25] (4,1) rectangle (5,2);
\fill[draw=none,fill=yellow,opacity=0.25] (1,2) rectangle (2,3);
\fill[draw=none,fill=yellow,opacity=0.25] (3,2) rectangle (4,3);
\fill[draw=none,fill=orange,opacity=0.25] (2,2) rectangle (3,3);
\fill[draw=none,fill=red,opacity=0.25] (4,2) rectangle (5,3);
\fillGridAt{2}{0}{$G$}
\fillGridAt{-0.25}{2+0.25}{\small$s_0$}
\fillGridAt{1-0.25}{2+0.25}{\small$s_1$}
\fillGridAt{2-0.25}{2+0.25}{\small$s_2$}
\fillGridAt{3-0.25}{2+0.25}{\small$s_3$}
\fillGridAt{4-0.25}{2+0.25}{\small$s_4$}
\fillGridAt{-0.25}{1+0.25}{\small$s_5$}
\fillGridAt{2-0.25}{1+0.25}{\small$s_6$}
\fillGridAt{4-0.25}{1+0.25}{\small$s_7$}
\fillGridAt{-0.25}{0.25}{\small$s_8$}
\fillGridAt{2-0.25}{0.25}{\small$s_9$}
\fillGridAt{4-0.25}{0.25}{\small$s_{10}$}
\end{tikzpicture}}}
    \caption{Physical environments for presented examples.}
\end{figure}

The agent has a limited visibility region, indicated by the green area, and can infer its state from observations and knowledge of the environment.
We define observations as Boolean functions that take as input the positions of the agent and moving obstacles, see Figure~\ref{fig:LTL_grid}.
Intuitively, the functions describe the 8 possible relative positions of the obstacles with respect to the agent inside its viewing range. The three problem settings are as follows:

\begin{enumerate}
	\item \textbf{Navigation with moving obstacles}---An agent and a single stochastically moving obstacle. 
	The agent task is to maximize the probability to navigate to a goal state $A$ while not colliding with obstacles (both static and moving): 
	$\varphi_1= \p_{\max}\left( \neg X\,\Until\,G_1\right)$ with $X = \hat{x}\cup\tilde{x}$,
	\item \textbf{Delivery without obstacles}---An agent and static objects (landmarks). The task is to deliver an object from $G_1$ to $G_2$ in as few steps as possible: $\varphi_2= \Ex_{\max}(\finally(G_1 \land \Ever G_2))$.
	\item \textbf{Slippery delivery with static obstacles}---An agent where the probability of moving perpendicular to the desired direction is $0.1$ in each direction. It attempts to maximize the probability to travel from location $G_1$ and to $G_2$ without colliding with the static obstacles $\hat{x}$: $ \varphi_3= \p_{\max} \left( \Always \Ever G_1 \land \Always \Ever G_2 \land \neg \Ever X \right)$, with $X = \hat{x}$.
\end{enumerate}

\paragraph{Problems settings for maximizing expected reward.}
For comparison to existing benchmarks, we extend a well-known \gls{pomdp} example \emph{Maze}($\gridsizeparam$) for an arbitrary-sized structure. These problems are quite different to the LTL examples, in particular the significantly smaller observation spaces, see \citep{Carr0T21} for details.

\begin{enumerate}
\item\emph{\textbf{Grid-based Maze($\gridsizeparam$) with $\gridsizeparam+2$ rows}}---The agent can only detect its neighboring walls and attempts to reach a goal state $G$ in as few steps as possible, see Figure~\ref{fig:maze} for Maze(1). Extra rows add uncertainty over the agent's position in the corridors, see the blue observations in Figure~\ref{fig:maze}.
\item \emph{\textbf{Grid($\gridsizeparam$) with restricted vision}}---A square grid with length $\gridsizeparam$ where the agent attempts to reach a goal state $G$ at the top right of the square. The agent is placed in the grid according to a uniform distribution of the states and can only observe its exact location when it reaches the goal state.
\end{enumerate}

\input{./figures/09-learn/FSC_memory}

\label{ssec:policy_improve}

\begin{table}[t!]
	\scriptsize\setlength{\tabcolsep}{2pt}
	\centering
	\scalebox{1.05}{\begin{tabular}{@{}lrc|rr|rr@{}}
		\toprule
		&&&   \multicolumn{2}{c}{PRISM-POMDP} &  \multicolumn{2}{p{32.5mm}}{RNN-based Policy} \\
		Problem & States     & Type, $\varphi$    &  Res. & Time (s)  &  Res. & Time (s)  \\
		\midrule
		Navigation (3)& 333 &$\p_{\max}^{\pomdp}$, $\varphi_1$  & \textbf{0.84} & \textbf{73.88} & 0.80  & 123.14 \\
        Navigation (4) & 1088 &$\p_{\max}^{\pomdp}$, $\varphi_1$  & \textbf{0.93}$^\dagger$ & 1034.64  & 0.92 & 160.32 \\
		Navigation (5) & 2725 &$\p_{\max}^{\pomdp}$, $\varphi_1$   & MO & MO & \textbf{0.95} & 311.65  \\
		Navigation (10) & 49060 &$\p_{\max}^{\pomdp}$, $\varphi_1$  & MO & MO & \textbf{0.85}  & 2561.02 \\
		Navigation (20) & 798040 &$\p_{\max}^{\pomdp}$, $\varphi_1$  & MO & MO & \textbf{0.98}* & 8173.03*  \\
		Navigation (30) & 4045840 &$\p_{\max}^{\pomdp}$, $\varphi_1$  & MO & MO & \textbf{0.97}* &  61350.34*  \\
		Navigation (40) & -- &$\p_{\max}^{\pomdp}$, $\varphi_1$ & MO & MO & TO  & TO \\
		\rowcolor{gray!25} Delivery (4) & 80 &$\Ex_{\max}^{\pomdp}$, $\varphi_2$  & \textbf{-6.0} & \textbf{28.53} & -6.04 & 94.32    \\
		\rowcolor{gray!25} Delivery (5) & 125  &$\Ex_{\max}^{\pomdp}$, $\varphi_2$ & \textbf{-8.0} & 102.41 & -8.13  & 150.44   \\
		\rowcolor{gray!25} Delivery (10) & 500 &$\Ex_{\max}^{\pomdp}$, $\varphi_2$  & MO & MO & -18.13 & 347.98  \\
		Slippery (4) & 460 & $\p_{\max}^{\pomdp}$, $\varphi_3$  & \textbf{0.90} & \textbf{5.10} & 0.80 & 180.15 \\
		Slippery (5) & 730 & $\p_{\max}^{\pomdp}$, $\varphi_3$  & \textbf{0.93} & \textbf{83.24}  & 0.89 & 212.79\\
		Slippery (10) & 2980 & $\p_{\max}^{\pomdp}$, $\varphi_3$    & MO & MO & \textbf{0.98} & 280.55 \\
		Slippery (20)  & 11980 & $\p_{\max}^{\pomdp}$, $\varphi_3$   & MO & MO  &\textbf{0.99} & 2384.56 \\
		\rowcolor{gray!25} Maze (1) & &$\Ex_{\max}^{\pomdp}$  & \textbf{-4.30} & \textbf{0.09} & -4.33 & 80.31 \\
		\rowcolor{gray!25} Maze (2)& &$\Ex_{\max}^{\pomdp}$  &   -5.23  & 2.176 & -5.34 & 114.23\\
		\rowcolor{gray!25} Maze (5)& &$\Ex_{\max}^{\pomdp}$     & -13.00$^\dagger$ & 4110.50  & -13.29 & 160.12\\
		\rowcolor{gray!25} Maze (10)& &$\Ex_{\max}^{\pomdp}$  & MO & MO & \textbf{-23.02} & 210.01  \\
		 Grid (3)& &$\Ex_{\max}^{\pomdp}$  & -2.88  & 2.332 & -2.90 & 87.31 \\
		Grid (4)&  &$\Ex_{\max}^{\pomdp}$  	& -4.13    &  1032.53  & -4.20 & 124.31 \\
		Grid (5)& &$\Ex_{\max}^{\pomdp}$ & MO & MO & -5.91 & 250.14  \\
		Grid (10)& &$\Ex_{\max}^{\pomdp}$  & MO & MO  & \textbf{-12.92} & 1031.21\\
		Grid (25)&  &$\Ex_{\max}^{\pomdp}$  & MO & MO & \textbf{-35.32} & 6514.30 \\
		\bottomrule
	\end{tabular}}
\caption{Computing policies for examples with LTL specifications.}
\label{tab:LTL_Prop}
\end{table}

\paragraph{Increasing the number of memory nodes improves performance.}
In Figure~\ref{fig:FSC_memory}, we show that increasing the number of memory nodes in the FSC produces higher performing policies, both in the form of higher probabilities of satisfying the specification and higher undiscounted expected rewards.
A noticeable characteristic is that for each FSC in Figure~\ref{fig:FSC_memory}, there is a point of diminishing returns where the additional memory does not produce higher quality policies. In most cases, this point falls between 6 and 8 memory nodes. As a consequence for the set of benchmarks, unless otherwise specified, we set the upper bound for the number of memory nodes at $\overline{B_h} = 8$.

\paragraph{Using counterexample data improves the quality of the extracted policies.}
Figure~\ref{fig:Refinement} compares the number of critical states in a set of counterexamples in relation to the probability of satisfying an LTL specification in each iteration of re-training for the proposed method. 
In particular, we depict the size of the set of critical states  $\mathit{Crit^{\pomdp}_{\fsc_{\hat{\gamma}}}} \subset S$ regarding the specification $\varphi$.
In Figure~\ref{fig:Refinement}, as the satisfaction probability and the expected reward increases, the number of the critical states identified by the verification decreases. 
In particular, the retraining of the RNN-based policy on the sequences of data generated using the local improvement step is effective in improving the policy with each iteration.

\input{figures/09-learn/refinement.tex}

\begin{figure}[t!]
	\centering
	\input{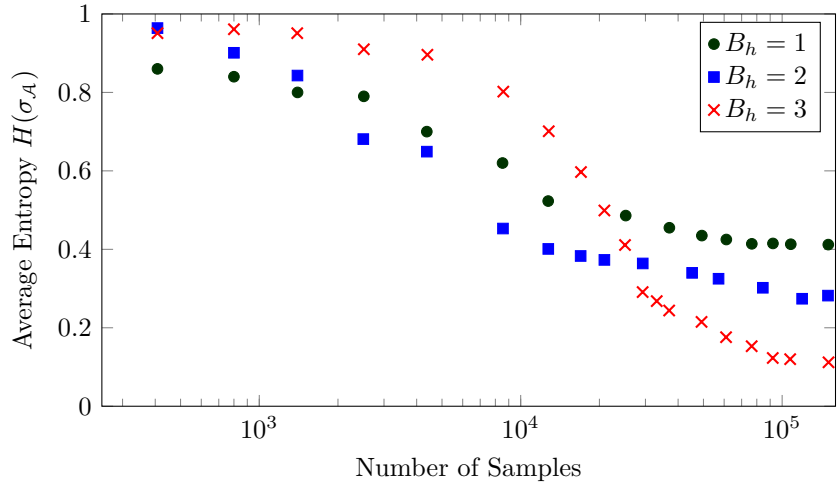}
		\caption{Entropy of the extracted FSCs for \emph{Navigation}($5$) from an RNN as it is trained with more samples. Each point represents an extracted FSC, for color sequence we fix the discretization and add more samples guided by the counterexamples. \label{fig:nav_entropy}\label{fig:entropy_plot}}
\end{figure}

\paragraph{Using counterexample data generates policies that make less arbitrary decisions.}
In Figure~\ref{fig:entropy_plot}, we ignore the decision at the entropy check, fix the memory precision, and iteratively add more sequences of data generated using the counterexamples.
Each point in Figure~\ref{fig:entropy_plot} represents one instance of verification in the loop in Figure~\ref{fig:high-level}.
As the RNN-based policy iteratively trains on additional sequences of data, the subsequent extracted policy makes less arbitrary decisions, shown in Figure~\ref{fig:entropy_plot} by the decrease in entropy of the FSC as the RNN-based policy is trained on larger sets of training sequences.

\paragraph{Increasing the number of memory nodes generates policies that make less arbitrary decisions.}
In Figure~\ref{fig:entropy_plot}, we also compare how the number of memory nodes in the extracted FSC correlates to the entropy of the decisions at critical states.
When trained on a large set of sequences of training data, the FSCs with a higher number of memory nodes have a lower entropy than those without.
This behavior is likely due the fact that extracted FSC with more memory nodes can better approximate the RNN-based policy, which itself is making less arbitrary decisions due to the larger training set.
Meanwhile, when the extracted FSCs are approximating RNN-based policies trained on smaller sets of training sequences, they generally make arbitrary decisions (see top left of Figure~\ref{fig:nav_entropy}). In these cases, the FSC with more memory nodes tend to make more arbitrary decisions than those with less, which is likely a function of an under-defined hidden state update $\hat{\delta}$ in the RNN-based policy.

\begin{figure}[t!]
	\centering
	\input{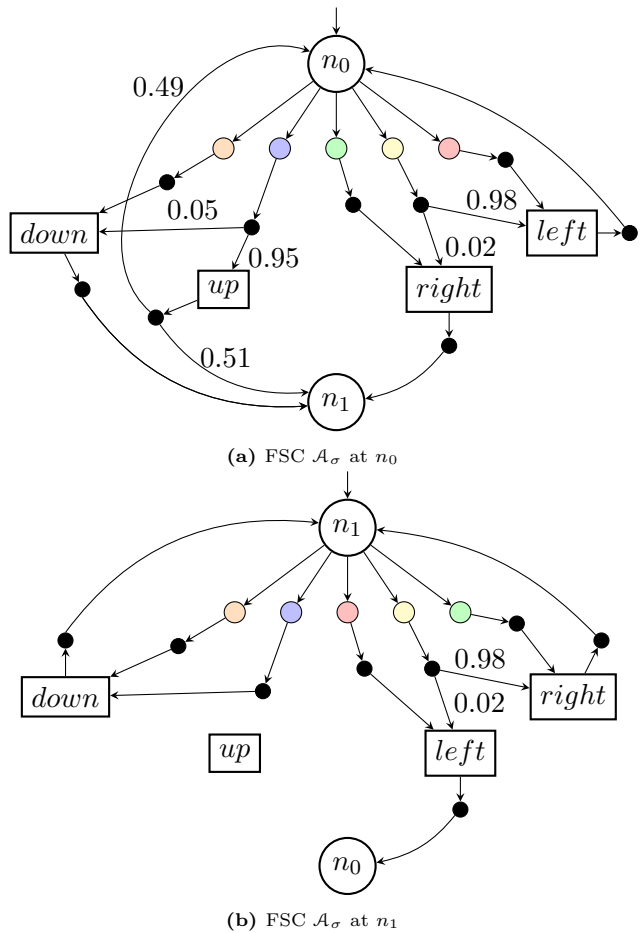}
	\caption{The corresponding FSC $\fsc_\gamma$ for Maze(1) example. The agent's initial state $s_I\in \states \backslash \lbrace s_{9}\rbrace$ is allocated over a uniform distribution and each color represents a different observation. The FSC $\fsc_\gamma$ has two memory nodes ($n_0$ and $n_1$), we prune action mappings and memory updates with low probabilities from \ref{fig:fscn0} and \ref{fig:fscn1}.}
	\label{fig:maze_fsc}
\end{figure}

\paragraph{Limiting the number of memory nodes creates a precision-performance trade-off.}
Increasing the number of memory states in the FSC produces policies with higher probabilities of satisfying the specification and greater expected rewards.
In Table~\ref{tab:LTL_Prop}, we include the sizes of the FSCs for the handcrafted procedure to demonstrate the trade-off between computational tractability and expressivity: a larger FSC means that the policy can store more information, which may lead to better decisions.
However, larger FSCs require more computational effort and may require more sequences of data for training the RNN-based policy.
Figure~\ref{fig:maze_fsc} shows the automatically extracted FSC for the \emph{Maze}($1$) environment. Note that a 2-FSC can represent the optimal \emph{Maze}($1$) policy. The FSC shown in Figures~\ref{fig:fscn0} and~\ref{fig:fscn1} is very close to this optimal policy.
The stochastic action choices at $(n_0,blue)$ and at $(n_1,yellow)$ create the suboptimality in this example with the optimal policy taking the respective $up$ and $right$ actions at these points.

\section{Case Study: Autonomous Driving in Traffic Lights}

\begin{figure}[t!]
	\centering
	\begin{subfigure}[b]{0.3\textwidth}
		\input{figures/09-learn/traffic}
		\caption{Environment}
		\label{fig:Traj_Env}
	\end{subfigure}%
	\begin{subfigure}[b]{0.3\textwidth}
	\begin{tikzpicture}
\node[] at (0,0) {\includegraphics[width=\textwidth]{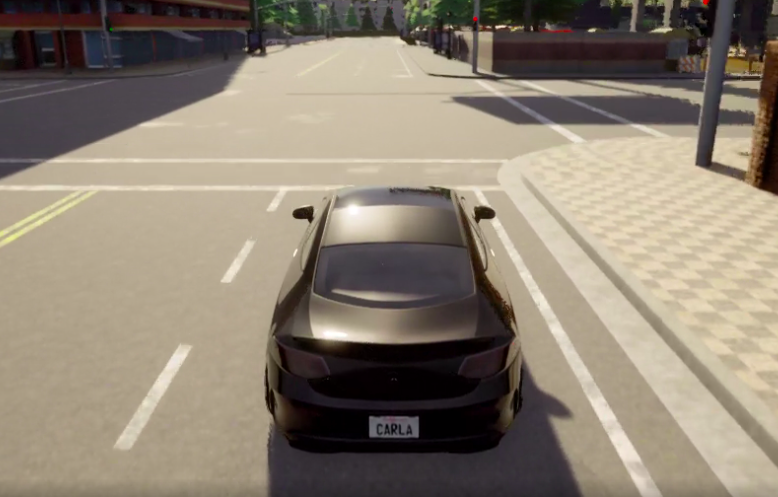}};
\begin{scope}[scale=0.5,yshift=-2.75cm,xshift=-0.5cm]
\fill[white,line width=1pt] (2,4) rectangle (4,5);
\draw[black,line width=1pt] (2,4) grid[step=1] (4,5);
\fill [orange,opacity=0.5]  (2,4) rectangle (3,5);
\fill [green,opacity=0.5]  (3,4) rectangle (4,5);
\node[] at (1,4.5) {\textcolor{white}{$\ObsFun(s)$}};
\end{scope}
\end{tikzpicture}
	    \caption{Sample observation}
	\end{subfigure}
	~~
	\begin{subfigure}[b]{0.3\textwidth}
		\scalebox{1.0}{\input{figures/09-learn//traffic_light}}
		\caption{Traffic light transitions with probability $p$.}
		\label{fig:Traf_Light}
	\end{subfigure}
	\caption{Autonomous car operating in an urban environment modeled as a gridworld and traffic lights. Each intersection has a decision point for the car who moves twice for each pedestrian move.}
	\label{fig:Traffic}
\end{figure}

Consider the scenario, pictured in Figure~\ref{fig:Traffic}, of an autonomous vehicle operating in a city with the following sensors:
\begin{enumerate}
    \item Navigation System (GPS),
    \item Optical Camera (Traffic Light Identification),
    \item Lidar (Proximity),
    \item Radar (Speed).
\end{enumerate}
At each intersection is a set of traffic lights that restrict the available action choices of the vehicle:
\begin{itemize}
    \item Red: $\lbrace \mathit{Stop} \rbrace$,
    \item Yellow: $\lbrace\mathit{Stop},\mathit{Straight}\rbrace$,
    \item Green: $\lbrace\mathit{Stop},\mathit{Straight},\mathit{Left},\mathit{Right}\rbrace$.
\end{itemize}
The state of each traffic light can be modeled as a Markov chain where the system switches from red to green and green to orange with probability $p=0.1$, see Fig.~\ref{fig:Traf_Light}.
In this environment there is a pedestrian, who moves stochastically but with reduced speed.
Each turn, the vehicle takes two actions while the pedestrian can move one intersection at a time.
While the vehicle must move according to the status of the lights, the pedestrian is under no such restriction and will ignore them.

\paragraph{State space description.} A state in this system 
is $s=(x,y,\theta,obs_x,obs_y,L)$, which is the location and direction of the vehicle and the location of the pedestrian as well as the status of the lights $L$ at each intersection.

\paragraph{Partial observability for the vehicle.} The car is only able to see the traffic lights that are directly in front of it.
For example in Figure~\ref{fig:Traj_Env}, the vehicle facing to the right can observe the lights of two intersections (the middle and the center right).
The vehicle uses its optical sensors to determine the status of the lights.
The task of navigating from an initial position $A_1$ to a goal location $A_2$ is modeled as a POMDP, with a belief on each status of the unseen intersection lights.

\paragraph{Specifications for the vehicle.} We demonstrate the effect of different specifications on feature splitting.
In particular, we describe three scenarios where the POMDP problem differs based on the safety task and the sensors that autonomous vehicle uses to measure the relevant features.

\begin{itemize}
    \item \textbf{Fastest trip to goal}---The vehicle attempts to navigate the changing lights to ensure the vehicle reaches the goal location $A_2$ ($\psi = \Ever A_2$).
    \item \textbf{Pedestrian never follows car}---The vehicle attempts to ensure the pedestrian cannot read the number plate of the back of the car. To read the plate the pedestrian must be behind the car (for example when the car faces east the label is defined by $B=\left( (x-obs_x)=1\land(y-obs_y)=0 \right)$) for two consecutive turns—described by $\psi_F = \neg \Ever(B \land \Next B) \Until A_2$.
    The vehicle uses the lidar proximity sensor to construct the relevant feature that determines its relative position to the pedestrian. 
    \item \textbf{Vehicle cannot speed within vicinity of pedestrian}---The vehicle moves twice for every one action cycle. Accordingly, we define ``speeding'' as moving straight $a_1$ twice through the green lights.
    For this specification, we utilize a lidar to sense the relative position as well as the radar to test speed. $\psi_G = \neg \Ever ((a_1 \land \Next a_1)\land \textrm{ped}  \Until A_2$ where $\textrm{ped}$ is when the pedestrian within one space of the vehicle, defined by $\textrm{ped}=|x-obs_x| + |y-obs_y| \leq 1$.
\end{itemize}

\begin{figure}[t!]
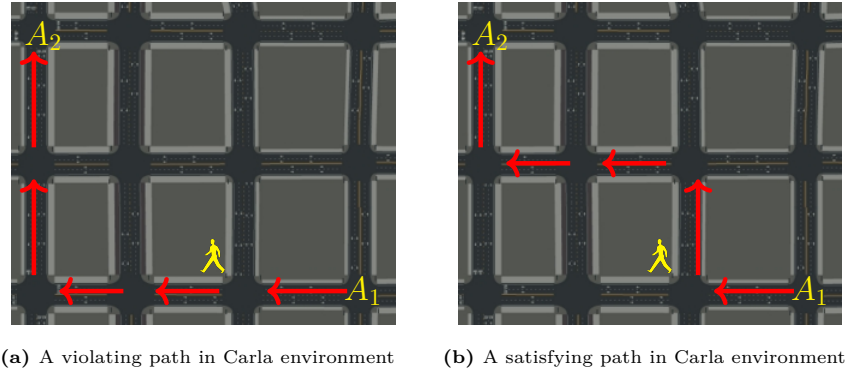

\centering
		\subfloat[A violating path in Carla environment \label{fig:Carla_unsat}]{
			\scalebox{0.85}{\input{./figures/09-learn/town0}}}
\quad
		\subfloat[A satisfying path in Carla environment 
            \label{fig:Carla_sat}]{
			\scalebox{0.85}{\input{./figures/09-learn/town1}}}
			\caption{Example paths in the autonomous vehicle case study. 
	The vehicle must avoid speeding past the pedestrian which would violate specification $\varphi_G$.}
	\label{fig:Carlashots}
\end{figure}

Figure~\ref{fig:Carlashots} shows the difference between a vehicle path that violates $\psi_G$ and one that satisfies $\psi_G$. In Figure~\ref{fig:Carla_unsat}, the vehicle at the intersection with the pedestrian, having previously taken $\textrm{straight}$, takes the same action again. This action causes the label $((a_1 \land \Next a_1)\land \textrm{ped})$ to be True and since it occurs prior to the vehicle reaching $A_2$, this path violates specification $\psi_G$. In Figure~\ref{fig:Carla_sat}, an alternate satisfying path has the vehicle turning right at this intersection.

\subsection{Experiments on the CARLA High-Fidelity Driving Simulator }

For the high-fidelity autonomous vehicle simulation, we implement the computed policies on the open-source simulator CARLA~\citep{DosovitskiyRCLK17}.
We run CARLA 0.9.11 on a 3.1 GHz machine with a GeForce RTX2060 graphics and 32GB of memory.
Table~\ref{tab:CarlaSynthTimes} contains the model sizes, synthesis compute times and the expected values for city-grids of size 3 and 4.

\begin{table}[t!]
	\caption{Synthesis times for autonomous vehicle POMDP}
	\label{tab:CarlaSynthTimes}
	\centering
	\scalebox{0.9}{
	\begin{tabular}{@{}lllrrr@{}}
		\toprule
		Case (grid size)  & Spec. & Size (states) & Time (s) & Result $V^{\osched}(b_0)$  \\
		\midrule
		Autonomous Vehicle (3) & $\psi$ & 2916 & 50.65 &  3.75\\
		Autonomous Vehicle (3) & $\psi_F$ & 26244 & 350.23 & 6.42   \\
		Autonomous Vehicle  (3) & $\psi_G$ & 26244 & 475.23  & 4.5 \\
		Autonomous Vehicle  (4) & $\psi$& 708588 & -TO- & -TO- \\
		Autonomous Vehicle  (4) & $\psi_F$& 6377292 & -TO- & -TO- \\
		Autonomous Vehicle  (4) & $\psi_G$& 6377292 & -TO- & -TO- \\
		\bottomrule
	\end{tabular}}
\end{table}

\begin{figure}[t!]	\centering
\input{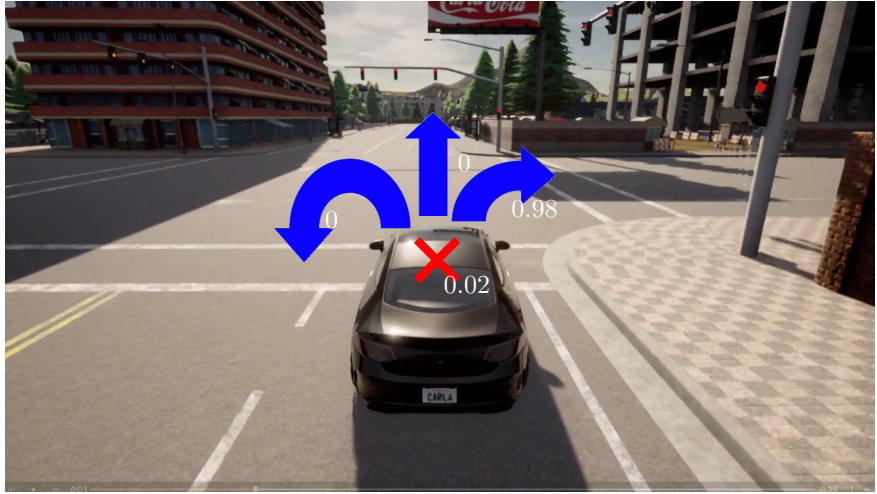}
			\caption{Output policy for vehicle at $s=(2,0,\textrm{west},2,0,L)$ (or at $t=1$ in Figure~\ref{fig:Carlashots}). The $\textrm{straight}$ action would result in a violation of specification $\varphi_G$ while taking $\textrm{right}$ leads to a path like that in Figure~\ref{fig:Carla_sat}.}
			\label{fig:Carla_policy}
\end{figure}

In Figure~\ref{fig:Carla_policy}, we see the synthesized policy for the situation that occurs at the pedestrian intersection in Figure~\ref{fig:Carlashots}. The computed policy rules out the choices of $\textrm{straight}$ and $\textrm{left}$. 

\chapter{Some Future Directions}
\glsresetall

We have seen how formal methods can assist with validating system models and generating provable guarantees. However, the increasingly complex structures necessary for implementing today's systems open new problems. In this section, we examine some paths for further developing formal methods that still contain a rich set of questions to answer. In fact, we show that formal methods must intersect with other areas to produce formalisms and tools for better validating and confidently assuring engineered systems. These intersections include but are not limited to fields such as learning, security, regulation, and certification. 

\section{Formal Methods for Reinforcement Learning} %

\Gls{rl} algorithms search for policies that are optimal with respect to user-specified objectives.
These algorithms allow for goal-oriented descriptions of complex behaviors and they provide a high degree of flexibility; they can be applied even when the system dynamics are high-dimensional, stochastic, and unknown \citep{sutton2018reinforcement, bertsekas2019reinforcement, powell2022reinforcement}.

Recently, deep \gls{rl} algorithms---which use neural networks to parameterize value and policy functions---have demonstrated empirical success in a variety of applications: e.g., controlling plasma configurations in a nuclear fusion reactor~\citep{degrave2022magnetic} and playing \textit{Chess}, \textit{Shogi}, and \textit{Go} at superhuman levels~\citep{silver2018general}.
In these examples, the application of neural networks enables approximate solutions to problems in decision and control that would otherwise not be possible; the state and action spaces of these examples are too large for exact solution computed via dynamic programming algorithms.

However, there remain barriers to the deployment of \gls{rl} algorithms in many engineering applications.
Autonomous vehicles, power systems management, and robotic systems are examples of complex application domains that require strict adherence of the system's behavior to stakeholder requirements.
However, the verification of \gls{rl} policies is difficult.
This is particularly true for deep \gls{rl} algorithms, which typically only output the learned policy and its estimated value function, rendering their resulting behaviors opaque to further verification and analysis.
The introduction of techniques borrowing from formal methods is necessary for the development of \gls{rl} algorithms that yield behaviors with verifiable properties.
To achieve this aim, we require frameworks to incorporate verifiable models into \gls{rl} algorithms.

For example, given some temporal logic specification \citet{alshiekh2018safe} synthesize reactive systems called \textit{shields}, which prevent reinforcement learning systems from taking unsafe actions with respect to the specification.
Meanwhile, a number of works have studied the use of \gls{rl} as a method of controller synthesis for temporal logic objectives \citep{hahn2019omega, hasanbeig, bozkurt2020control, hasanbeig2019}.
While \citet{wenLearning, djemou2021task} use linear temporal logic formulas as \textit{side information} that constrains the outputs of inverse reinforcement learning algorithms. 

A general framework that brings verifiable models into reinforcement learning algorithms is that of the \textit{reward machine}---a type of finite state machine used to encode reward functions in \gls{rl} problems \citep{toro2022reward}.
These structured task representations can encode non-markovian tasks, and they can be exploited to improve learning efficiency \citep{icarte2018using, xu2019joint, icarte2019learning} and to automatically translate specifications given in formal languages, such as linear temporal logic, into reward functions \citep{camacho2019ltl}.
In the context of multiagent \gls{rl}, \citet{neary2020reward} use reward machines to specify multi-agent tasks, and to decompose these tasks into subtasks to be learned by individual agents through decentralized \gls{rl} algorithms.
The authors use the structure of the reward machines to prove conditions under which the resulting learned behavior is guaranteed to accomplish the original task.

Beyond the framework of reward machines, compositional design of RL systems can be used to greatly reduce the complexity of, and to more easily verify, individual subsystems.
By creating well-defined interfaces between subsystems, system-level requirements may be decomposed into component-level ones. 
Conversely, each component may be developed and tested independently, and the satisfaction of component-level requirements may then be used to place assurances on the behavior of the system as a whole.
Towards these ends, \citet{neary2021verifiable} propose a framework to verify compositional \gls{rl} systems against probabilistic task specifications.
The framework builds a \textit{high-level} system model, represented as a Markov decision process, which is used for high-level planning and to automatically generate subtask specifications for the \textit{low-level} subsystems, each of which is implemented as an independent \gls{rl} agent.
\citet{jothimurugan2021compositional} similarly build compositional \gls{rl} systems by using a graph-based representation of high-level tasks, and by using \gls{rl}-based controllers to accomplish all necessary subtasks.

The above references provide examples of ways in which properties of \gls{rl}-trained policies may be verified.
Future work that continues to develop such frameworks and that applies them in experiments is necessary for the eventual deployment of trustworthy autonomous systems that incorporate \gls{rl}-trained components.

\section{Operating Under Limited Information and Concealing Information}
The operation of autonomous systems relies on the information flow both within the components of the system and between the system and its environment. Most of the formal methods for autonomous systems discussed in this review inherently assume that this information flow is perfect. For example, numerical operations can be carried out with full accuracy, the sensor inputs are not quantized, and the communication channels are not noisy. However, in reality, the information is distorted in many ways due to both the system's internal design and the environmental factors. For example, cyber-physical systems typically have bandwidth limitations that require the sensor and controller outputs to be quantized~\citep{franceschetti2014elements}. In addition to these naturally occurring distortions, the information flow to the system may be adversarially modified. 

The dependency on information brings multiple questions: What are the possible sources of distortion? What is the lowest amount of information that can ensure the safe operation of a system? How can we modify the existing formal methods for autonomous systems to account for these distortions? As an answer to these questions, early works from control theory show that the stability of a dynamical system imposes a lower bound on the information rate of a system~\citep{nair2004stabilizability,franceschetti2014elements}. For the discrete dynamical systems, \citet{tanaka2021transfer} and \citet{eysenbach2021robust} provided methods that limit the information flow between the components of a system that is modeled with a Markov decision process. In the context of multi-agent systems, \citet{wang2020learning} and \citet{karabag2022planning} minimize the information shared between the agents to improve the performance under communication loss. In the case of an adversarial corruption of information flow in deterministic systems, we can represent the adversary as an additional player, represent the synthesis problem as a two-player game, and utilize the reactive synthesis methods mentioned in Section \ref{sec:reactive-synthesis}. 

On the flip side, an autonomous system must not leak critical information about itself to maintain the success of its operation. There is a growing literature on minimizing the information leakage of a system by considering concepts such as opacity~\citep{saboori2007notions,tong2018current,berard2015quantifying}, privacy~\citep{such2017privacy,gohari2020privacy}, deceptiveness~\citep{zhang2018hypothesis,karabag2021deception}, and estimation error~\citep{farokhi2017fisher,karabag2019least}. There is a trade-off between minimizing the information leakage by considering such concepts and the performance of the system. However, minimizing the information leakage gives robustness to the system against its adversaries, thereby maximizing the performance in the long run.

\section{Safeguarding Information Privacy in Autonomous Decision-Making Systems}

In this survey, we mainly addressed the safety implications of autonomous decision-making systems. In particular, we surveyed papers that generally study the problems of policy synthesis and verification with respect to certain mathematically and formally specified objectives; for example, we frequently revisited the case study of an autonomous vehicle that must avoid crashing into obstacles and adhere to certain traffic rules formulated via temporal logic. While such developments are crucial to safeguarding the safety of the individuals whose daily lives will be affected by the deployment of autonomous decision-making systems, the societal impacts of these systems may extend far beyond the matters discussed in this survey; these systems often incorporate confidential, proprietary, operational, personal, or otherwise sensitive information in their decision-making algorithms, which raises privacy concerns.

Markov decision processes have been a major part of the formulation of many policy synthesis and verification problems discussed in this paper. \citet{gohari2020privacy} study a policy-synthesis problem in which it is within the privacy interests of a decision-maker to keep the transition probabilities of the underlying Markov decision process confidential while publicly taking actions according to synthesized policy. The paper uses the framework of differential privacy to obfuscate the transition probabilities and then synthesizes a policy based on the obfuscated transition probabilities using dynamic programming. Then, the differential privacy of the overall synthesis algorithm becomes immediate due to differential privacy's immunity to post-processing. 
Although the proposed policy-synthesis algorithm addresses some of the named privacy concerns, it is not clear how the algorithm can satisfy a set of safety specifications, especially that differentially private algorithms are known to often trade off utility. Future work that incorporates the approaches discussed in this survey may be a potential solution to this issue.

Recall that, in Section \ref{chapter:learning}, we reviewed an approach for policy synthesis and verification in which the policy is represented via recurrent neural networks. There exists mounting evidence that processing data in the form of training neural networks has privacy ramifications for the training data \citep{mireshghallah2020privacy}. \citet{yang2021privacy} show that the privacy ramifications of training recurrent neural networks can be worse than conventional feed-forward neural networks, especially in the task of deep reinforcement learning which is closely related to the policy-synthesis problem discussed in Section \ref{chapter:learning}. The authors further study mitigation methods that leverage the promise of differential privacy by perturbing the trainable parameters of the neural network under protection. In this case, it must be further studied how the perturbations in the name of differential privacy affect a policy's ability to satisfy a given set of safety specifications.

\section{Explainability in Verification and Synthesis}

Formal methods such as model checking~\citep{BK08} are capable of verifying human-generated robotic mission plans against a set of requirements~\citep{Humphrey+2013}.
In cases in which the plans may violate the requirements, such techniques generate \emph{counterexamples} that illustrate requirement violations and provide valuable diagnostic information~\citep{WJA+14,feng2016human}.
Nevertheless, these artifacts may be too complex for humans to understand, because existing notions of counterexamples are defined as either a set of finite paths or an automaton typically with large number of states and transitions.

Counterexample generation for model checking Markov decision processes has been studied in several works using different representations of counterexamples:
\cite{han2009counterexample} computes the smallest number of paths in a Markov decision process whose joint probability mass exceeds the threshold and formulates the counterexample generation as a k-shortest path problem;
\citep{WJA+14} computes a critical subsystem of a Markov decision process with the minimal number of states and proposes solutions based on \gls{milp} and SAT-modulo-theories.
There are several attempts to generate human-readable counterexamples:
\citep{wimmer2013high} computes a minimal fragment of model description in some high-level modeling language (e.g., probabilistic guarded commands),
while \citep{feng2016human} computes structured probabilistic counterexamples as a sequence of ``plays'' that capture the high-level objectives in UAV mission planning.
\citet{feng2018counterexamples} define a notion of explainable counterexample, which includes a set of structured natural language sentences to describe the system behavior that lead to a requirement violation in an MDP model of a robotic mission plan.
They propose an approach based on \gls{milp} for generating explainable counterexamples that are minimal, sound and complete.

The study of natural language for robotic applications has mostly focused on translating human instructions expressed in natural language to robotic control commands.
For example, \citet{hayes2017improving} considers the problem of synthesizing natural language descriptions of robotic control policy. \citet{lignos2015provably} present an integrated system for synthesizing reactive controllers using natural language specifications which are translated into linear temporal logic formulas. If unsynthesizable, the minimal unsynthesizable core is returned as a subset of the natural language input specifications.

\section{Regulation}
Regulatory requirements have to be incorporated when deriving specifications for safety-critical
systems.
Autonomous driving, for instance, is one of the fast-growing domains where regulation plays a significant
role.
Regulatory requirements, in this case, include the rules of the road and traffic regulations.
Some rules, however, are ambiguously stated and include implicit assumptions related to the
capability and rationality of human drivers.
Such ambiguity potentially leads to inconsistent interpretation of the rules among different
developers, complicates the design process, and ultimately jeopardizes the safety and efficiency of the system.

The lack of precise specifications has been acknowledged both by industry and academia
\citep{ShalevShwartz:2017:Formal,Phan-Minh:2019:Towards,Censi:2019:Liability,Harel:2022:Creating}
and is a significant impediment towards formal methods realizing their full potential in this domain.
Furthermore, without a clear description of how autonomous vehicles should
behave, especially around dynamic objects such as humans and other vehicles, a common practice is to
start with an implementation of specific behaviors and evaluate it based on simulation and testing. This
approach is costly, as evidenced by the delay of deploying autonomous vehicles from the original estimate
of 2020 \citep{Spectrum:2020:Surprise}.
Having a precise specification not only speeds up the development process but also ensures transparency
and predictability of the system while giving the developers the flexibility of picking an implementation
that leads to admissible behaviors as defined by the specification. Finally, it equips the regulators with
an objective measure to validate autonomous vehicles.

\citet{Censi:2019:Liability} make an initial attempt to tackle this problem by introducing frameworks for describing these specifications.
The focus of these initial works is on the structure
of the specifications that allows trajectories to be evaluated based on the violation of individual rules, which
incorporate different factors such as safety, law, ethics, culture, etc. Systematic approaches to derive the
specifications, taking into account these factors, however, remain an open problem.

\section{Dynamic Certification}

In addition to more conventional versions of regulation that focus on \emph{regulatory requirements}, autonomous systems require a more proactive approach to certification. Such dynamic approaches use formal methods as a form of design guide, to give a quick and inexpensive way of circling through different scenarios and contexts where the system ought to behave as expected. This type of \emph{dynamic certification} is the iterative revision of permissible $\langle \text{use}, \text{context} \rangle$ pairs for a system---rather than prespecified tests that a system must pass to be certified~\citep{bakirtzis:2022}, meaning that we can only certify for particular operational requirements and environments and can never fully guarantee any system-level requirements preserve in \emph{any} deployment scenario. 

Specifically, dynamic certification is based on applying testing that is informative to the different stages of a systems lifecycle, from early to transitional to finally confirmatory. These testing phases are not executed linear and testing is not expected to certify for \emph{any} context. Rather, the expectation of testing, compared to more traditional methods within certification where the goal is \emph{to be} certified, is continual and inputs of new contexts must again be tested through a series of living documents (e.g., capturing changing requirements within formal models), simulations (e.g., congested environments vs. open areas), and controlled environments before deployment (e.g., testing car tracks). 

The output of dynamic certification is acceptable scenarios of operation, including but not limited to identified and potentially mitigated hazardous states, admissible environments of operation (e.g., congested vs. non-congested streets for an autonomous car), and modified system-level requirements. While the word output might suggest that the process terminates, the expectation is closer to ``terminates yet starts again'': Given that autonomous systems operate in changing environments it is natural to suggest that dynamic certification must apply to those changes. Having done dynamic certification for a $\langle \text{use}, \text{context} \rangle$ pair, however, only needs to document and test the changed parameters and do a form of regression testing in models, simulations, and controlled environments to make sure those changes do not break our overall assumptions and guarantees.

Autonomous systems rely on two types of decision-making: design and deployment decisions. Design decisions concern the structure and intended operation of the autonomous system, which can include criteria and metrics of correct behavior and revisions to system requirements, behaviors, and architectures (reflected to formal models, testings procedures, and controlled environment parameters). Deployment decisions concern the contexts and uses for the autonomous system, including what hazardous states ought to be eliminated or mitigated against. Dynamic certification considers both.

Further research in autonomy is paramount to achieve a level of granularity that accounts for design and deployment decisions and their associated models and testing, particularly when dynamic certification is partially implemented with formal methods. Such thrusts can include finding more economical ways to compute formal models, addressing uncertainty to better match environmental parameters, and accounting for learning algorithms that are composed with traditional control-theoretic methods.

\begin{acknowledgements}
The work this article reported has been carried out in collaboration with a large group of researchers. We are thankful to all of them.
  
The preceding sections draw material from several of our earlier publications heavily. We would like to express our appreciation to our co-authors on those publications: Erdem Arinc Bulgur, Kai-Wei Chang, Rayna Dimitrova, Lu Feng, Nils Jansen, Sebastian Junges, Joost-Pieter Katoen, Hadas Kress-Gazit, Ahmadreza Marandi, Richard Murray, Marnix Suilen.

Over the course of preparing this article, the authors affiliated with The University of Texas at Austin have been supported by the following grants: NASA 80NSSC21M0071, NSF CNS-1836900, NSF 1646522, NSF 1652113, NSF 2141153, ARL ACC-APG-RTP W911NF1920333, AFRL FA9550-19-1-0169, and AFOSR FA9550-19-1-0005. 

Wongpiromsarn is supported by NSF awards 2141153 and 2211141.
\end{acknowledgements}

\backmatter  %

\printbibliography
\printglossaries
\end{document}